\documentclass[paper]{elsarticle}

\usepackage{hyperref}
\usepackage{graphicx}
\usepackage[utf8]{inputenc}
\usepackage{color}
\usepackage{amsmath}
\usepackage{amssymb}
\usepackage[]{algorithm2e}

\newcommand{\msc}[1]{{\color{black}#1}}
\newcommand{\mscc}[1]{{\color{black}#1}}

\journal{Computer Vision and Image Understanding}

\begin{document}

\begin{frontmatter}

\title{3D non-rigid registration using color: Color Coherent Point Drift\tnoteref{mytitlenote}}
\tnotetext[mytitlenote]{Authors acknowledge the projects from University of Alicante (Gre16-28) and Spanish Ministry (TIN2017-89069-R) supported with Feder funds.}

\author[mymainaddress]{Marcelo Saval-Calvo\corref{mycorrespondingauthor}}
\cortext[mycorrespondingauthor]{Corresponding author}
\ead[url]{www.dtic.ua.es}
\ead{msaval@dtic.ua.es}

\author[mymainaddress]{Jorge Azorín-López}
\author[mymainaddress]{Andrés Fuster-Guilló}
\author[mymainaddress]{Víctor Villena-Martínez}
\author[mysecondaryaddress]{Robert B. Fisher}

\address[mymainaddress]{University of Alicante, Carretera Sant Vicent del Raspeig s/n, 03690,Spain}
\address[mysecondaryaddress]{School of Informatics, Univ. of Edinburgh, 10 Crichton St, Edinburgh EH8 9AB, UK}

\begin{abstract}
Research into object deformations using computer vision techniques has been under intense study in recent years. A widely used technique is 3D non-rigid registration to estimate the transformation between two instances of a deforming structure. Despite many previous developments on this topic, it remains a challenging problem. In this paper we propose a novel approach to non-rigid registration combining two data spaces in order to robustly calculate the correspondences and transformation between two data sets. In particular, we use point color as well as 3D location as these are the common outputs of RGB-D cameras. We have propose the Color Coherent Point Drift (CCPD) algorithm (an extension of the CPD method \cite{Myronenko2010}). Evaluation is performed using synthetic and real data. The synthetic data includes easy shapes that allow evaluation of the effect of noise, outliers and missing data. Moreover, an evaluation of realistic figures obtained using Blensor is carried out. Real data acquired using a general purpose Primesense Carmine sensor is used to validate the CCPD for real shapes. For all tests, the proposed method is compared to the original CPD showing better results in registration accuracy in most cases.

\end{abstract}


\end{frontmatter}


\section{Introduction}
\label{sec:ccpd:intro}

The study of the evolution of shapes over time is under intense study in many areas, such as biology, health, etc. During evolution, objects are affected by multiple changes, disturbing both shape and appearance. To measure all the changes is a difficult and tedious task, due to the complexity of some shapes and the large amount of data necessary to have a complete study. Computer vision techniques can help provide methods which, given a set of data from a sensor, estimate the changes. In this paper, we propose a method to robustly estimate the deformation observed in an object. Concretely, non-rigid registration methods estimate the transformation between two shapes aligning the data using non-rigid transformations.  

\msc{There are many applications that require non-rigid alignment. For instance, face or body motion recovery where the different parts need to be tracked to perceive the motion or identify the action. Applications where shape evolution is studied require deformable alignment as well, and may also involve appearance changes, which commonly include color variations. Using machine intelligence to evaluate those changes requires using methods than can perceive them regardless the nature of the change. For example, intelligent farms can use these techniques to improve the quality of their products since they can be constantly supervised while growing. In health, automatic analysis of human body change will help specialists in treatment supervision (eg. for cancer therapy).}

There exist various kind of deformations: isometric deformation, where both topology and distances are preserved (e.g. articulated changes or flag movements); elastic deformation, where the topology is kept but distances can vary (e.g. balloon inflation); and free deformations where both topology and distances can change (e.g. growing objects or breaking situations).

In this paper we focus on 3D point clouds without any previous filtering, only downsampling if necessary. For the specific case of this paper, the data comes from a low-cost RGB-D sensor, such as a Microsoft Kinect, which provides color and 3D information. The sensitivity of these sensors may be \msc{lower than} the requirements of the problem, and usage \msc{may be difficult for some tasks}. Nonetheless, they are widely used and a contribution suitable for working with data from both low and high quality sensors will be useful in many research tasks and industrial applications. 


The deformations considered in this work are not constrained. That is, they do not assume any prior restrictions to the deformations such as topology/size constraints, larger/smaller variations, etc. The objective is to develop a non-rigid registration method for non-constrained free deformations. 

\msc{Non-rigid registration methods for 3D point sets, such as the well-known Coherent Point Drift (CPD) \cite{Myronenko2010}, only use spatial 3D information (or location information) to register the data. Ignoring other information, such as color, increases the probability of misalignment. For instance, in cases where the object grows the number of points may increase or decrease in an irregular distribution. If only 3D spatial data is taken into account, those irregularities are harder to register. Those are situations where additional information can be used to robustly register. A practical example is the plant growth, where leaves change shape differently over their surface. Commonly, the central region remains similar whilst the edges enlarge significantly, but in the spatial data this variation in growth is not as obvious. It is necessary to use color information to perceive this difference. The leaf growth problem  motivates our work, which extends the CPD algorithm to include color information in the process of registration to improve the estimation of the deformations. Although originally motivated by the leaf growth problem, the developed Color Coherent Point Drift (CCPD) algorithm is a general algorithm usable for registering deforming colored point clouds.}

\msc{The main contribution of this paper is a novel approach for colored point cloud non-rigid registration combining various inputs in the correspondence estimation step. To handle real and adverse situations, the method has to deal with noise, outliers and missing data, common issues in real applications. The proposal makes use of the basis proposed in the CPD algorithm \cite{Myronenko2010}, because of its generality and because it has shown good results in point set non-rigid registration in presence of noise and outliers.}

The rest of the paper is organized as follow: Section \ref{sec:soa} presents a review of the State-of-the-Art in 3D non-rigid registration methods for point sets. Section \ref{sec:ccpd:method} details the proposed CCPD method. The evaluation is shown in Section \ref{sec:ccpd:exp} where synthetic and real data are used to validate the proposal. Finally, some discussion and conclusions are presented in Section \ref{sec:conclusions}. 

\section{Previous research}\label{sec:soa}

Recently, the increasing interest in non-rigid registration has produced much research that improves existing algorithms or introduces new methods, but this is still a challenging problem to be solved. This interest comes from the need to improve reconstruction, mapping or other computer vision problems, where dynamic objects are treated. Tam et al. \cite{Tam2013} surveyed different methods for point cloud and mesh registration, in both rigid and non-rigid situations. 

Chui and Rangarajan \cite{Chui2000,Chui2003} proposed the TPS-RPM non-rigid registration method for 3D point clouds based on Thin Plate Splines to stabilize the displacement of the points during the process of registration. This method uses softassign matches between each point set \cite{Rangarajan1997}. Softassign refers to the use of non-binary correspondences to handle noise and outliers because there is no imposing of a unique matching per point. Deterministic annealing \cite{Ueda1998} is also used in the kernel of TPS-RPM to gradually allow a less constrained movement of the individual points. \msc{Their proposal outperforms ICP in 2D, and also achieves better results in 3D than the main state-of-the-art methods. Yang revisited TPS-RPM in \cite{Yang2011} demonstrating limited performance when outliers are present in both point sets simultaneously. He proposed a double-sided outlier handling approach obtaining better registration results.}

Li et al. \cite{Li2008} presented a non-rigid registration method that simultaneously estimated confidence weights, that measure the reliability of each correspondence, and identified non-overlapping areas. A warping field brings the source scan into alignment with the target geometry.

Sang et al. \cite{Sang2013} proposed the FDMM non-rigid registration method based on GMM and the use of features, that they called Gaussian soft shape context, based on radial distribution of the neighbourhood. This feature was initially presented in \cite{Belongie2000,Belongie2002}, and they modified it adding a Gaussian distribution for avoiding the problem of non-real similarities. The algorithm takes into account the relative distribution of all points with respect to the analysed point, making a histogram, which adds information to the registration process. Comparison to CPD, RPM and BEM \cite{Sang2012} is provided using 2D data, outperforming the previous results. Yawen et al. \cite{Yawen2014} proposed also the use of this feature enhancement with CPD to handle noise and outliers with better results. 

Wang and Fei \cite{Wang2008} proposed B-spline-based point matching (BPM), an extension of RPM, using a deterministic annealing scheme to regularize the registration process. The method was evaluated in different situations with accurate results in 2D and 3D data. Yang et al. proposed in \cite{Yang2014} GLMD, a two step non-rigid registration method for point sets. They proposed the use of local and global distances combined to estimate the binary correspondences, and the transformation using the TPS kernel. The local distances are measured using a certain neighbourhood, which is provided initially. Experiments were provided using the proposed method against CPD, TPS-RPM and GMMreg for different levels of noise, outliers and rotations. 

Recently Chen et al. \cite{Chen2015} proposed the Coherent Spatial Mapping (CSM) algorithm. They used the shape context \cite{Belongie2000} which describes the shape using a histogram of each point relative position to the others, and calculated correspondences with this information. The Hungarian method is also used to estimate the initial correspondences. The transformation is iteratively estimated with the EM method using a spatial mapping function of the correct matches, and TPS to provide smooth deformations. Hence, the improvement comes from the matching estimation. \msc{They compare CSM to CS \cite{Belongie2002}, CPD, COA-RPM \cite{Lian2012} and TPS-RPM with 2D data achieving better alignment with lowest RMS error with different levels of noise and outliers. In 3D they compare against CPD achieving lower registration error.}

S. Lin et al. presented in \cite{Lin2016} a proposal for incorporating color in the registration process, both in rigid and non-rigid registration. The non-rigid approach is based on the paper of Li et al. \cite{Li2008}, incorporating the color information in the vertex selection by evaluating 3D location and color distance in Euclidean space, using a neighborhood to improve robustness, between the two views. Moreover, after estimating the descriptors (Gabor and HOG) from the vertex, color is also used for rejecting wrong correspondences. This paper considers small deformations mainly related to orientation of views which deform the shapes due to the RGB-D sensor pattern projection.

\subsection{Coherent Point Drift variants}
One of the most common algorithms used for non-rigid registration is the Coherent Point Drift (CPD) proposed by Myronenko et al. in \cite{Myronenko2010,Myronenko2006}. This method is based on a Gaussian Mixture Model (GMM) and Expectation Maximization (EM) to calculate the correspondences, and then the transformations, of the points to map one set of points into another. They used a GMM to represent the moving point set to be registered, and EM to evaluate the new parameters of the GMM and hence, the new position of the points. Moreover, in order to constrain the movement, they made use of Coherent Motion Theory that helps the translation of points to be regular. They compared their results to the TPS-RPM outperforming the registration for 2D and 3D cases.
Wang et al. \cite{Wang2011} proposed an extended version of CPD to automatically evaluate the outlier percentage parameter, which is manually provided in the original version. They used a combination of Nelder-Mead simplex and genetic algorithms. The genetic algorithm provides good initial values for this parameter, while the Nelder-Mead simplex optimizer attempts to find an optimal solution. The experimentation \msc{showed} an improvement of the original version for different levels of noise, where they initialized the outlier parameter to 0.7.

A different approach called GMMreg was presented by Jian et al. in \cite{Jian2005,Jian2010}. Instead of representing a point set with a GMM and registering it to a point cloud using the EM technique, they align two GMMs each representing one of the point sets to be registered. They calculate the displacement between Mixtures of Gaussians and iteratively align them using the L2 distance. They provided rigid and non-rigid results for 2D and 3D data compared to the original CPD, LM-ICP \cite{Fitzgibbon2003}, and TPS-RPM among others, resulting in more accurate results. \msc{Additionally, they apply the L2 distance to TPS and to Gaussian radial basis functions, improving the results.}

Gerogiannis et al. \cite{Gerogiannis2012} proposed a different matching method using the Hungarian Algorithm instead of the posterior distribution used in CPD and RPM. Moreover, they used Bayesian regression for the Maximization step (i.e. the registration or transformation part). The experiments compared the proposed method with CPD, RPM and GMMreg for 2D and 3D cases. 

Gao et al. studied in \cite{Gao2013} the main drawbacks of CPD related to outliers, which are a consequence of the way CPD keeps the distribution of outliers, and the input parameter for the outlier ratio. They proposed an Expectation-Maximization solution to iteratively evaluate the outlier ratio. TPS-RPM and the original CPD algorithms show less accurate results when the outlier ratio grows. The main advantage of this method is to avoid the need to indicate the outlier ratio initially.

Ge et al. \cite{Ge2014} presented a similar approach to the previous one, called Global-Local Topology Preservation (GLTP). The main motivation of this work is to handle non-rigid articulated deformations such as those of human movements. They added the principle of Local Linear Embedding to the original CPD to take into account local deformation coherence, apart from the global coherence intrinsic in the CPD algorithm. With large articulated deformations GLTP works better than the original CPD, which is not able to find a good registration.

De Sousa and Kropatsch \cite{DeSousa2014} proposed a variant of Coherent Point Drift (CPD) by integrating centrality information, a concept initially applied in social networks. It creates a graph (e.g. Delaunay triangulation), and applies different centralities (node degree, betweenness, eigenvector ...) to evaluate which results in a better solution. The proposal shows good performance with noisy data, improving the original CPD.

Another variation of CPD was presented by Zhou et al. \cite{Zhou2014a} using Student’s mixture model, which they claim to be more robust in the presence of high amounts of noise. The comparison they made against CPD and TPS-RPM shows better performance when the noise rate grows. Moreover, they automatically estimated the probability of outliers whereas in CPD it is manually indicated.

In conclusion, many studies have been done for non-rigid registration of point sets. Most of them focused their attention on outliers and noise handling. In order to do this, they proposed techniques to estimate automatically the outlier ratio or used descriptors which use point distributions to improve the matching. However, there still exist problems when there are large deformations. Another issue not studied is where the data does not have to move coherently in the whole space. For example, situations in which one set is a full model and the other is just a region. Moreover, there are no general proposals facing the problem from a generic perspective including several sources of data using individually the different spaces, e.g. using color and 3D location without using them as a 6D data set but being independent in the process for a more robust and generic combination.

\section{Color Coherent Point Drift}
\label{sec:ccpd:method}

In this section, a framework for non-rigidly registering 3D colored points based on CPD \cite{Myronenko2010,Myronenko2006} is presented. We use the optimization algorithm of the original CPD algorithm, only replacing the original similarity matching formulation with one that takes account of having colored 3D points.

The proposed Color Coherent Point Drift (CCPD) algorithm registers 3D points by using color and shape spaces to jointly estimate the best match. It improves upon the CPD algorithm by using the two input spaces together to handle situations where point position is not sufficient to adequately estimate the matches, {\it e.g.} aligning shapes with missing parts, or non-linear growth of the shape.  


In any registration problem it is normal to have one point set used as the anchor or reference point set which we will call \textit{Anchor}, and the other as the moving points called \textit{Moving}. The \textit{Moving} set will be deformed and moved until it aligns with the \textit{Anchor}. CCPD (following the basics of CPD) models the \textit{Moving} set using a Gaussian Mixture Model (GMM) and estimates the transformation of the \textit{Moving} set using the Expectation-Maximization (EM) technique. The use of a GMM to represent the \textit{Moving} set will give soft correspondences, {\it i.e.} they are not binary, allowing a more robust estimation of the displacement by not requiring one-to-one matching. Moreover, in order to smooth the displacement, the Coherent Motion Theory is used to regularize the motion of the points in the process of the transformation. 

Here, we introduce the combination of color and shape (3D positions) spaces for non-rigid registration. Let $A^S$, $A^C$, $M^S$ and $M^C$ (Eq. \ref{eq:tuples}) be four data sets representing two spaces (shape and color) of two data sets. $A^S$ and $A^C$ are the shape and color values of the \textit{Anchor} set and $M^S$ and $M^C$ are the shape and color values of the \textit{Moving} set. To simplify the notation, we will refer to $A^S,~A^C$ as $A$, and $M^S,M^C$ as $M$ when we refer to both spaces together.

\begin{equation}\label{eq:tuples}
\begin{array}{l}
A^S = \{a^S_1, \cdots, a^S_N \}\\
A^C = \{a^C_1, \cdots, a^C_N \}\\
M^S = \{m^S_1, \cdots, m^S_M \}\\
M^C = \{m^C_1, \cdots, m^C_M \}\\
\end{array}
\end{equation}
where $a^S_i, m^S_i \in \mathbb{R}^{D_S}$ and $a^C_i, m^C_i \in \mathbb{R}^{D_C}$.
$N$ and $M$ are the number of points in the  \textit{Anchor} and \textit{Moving} point sets. $M^S$ and $M^C$ are the \textit{Moving} to be aligned with the reference \textit{Anchor} $A^S$ and $A^C$. Each space has its own dimension {\it e.g.}
$D_S=3$ for shape (3D points), but $D_C=1$ for monochrome or $D_C=3$ if 
we use 3 color components. The points $M^S$ and $M^C$ are appended to form the centroids
of the components of a Gaussian Mixture Model (GMM) (m = 1..$M$) that encodes the 
probability of the \textit{Moving} point set, as described in Eq. \ref{eq:gmmini}. 
$x$ and $m_i$ are vectors with the point's position and color appended,

\begin{equation} \label{eq:gmmini}
p(x) = \sum_{i = 1}^M w(m_i)p(x|m_i)
\end{equation}
$w(m_i)$ is the weight of each GMM component.
Here, all points are treated equally, so $w(m_i) = \frac{1}{M}$.

Let $D = D_S + D_C$
and $\Lambda$ be the $D$ dimensional covariance matrix.
Then, each Gaussian is modelled using Eq. \ref{eq:postprob}.

\begin{equation} \label{eq:postprob}
p(x|m_i) = \frac{1}{(2\pi)^\frac{D}{2}} \frac{1}{det(\Lambda)^{D}} e^{ -\frac{1}{2} (x - m_i)' \Lambda^{-1} (x - m_i)}
\end{equation}

Eq. \ref{eq:postprob} will be modified later as 
all components have equal isotropic variance $\sigma^2_S$ (for the 
shape components) and $\sigma^2_C$ (for the color components).
The shape (S) and color (C) covariance matrices for the 
old and new (for $z \in \{o,n\}$ for o:old and n:new which will be defined below) Gaussian distributions are:
$\Lambda_S^z = (\sigma_S^z)^2 {\rm I}_{D_S} $,
$\Lambda_C^z = (\sigma_C^z)^2 {\rm I}_{D_C} $.
From these, we get 
$(\Lambda_S^z)^{-1} = (\sigma_S^z)^{-2} {\rm I}_{D_S} $,
$(\Lambda_C^z)^{-1} = (\sigma_C^z)^{-2} {\rm I}_{D_C} $,
and
$det(\Lambda_S^z) = (\sigma_S^z)^{2 D_S}$,
$det(\Lambda_C^z) = (\sigma_C^z)^{2 D_C}$.

In order to handle noise and outliers, an additional probability distribution
$\frac{1}{N}$, where $N$ is the number of \textit{Anchor} points, is included
which is weighted with a predefined parameter $\alpha$. Thus, Eq. \ref{eq:gmmini:noise} 
is the complete probability of the fit of the \textit{Anchor} points to the
\textit{Moving} points.

\begin{equation} \label{eq:gmmini:noise}
p(x) = \alpha\frac{1}{N} + (1-\alpha)\sum_{i = 1}^M \frac{1}{M} p(x|m_i) = \sum_{i = 1}^{M+1} w(i) p(x|m_i)
\end{equation}
\noindent
where $w(M+1) = P(X|m_{M+1}) =  \frac{\alpha}{N}$ and otherwise $w(i) = \frac{1-\alpha}{M}$.

The GMM is parametrized by a set of parameters 
$(\theta_S, \sigma_S, \sigma_C)$ which specify
the transformation of the \textit{Moving} point set ($\theta_S$), the
standard deviation ($\sigma_S$) of the points' positions, and the standard deviation ($\sigma_C$) of the points' colors.

Expectation-Maximization (EM) is used to register the
\textit{Moving} points to the \textit{Anchor} points.

The function $E$ finds the parameters ($\theta_S, \sigma_S$) that maximize the likelihood, or equivalently, minimize the negative log-likelihood (Eq. \ref{eq:E:loglikelihood}). In this paper we are registering only the shape vectors, but not the color vectors.
We are using shape and color information in the similarity score to make the matching estimation more robust. 
Thus, the set of parameters is ($\theta_S, \sigma_S, \sigma_C$), where $\theta_S$ are the parameters 
that control the position of the \textit{Moving} points.

\begin{equation} \label{eq:E:loglikelihood} 
E(\theta_S, \sigma_S) = -\sum_{n=1}^N log (\sum_{i = 1}^{M+1} w(i) p(a_n|m_i) )
\end{equation}

Following the original formulation of CPD, the probability of correct 
correspondence between model point $m_i$ and anchor point $a_n$ is the posterior probability of the GMM centroid given the anchor point: $p(m_i|a_n)$,
which by Bayes' Rule equals $p(m_i)p(a_n|m_i)/p(a_n)$.
Since the objective of the registration is to find the parameters to 
make model $M$ best fit anchor $A$, the Expectation-Maximization (EM)
algorithm is used. 
Given the value of the ``old'' (superscript `o') position and tolerance parameters,
we use Bayes' theorem to estimate the posterior probability $p^{o}$ (Eq. \ref{eq:Qp5}), known as Expectation or E-step; 
then we find the new parameters  that Maximize (M-step) the probability.
Here, we minimize the negative log-likelihood:

\begin{equation}
\label{eq:Q}
\begin{aligned}
Q(\theta_S, \sigma_S) = -\sum_{n=1}^N\sum_{i=1}^{M+1} w(i) p^{o}(m_i|a_n)log(p^{n}(m_i) p^{n}(a_n|m_i) )
\end{aligned}
\end{equation}

Before we manipulate $Q$, we need some useful terms.
Recalling that $M+1$ refers to the background model:
$p(m_{M+1}) = 1$ and otherwise $p(m_i) = 1$ and $p(x|m_{M+1}) = \frac{1}{N}$.

The multivariate Gaussian distributions that we need 
for the shape term is ($z \in \{o,n\}$ for o:old and n:new):
\begin{equation}
\label{eq:sgauss}
\begin{aligned}
&p^z_S(x_S|m_{i,S}) = \frac{1}{(2\pi)^\frac{D_S}{2}} \frac{1}{(\sigma_S^z)^{D_S}}
e^{
-\frac{1}{2(\sigma_S^z)^2} ||x_S - \tau(m_{i,S},\theta_S^z)||^2
}
\end{aligned}
\end{equation}

\noindent
and for the color term is:
\begin{equation}
\label{eq:cgauss}
\begin{aligned}
& p^z_C(x_C|m_{i,C}) = \frac{1}{(2\pi)^\frac{D_C}{2}} \frac{1}{(\sigma_C^z)^{D_C}}
e^{
-\frac{1}{2(\sigma_C^z)^2} ||x_C - m_{i,C}||^2
}
\end{aligned}
\end{equation}
\noindent
where $\tau(m,\theta_S)$ transforms the position of point $m$ given the
\textit{Moving} point set pose parameters $\theta_S$.
Here, the transformation is only a Euclidean rigid motion.
Note the color matching probability $p^z_C(x_C|m_{i,C})$ uses the
distance between the colors without any transformation. Combining Eq. \ref{eq:sgauss} and \ref{eq:cgauss} we get $P^z(x|m_i) = P^z_S(x_S|m_{i,S}) \cdot P^z_C(x_C|m_{i,C})$.

The first manipulation addresses the background term $M+1$.
We split out the $M+1$ term from the rest and analyze it:
\begin{equation}
\label{eq:Qp}
\begin{aligned}
Q(\theta_S, \sigma_S) = Q'(\theta_S, \sigma_S)-\sum_{n=1}^N w(M+1) p^{o}(m_{M+1}|a_n)log(p^{n}(m_{M+1})p^{n}(a_n|m_{M+1}))
\end{aligned}
\end{equation}
\noindent 
We have:
$w(M+1) = \frac{\alpha}{N}$, $p^{o}(m_{M+1}) = p^{n}(m_{M+1}) = 1$,
$p^{o}(a_n|m_{M+1}) = p^{n}(a_n|m_{M+1}) = \frac{1}{N}$,\\
$p^{o}(m_{M+1}|a_n) = \frac{p^{o}(a_n|m_{M+1}) p^{o}(m_{M+1})}{p^{o}(a_n)} = \frac{1}{N}\frac{1}{p^{o}(a_n)}$.\\
Substituting, this gives:
\begin{equation}
\label{eq:Qp2}
\begin{aligned}
& Q(\theta_S, \sigma_S) = Q'(\theta_S, \sigma_S) + \frac{\alpha log(N)}{N^2} \sum_{n=1}^N \frac{1}{p^{o}(a_n)}
\end{aligned}
\end{equation}

The latter term becomes small as $N$ grows. Further, there are none of the `new' parameters to optimize
in that term. So, we can ignore it and find the parameters $(\theta_S^{n}, \sigma_S^{n})$ that minimizes only $Q'$:
\begin{equation}
\label{eq:Qp3}
\begin{aligned}
Q'(\theta_S^{n}, \sigma_S^{n}) = -\sum_{n=1}^N\sum_{i=1}^{M} w(i) p^{o}(m_i|a_n) log(p^{n}(m_i)p^{n}(a_n|m_i))
\end{aligned}
\end{equation}

Since $log(p^{n}(m_i)p^{n}(a_n|m_i))$ \\ $ = log(p^{n}(m_i)) + log(p^{n}(a_n|m_i))$ and
$log(p^{n}(m_i)) = log(\frac{1}{M})$ has none of the optimization parameters, 
even when multiplied by $p^{o}(m_i|a_n)$, we can
ignore this term. Similarly, $w(i) = \frac{1-\alpha}{M}$ so it is ignored. 
Thus, we need to optimize:
\begin{equation}
\label{eq:Qp4}
\begin{aligned}
Q''(\theta_S^{n}, \sigma_S^{n}) = -\sum_{n=1}^N\sum_{i=1}^{M} p^{o}(m_i|a_n) log(p^{n}(a_n|m_i))
\end{aligned}
\end{equation}

By Bayes's rule:
\begin{equation*} \label{eq:Pold}
\begin{split}
& p^{o}(m_i|a_n) = \\
& \frac
{p^{o}(a_n|m_i) p^{o}(m_i)}
{\sum_{j=1}^{M} w(j) p^{o}(a_n|m_j) p^{o}(m_j) + w(M+1) p^{o}(a_n|m_{M+1}) p^{o}(m_{M+1})}
\end{split}
\end{equation*}

Simplifying, we get:
\begin{equation}
\label{eq:Qp5}
\begin{aligned}
& p^{o}(m_i|a_n) = 
\frac{M}{1-\alpha}
\frac
{p^{o}(a_n|m_i)}
{\sum_{j=1}^{M} p^{o}(a_n|m_j) + \frac{\alpha}{1-\alpha} \frac{M}{N}}
\end{aligned}
\end{equation}
This is evaluated using the `old' parameters and does not change with the
current optimization iteration. The initial $\frac{M}{1-\alpha}$ can also be omitted as
an inessential scaling factor.

Finally, we need to consider $p^{o}(a_n|m_i)$ and $ p^{n}(a_n|m_i)$.
We will analyze both of these together for $z \in \{o,n\}$ (for o:old and n:new).

We assume that point shape and color are independent, and that the
optimization affects only the position of the points, but not the color.
Therefore, $p^{z}(a_n|m_i) = p^z_S(a_n|m_i) p^z_C(a_n|m_i) $,
and these terms were defined above.
For operational reasons, we choose to weight the shape and color components
with $w_S$ and $w_C$. So our formula is: $p^{z}(a_n|m_i) = [p^z_S(a_n|m_i)]^{w_S} [p^z_C(a_n|m_i)]^{w_C}$.

Substituting these derivations into Eq. \ref{eq:Qp4}, we get
(where the first term is evaluated before optimization using Eq. \ref{eq:Qp5}):
\begin{equation*}
\begin{split}
& Q''(\theta_S^{n}, \sigma_S^{n}) =
-\sum_{n=1}^N\sum_{i=1}^{M}
p^{o}(m_i|a_n) \times \\
& log(
[p^{n}_S(a_n|m_i)]^{w_S} [p^{n}_C(a_n|m_i)]^{w_C}
) 
\end{split}
\end{equation*}

Applying the `log' function  and then simplifying:
\begin{equation*}
\begin{split}
& Q''(\theta_S^{n}, \sigma_S^{n}) =
-\sum_{n=1}^N\sum_{i=1}^{M}
p^{o}(m_i|a_n) \times  \\
& \left[
{w_S} log(p^{n}_S(a_n|m_i))
+
{w_C} log(p^{n}_C(a_n|m_i))
\right]
\end{split}
\end{equation*}

And then applying the substitutions from Eq. \ref{eq:sgauss} and Eq. \ref{eq:cgauss},
and then simplifying:
\begin{equation*}
\begin{split}
& Q''(\theta_S^{n}, \sigma_S^{n}) =
-\sum_{n=1}^N\sum_{i=1}^{M}
p^{o}(m_i|a_n) \times \\
& [
{w_S} [ log(\frac{1}{(2\pi)^\frac{D_S}{2}} \frac{1}{(\sigma_S^{n})^{D_S}}) 
- \frac{1}{2(\sigma_S^{n})^2} ||a_{n,S} - \tau(m_{i,S},\theta_S^{n})||^2] \\
& +
{w_C} [ log(\frac{1}{(2\pi)^\frac{D_C}{2}} \frac{1}{(\sigma_C^{n})^{D_C}})
- \frac{1}{2(\sigma_C^{n})^2} ||a_{n,C} - m_{i,C}||^2]
]
\end{split}
\end{equation*}


Simplifying again and
removing terms not involving the optimization parameters, we get
Eq; \ref{eq:Qp6} to be optimised in the EM `M' step
over the parameters: $(\theta_S^{n}, \sigma_S^{n}, \sigma_C^{n})$:

\begin{equation}
\label{eq:Qp6}
\begin{aligned}
& Q''(\theta_S^{n}, \sigma_S^{n}) =
\sum_{n=1}^N\sum_{i=1}^{M}
p^{o}(m_i|a_n) \times \\
& [
 w_S D_S log(\sigma_S^{n})
+ \frac{w_S}{2(\sigma_S^{n})^2} ||a_{n,S} - \tau(m_{i,S},\theta_S^{n})||^2 \\
& 
+ w_C D_C log(\sigma_C^{n})
+ \frac{w_C}{2(\sigma_C^{n})^2} ||a_{n,C} - m_{i,C}||^2 ]
\end{aligned}
\end{equation}

Since the parameters of the color GMM are not optimized in the EM process, the second term in the addition in Eq. \ref{eq:Qp6} becomes a constant and can be removed along with the weighting operators. Thus, we end up with a simpler $Q''$ as next:

\begin{equation}
\label{eq:Qp7}
\begin{aligned}
& Q''(\theta_S^{n}, \sigma_S^{n}) =
\sum_{n=1}^N\sum_{i=1}^{M}
p^{o}(m_i|a_n) \times \\
& [ D_S log(\sigma_S^{n}) + \frac{w_S}{2(\sigma_S^{n})^2} ||a_{n,S} - \tau(m_{i,S},\theta_S^{n})||^2 
\end{aligned}
\end{equation}

Therefore, the color information is involved only in the `old' probability. Recalling that $p^{z}(a_n|m_i) = [p^z_S(a_n|m_i)]^{w_S} [p^z_C(a_n|m_i)]^{w_C}$ for $z \in \{o,n\}$ (for o:old and n:new) and those terms were defined in Eq. \ref{eq:sgauss} and \ref{eq:cgauss}, we substitute Eq. \ref{eq:Qp5} by Eq \ref{eq:Pcomb}:

\begin{equation}\label{eq:Pcomb}
\begin{aligned} 
 P^{o}(m_i|a_n) = \frac{[p^z_S(a_n|m_i)]^{w_S} [p^z_C(a_n|m_i)]^{w_C}}{(\sum_{j=1}^M p^z_S(a_n|m_j)]^{w_S} (\sum_{j=1}^M  p^z_C(a_n|m_j)^{w_C} + o_C + o_L}
\end{aligned}
\end{equation}

Outlier biases $o_C$ are calculated with Eq. \ref{eq:oC} and $o_L$ with the outlier probability $\frac{\alpha}{1-\alpha} \frac{M}{N}$

\begin{equation} \label{eq:oC} 
o_C =  \frac{M}{\sigma_C \sqrt{2 \pi}} \cdot exp^{-\frac{1}{M}\frac{ \left \|\sum^M_m P^o_C(a_C|m^C_i)\right \| ^2}{2\sigma_C^2}}
\end{equation}

\msc{The general process of registration is summarized in the next pseudo-code Algorithm \ref{alg:ccpd}. Since we focus on modifying the matching probability ($P^o$), the general procedure is similar to the original CPD, but with modifying step E:}

\msc{\begin{algorithm}[H] \label{alg:ccpd}
 \KwData{M and A pointsets, color M and color A information}
 
 Initialization: $W= o,\sigma^2=\frac{1}{DNM}\sum_{m,n=1}^{M,N}||x_n-y_m||^2$ \;
 Construct $G$: $g_{ij}=exp^{-\frac{1}{2\beta^2}||y_i-y_j||^2}$\;
 
 Expectation-Maximization\\
 \While{not converged}{
  E-step: Compute $P^o$, (contribution)\\
  
\begin{itemize}
  \item $P^{o}(m_i|a_n)=$
\end{itemize}
  $= \frac{[p^z_S(a_n|m_i)]^{w_S} [p^z_C(a_n|m_i)]^{w_C}}{(\sum_{j=1}^M p^z_S(a_n|m_j)]^{w_S} (\sum_{j=1}^M  p^z_C(a_n|m_j)^{w_C} + o_C + o_L}$ (see Eq. \ref{eq:Pcomb})\;
  
  M-step: \\
    
 \begin{itemize}
\item Solve $(G + \lambda\sigma^2 d(P1)^{-1})W=d(P1)^{-1}PX-Y$ (see \cite{Myronenko2010})\;
 \end{itemize} 
 }
 The result of alignment: $T=\tau(Y,W)=Y+GW$\;
 \caption{Pseudo-code of the proposed Color Coherent Point Drift}
\end{algorithm}
}

\section{Experiments}
\label{sec:ccpd:exp}
A set of tests have been carried out to evaluate the performance of the proposed CCPD \footnote{The code is available  at tech4d.dtic.ua.es} compared to the original version. First, the dataset of the original CPD (Subsection \ref{sec:ccpd:exp:synthetic}), the fish and the face, has been used (Figure \ref{fig:ccpd:fish:model}). The implementation of the code has been done in Matlab, using part of the toolbox provided by Myronenko \footnote{www.bme.ogi.edu/$\sim$myron/matlab/cpd}. Color information has been added to the original data. The distribution of colors on the shape has been done in this way to distinguish its different parts, i.e. a region with same color corresponds to a specific part of the shape (e.g. mouth in the face, or tail in the fish). It is important for the non-rigid registration with color because it gives meaning to the relationship between color and shape.

The second test (Subsection \ref{sec:ccpd:exp:blender}) presents two synthetic datasets with realistic color and shape (Figure \ref{fig:face:models}). A face and a flower are used, which have been deformed using Blender and acquired using a plugin called Blensor \cite{Blensor2011}. This plugin emulates different sensors, including the general purpose RGB-D sensor Kinect.

Finally, a real data evaluation using data provided by a Primesense Carmine RGB-D sensor is done in Subsection \ref{exp:real} to confirm that the algorithm is able to handle real data acquired from a general purpose RGB-D sensor (Figure \ref{fig:ccpd:real:data}).

\mscc{In this section we will use $X$ to refer to the \textit{Anchor} set and $Y$ to refer to the \textit{Moving} set.}

The experiments evaluate different aspects:
\begin{itemize}
\item Outliers: points which are in \textit{Anchor} $X$ but do not have real matches in \textit{Moving} $Y$. 

\item Missing data: the opposite of outliers. Points which are in \textit{Moving} $Y$ but do not have real matches in \textit{Anchor} $X$. This situation is not taken into account in the original CPD algorithm.

\item Large or non-linear deformation: deformations which involve a large displacement that may not be solved with traditional algorithms. Non-linear deformation could be seen as an abrupt change in the relative direction of the deformation.  

\end{itemize}

The experiments used Windows 7, an Intel i5 processor and 8 GB of RAM. The code was implemented in Matlab vR2013b. 

\subsection{Synthetic data experimentation}\label{sec:ccpd:exp:synthetic}
The tests consider four issues: outliers, missing data, color distribution changes and large deformations. First, points from $Y$ are removed. With this test the missing data handling is compared with the original CPD algorithm. Next, we remove data from \textit{Anchor} $X$ representing extra points, a situation which is not possible to parametrize in the original CPD (points in $Y$ do not have a real correspondence in $X$). In this case, CCPD uses the color information to improve the probability evaluation to avoid wrong matches. Another test evaluates a different displacement in the color with respect to the shape, which evaluates situations where the color distribution in $X$ and $Y$ are different. An example of this could be moving the eyebrows up and down, where the shape in 3D largely remains the same, but the color changes its position. Lastly, large deformations are evaluated to show how the color facilitates the registration when the transformation is complex or semi-coherent. It is important to highlight that the parameters have been adjusted individually to result in the best alignment for both the CPD and CCPD algorithms.

The main difference between the original CPD and the proposed CCPD method comes when the \textit{Moving} has missing data, which cannot be modelled as outliers in the CPD. As the color is a distinctive feature, the proposal is able to evaluate the correspondences properly and then provide better results.

\subsubsection{2D fish experimentation}\label{sec:ccpd:exp:fish}
\begin{figure}

  \centering
\includegraphics[width = 0.40\textwidth]{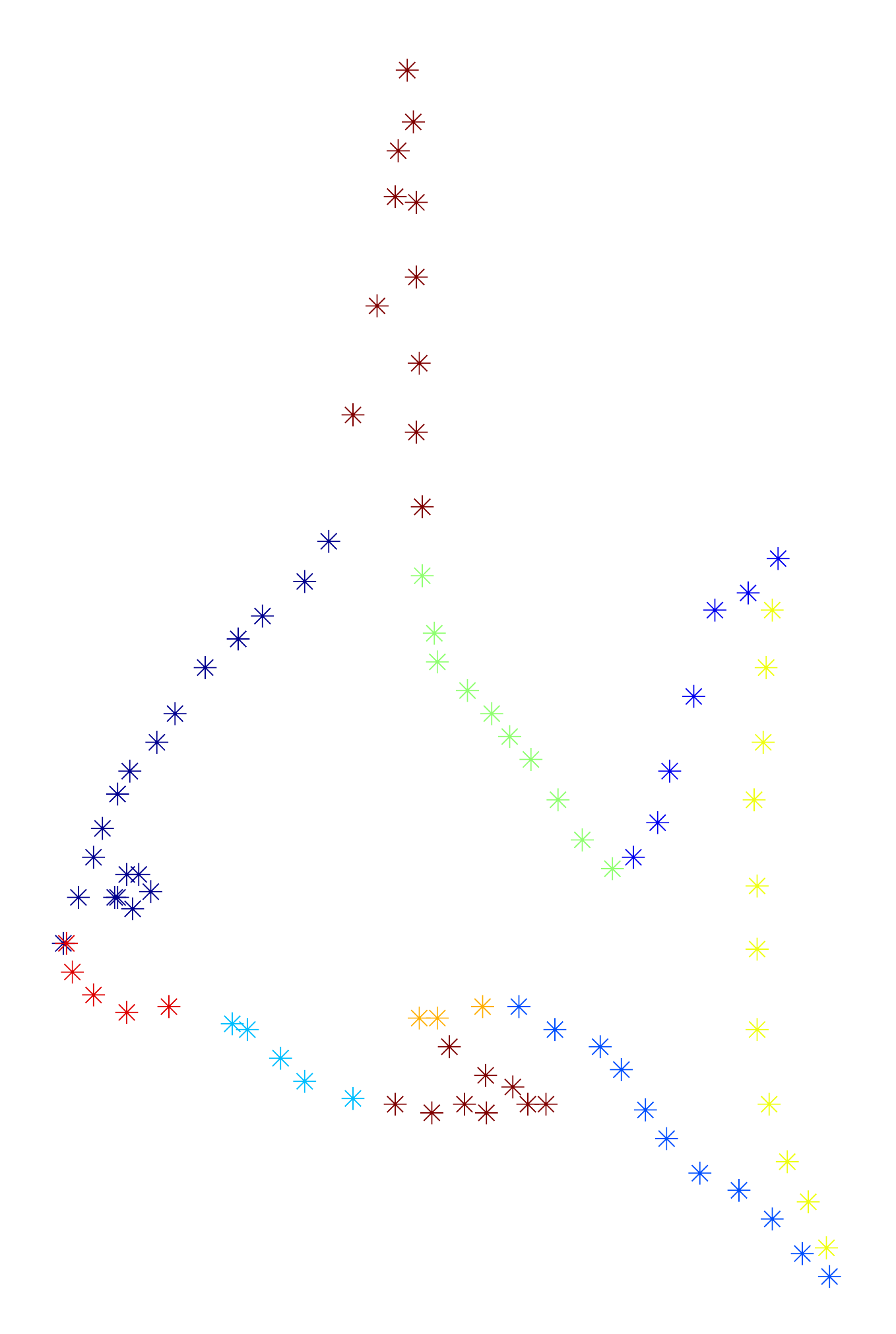} 
\includegraphics[width = 0.50\textwidth]{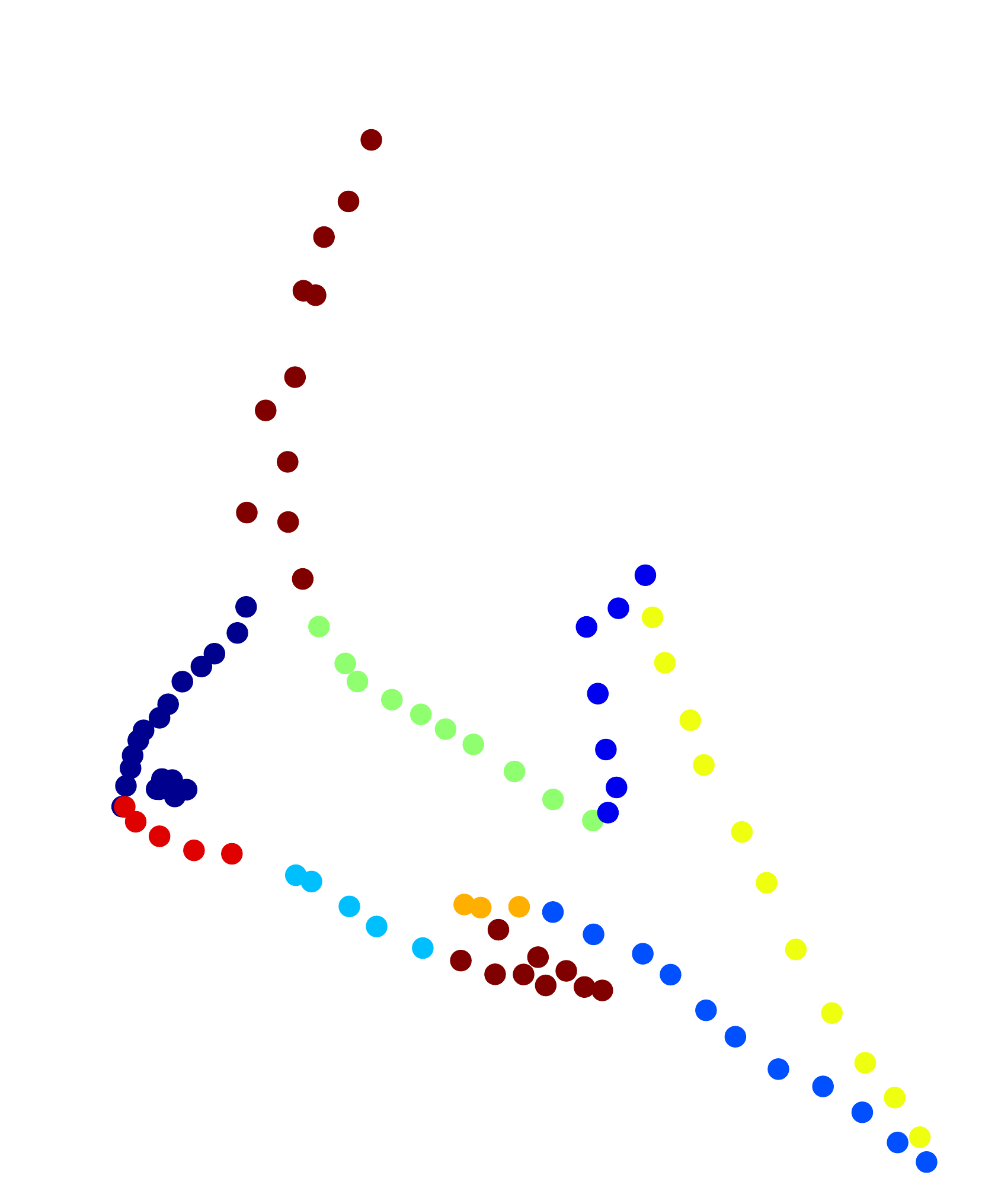} 
\caption{\textit{Anchor} $X$ (left) and \textit{Moving} $Y$ (right) fishes based on the original work of CPD \cite{Myronenko2010} including color information.}
\label{fig:ccpd:fish:model}
\end{figure}

The 2D tests use different \textit{Anchor} $X$ and \textit{Moving} $Y$ fishes based on two initial shapes (dataset from the original work of CPD \cite{Myronenko2010}). Nine colors using the H component of HSV are used to distinguish the different parts of the fish (see Figure \ref{fig:ccpd:fish:model}).

\begin{figure}
  \centering

\begin{tabular}{p{0.01\textwidth} p{0.2\textwidth} p{0.2\textwidth} p{0.2\textwidth} p{0.2\textwidth}}
& $X$ & $Y$ & CCPD & CPD \\
\hline
\end{tabular} \\

\includegraphics[width = 0.2\textwidth]{./cpd_exp_fish_model.pdf}
\includegraphics[width = 0.3\textwidth]{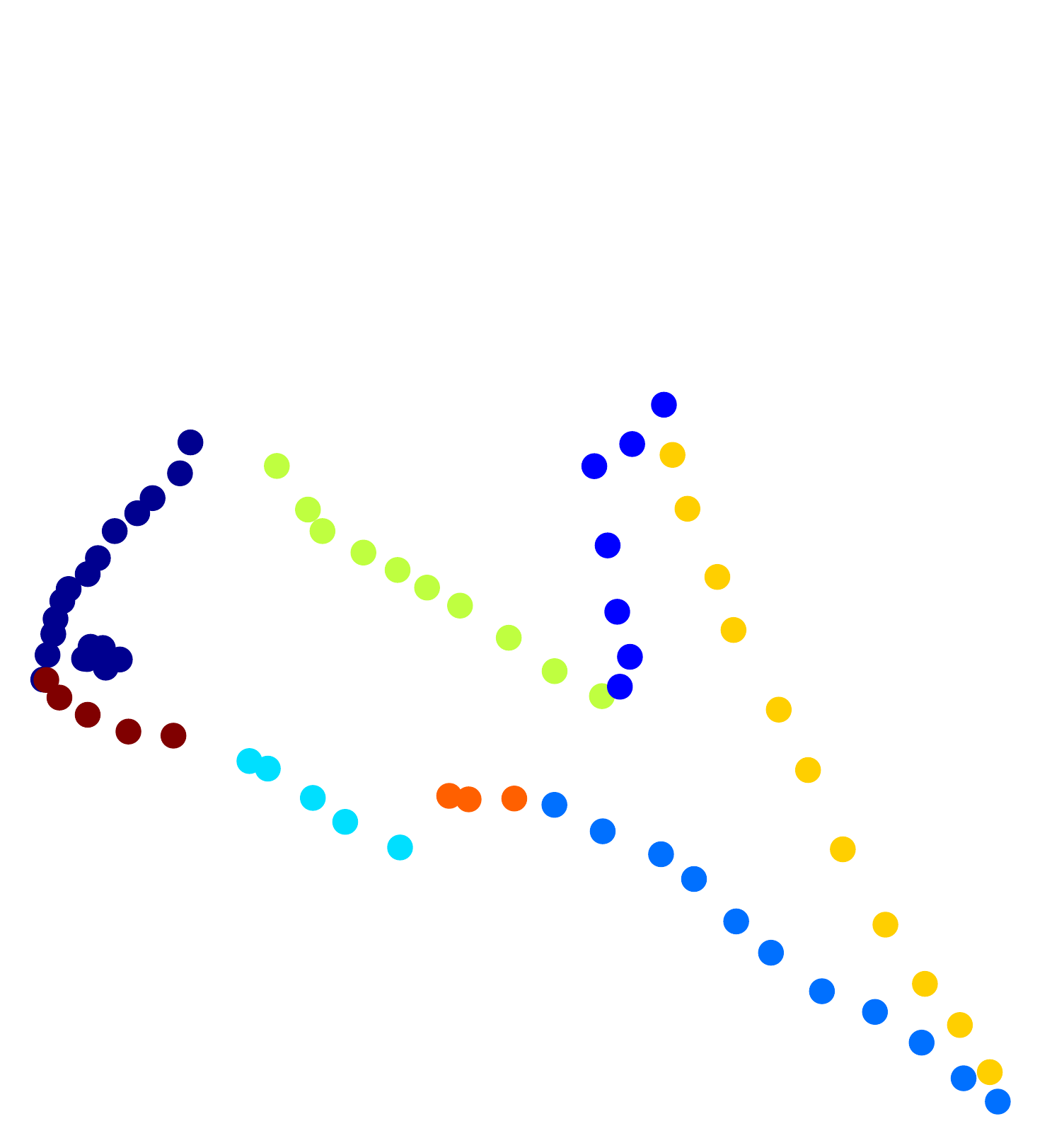}
\includegraphics[width = 0.2\textwidth]{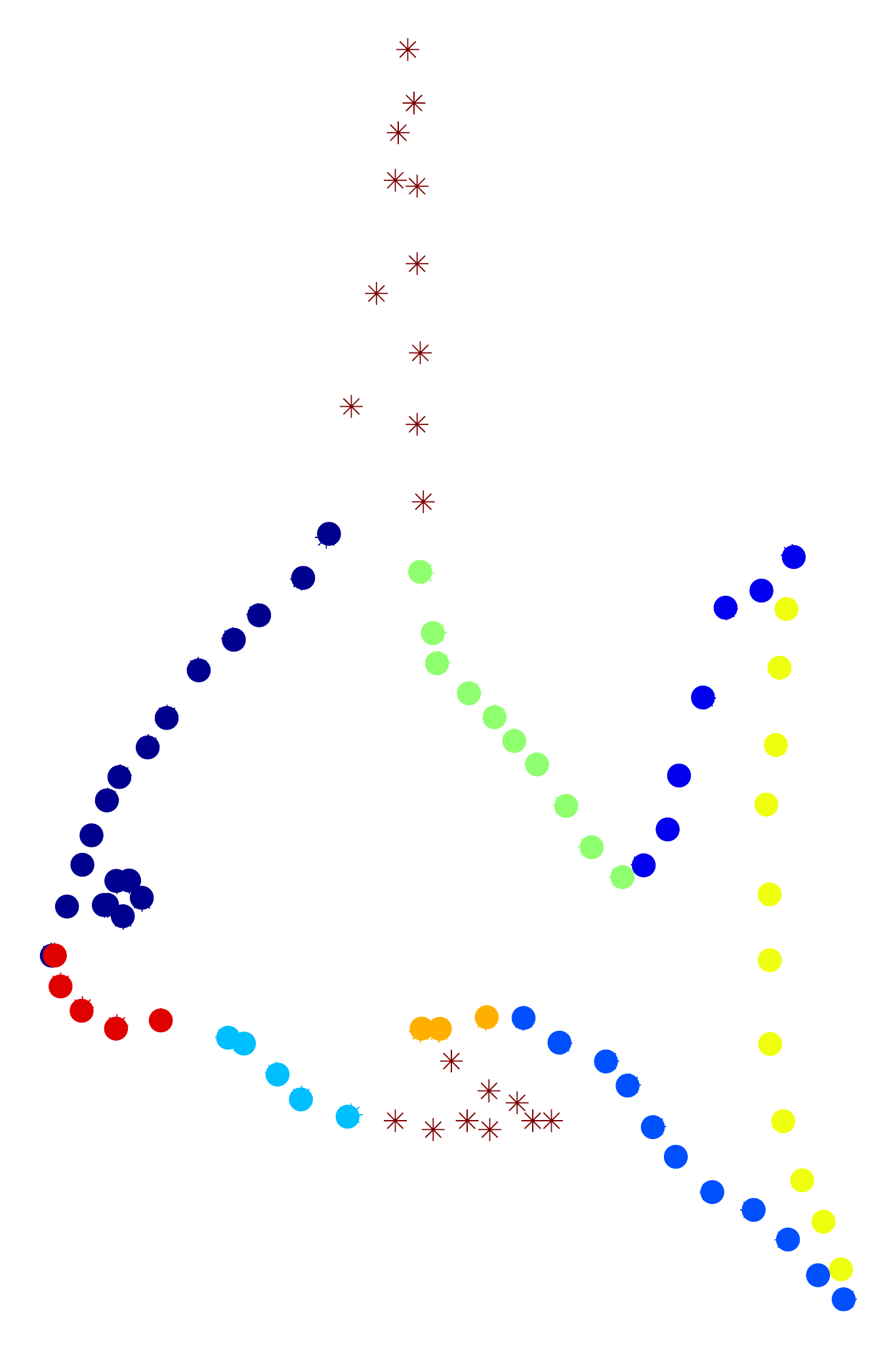} 
\includegraphics[width = 0.2\textwidth]{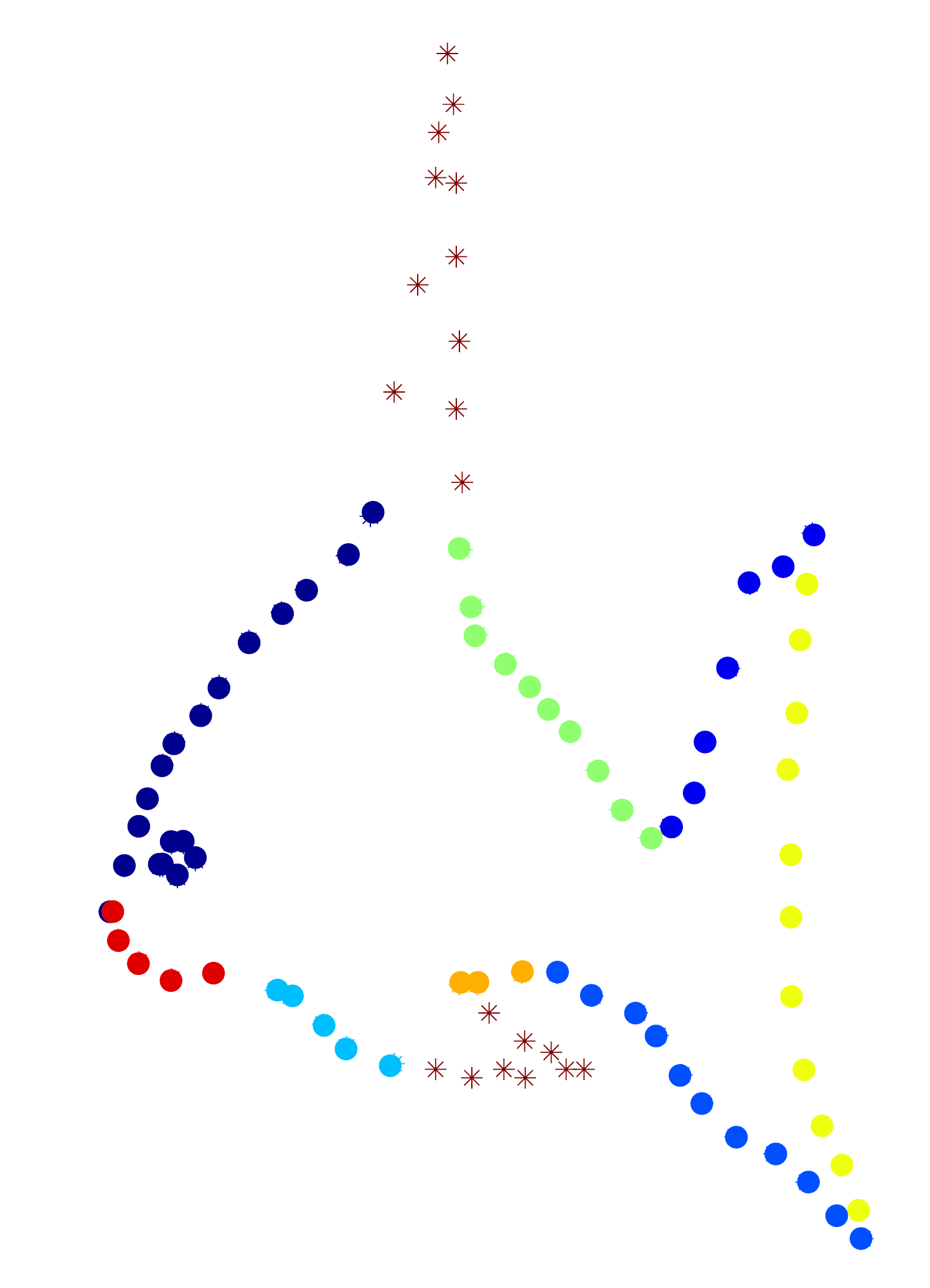} 
    \\


\includegraphics[width = 0.2\textwidth]{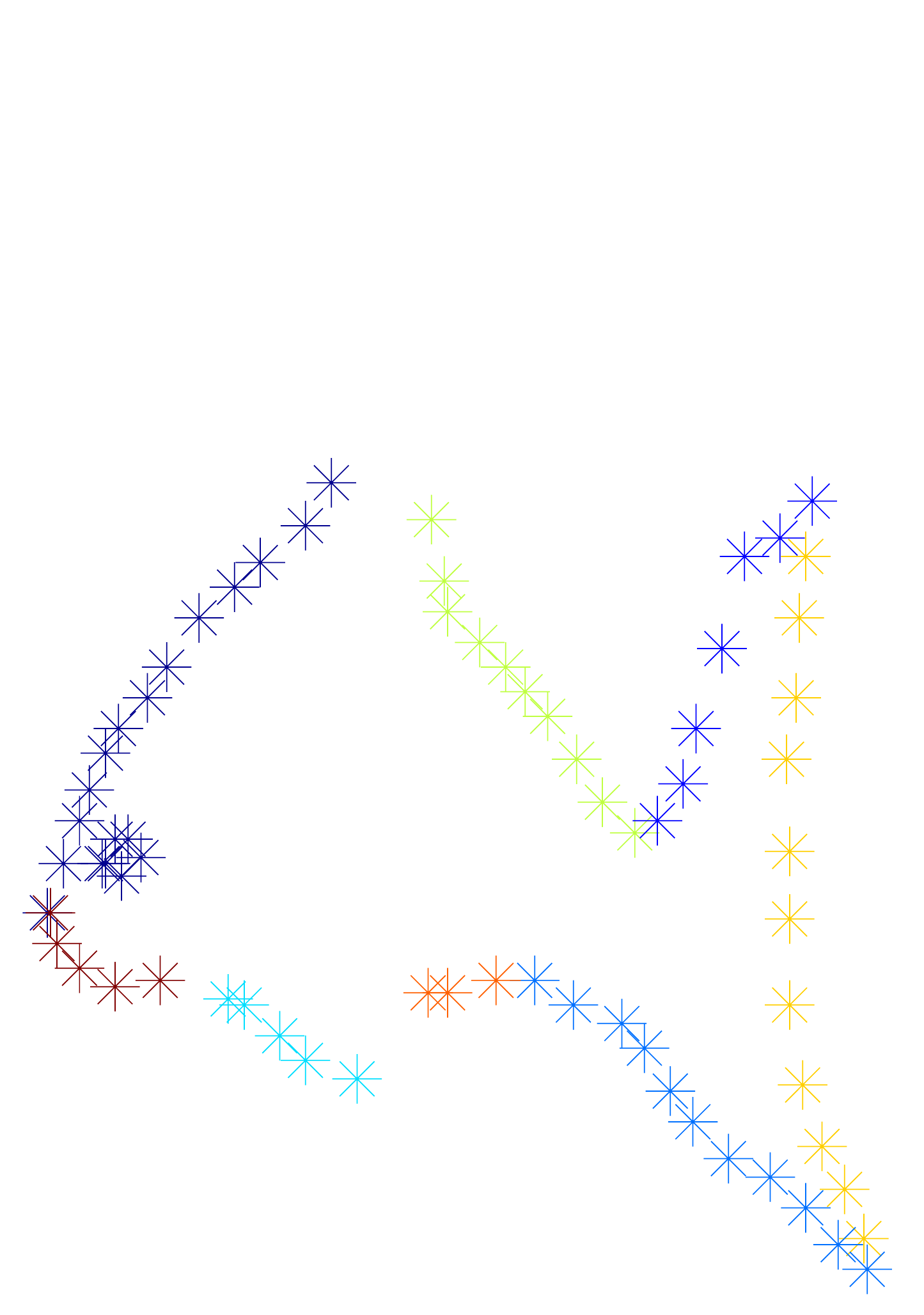}
\includegraphics[width = 0.3\textwidth]{./cpd_exp_fish_data.pdf} 
\includegraphics[width = 0.2\textwidth]{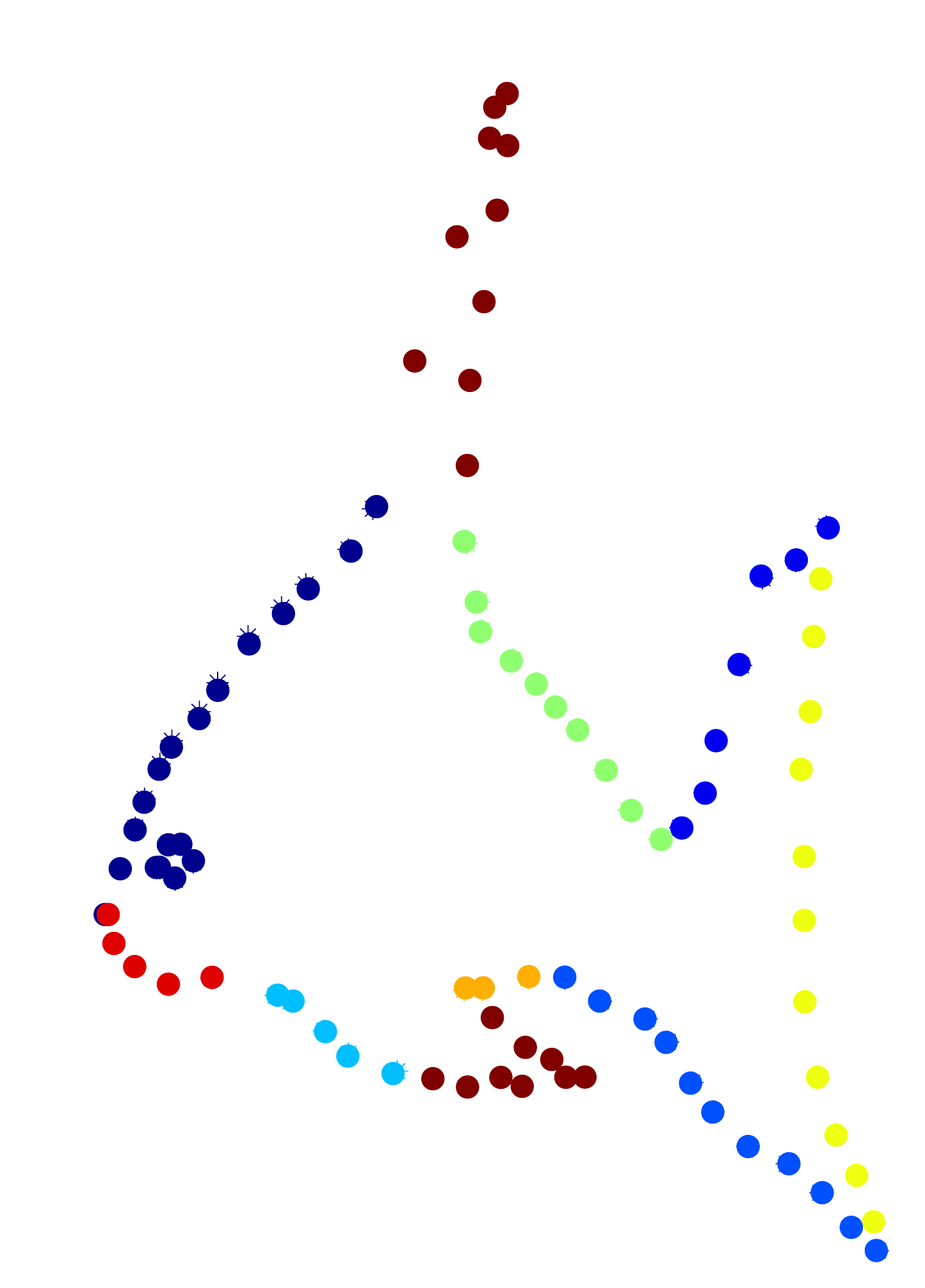} 
\includegraphics[width = 0.2\textwidth]{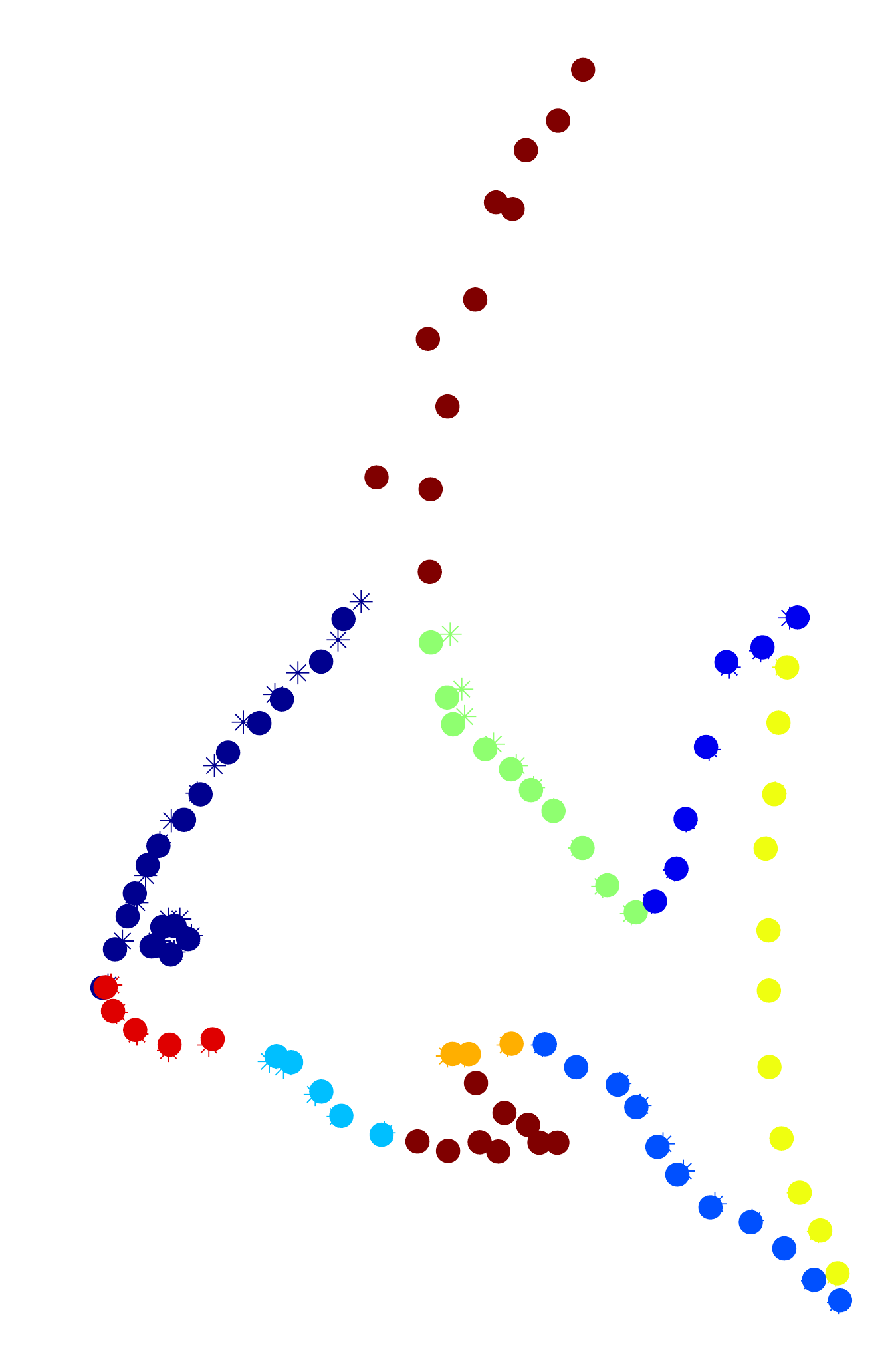}
    \\

\includegraphics[width = 0.2\textwidth]{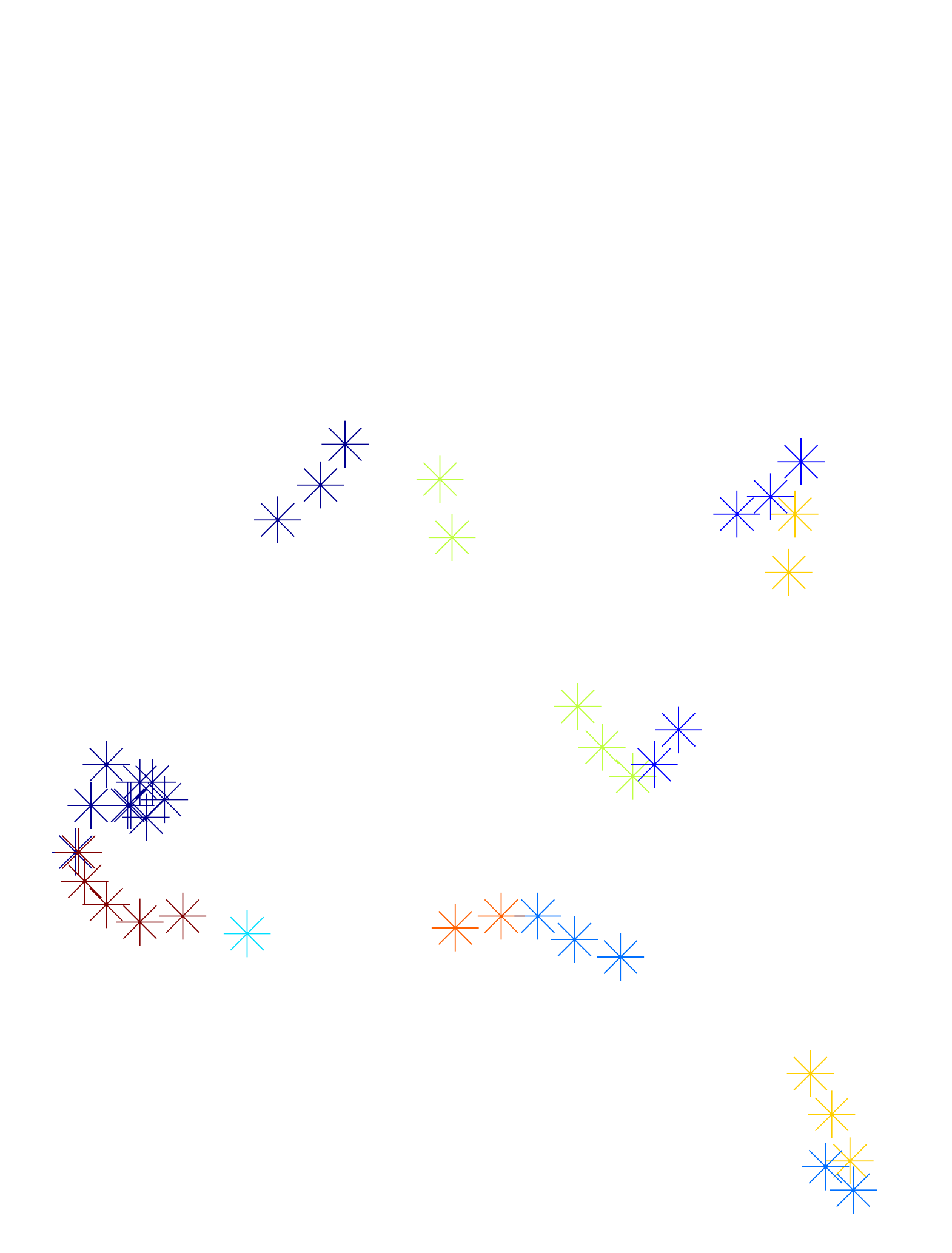} 
\includegraphics[width = 0.3\textwidth]{./cpd_exp_fish_data.pdf}  
\includegraphics[width = 0.2\textwidth]{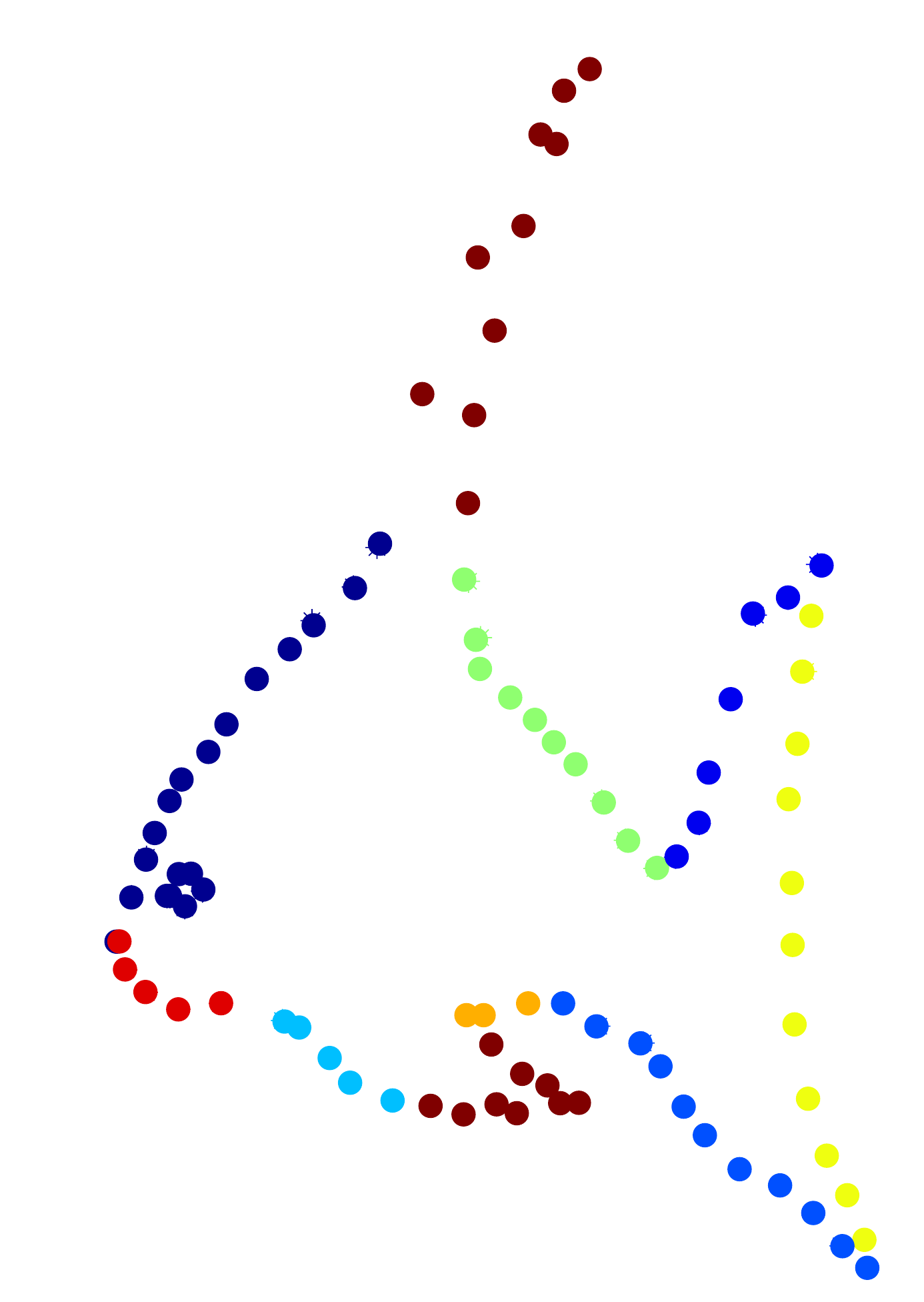}      
\includegraphics[width = 0.2\textwidth]{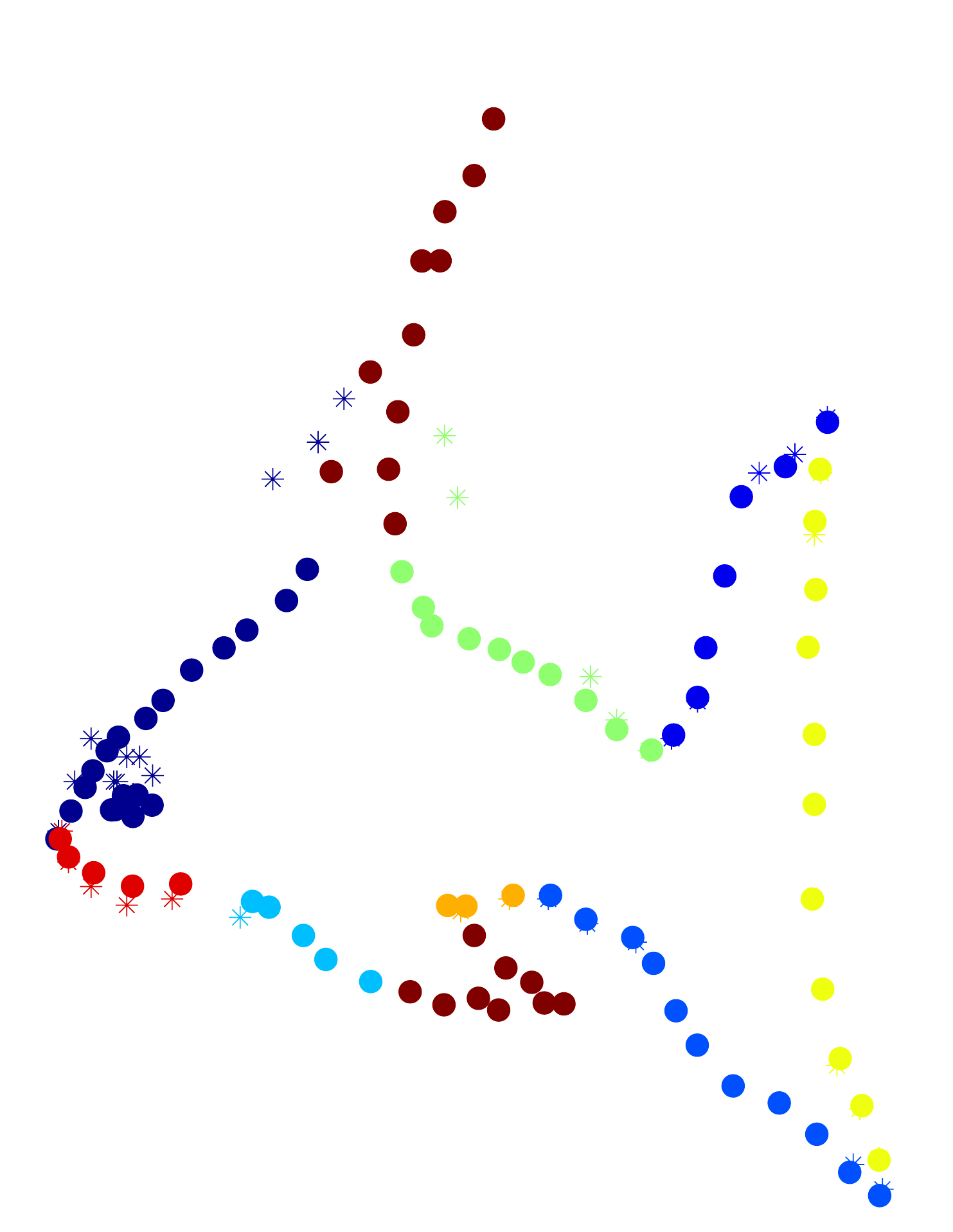} 
    \\

\includegraphics[width = 0.2\textwidth]{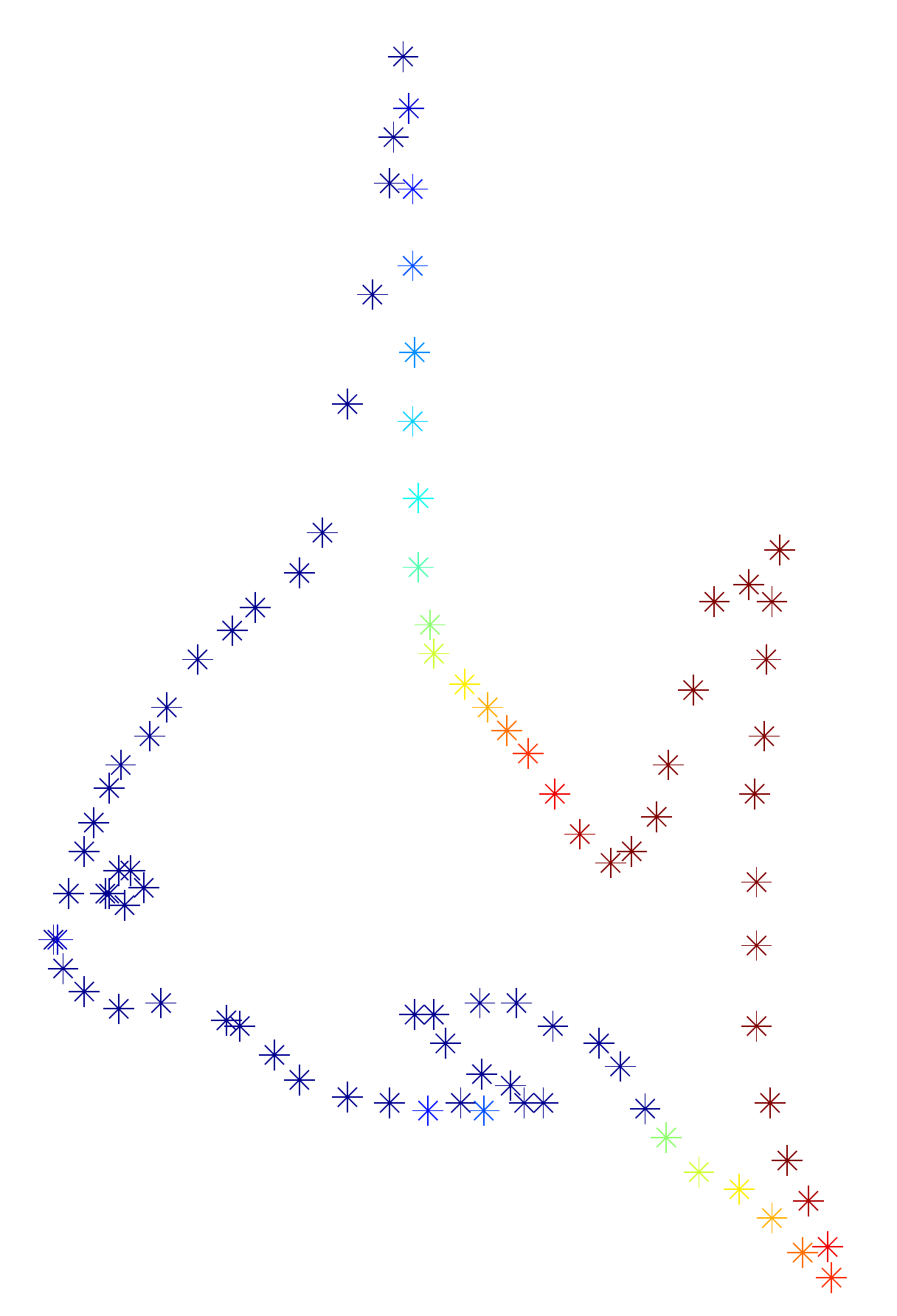} 
\includegraphics[width = 0.3\textwidth]{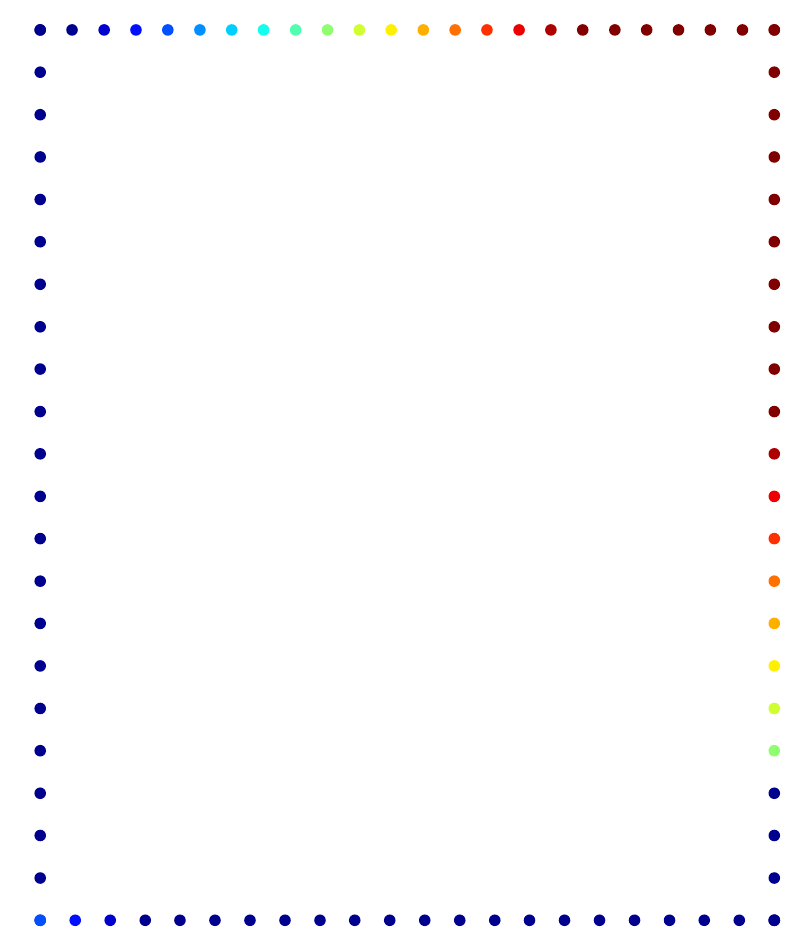} 
\includegraphics[width = 0.2\textwidth]{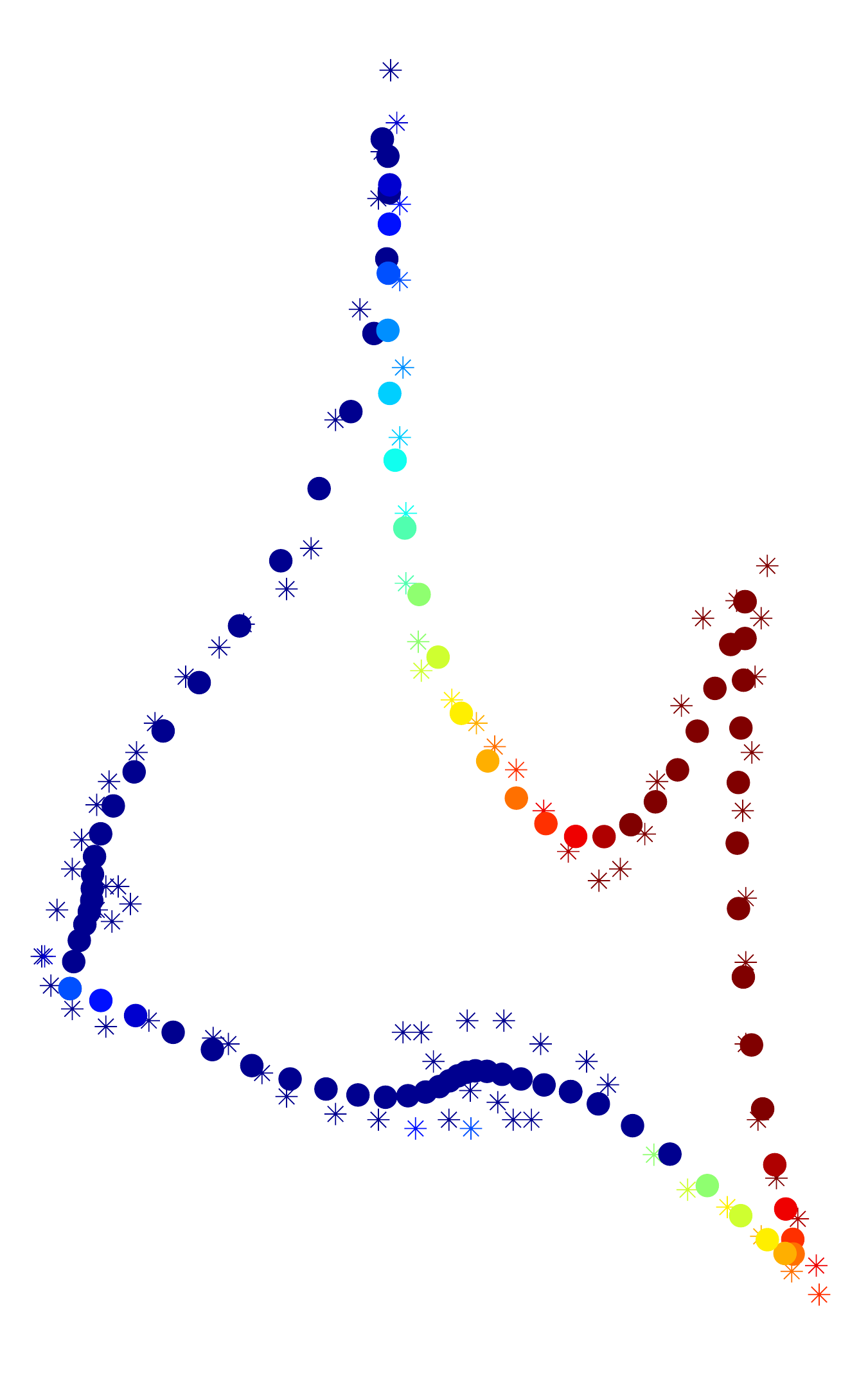} 
\includegraphics[width = 0.2\textwidth]{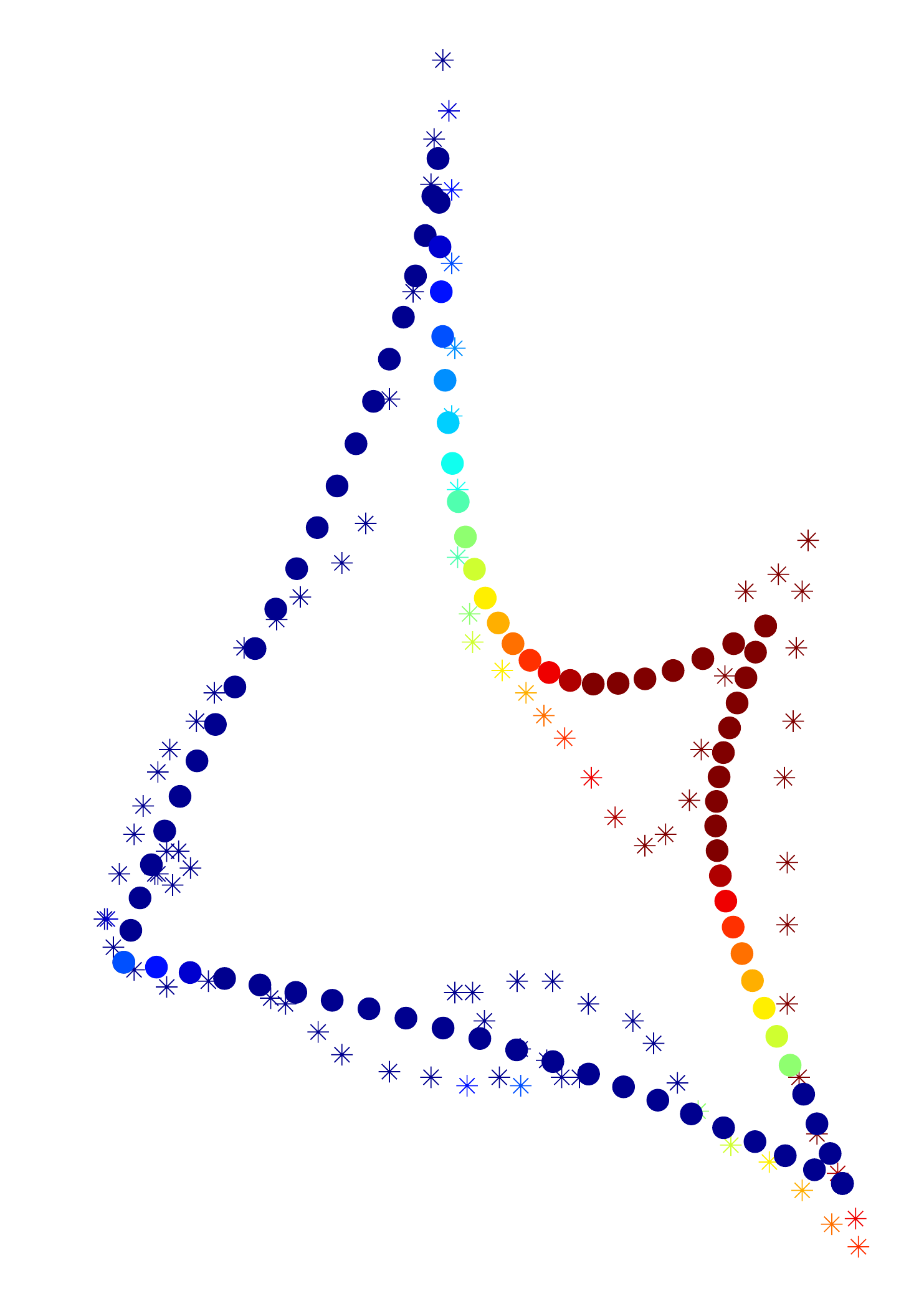} 
    \\

	\caption[Tests 1 to 4 of fish shape]{\mscc{Tests 1 to 4 of fish shape from top to bottom respectively. The columns represent from left to right the \textit{Anchor} $X$, the \textit{Moving} $Y$, the CCPD registration result and the CPD result.}}
  \label{fig:fishimages}
\end{figure}

Table \ref{tab:fish} presents the RMS error of the registration taking into account euclidean distances of real correspondences in location space. Figure \ref{fig:fishimages} shows the visual result of the tests. In general, the registration achieves better alignment (minimize the error distance) in the CCPD results. Test 1 evaluates the effect of outliers by removing from Y the top and bottom tip of the fish. In this case, the proposal returns a slightly better registration because the color feature provides a more robust matching estimation and hence registration. Test 2 and Test 3 correspond to missing data testing where points in \textit{Anchor} $X$ are removed, while $Y$ remains complete. The total amount of points is 91. For Test 2, 20 points are removed ($20/91 = 21.9\%$ of outliers) and for Test 3, 53 points are removed ($53/91 = 58.2\%$ of outliers). The results demonstrate the improved performance of CCPD in the alignment against CPD. Concretely, in Test 2, CCPD achieves 0.747E-02 RMS error in registration being 4.82 times lower RMS  than CPD, while in Test 3, CCPD achieves 0.624E-02 RMS error being 23.1 times lower than the original method. CCPD is more robust against outliers in the \textit{Moving} $Y$ (or missing data from the Data point of view).

A large deformation test has been considered by registering a square to the \textit{Anchor} fish in Test 4, where Matlab \textit{jet} colormap is used. This color map provides colors in RGB = [0,0,0.562] to [1,1,0], which in H component used here are H = [0 0.0625 0.1250 0.1875 0.2500 0.3125 0.3750 0.4375 0.5000 0.5625 0.6250 0.6875 0.7500 0.8125 0.8750 0.9375 1.0000]. The RMS error is 26.62E-02 in the CCPD method and 51.559E-02 in the original CPD, a 93.69\% of improvement of CCPD against CPD. Furthermore, CPD on the low tip of the back tail (Figure \ref{fig:fishimages} fourth-row right-image) misaligns the colors as it does not have this information, which also demonstrates the improvement in registration accuracy of the proposed color feature consideration in the registration process. 

\begin{table}[h]
  \centering
  \caption{RMS registration error of fish shape tests.}
    \begin{tabular}{lrr}
\hline
          & CCPD & CPD \\
\hline
    Test 1 & 0.52064E-02 & 0.53293E-02 \\
    
    Test 2 & 0.7468E-02 & 3.5967E-02 \\
    
    Test 3 & 0.6239E-02 & 14.406E-02 \\
    
    Test 4 & 26.622E-02 & 51.559E-02 \\

\hline
    \end{tabular}%
	
  \label{tab:fish}%
\end{table}%

\begin{figure}

  \centering%

\begin{tabular}{p{0.01\textwidth} p{0.2\textwidth} p{0.2\textwidth} p{0.2\textwidth} p{0.2\textwidth}}
& X & Y & CCPD & CPD \\
\hline
\end{tabular} \\

\includegraphics[width = 0.2\textwidth]{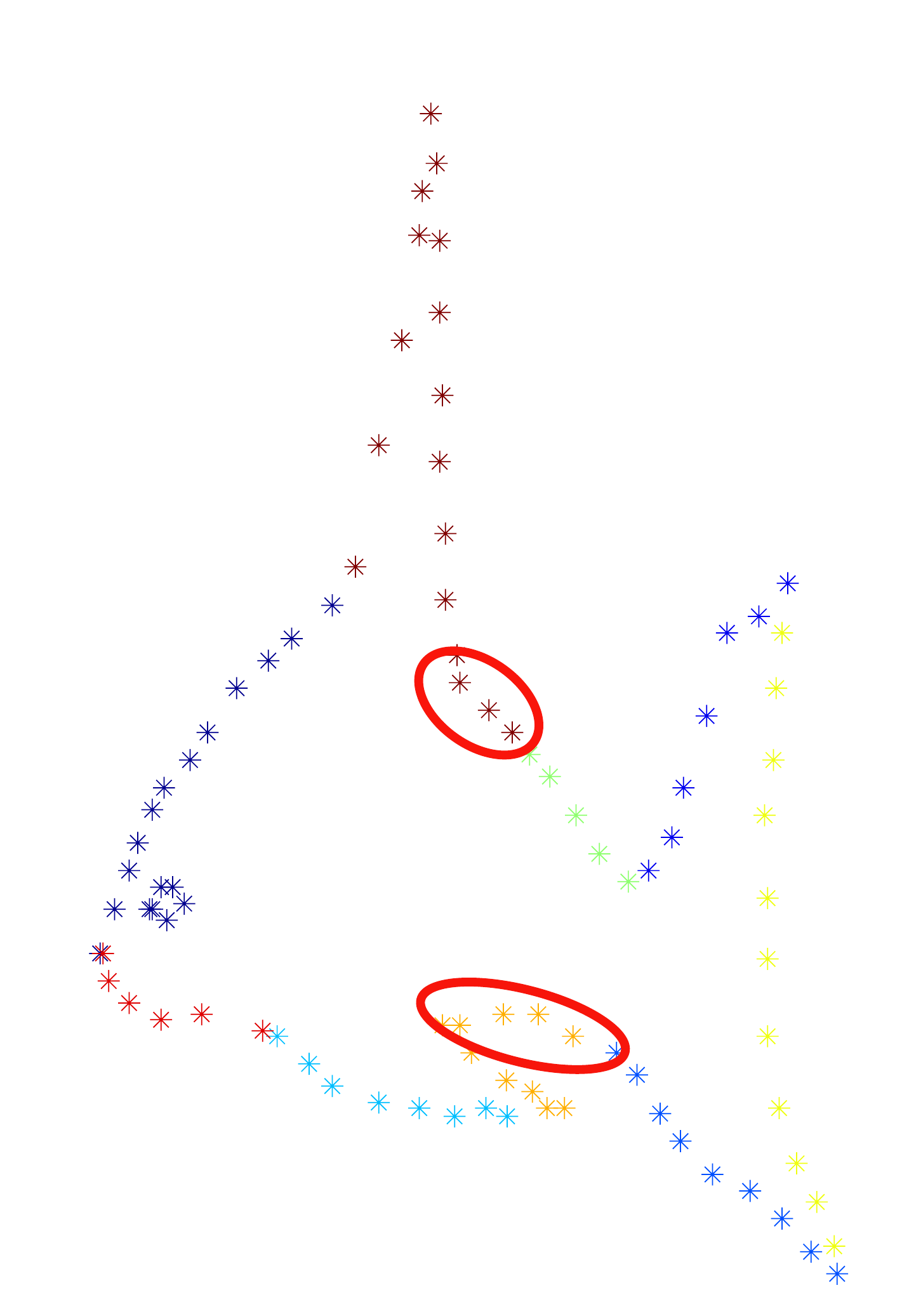}
\includegraphics[width = 0.2\textwidth]{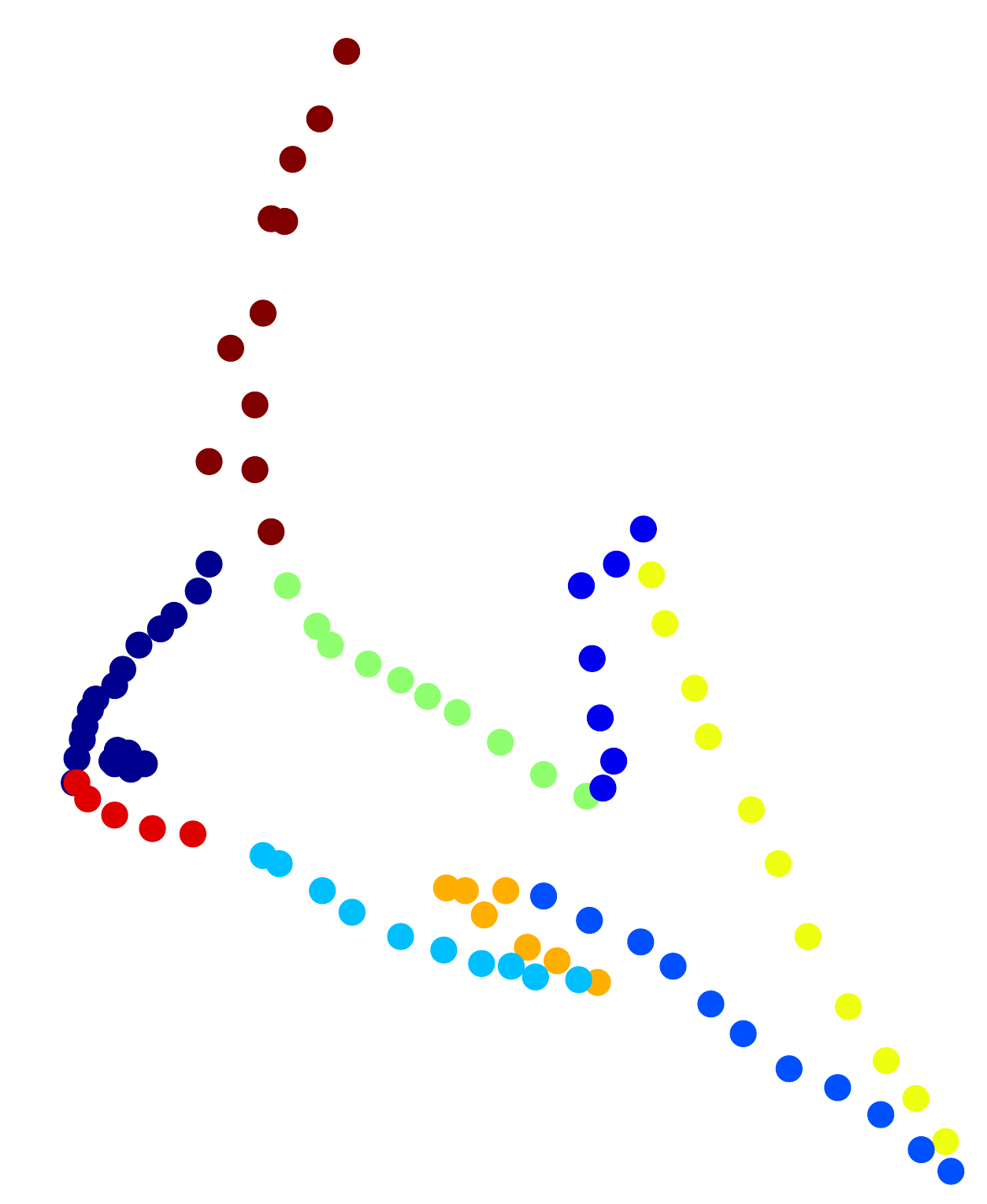} 
\includegraphics[width = 0.2\textwidth]{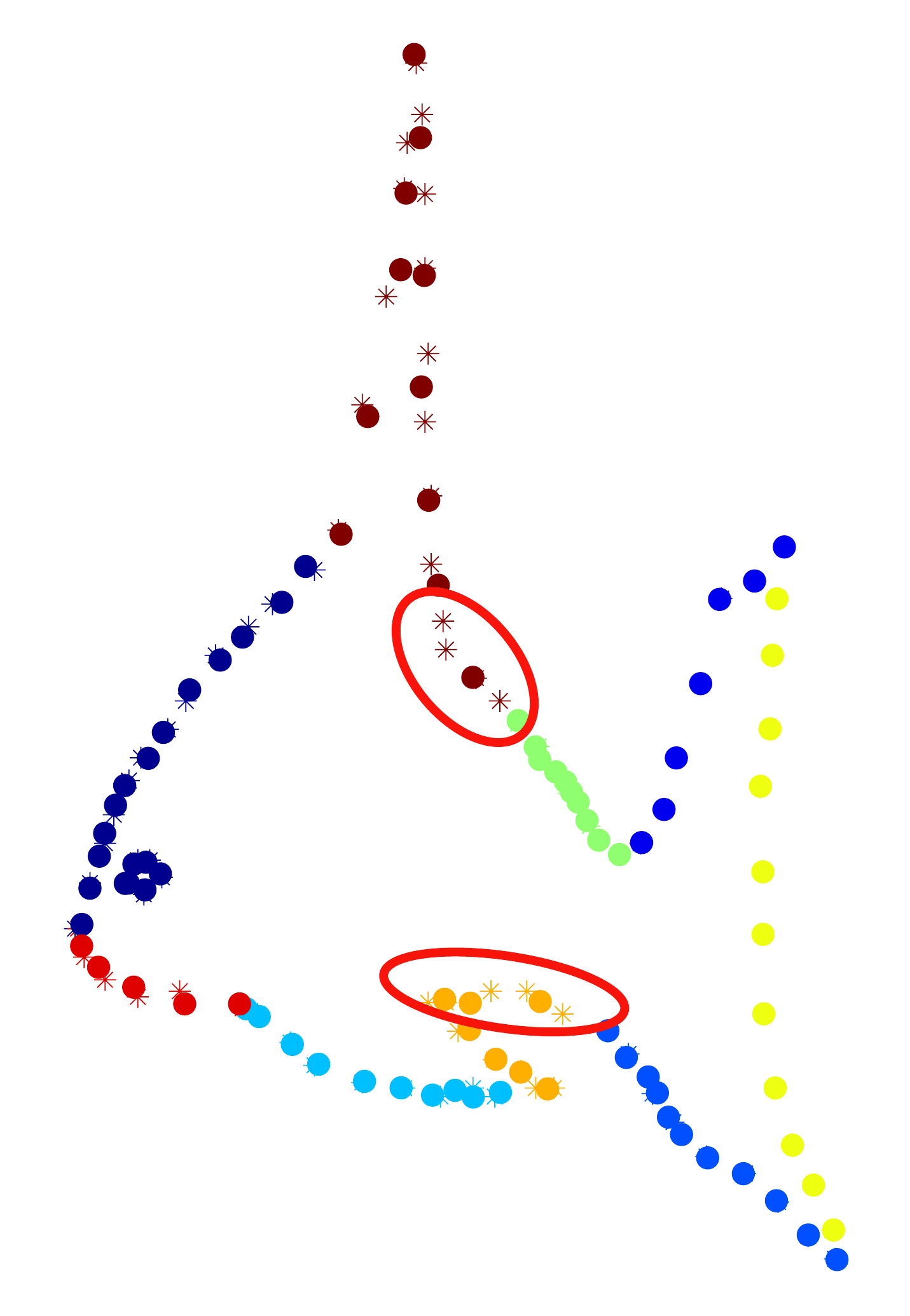}
\includegraphics[width = 0.2\textwidth]{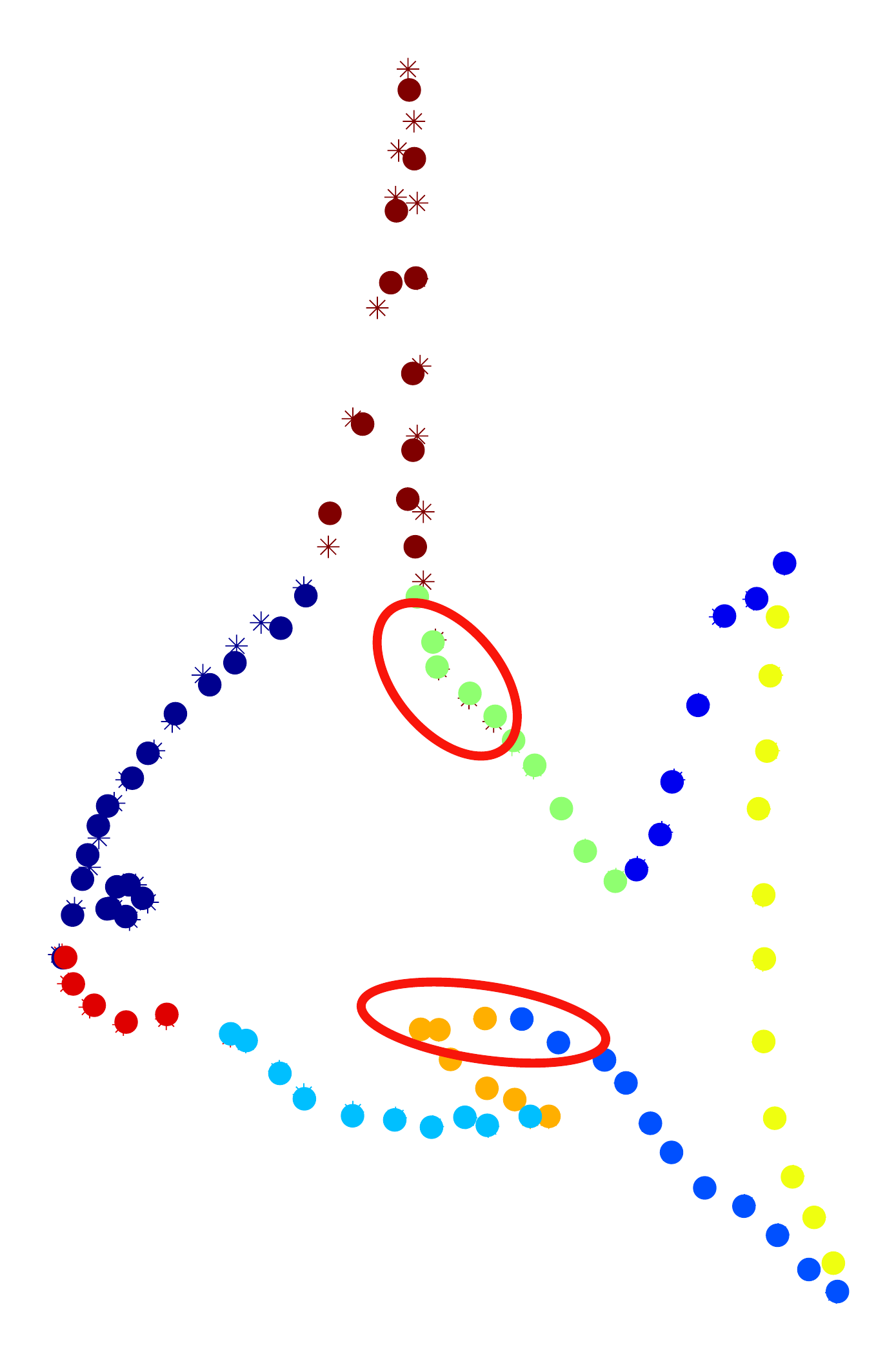}

 \caption[Registration result for different color in fish]{Registration result for different color distribution in \mscc{ \textit{Anchor} $X$  and \textit{Moving} $Y$ sets (see Fig. \ref{fig:fishimages}). The third and fourth columns are the results for the CCPD and the CPD algorithms. The red circles highlight the parts where the color distribution changes.}}
 \label{fig:fishdisplacement}
\end{figure}

The next test evaluates changes in the color distribution. In this situation both shapes have the same points as the original, but the colors are slightly different. The result is visually evaluated in Figure \ref{fig:fishdisplacement}. \msc{The regions where the colors do not coincide are marked with a red circle to simplify the visualization. At the lower part of the upper tip, $X$ has a larger region of brown towards the back while $Y$ is green from the end of the tip. CCPD registers adequately this part. Similarly, the lower tip has larger part of orange on the $X$ than in $Y$, and again the proposed method achieves better results.}

\subsubsection{3D face experiments}\label{sec:ccpd:exp:face}
\begin{figure}
  \centering%
\includegraphics[width = 0.3\textwidth]{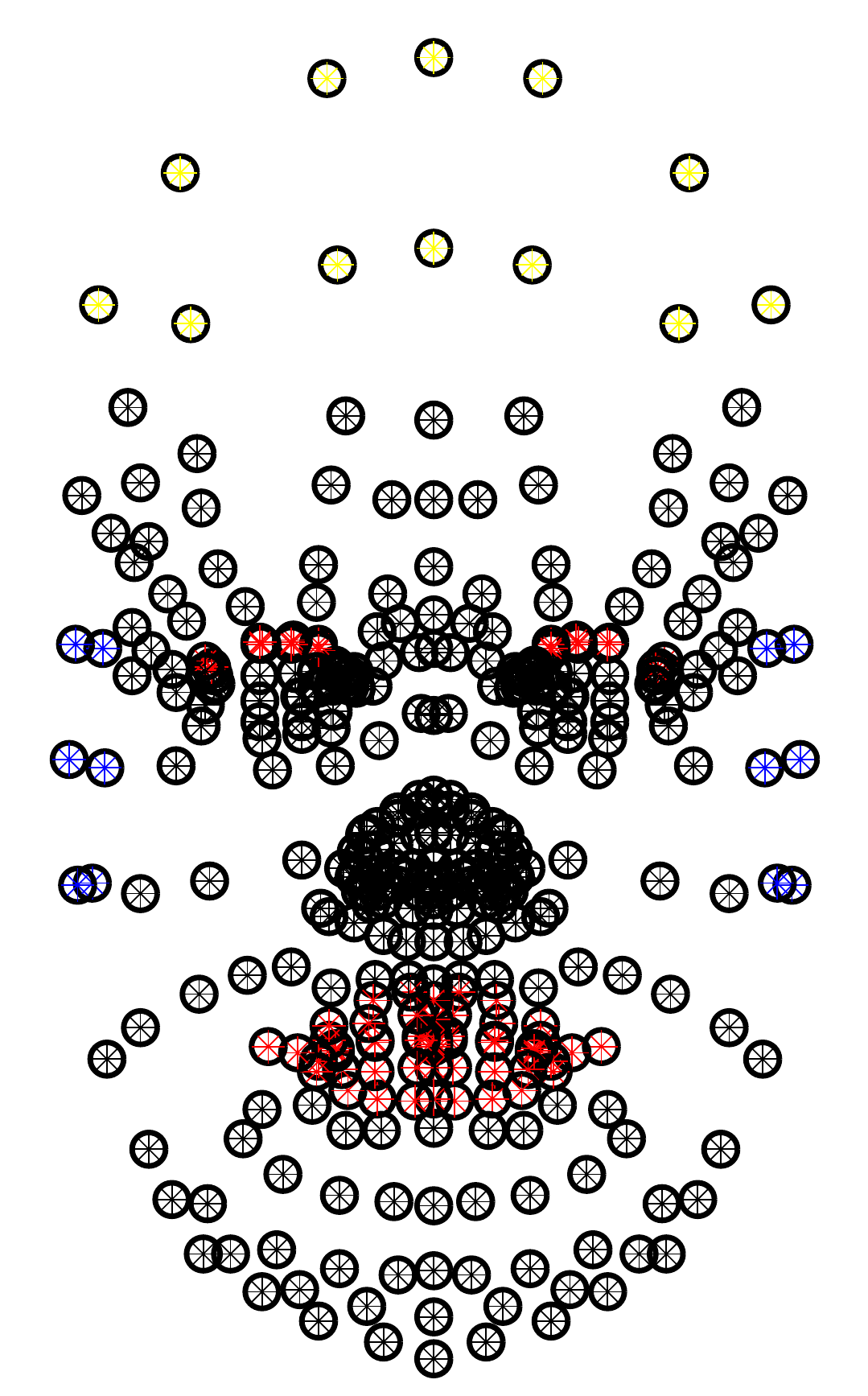} \qquad
\includegraphics[width = 0.3\textwidth]{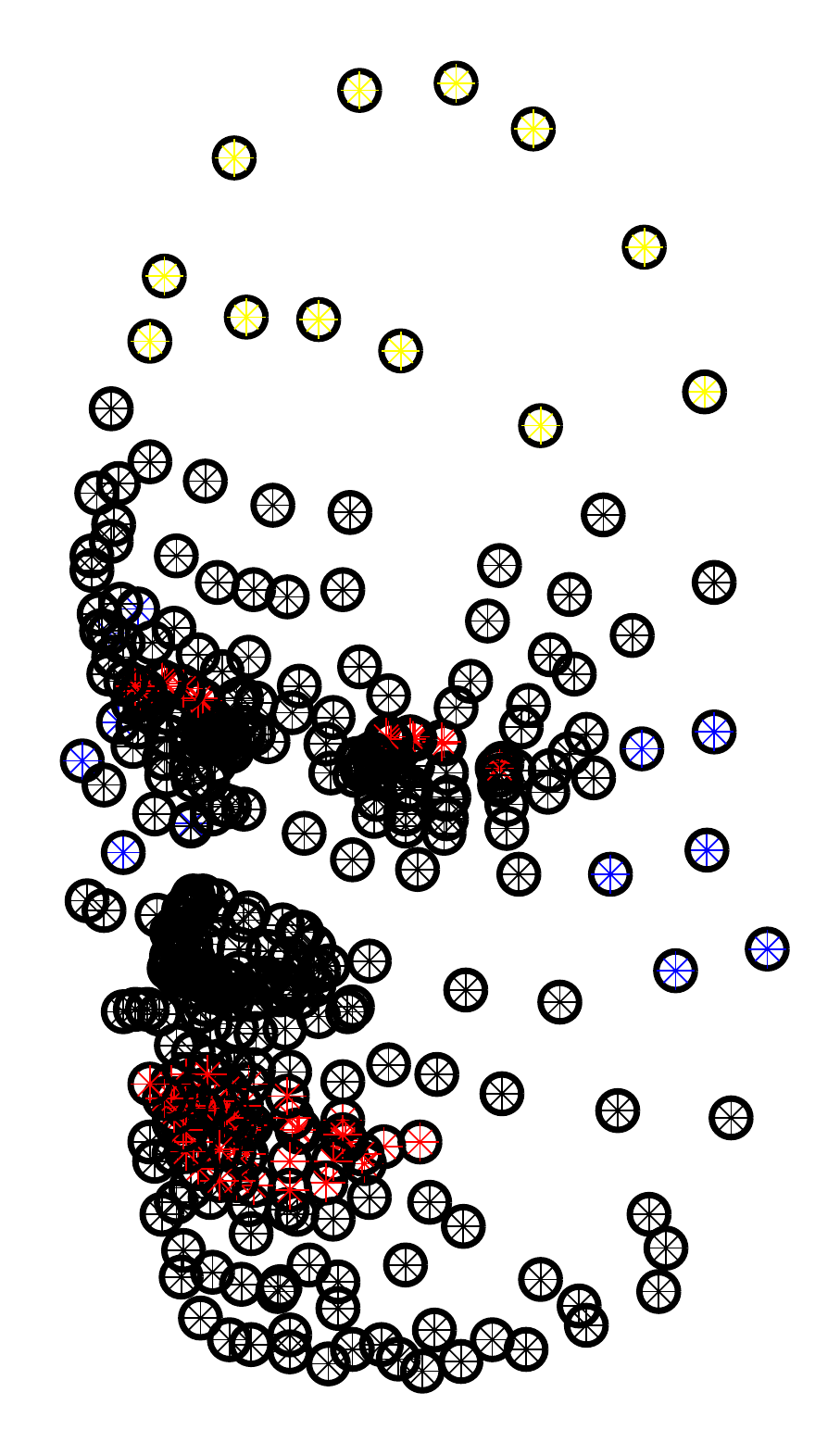}
\\
\includegraphics[width = 0.3\textwidth]{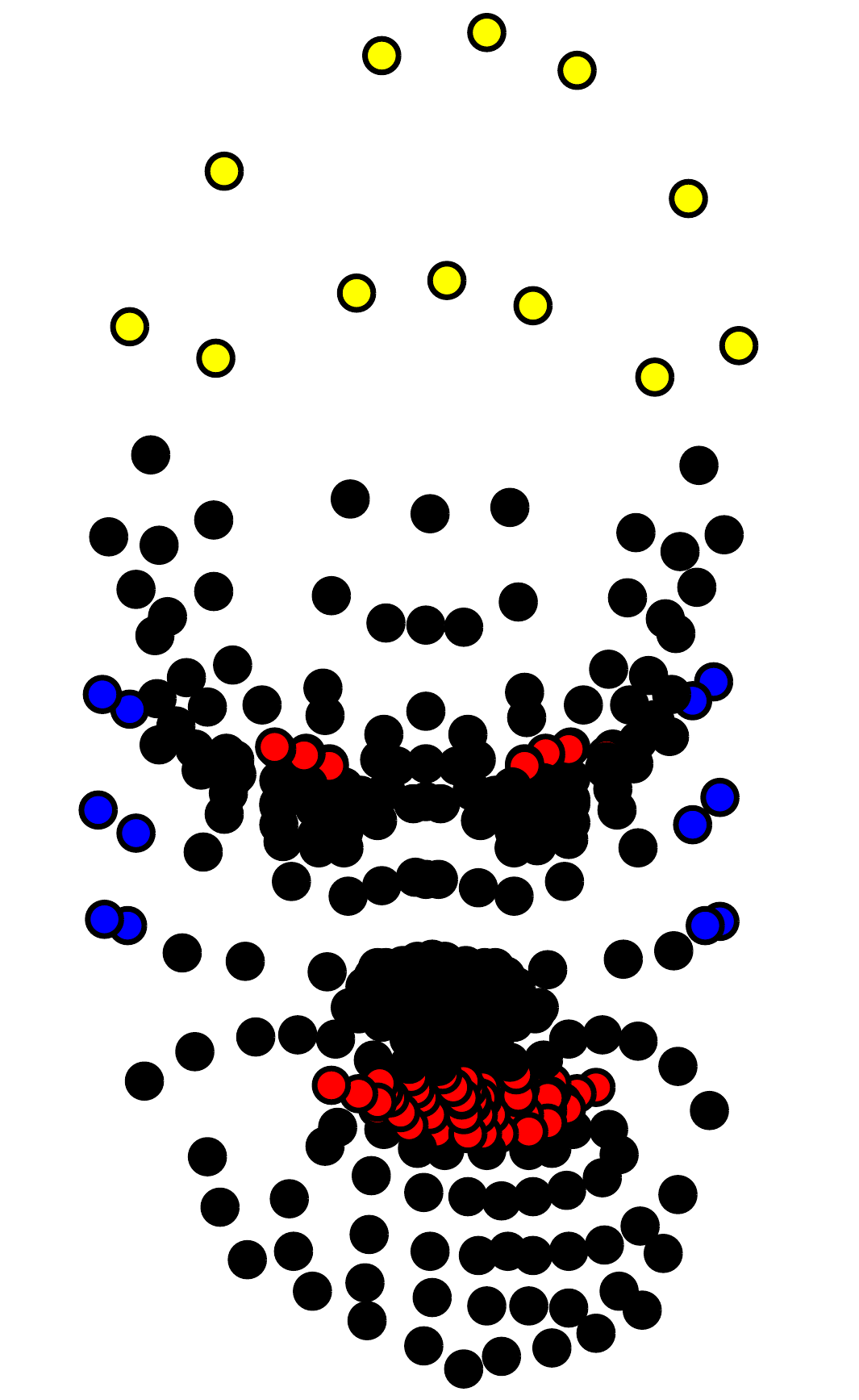} \qquad
\includegraphics[width = 0.3\textwidth]{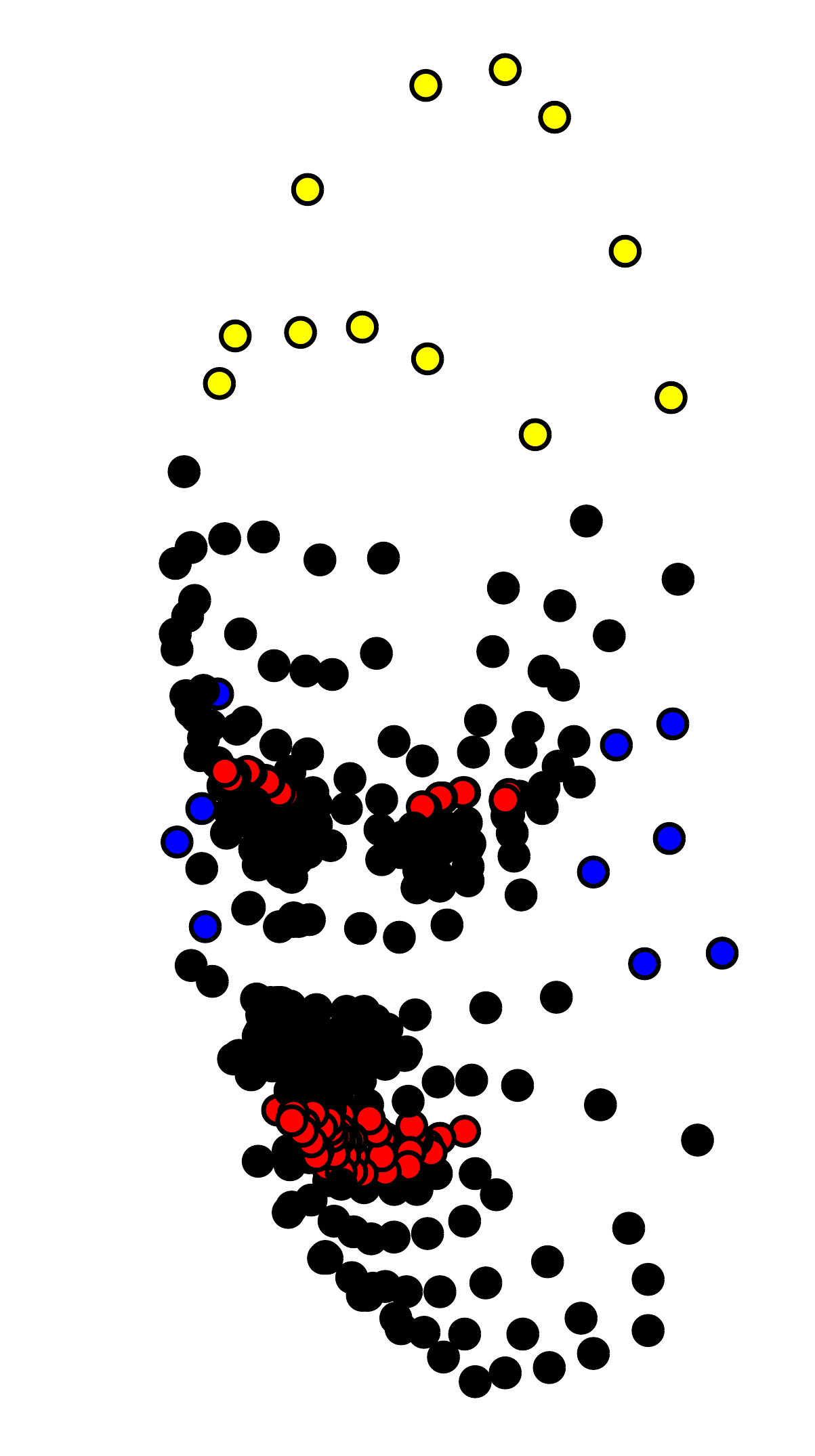}

 \caption{ \mscc{\textit{Anchor} $X$ (first row) and \textit{Moving} $Y$ (second row)} of face shape. There are 4 colors, yellow forehead, red eyebrows and lips, blue ears and the rest black.}
 \label{fig:ccpd:easy:face}
\end{figure}

The 3D face experiments are presented here. Different \textit{Anchor} $X$ and \textit{Moving} $Y$ points are used based on two initial positions (data obtained from the original work of CPD \cite{Myronenko2010}). The face coloring has been done using four tones in RGB, with main black part, red lips and eyebrows, blue ears and yellow forehead (see Figure \ref{fig:ccpd:easy:face}). 
  
Table \ref{tab:face} presents the RMS error for the 3D tests. In Test 1 outlier handling is evaluated by removing all data points from the forehead (yellow part) of $Y$. Tests 2, 3 and 4 correspond to the missing data evaluation. Test 2 is similar to Test 1, but removing the data from $X$. As the unmatched data cannot be parametrized as outliers, the original CPD is not able to register it properly. Test 3 removes all color parts except the black one obtaining better results for the CCPD proposal. Finally, in Test 4 the algorithm registers the non-black parts (i.e.: forehead, ears, lips and eyebrows), in the \textit{Anchor} $X$ with the complete \textit{Moving} $Y$. Similarly to the 2D experiments, the proposed method is able to register more accurately. 

\begin{table}[b]
  \centering
     \caption{RMS registration error of face shape tests.} 
    \begin{tabular}{lrr}
    \hline
          & CCPD & CPD \\
    \hline
    Test 1 & 0.37278E-02 & 1.62E-02 \\
    Test 2 & 0.28985E-02 & 4.078E-02 \\
    Test 3 & 0.21677E-02 & 12.051E-02 \\
    Test 4 & 0.5984E-02 & 10.407E-02 \\
    \hline
    \end{tabular}%

  \label{tab:face}%
\end{table}%

\begin{figure}
  \centering
\begin{tabular}{p{0.01\textwidth} p{0.2\textwidth} p{0.2\textwidth} p{0.2\textwidth} p{0.2\textwidth}}
& X & Y & CCPD & CPD \\
\hline
\end{tabular} \\

\scalebox{1}{
	\includegraphics[width=0.2\textwidth]{./cpd_exp_face_model.pdf} 
	\includegraphics[width=0.2\textwidth]{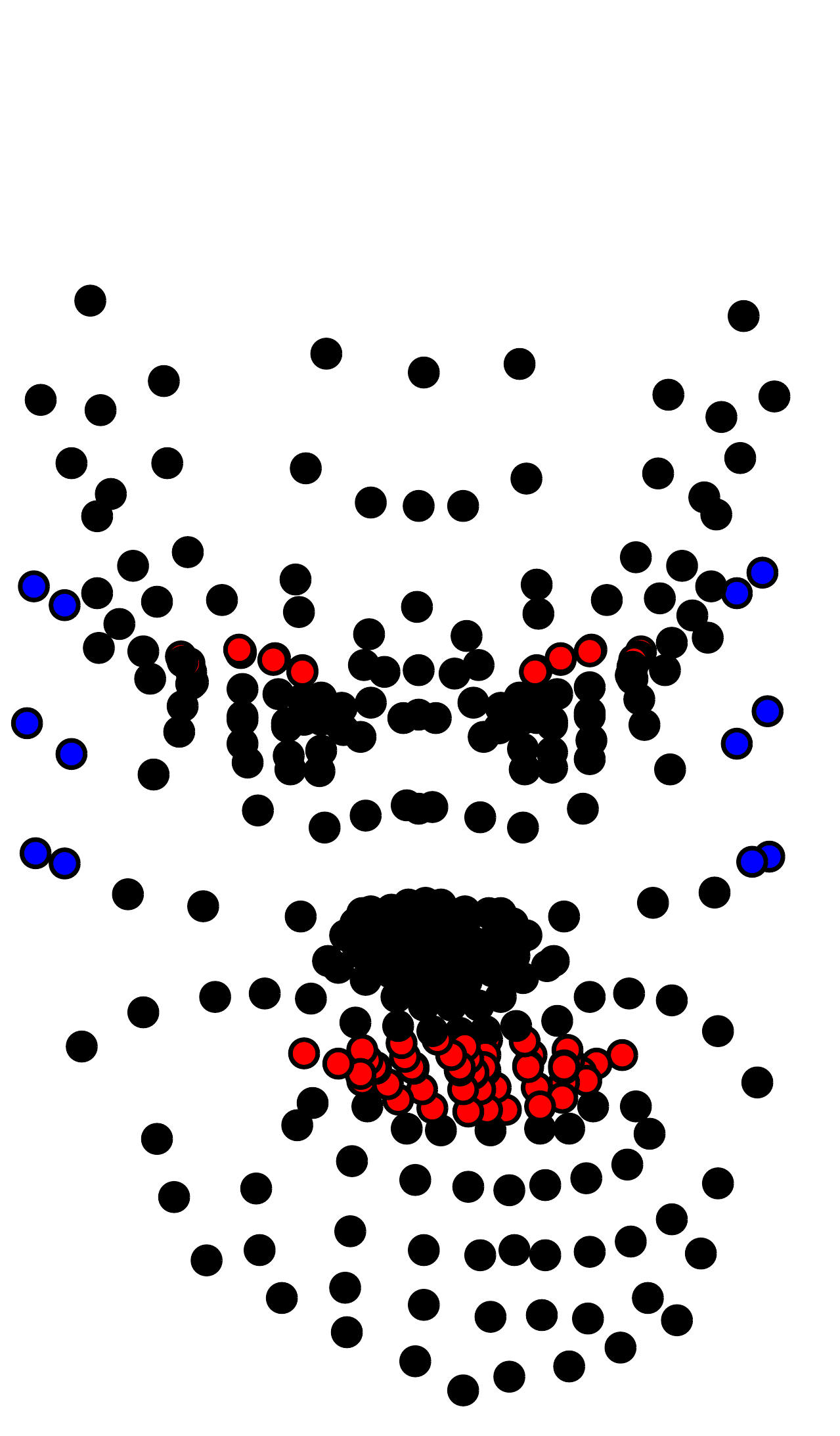}  	
	\includegraphics[width=0.2\textwidth]{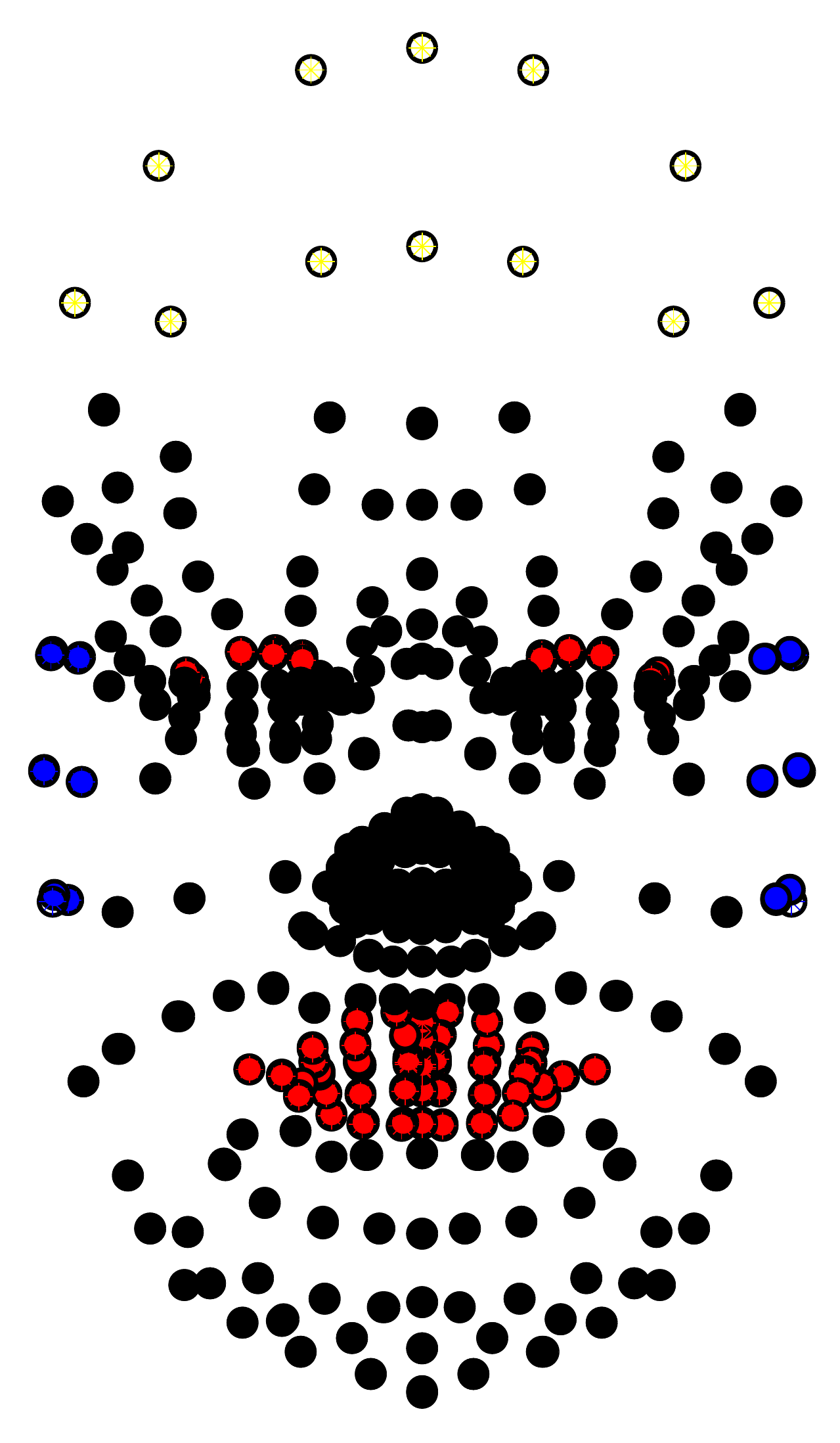}  
	\includegraphics[width=0.2\textwidth]{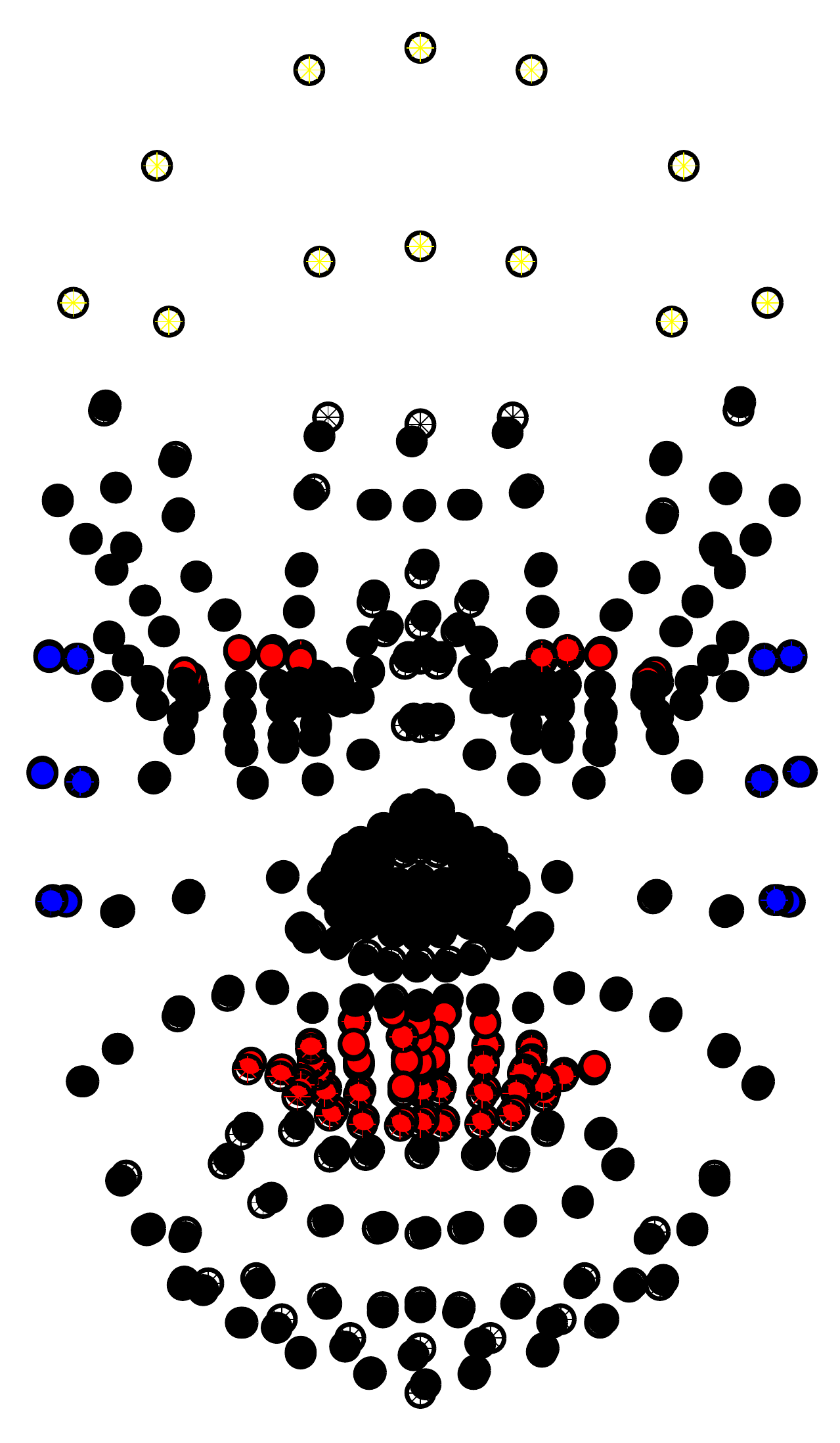}
}

\scalebox{1}{
	\includegraphics[width=0.2\textwidth]{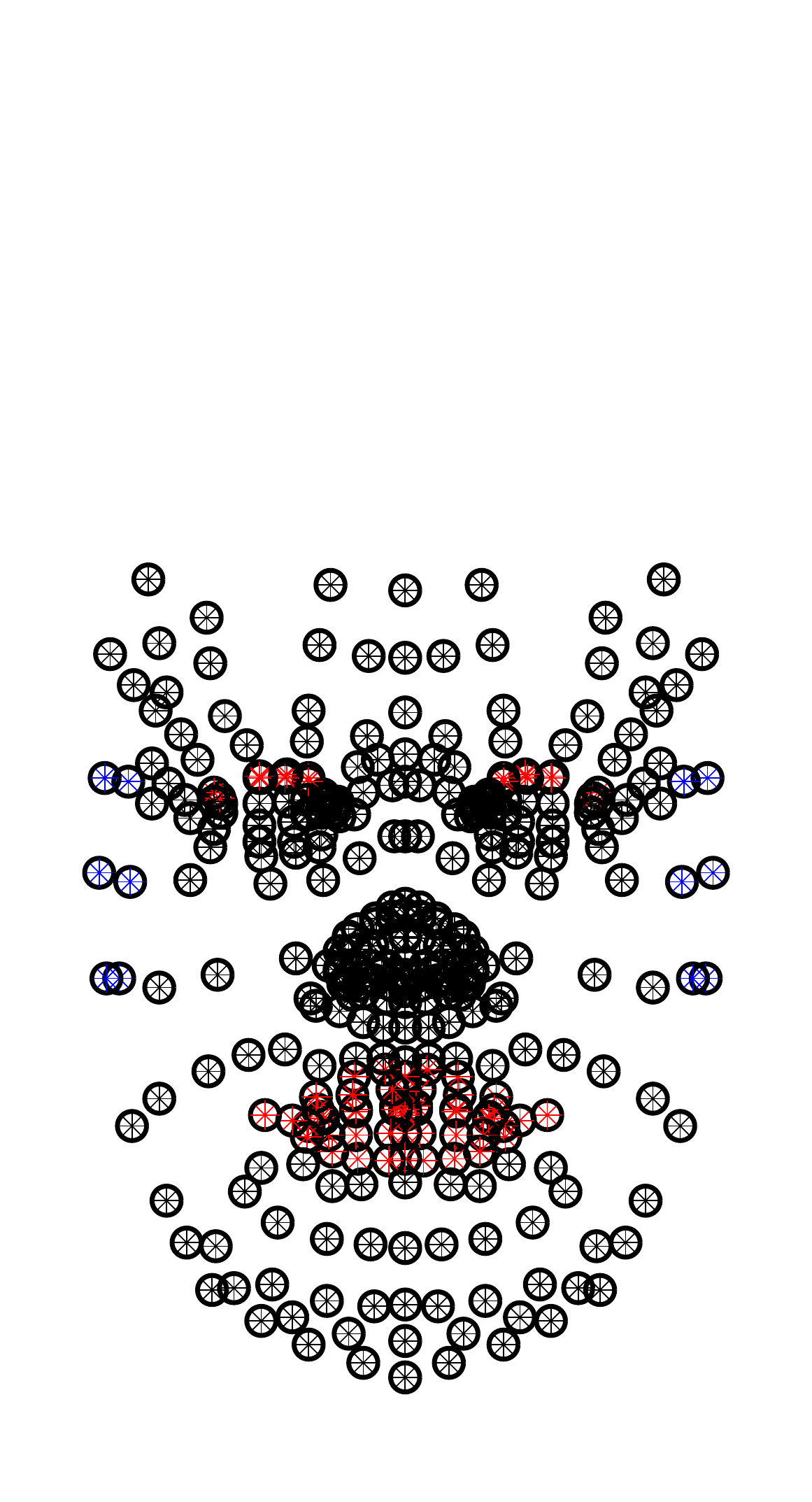}  
	\includegraphics[width=0.2\textwidth]{./cpd_exp_face_data.pdf}
	\includegraphics[width=0.2\textwidth]{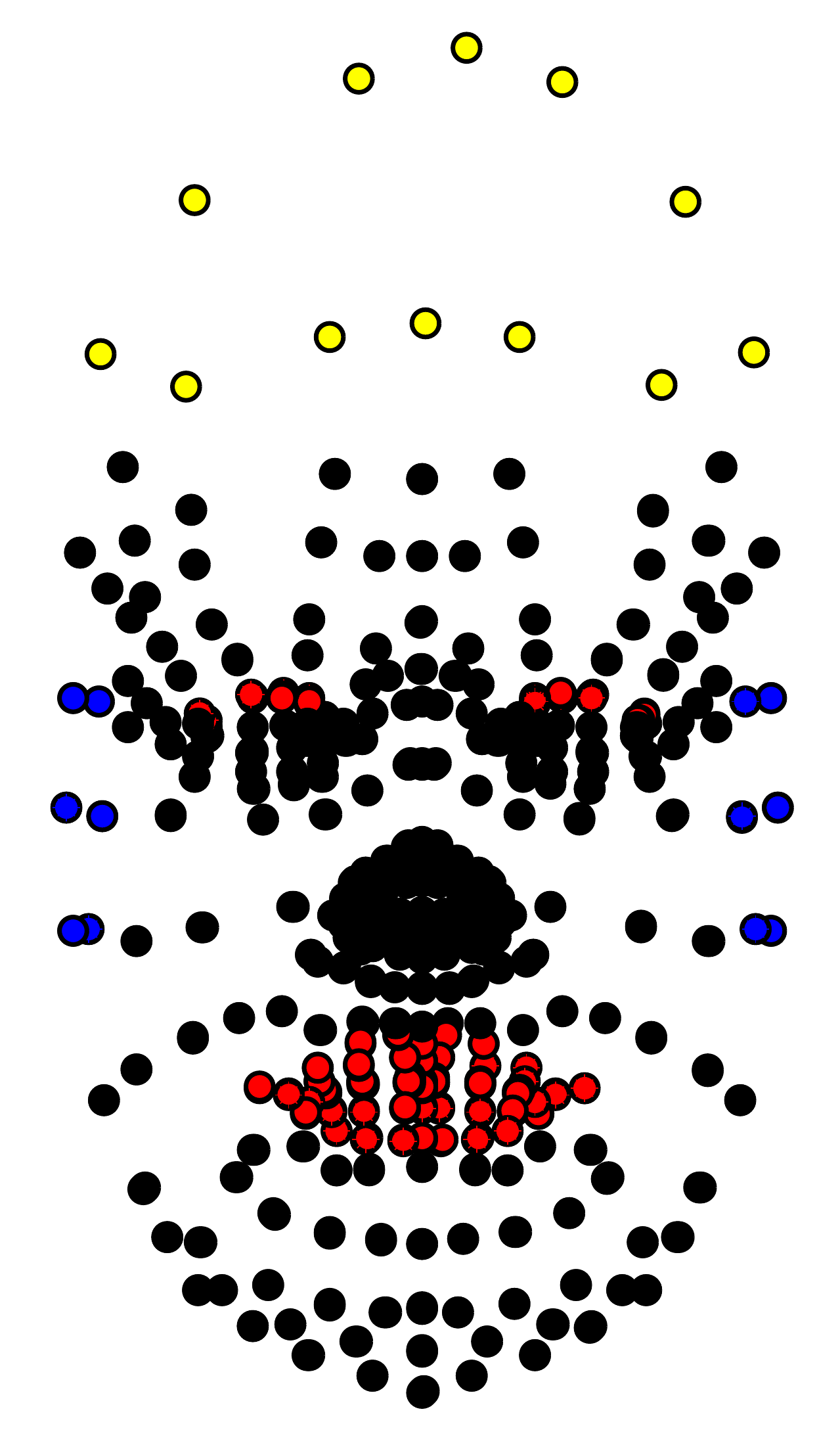}  
	\includegraphics[width=0.2\textwidth]{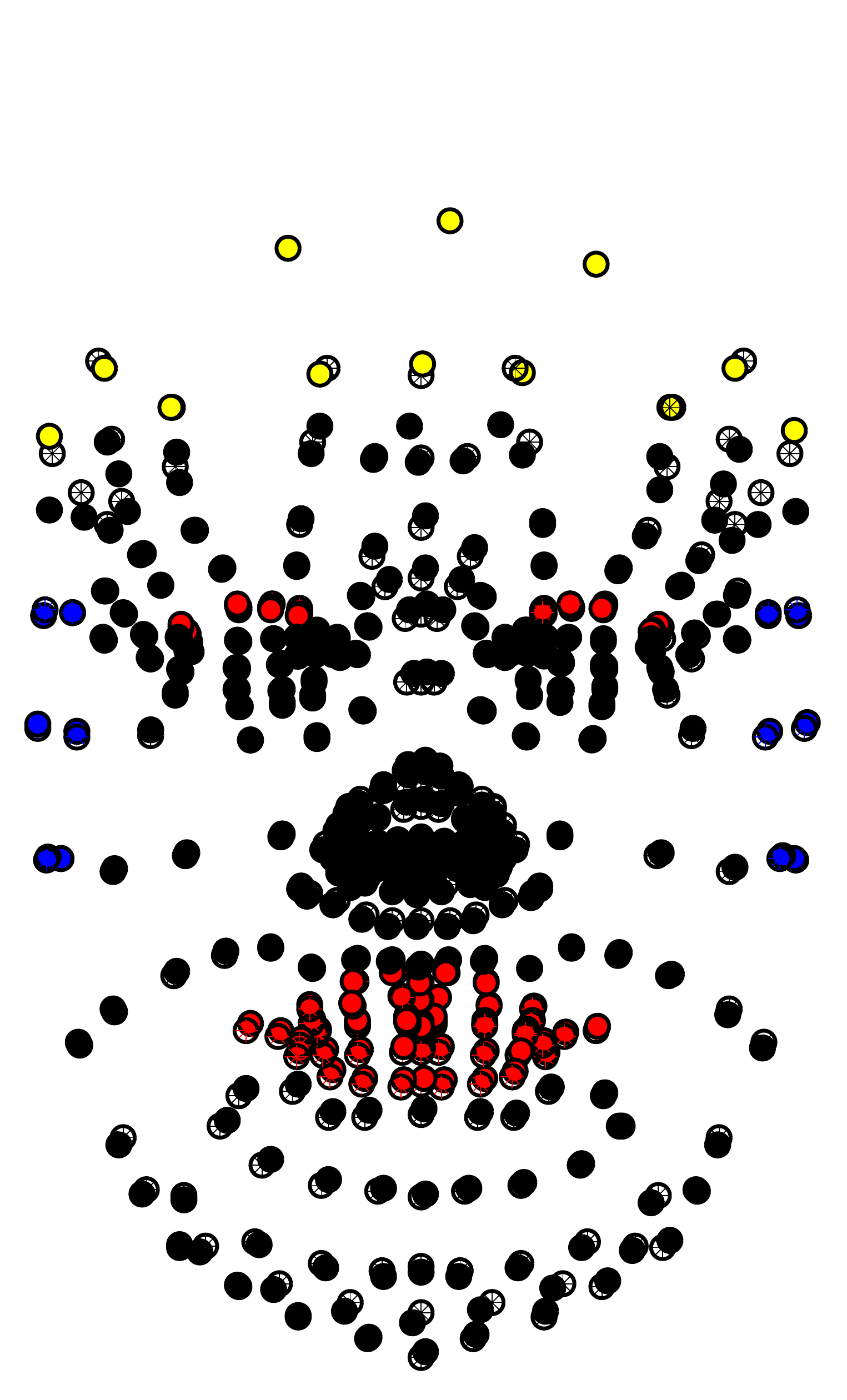}
}

\scalebox{1}{
	\includegraphics[width=0.2\textwidth]{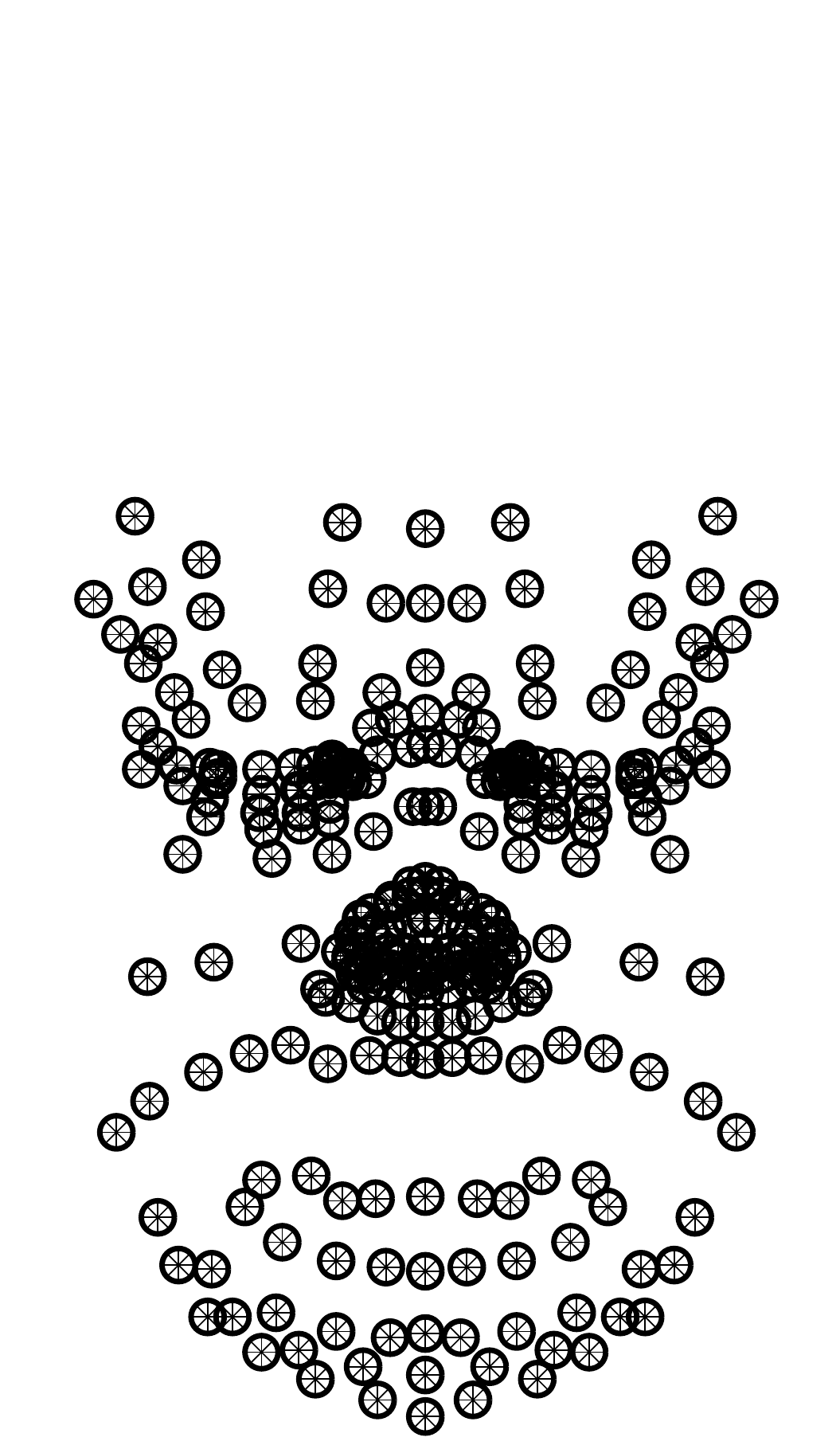}  
	\includegraphics[width=0.2\textwidth]{./cpd_exp_face_data.pdf}
	\includegraphics[width=0.2\textwidth]{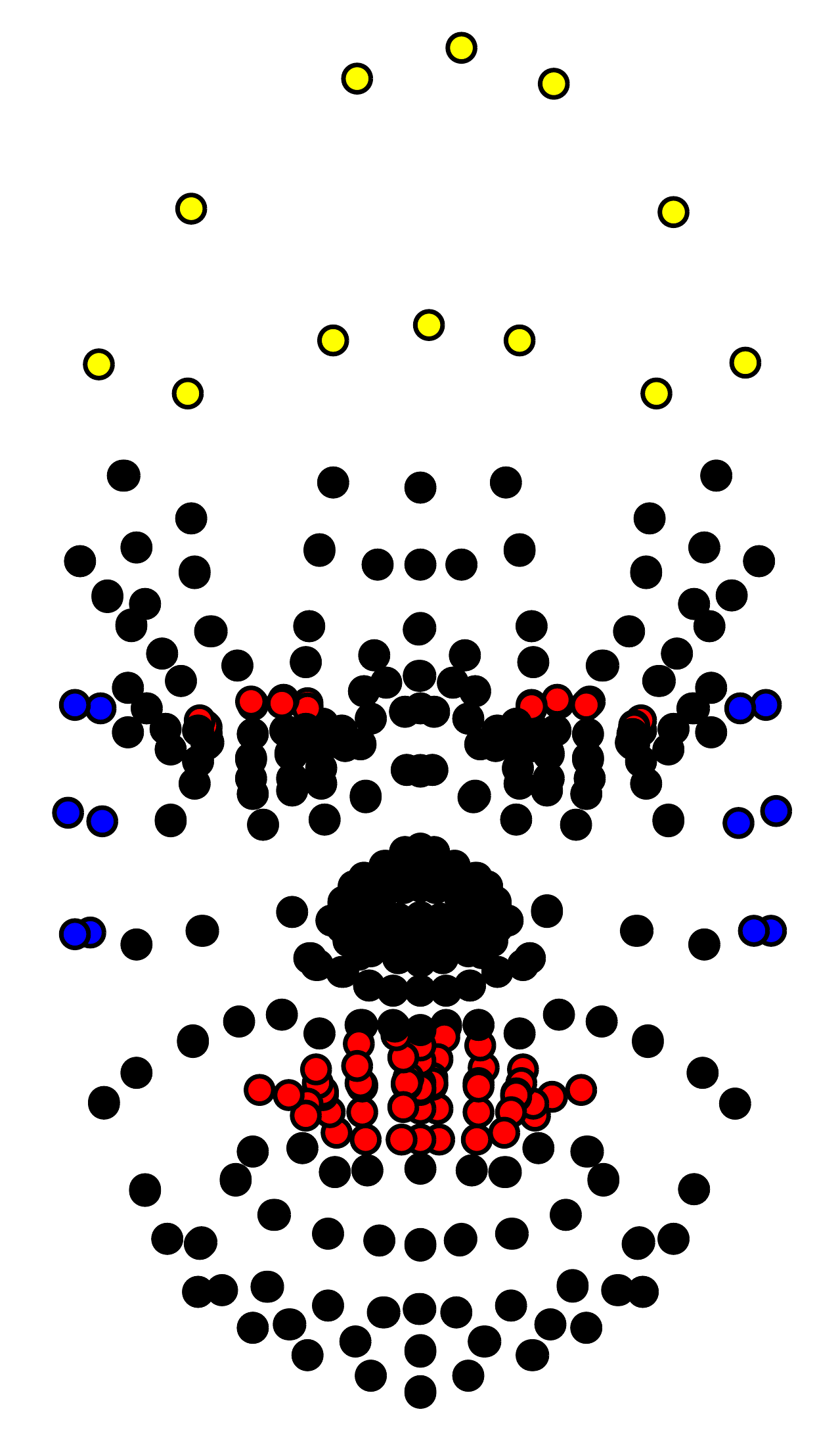} 
	\includegraphics[width=0.2\textwidth]{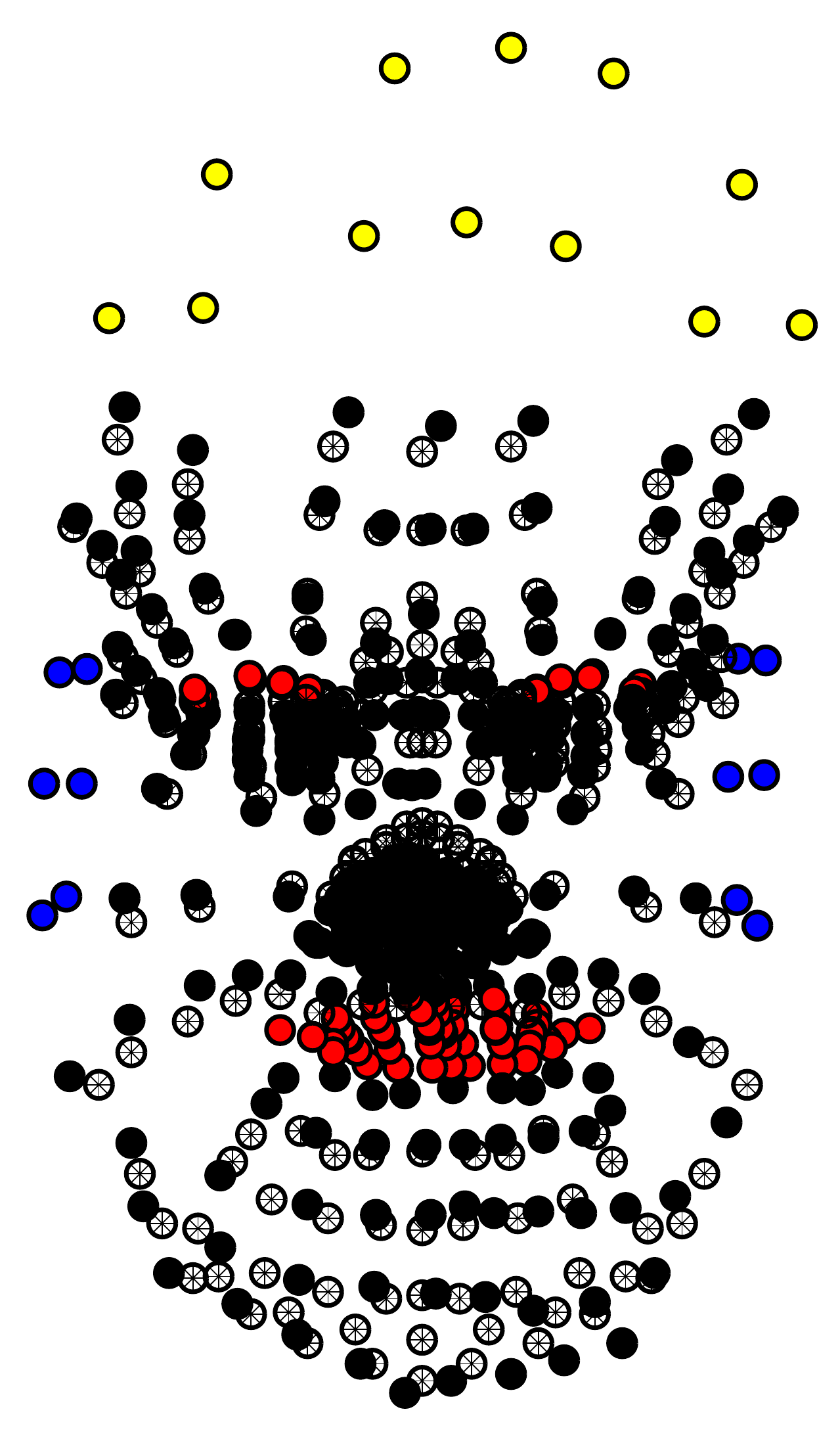}    
}

\scalebox{1}{ 
	\includegraphics[width=0.2\textwidth]{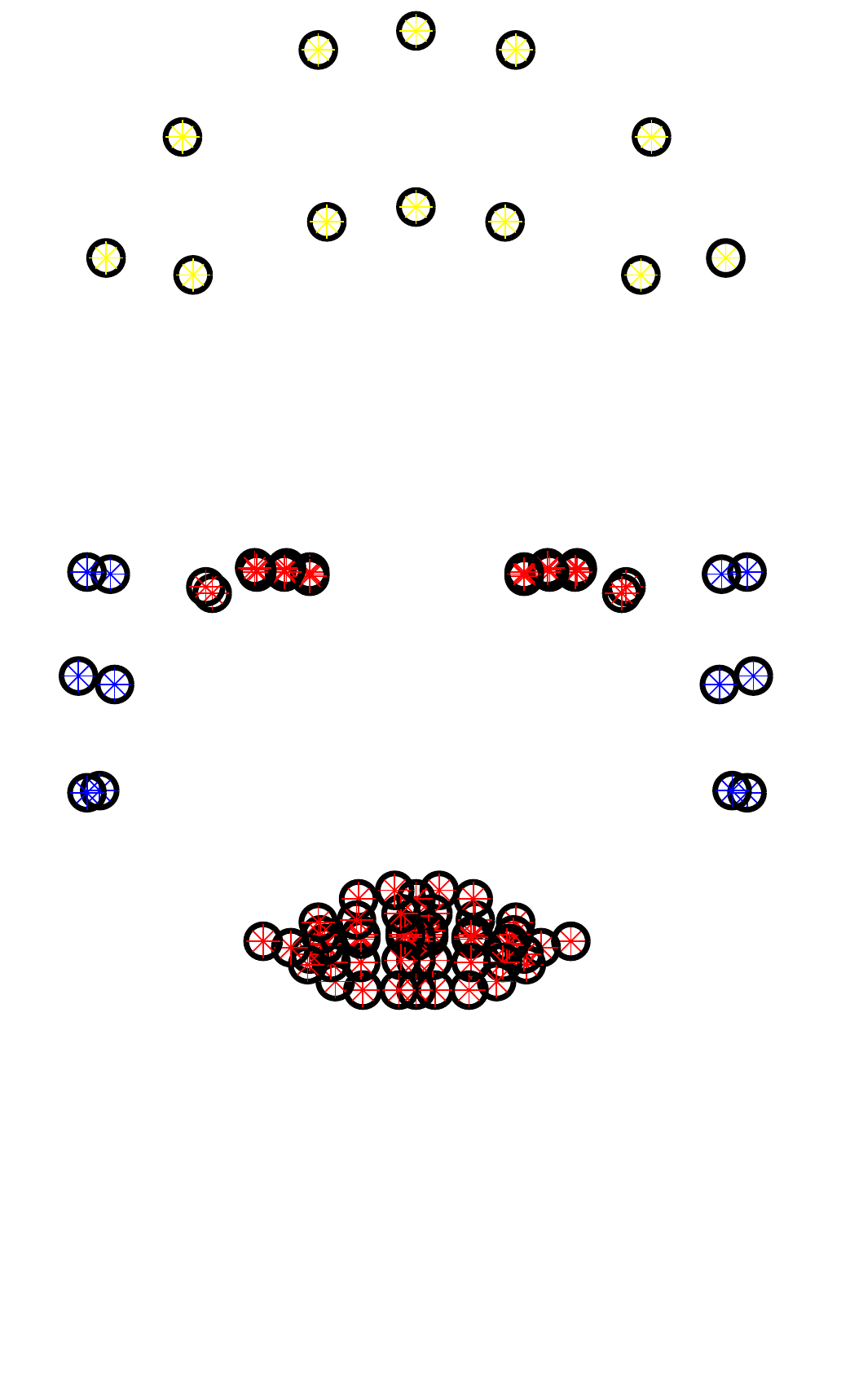}  
	\includegraphics[width=0.2\textwidth]{./cpd_exp_face_data.pdf}  
	\includegraphics[width=0.2\textwidth]{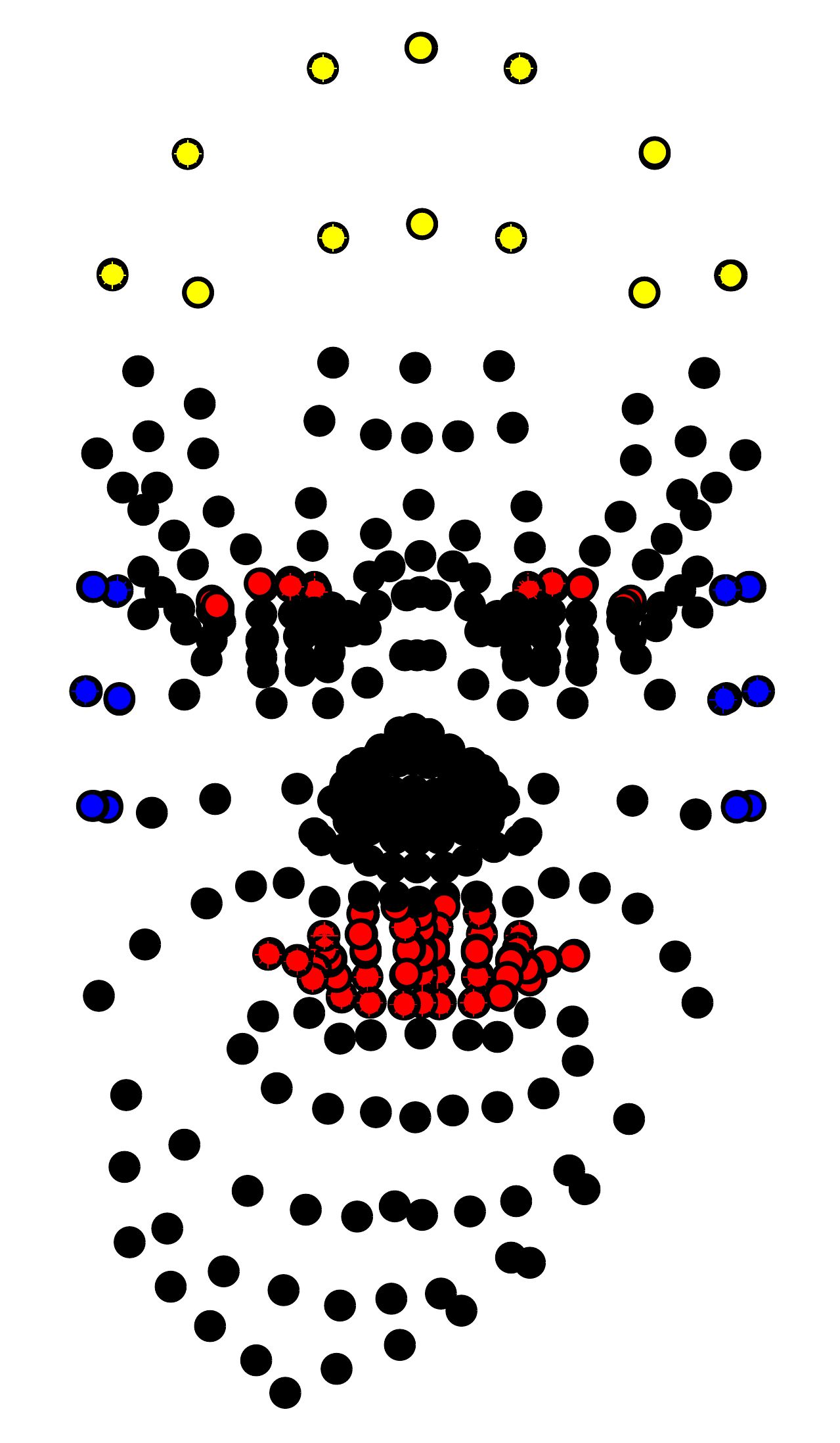} 
	\includegraphics[width=0.2\textwidth]{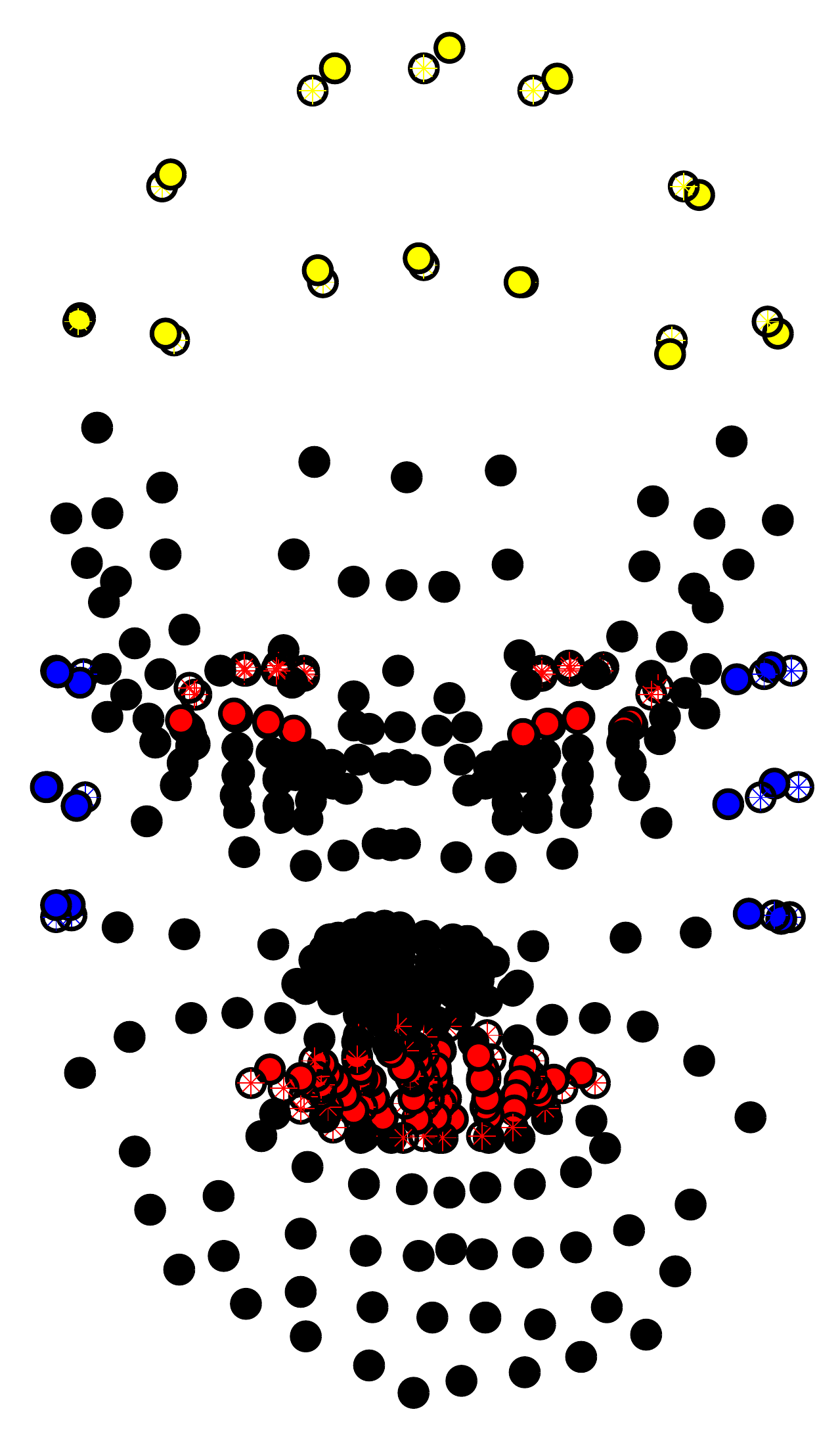}
}
	\caption[Rows 1 to 4 show Tests 1 to 4 of face shape]{\mscc{Rows 1 to 4 show Tests 1 to 4 of the face shape. The columns represent from left to right the \textit{Anchor} $X$, the \textit{Moving} $Y$, the CCPD result and the CPD result.}}
  \label{fig:faceimages}%

\end{figure}

Figure \ref{fig:faceimages} shows the result of the tests, where each row is a test from 1 to 4 respectively, to visually evaluate the performance of both methods. In the second row it is possible to see how CPD moves wrongly yellow points downward while the proposed method keeps the point in the top part as they do not have correspondences. The third row has only color points in the \textit{Anchor} $X$, without the black part. The proposal aligns properly these remaining parts while CPD cannot align the parts properly. Similarly the fourth test is correctly aligned by CCPD as the corresponding points in the \textit{Anchor} and \textit{Moving} are properly aligned, while CPD returns an inaccurate result.

\begin{table}
  \centering
	\caption{RMS registation error of 50 face shapes.}  
    \begin{tabular}{lrr}
    \hline
          & CCPD & CPD \\
    \hline

    Test 2 & 0.26E-02 & 3.93E-02 \\
    Test 3 & 0.32E-02 & 8.53E-02 \\
    Test 4 & 1.13E-02 & 11.54E-02 \\
    \hline
    \end{tabular}%

  \label{tab:50face}%
\end{table}%

A large test evaluation is presented in Table \ref{tab:50face} where a set of 50 different changes are registered (dataset available from Myronenko \cite{Myronenko2010}). The average RMS errors for the Tests 2, 3 and 4 are 0.36834E-02 for CCPD and 8.8453E-02 for CPD. The proposal's RMS is 24 times lower than original method. 

\begin{figure}
  \centering%
\includegraphics[width = 0.49\textwidth]{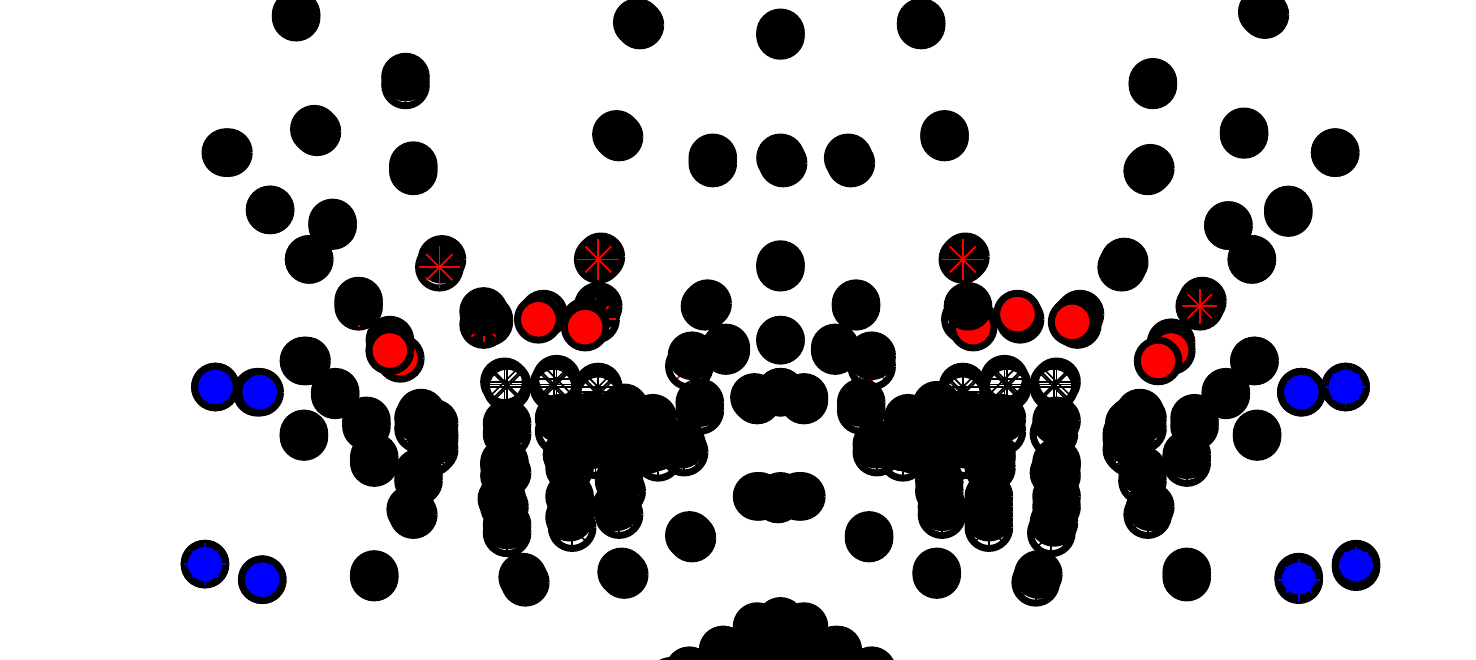} 
\includegraphics[width = 0.49\textwidth]{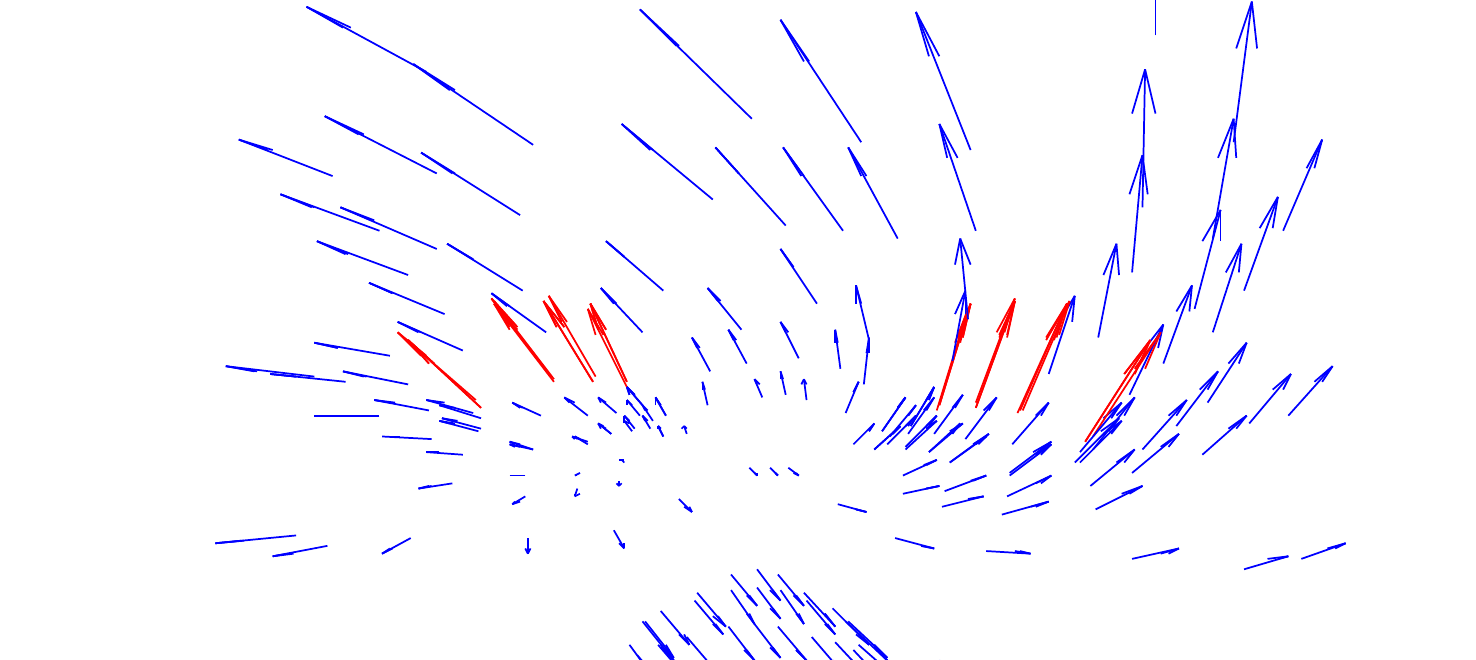}
\\

\includegraphics[width = 0.49\textwidth]{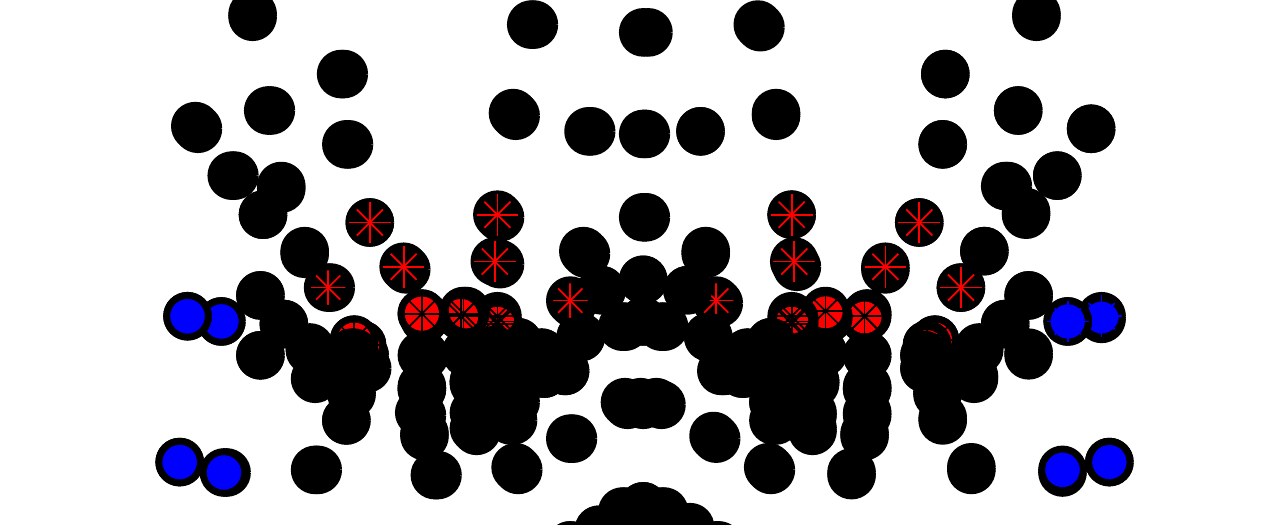}
\includegraphics[width = 0.49\textwidth]{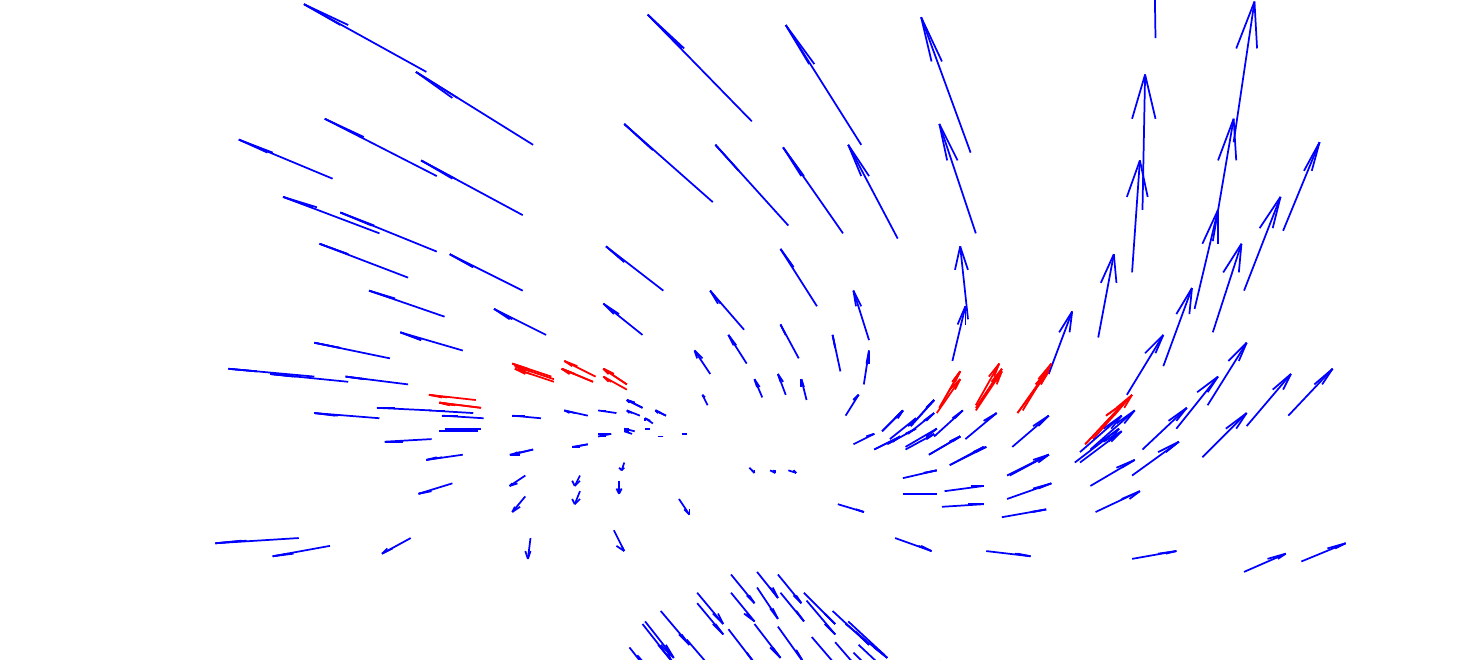}

 \caption{\mscc{Eyebrow movement test. From the top to the bottom, the CCPD result, the CCPD flow, the CPD result and the CPD flow. In the flow pictures the red arrows show the most significant displacement, i.e. the eyebrows.}}
 \label{fig:testeyebrow}
\end{figure}

A final test was carried out to evaluate a displacement of color and a large deformation. In this test, the eyebrows of $Y$ are lower than in $X$. The movement should displace the eyebrows upward. This is considered a large deformation or a non-linear deformation as the movement is not coherent in the shape data space, but coherent in the color data space. Figure \ref{fig:testeyebrow} shows this registration. In order to help in the visualization, a flow image is shown for both methods. The proposed method achieves a proper result moving up the eyebrows while the original CPD algorithm, as it does not take into account color, is not able to achieve the correct result.

\subsubsection{Experiments with noise and outliers in color space}
\begin{figure}
  \centering%
\includegraphics[width = 0.18\textwidth]{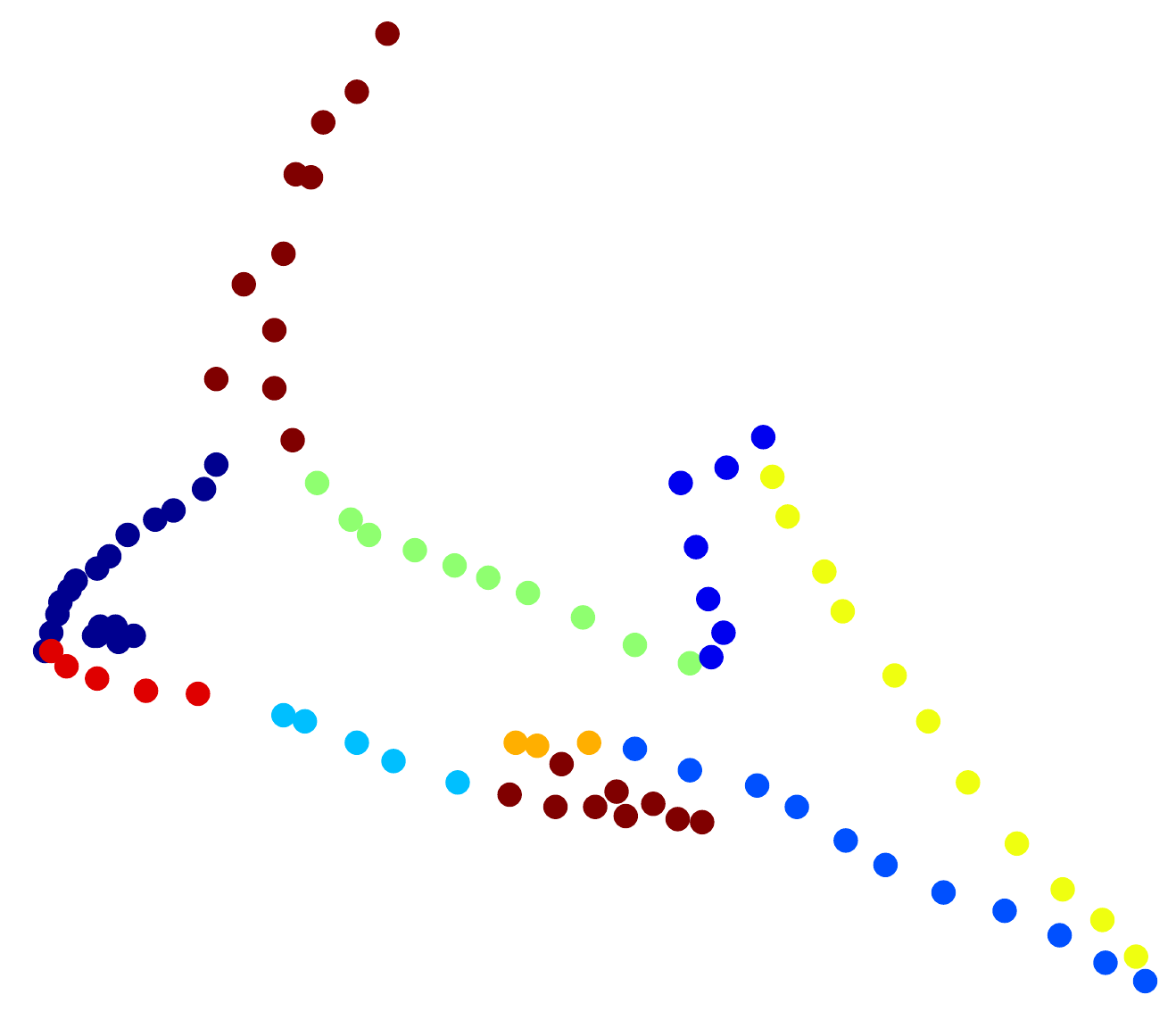} 
\includegraphics[width = 0.18\textwidth]{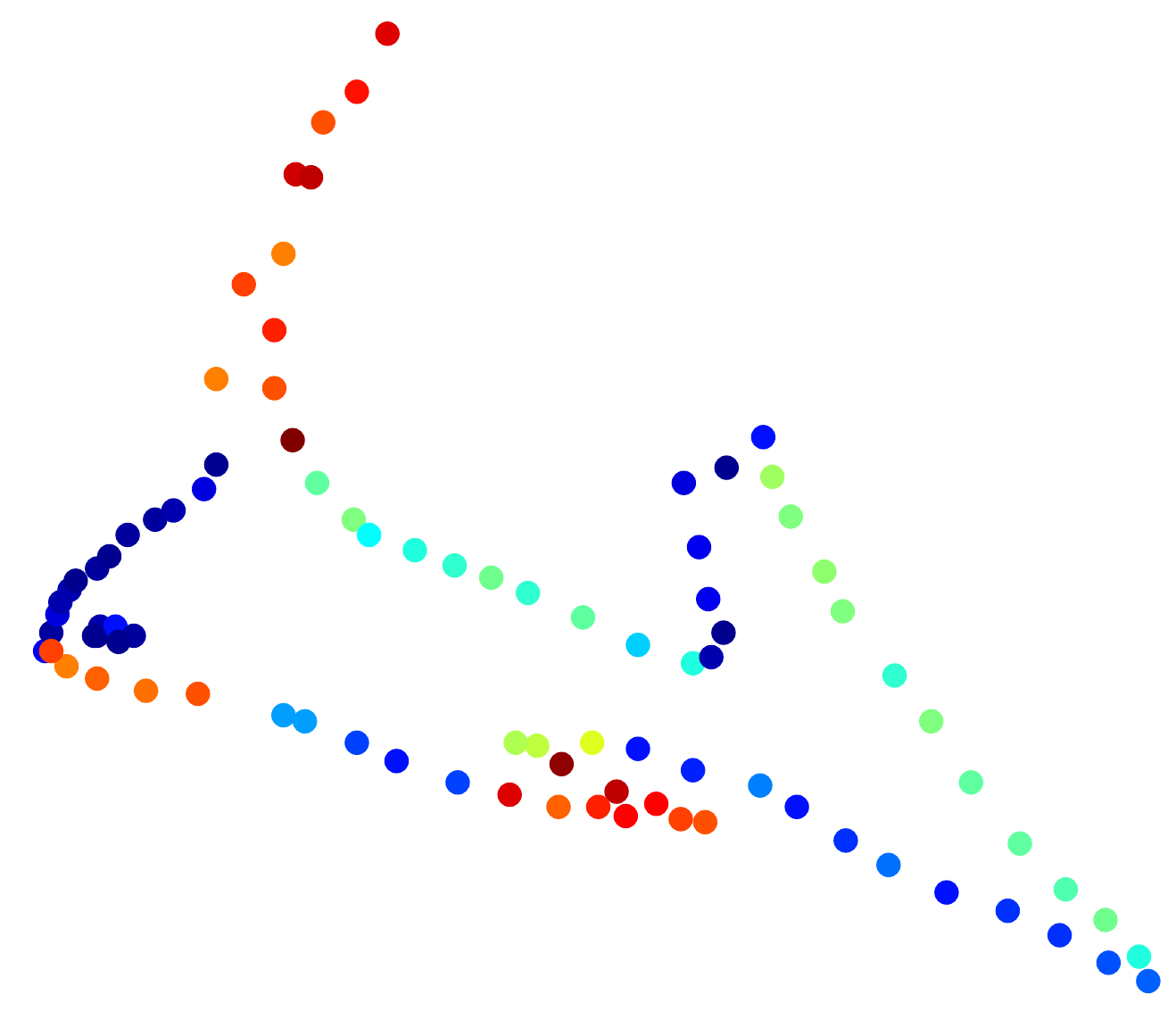}
\includegraphics[width = 0.18\textwidth]{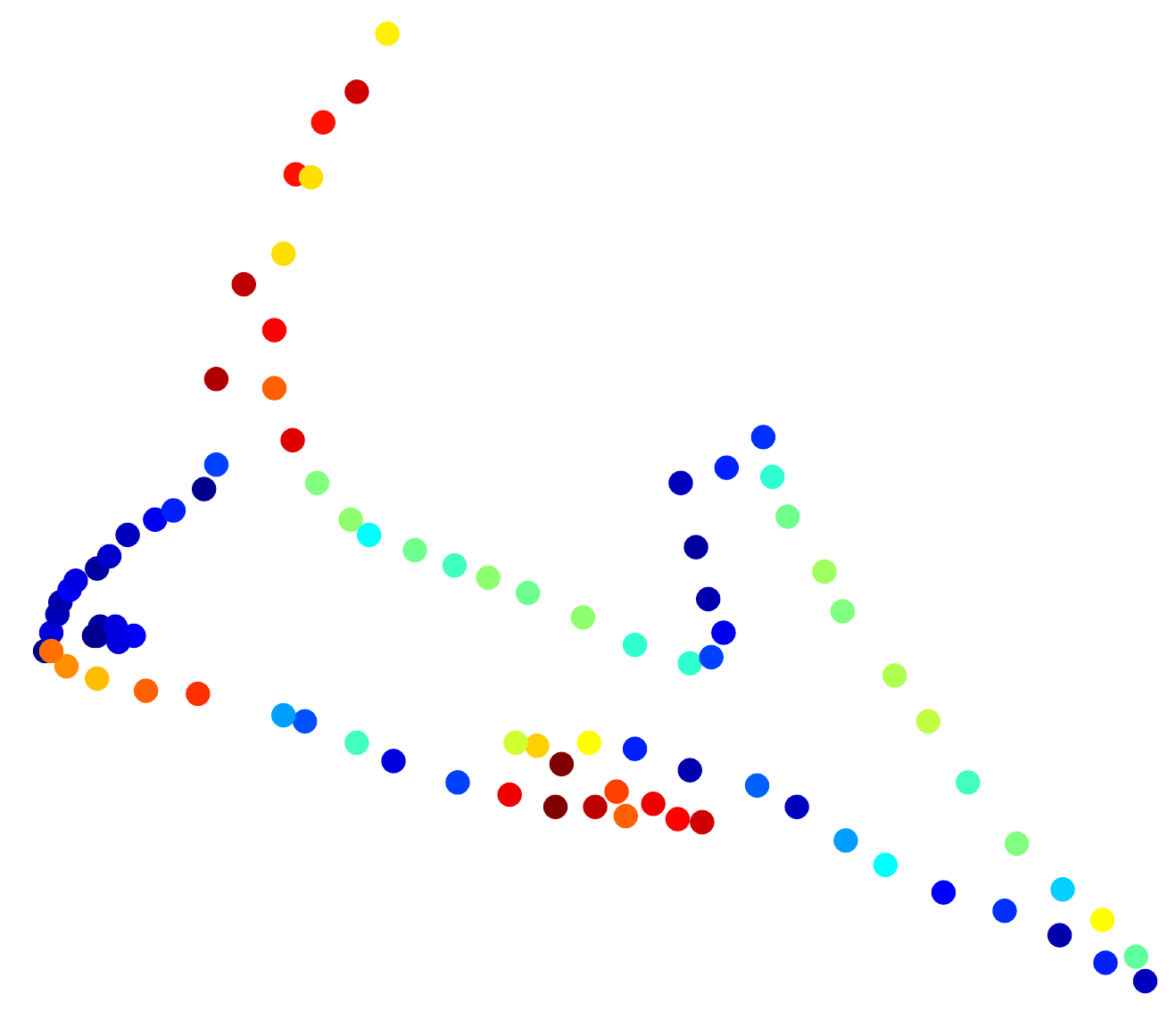}
\includegraphics[width = 0.18\textwidth]{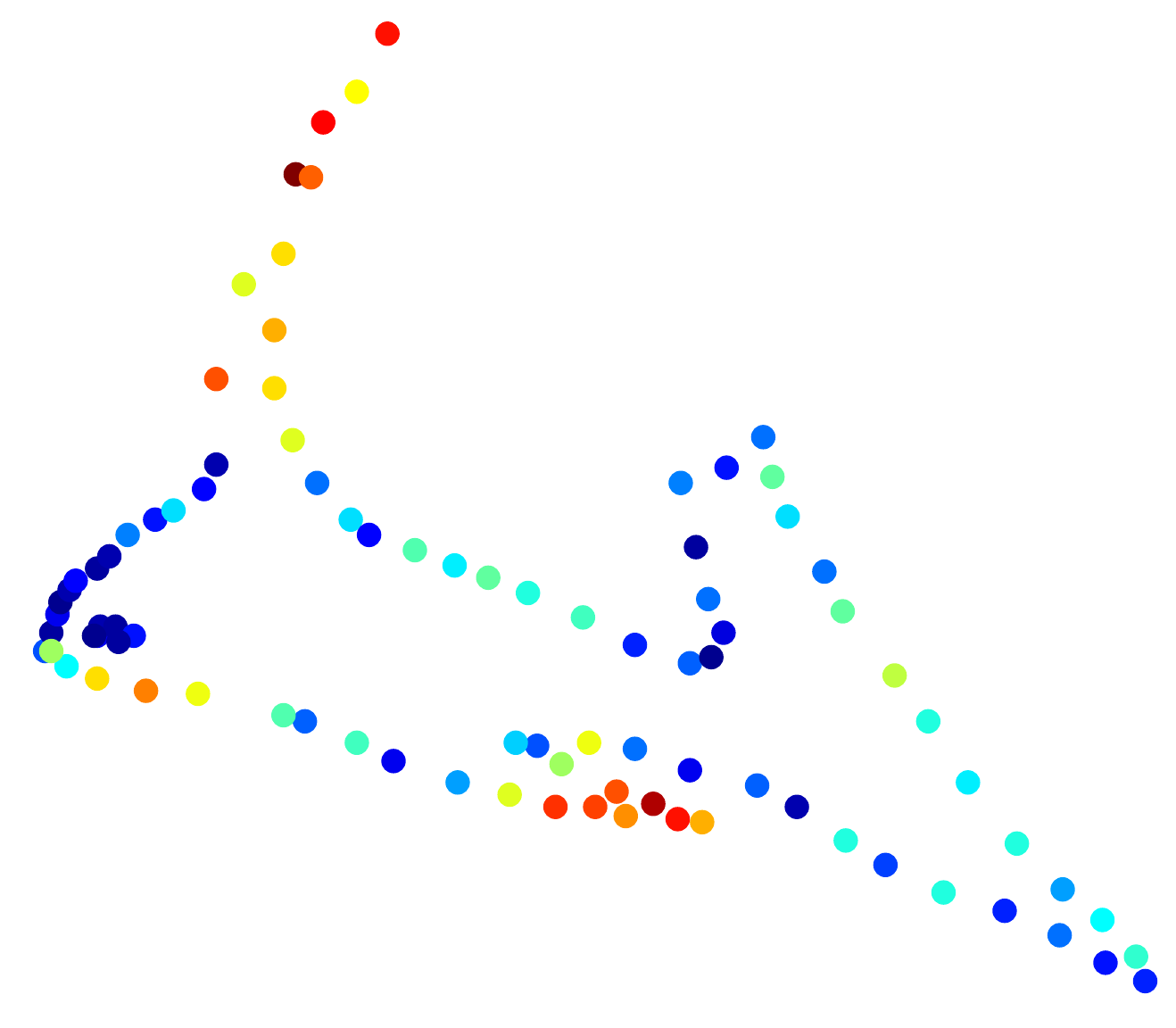}
\includegraphics[width = 0.18\textwidth]{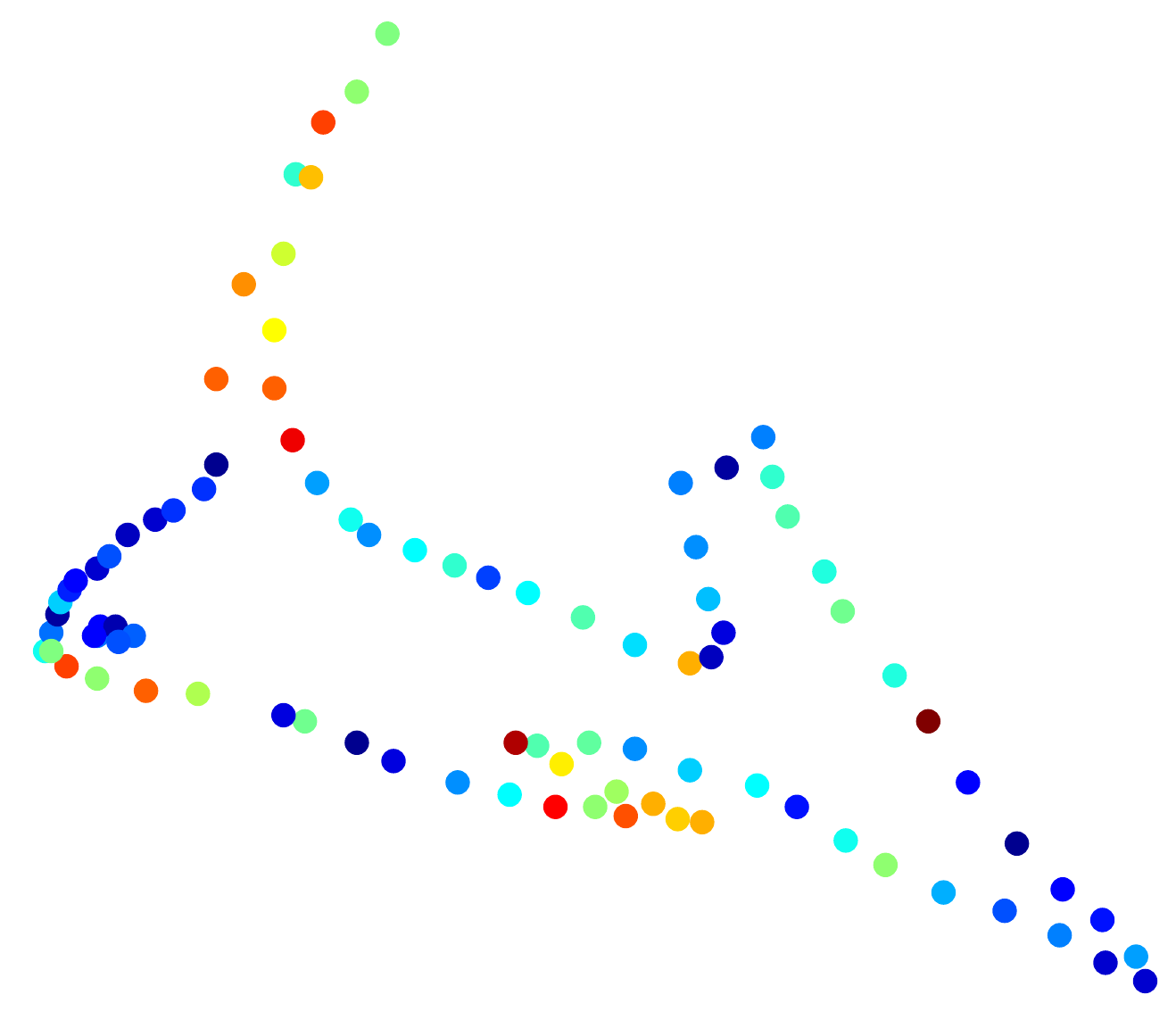}
\\

\includegraphics[width = 0.18\textwidth]{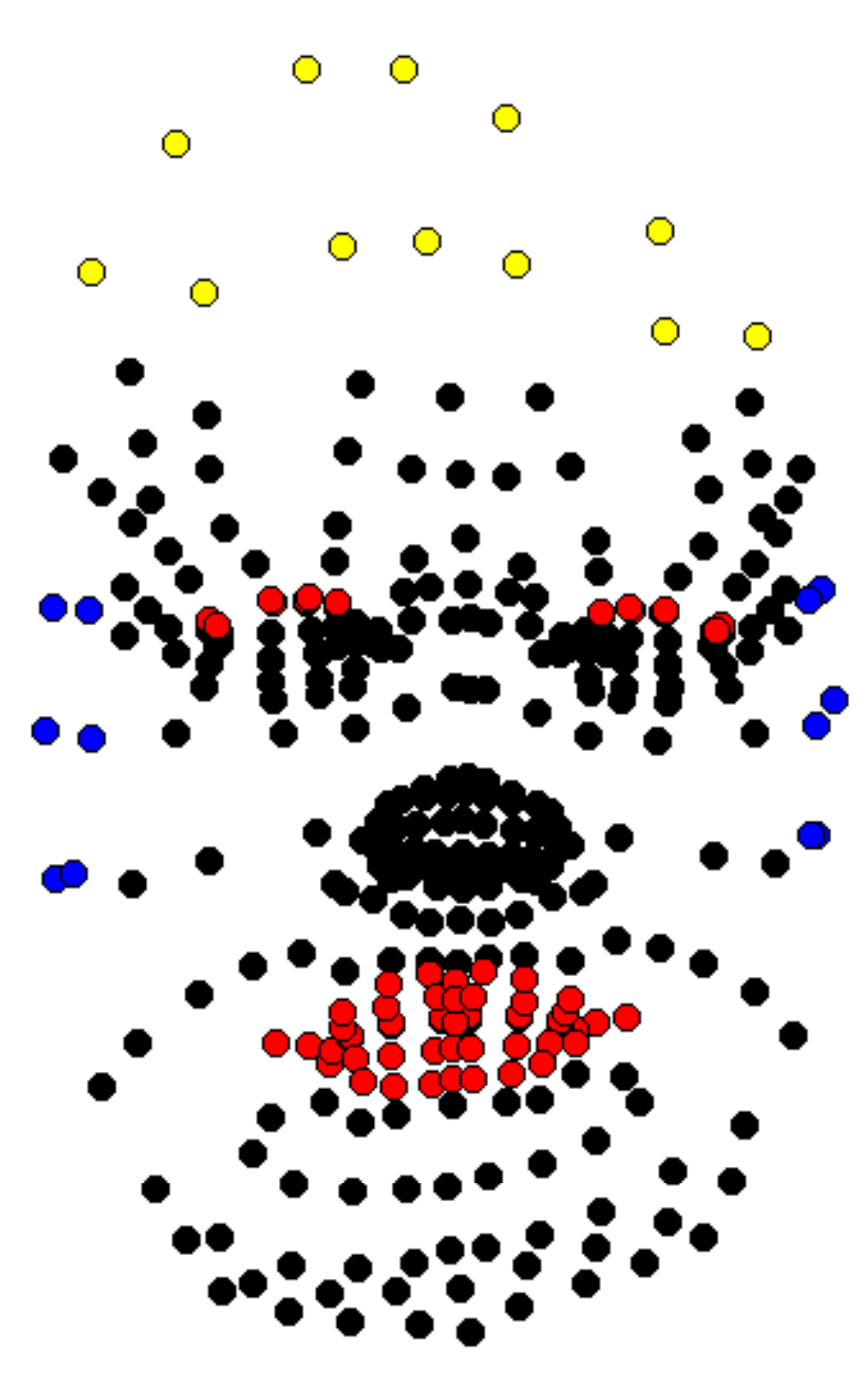} 
\includegraphics[width = 0.18\textwidth]{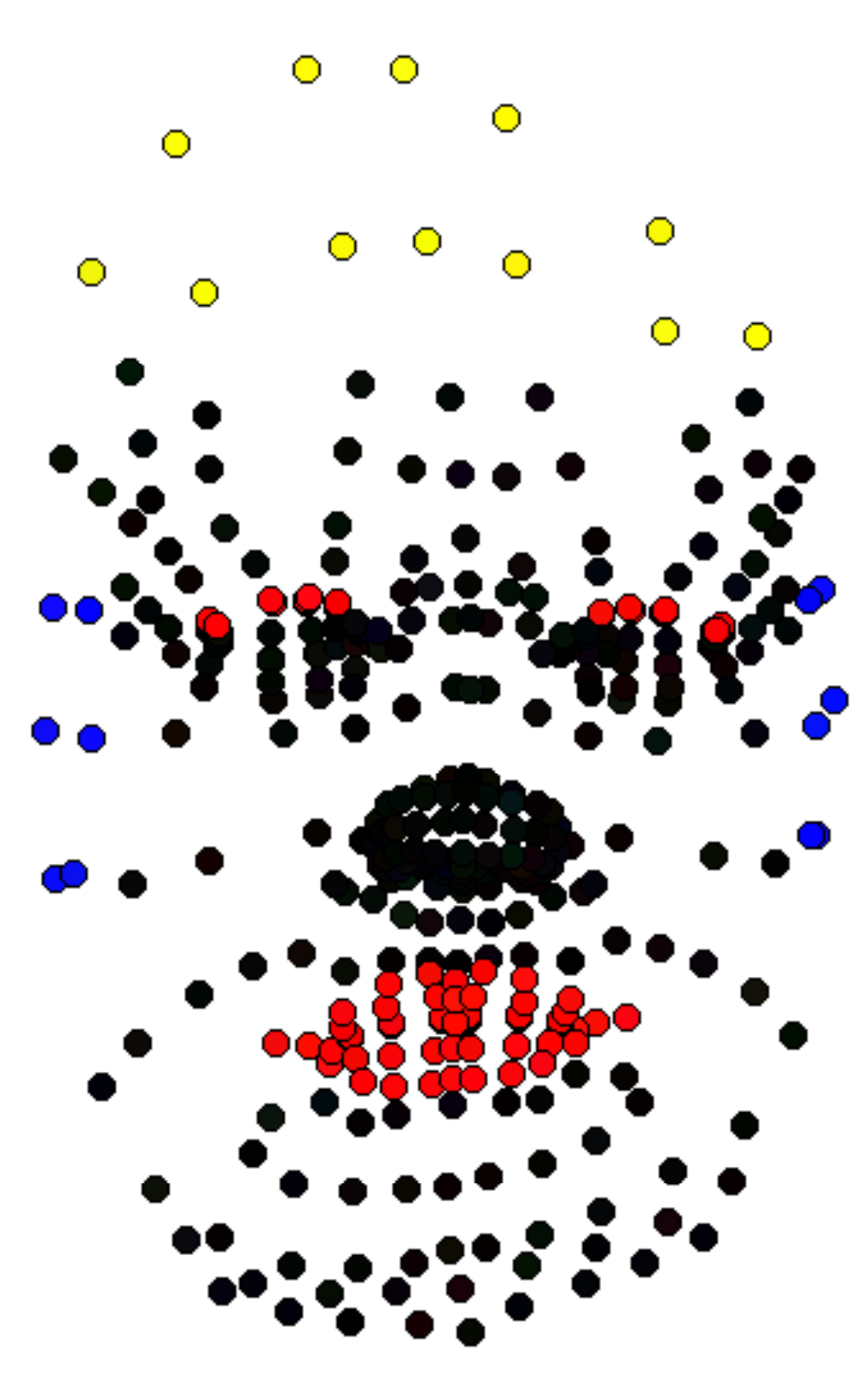}
\includegraphics[width = 0.18\textwidth]{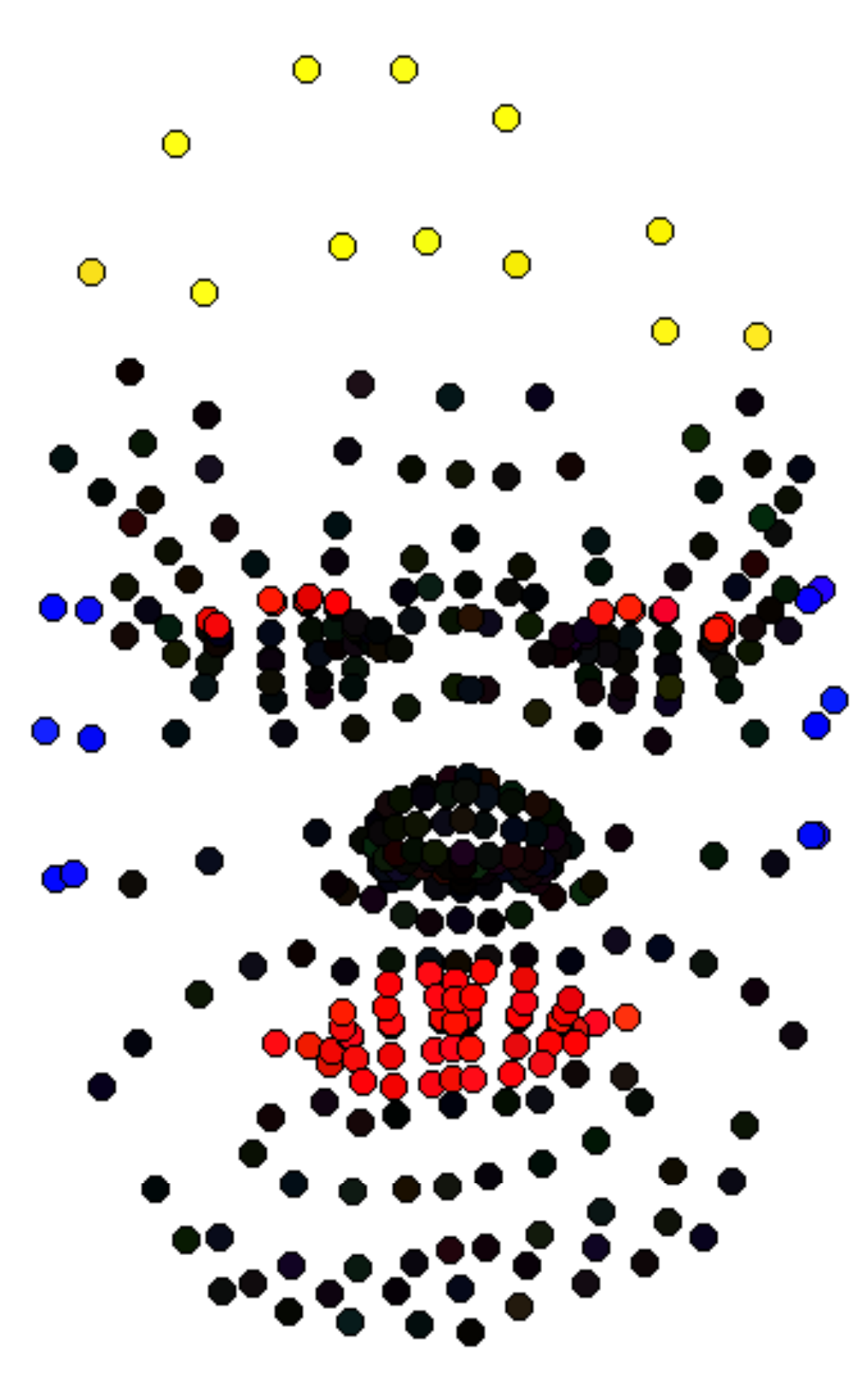}
\includegraphics[width = 0.18\textwidth]{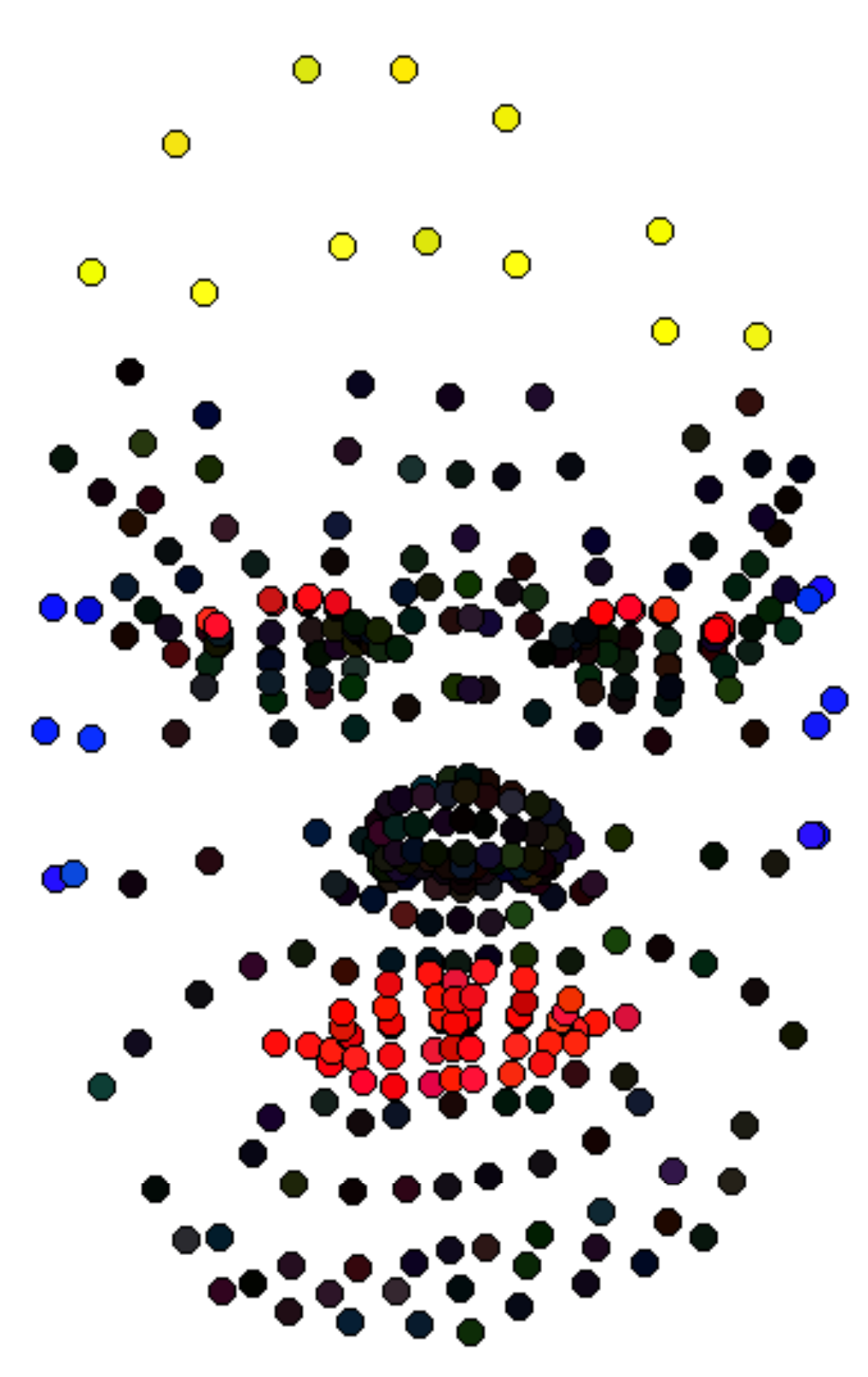}
\includegraphics[width = 0.18\textwidth]{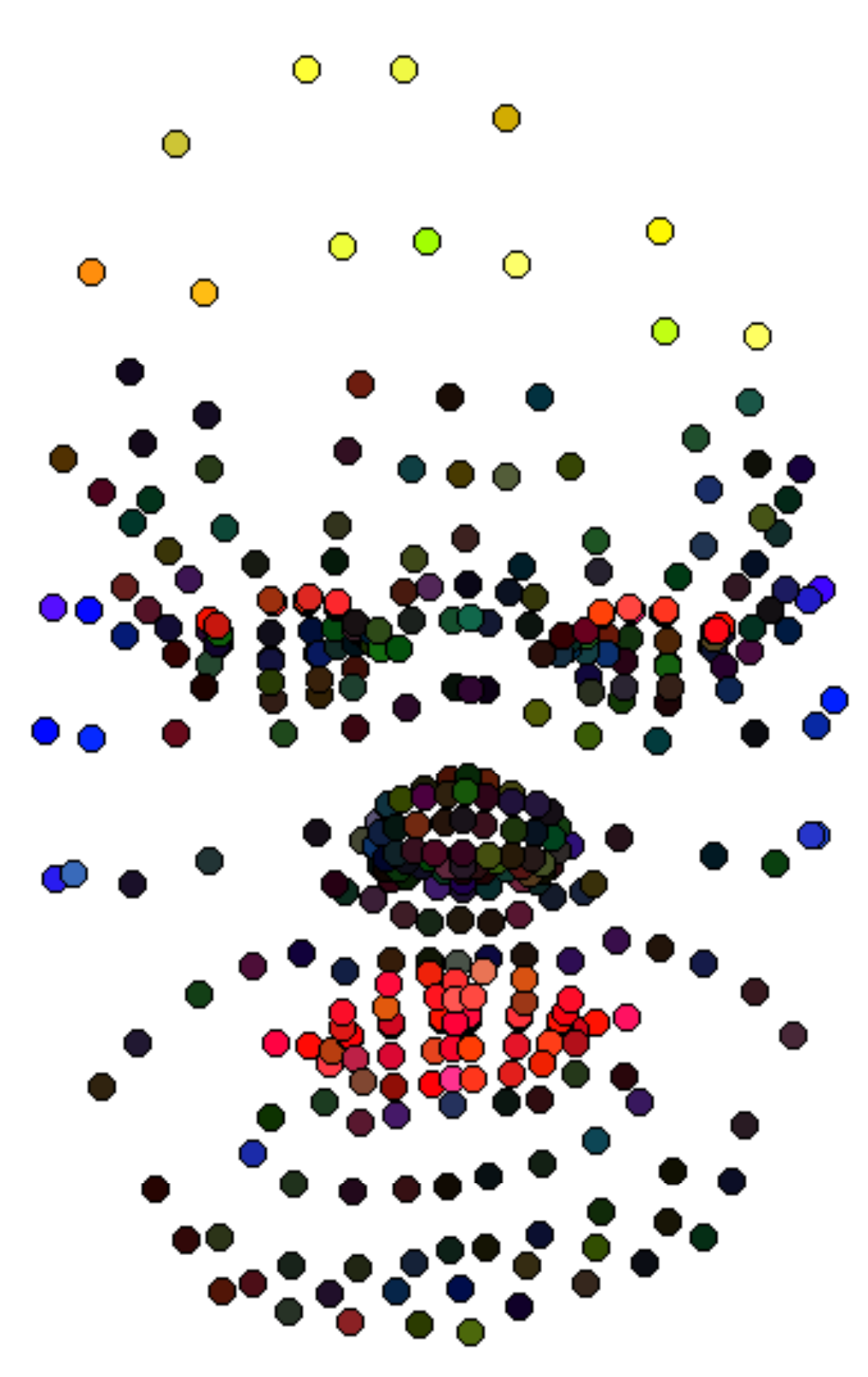}

 \caption{Noisy point-clouds of the fish and face corresponding to $Y$ the moving data. From left to right, the data without noise, 20, 15, 10 and 5 dB of SNR.}
 \label{fig:ccpd:noise:model}
\end{figure}

\msc{In this section we evaluate the effect of noise and outliers in the color space on the non-rigid registration with the proposed CCPD. The experimentation is carried out using the fish and face data used in the previous experiments.

The first experiment considers the noise in the color space adding random gaussian noise to each R, G, and B component in 4 different levels of Signal/Noise ratio (SNR): 20, 15, 10 and 5 dB (see Fig. \ref{fig:ccpd:noise:model}). Initially, the experiment analyses the effect of choosing suitable parameters for CCPD to compensate for the color noise using the fish data. Later, using the face data, the parameters are fixed to those giving the best CCPD performance in the experiments carried out in Sect. \ref{sec:ccpd:exp:face} in order to analyse the color noise effects and tolerance of the proposal against that noise. Since the noise is assigned randomly, 5 iterations per level of noise have been performed to calculate the averaged RMS as the registration error.}


\msc{The results for the initial color noise experiment  using the optimal set of parameters by experimentation are shown in Table \ref{tab:fish:colornoise}. As can be seen from the results, even with high levels of noise, the performance of the CCPD method remains high being hardly affected by the color noise (the order of the RMS is the same regardless the SNR).}

\begin{table}[h]
  \centering
  \caption{\msc{RMS registration error of fish shape with color noise. The Signal-to-Noise ratios are 20, 15, 10 and 5 dB.}}
    \begin{tabular}{cccc}
\hline
       20         & 15              & 10             & 5\\
\hline
     0.41080e-02  &  0.40603e-02    &  0.59438e-02   &  0.69721e-02\\
    
\hline
    \end{tabular}%
	
  \label{tab:fish:colornoise}%
\end{table}%

\begin{figure}
  \centering%
\includegraphics[width = 0.9\textwidth]{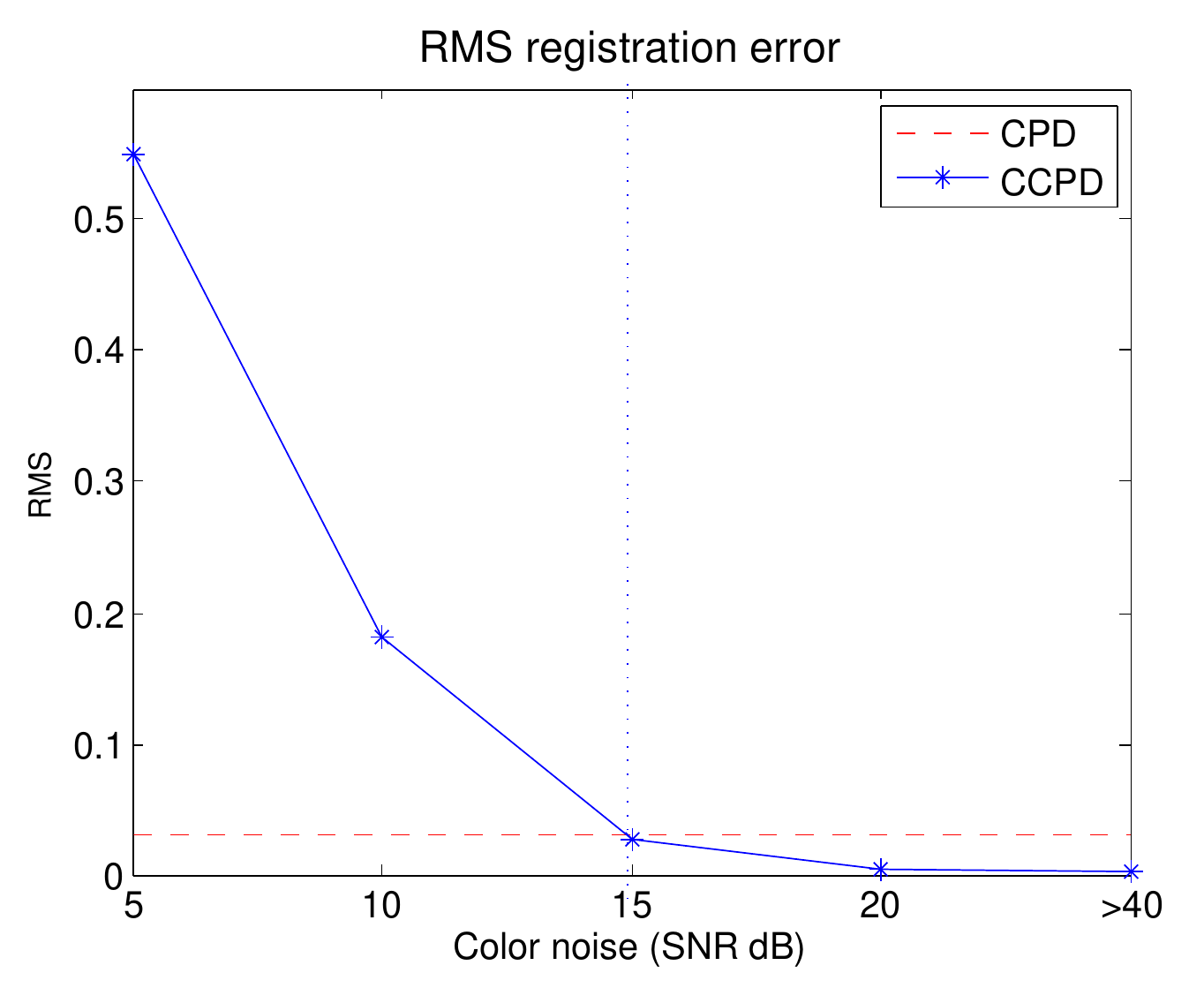} 

 \caption{\mscc{Average CCPD RMS registration error for 50 face deformations with 20, 15, 10 and 5 dB of color Signal-To-Noise ratio.}}
 \label{fig:face:50:colornoise}
\end{figure}

\msc{The results for the second color noise experiment (using the optimal set of parameters obtained for the CCPD without noise presented in Sect. \ref{sec:ccpd:exp:face}) are shown in Table \ref{tab:face:colornoise}. The data includes noise in the same four tests shown in Table \ref{tab:face} and Figure \ref{fig:faceimages}. For Tests 2 to 4 with 15 dB of SNR the error of CCPD is similar to CPD and decreases as SNR gets worse. In the case of Test 1, the performance is lower than CPD but remains similar for every level of noise due to the outliers are modelled with the original Eq. of outliers $o_L$ from the CPD. If we assign high $\sigma_C$ or low $w_C$, we will have the results similar to CPD. Furthermore, Figure \ref{fig:face:50:colornoise} shows the same results for the full set of 50 faces that is part of the original CPD synthetic dataset. This experiment shows similar results as the previous one, confirming the results in a large set of deformations. On average, the CCPD method outperforms the CPD results even for large color noise (about 15 dB).}

\begin{table}[h]
  \centering
     \caption{\mscc{CCPD RMS registration error from the face shape tests with 20, 15, 10 and 5 dB Signal-To-Noise ratio. Columns CCPD and COD are from Table \ref{tab:face}. }} 
    \begin{tabular}{lrrrrrr}
    \hline
         & CCPD & CPD & 20 & 15 & 10 & 5 \\
         & no noise & no noise & & & & \\
    \hline

    
    Test1	& 0.0037   & 0.0162    & 0.2176	& 0.2141	& 0.2150    & 0.2159\\
    Test2	& 0.0029   & 0.0408    & 0.0308	& 0.0617	& 0.1823	& 0.4560\\
    Test3	& 0.0022   & 0.1205    & 0.0332	& 0.0830    & 0.2606	& 0.4304\\
    Test4	& 0.0060   & 0.1041    & 0.1365	& 0.1372	& 0.1483	& 0.2092\\
    \hline
    \end{tabular}%

  \label{tab:face:colornoise}%
\end{table}%

%
%

\msc{Finally, the effect of outliers in the color data is evaluated. In this case, the outliers are in the color space, hence to generate them we have chosen the color that is the furthest to the rest of colors, which in this case is white. We have randomly generated, over the data set, four percentages of outliers: 5\%, 25\%, 50\% and 75\%. The results are presented in Table \ref{tab:face:coloroutliers}.}

\begin{table}[h]
  \centering
     \caption{\msc{RMS registration error of face shape for four levels of color outliers, 5\%, 25\%, 50\% and 75\% compared to the CCPD without noise (BL: Baseline).}} 
    \begin{tabular}{lrrrr}
    \hline
          BL         & 5\%        & 25\%        & 50\%        & 75\% \\
    \hline
        	  0.0029    & 0.0222    & 0.0783     & 0.1311     & 0.2579  \\
    \hline
    \end{tabular}%

  \label{tab:face:coloroutliers}%
\end{table}%

\subsection{Synthetic realistic experiments}\label{sec:ccpd:exp:blender}
In this section, we present the experiments to evaluate the method for non-rigid registration using realistic shapes. The dataset includes two different objects: a flower\footnote{\url{https://www.turbosquid.com/3d-models/pink-primrose-flowering-3d-obj/516226} (last access: 11/08/2017)} and a face\footnote{\url{http://eat3d.com/forum/art-gallery/models-face} (last access: 11/08/2017)}. The synthetic models have been acquired using the Blensor tool \cite{Blensor2011}, a Blender plugin which simulates a Microsoft Kinect RGB-D sensor. \msc{This tool uses raytracing to simulate 3D sensors, that in this particular case is an RGB-D, providing a PCD file with all the spatial coordinates of the points and the color information. The virtual sensor is oriented in the direction as it would be done with a real one. The only preprocessing is to deform the models using the Blender tools, to have in this case three shapes, original, small deformation and large deformation.}

\begin{figure}
  \centering

\begin{tabular}{c c c}
	\includegraphics[width = 0.3\textwidth]{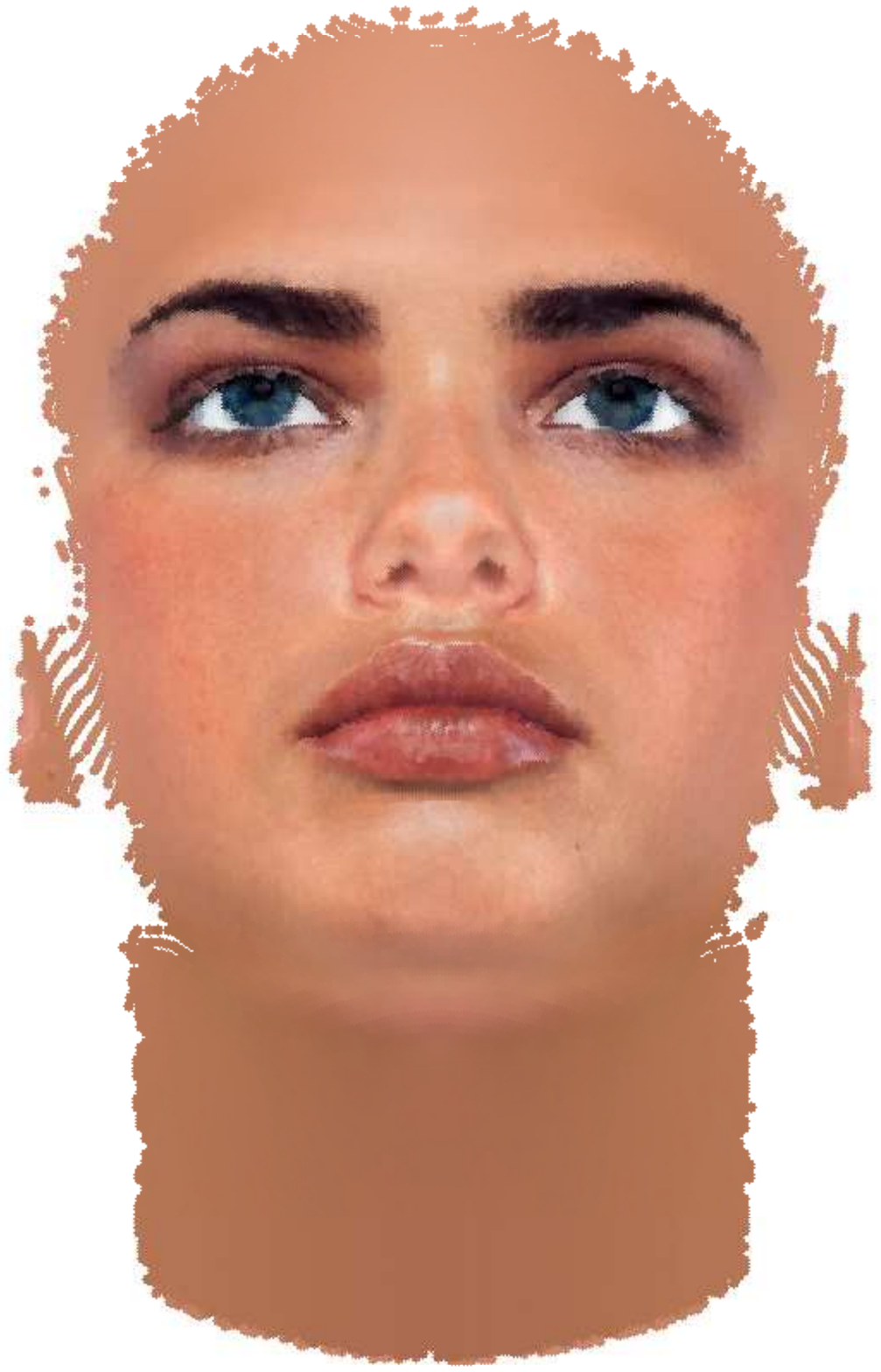} & \includegraphics[width = 0.3\textwidth]{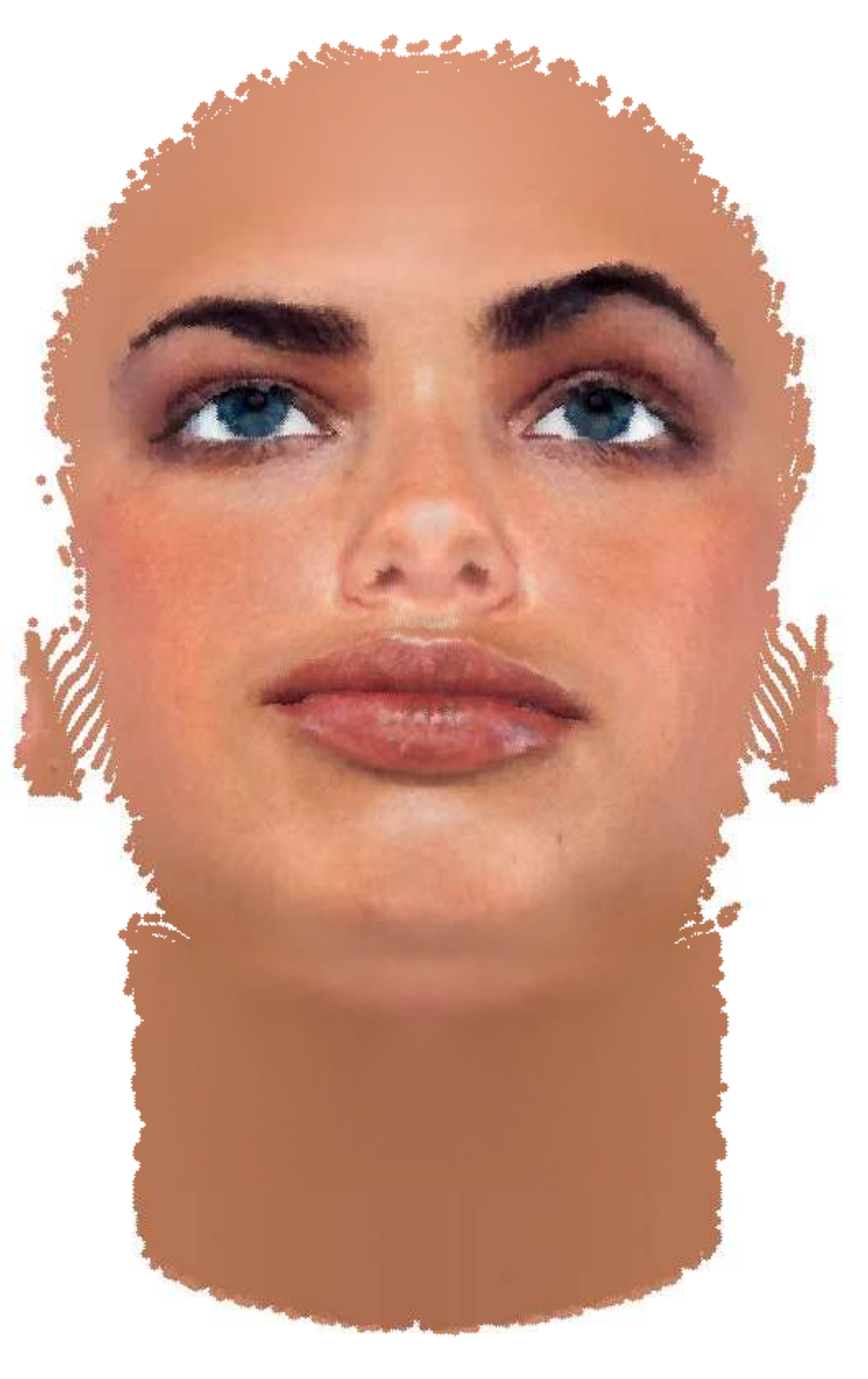} & \includegraphics[width = 0.3\textwidth]{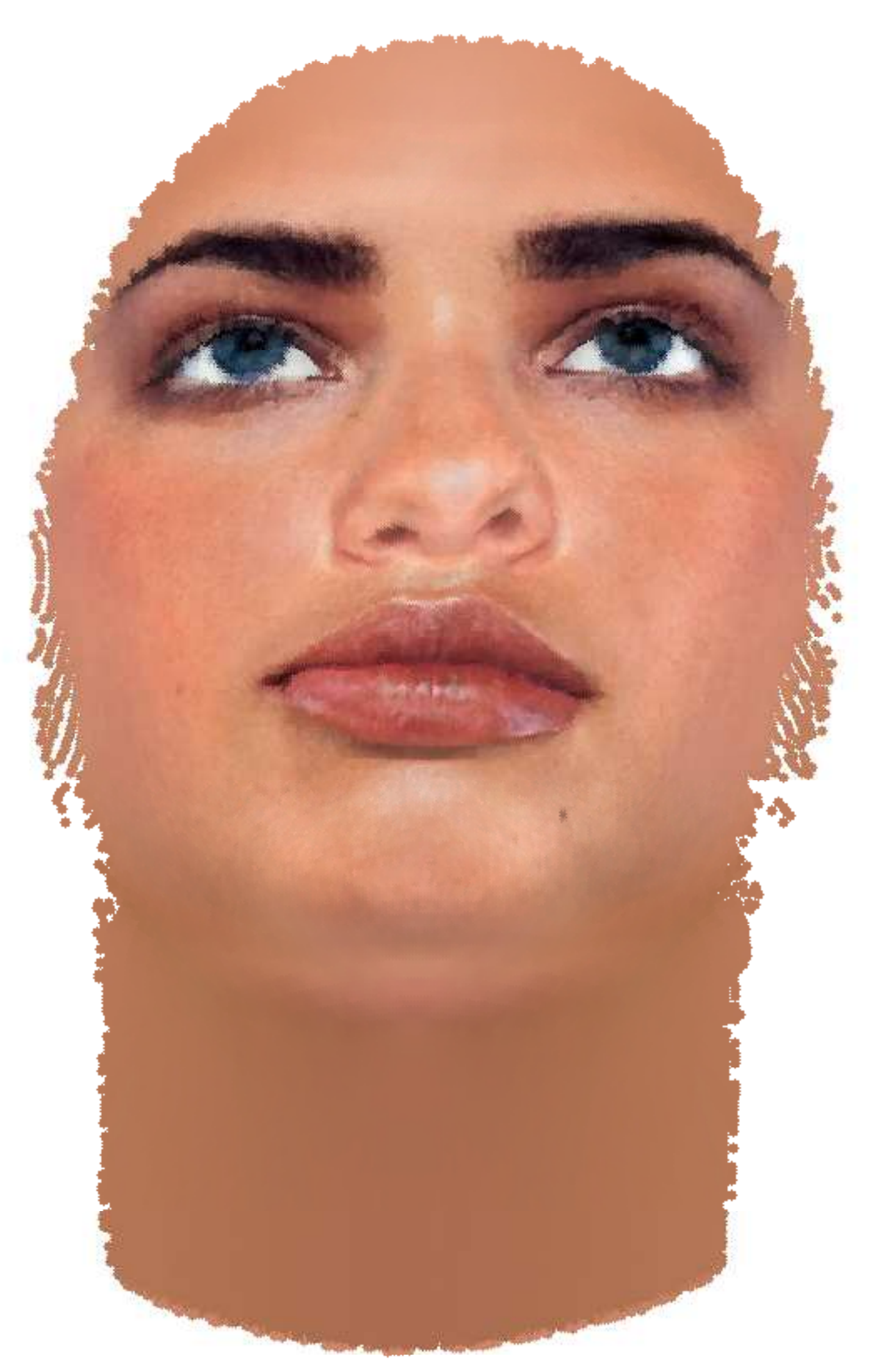}\\
	
	\includegraphics[width = 0.3\textwidth]{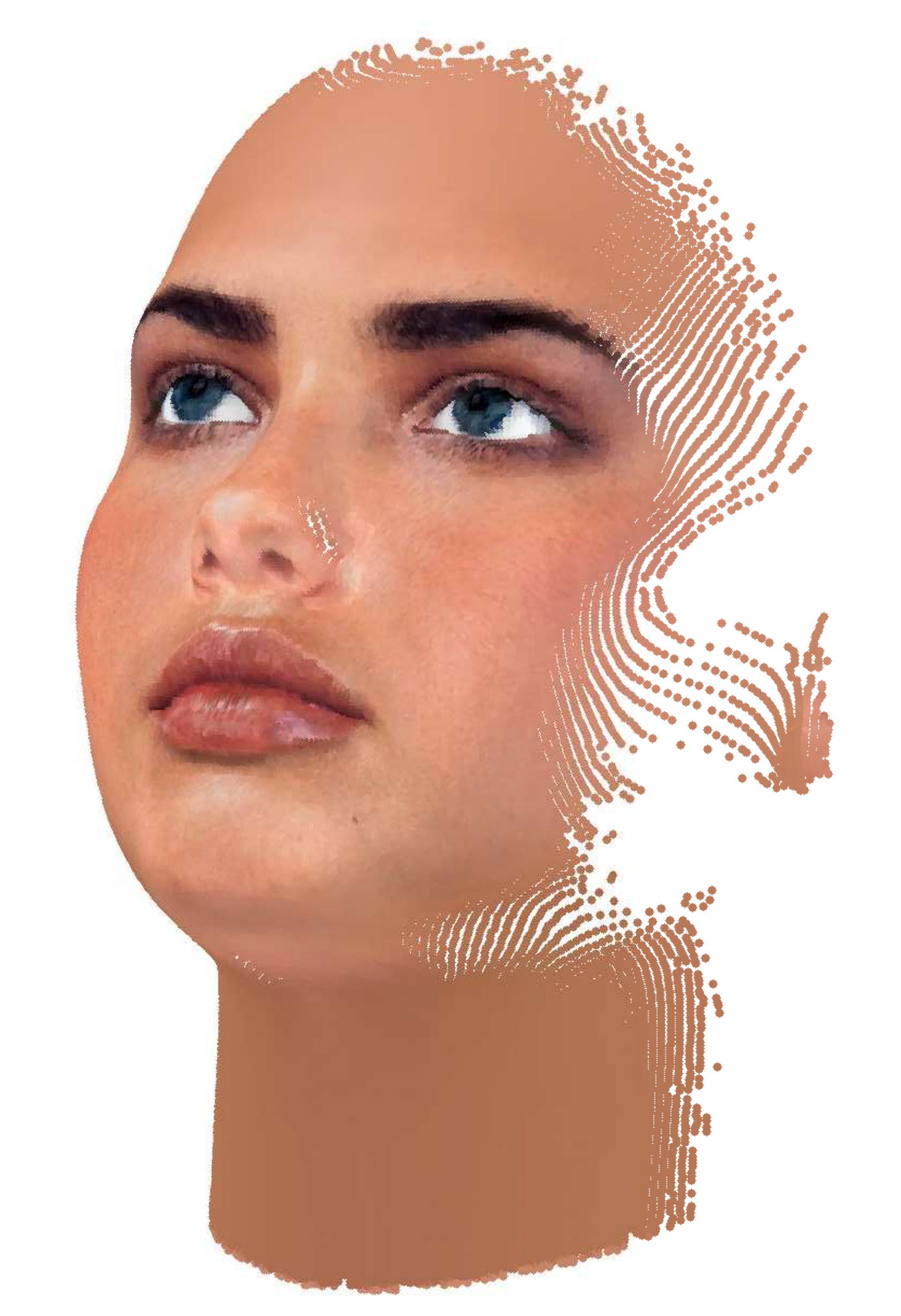} & \includegraphics[width = 0.3\textwidth]{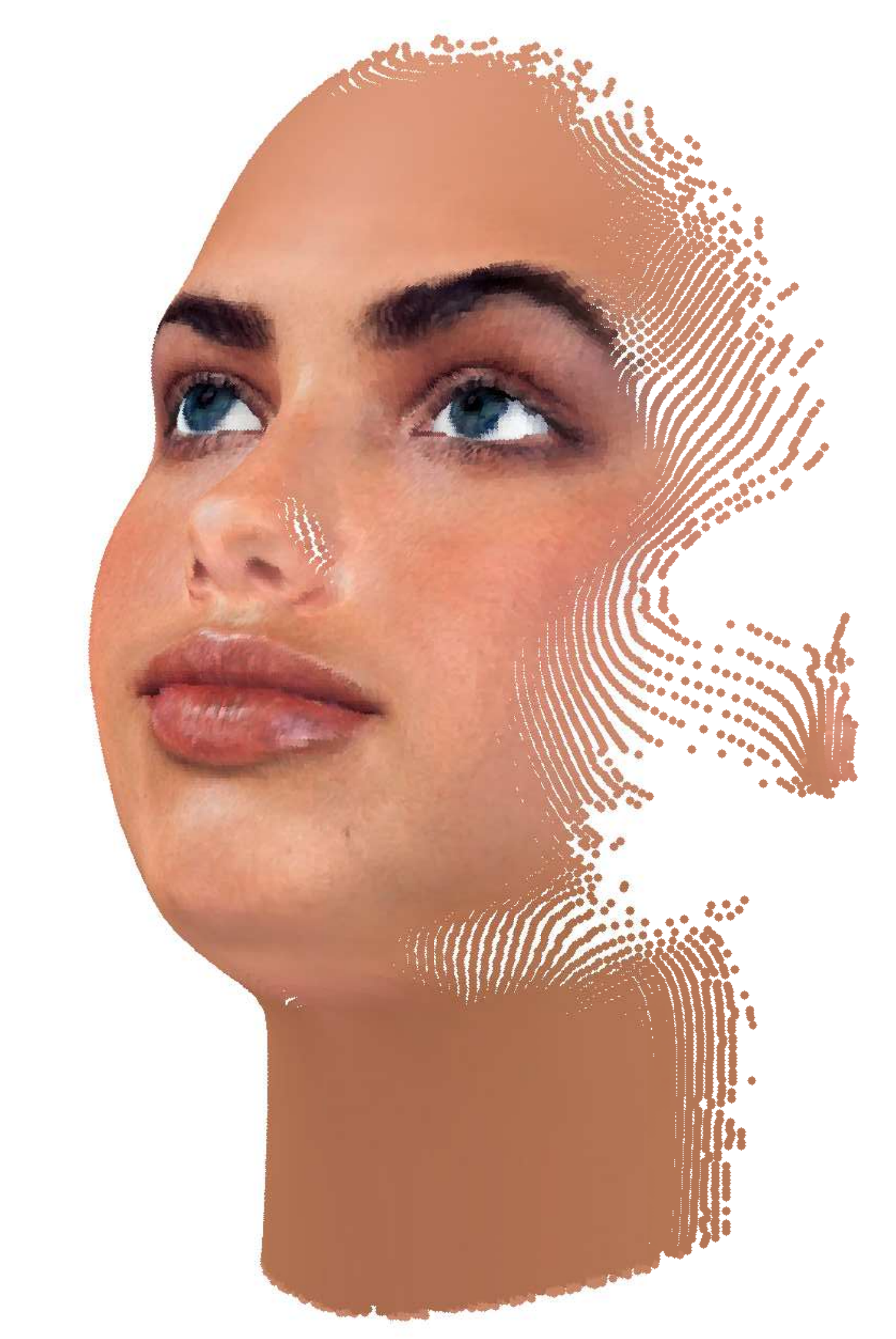} & \includegraphics[width = 0.3\textwidth]{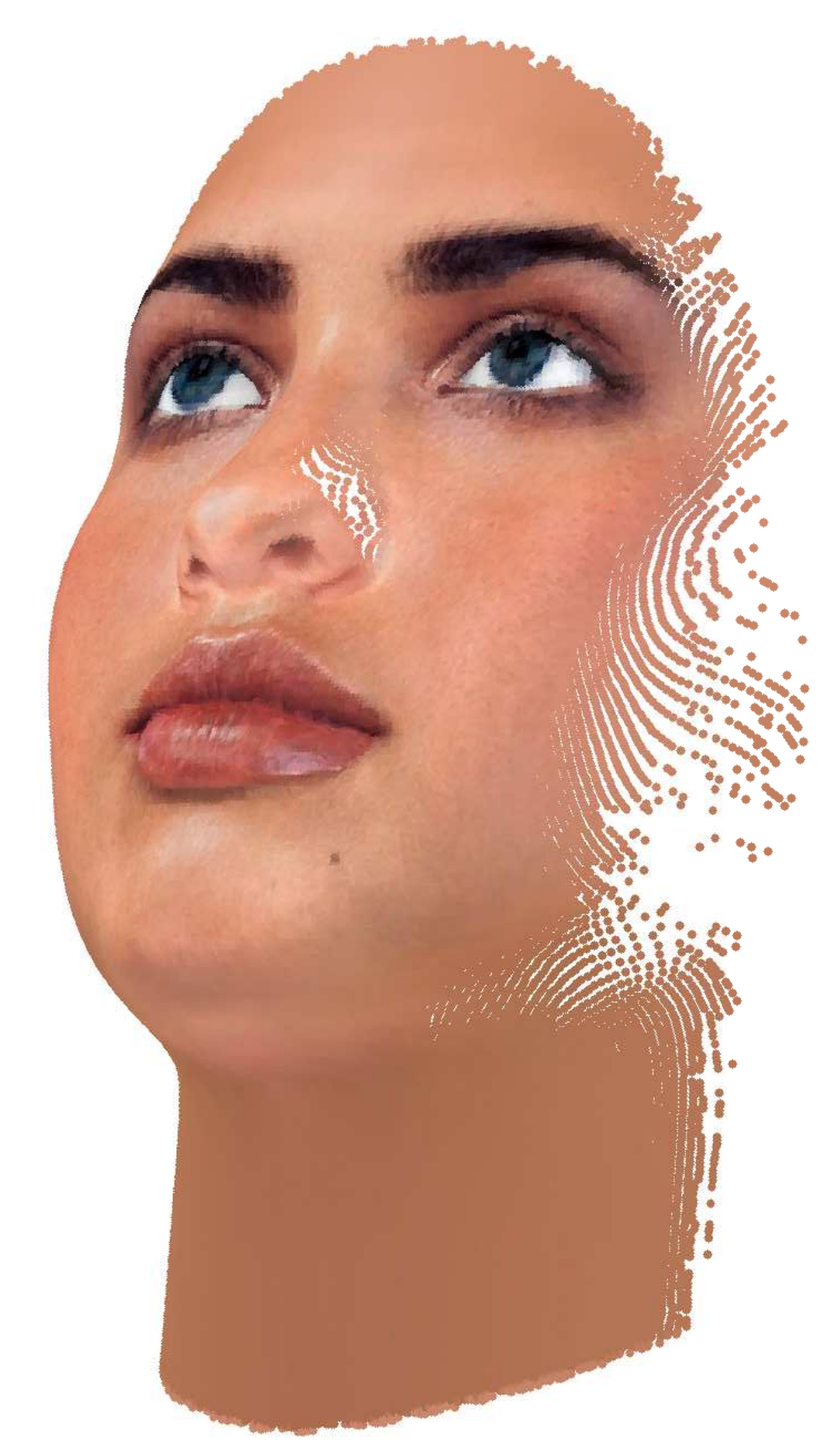}
\end{tabular}
	
	\caption[The face model used in the experiments]{\mscc{The face model used in the experiments. Two viewpoints (each per row) of the faces used. From left to right, the original face shape as target for the small and larger deformations in second and third columns.}}
  \label{fig:face:models}
\end{figure}

\begin{figure}
  \centering

\begin{tabular}{c c c}
	\includegraphics[width = 0.3\textwidth]{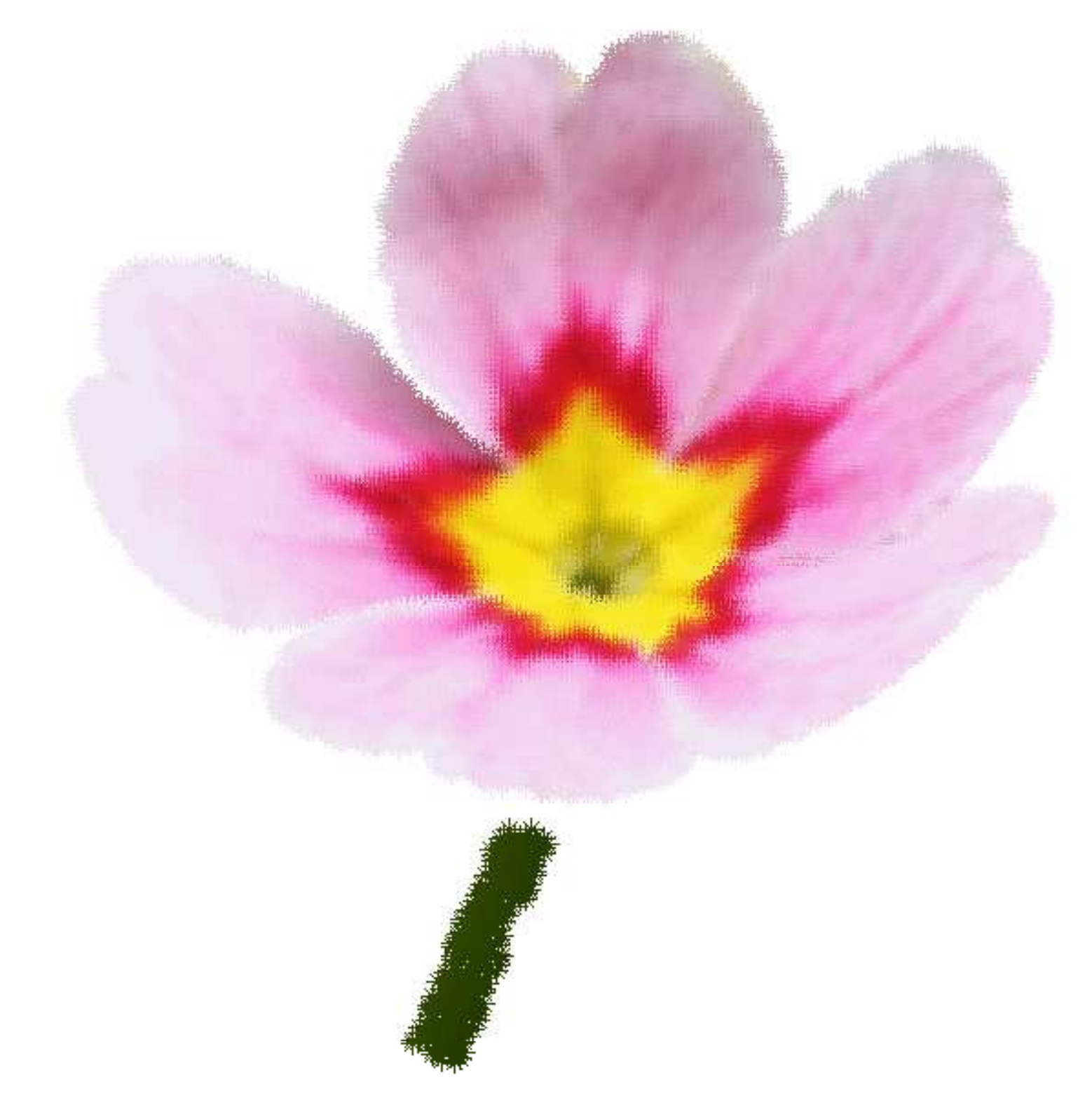} & \includegraphics[width = 0.3\textwidth]{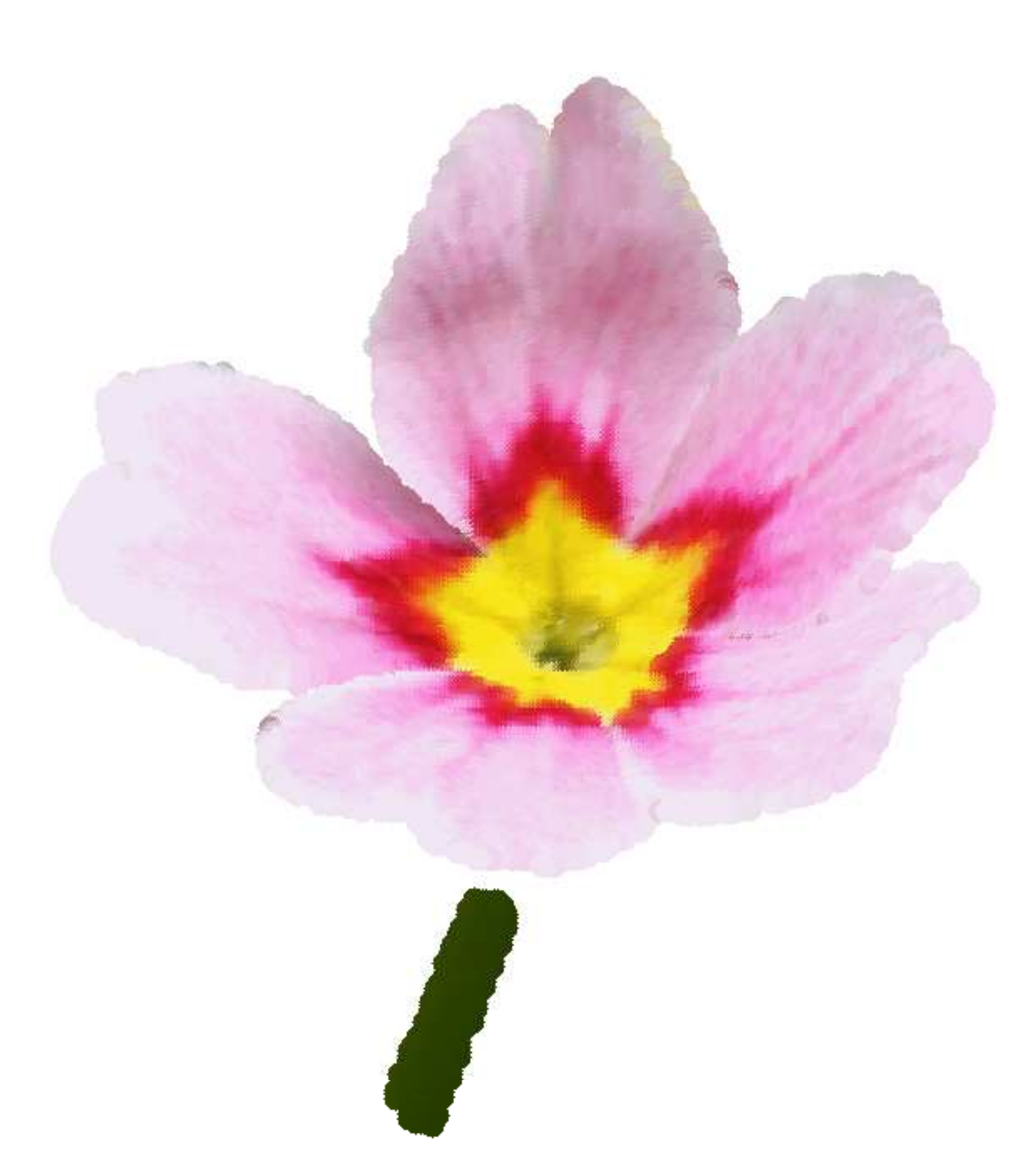} & \includegraphics[width = 0.3\textwidth]{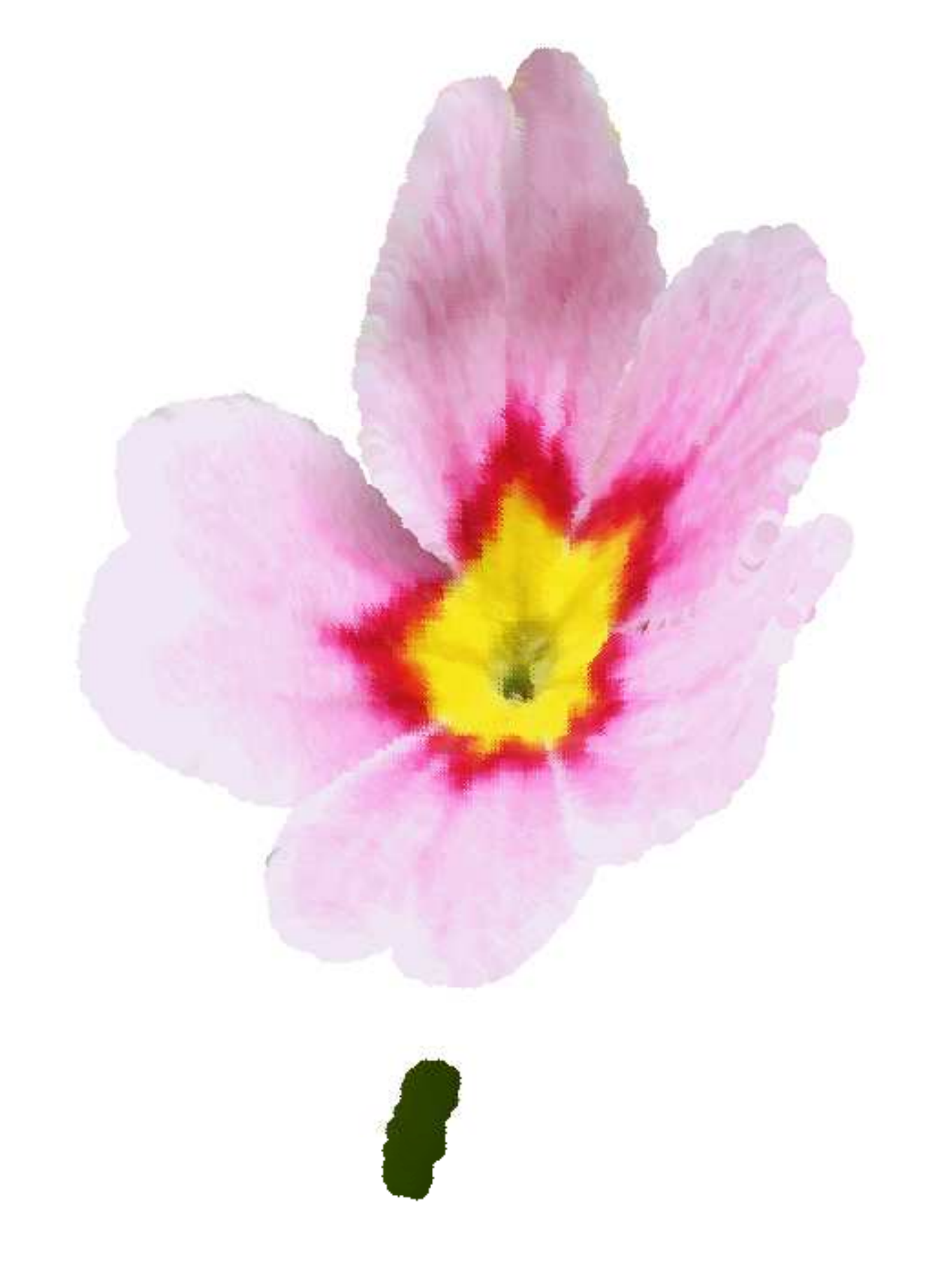}\\
	
	\includegraphics[width = 0.3\textwidth]{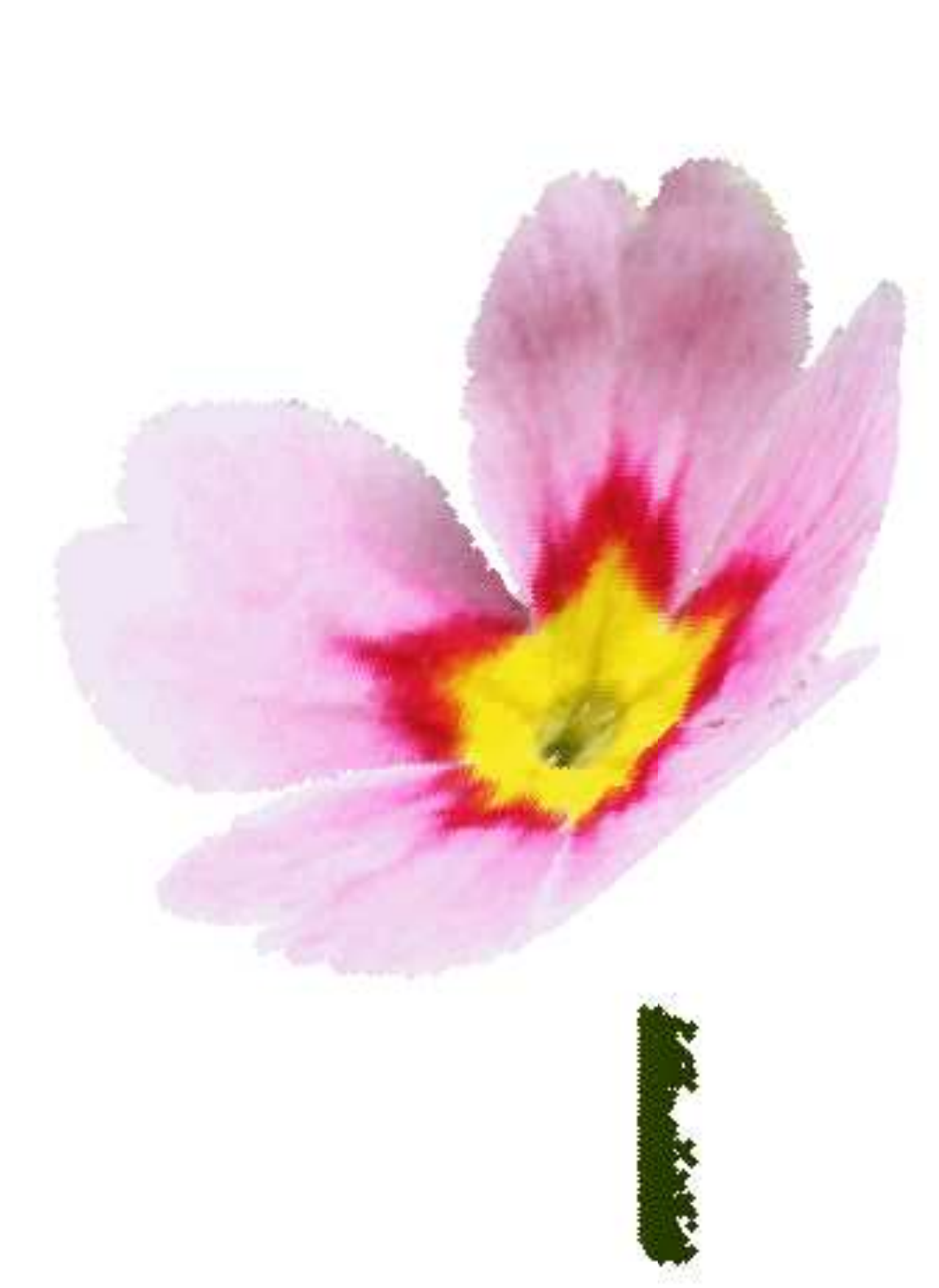} & \includegraphics[width = 0.3\textwidth]{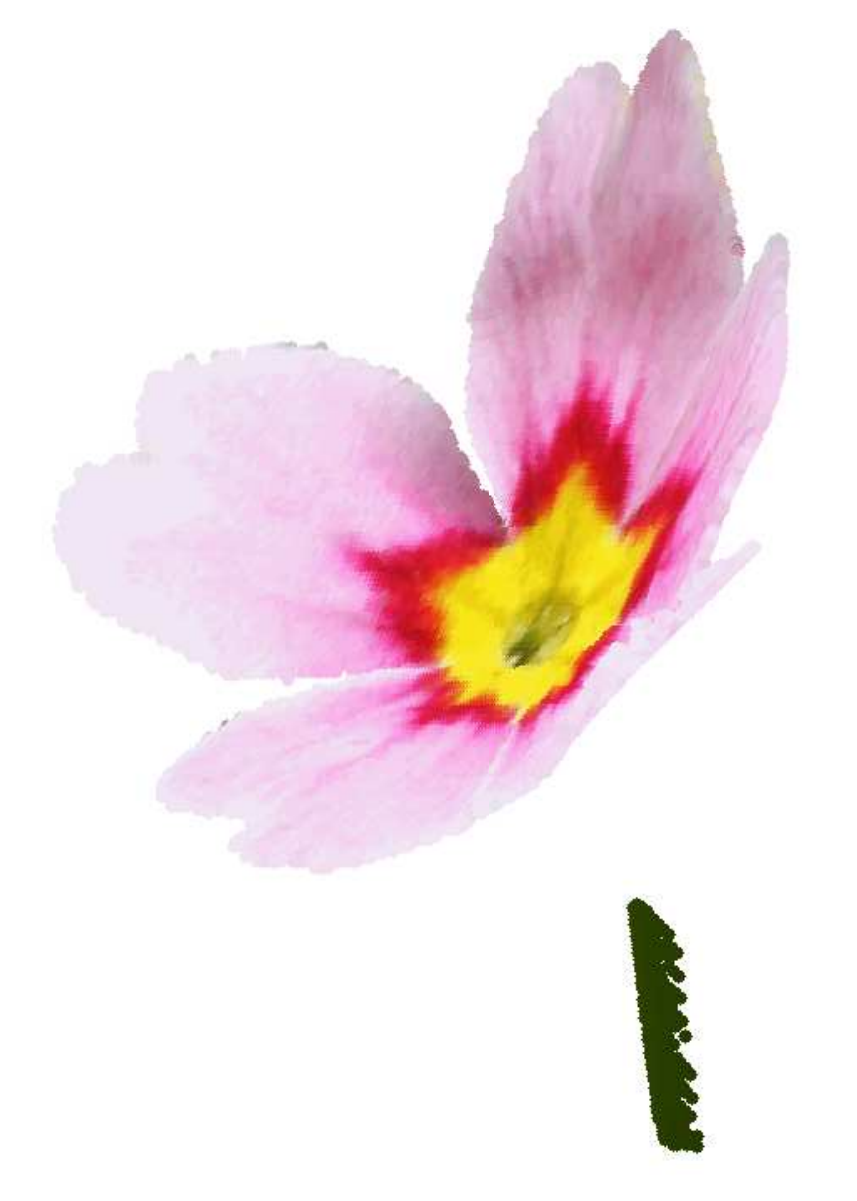} & \includegraphics[width = 0.3\textwidth]{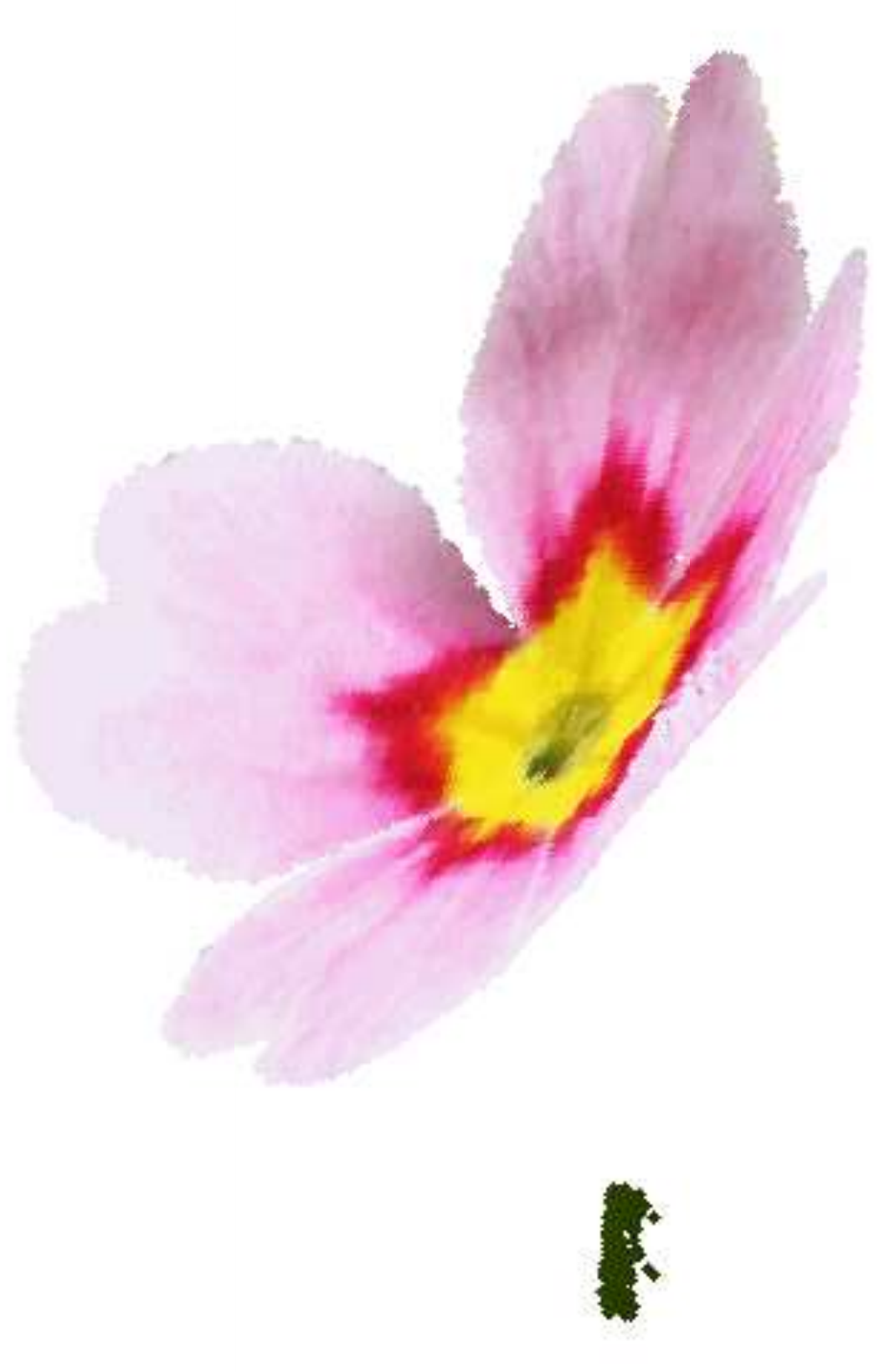}
	
\end{tabular}
	\caption[The flower model used in the experiments]{\mscc{The flower model used in the experiments. Two viewpoints (each per row) of the flower used. From left to right, the original flower shape as target for the small and larger deformations in second and third columns.}}
  \label{fig:plant:models}
\end{figure}

Figure~\ref{fig:face:models} and \ref{fig:plant:models} show the face and flower models used for the experiments. The images are from left to right: the target, a first deformation, and a second larger deformation. The face deformations could be seen as elastic deformations, because the face remains the same except for displacement of some parts. The first deformation is a eyebrow rise and a mouth change. The second moves both eyebrows and the mouth, changes the nose and the chin. For the flower, it could be seen as growth deformations due to the size of the object changes. The first deformation enlarges a little the leaves and the second is a larger deformation. 

\begin{figure}
\centering

 
\includegraphics[width=\textwidth]{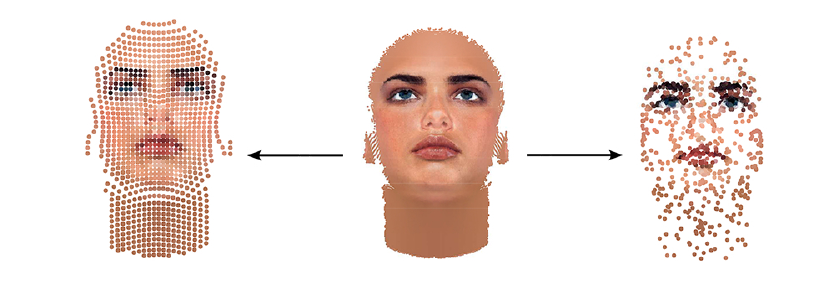}

 
\caption{Two sampling examples. The image in the middle represents a point set of a face shape. At the left is a uniform sampling. At the right side is a representation of a color-based sampling, which provide higher density of points at salient features, such as eyebrows, eyes or lips. }
\label{fig:ccpd:prop:functionUpsilonS}       
\end{figure}

In order to reduce and enhance the data for the registration purpose, we have used different downsampling techniques to sample the data. Figure \ref{fig:ccpd:prop:functionUpsilonS} shows two different kinds of sampling. The figure has in the middle the face example. At the left side a uniform sampling is presented, while at the right side a representation of a color-based sampling, which provide higher density of points at salient features, such as eyebrows, eyes or lips. In previous works, we have studied the use of downsampling as a method to enhance the quality of the data. These studies have been published in \cite{Saval-Calvo2015c} \cite{Saval-Calvo2016a}. In this paper we use the same methods, including: bilinear interpolation, normal-based sampling, color-based sampling, a combination of color and normal based technique, and GNG sampling approach proposed in \cite{Orts-Escolano2013}.

\subsubsection{Non-rigid registration evaluation}\label{exp:non-rigid}
Here we present a comparative evaluation of CCPD and CPD registration using synthetic realistic subjects. The color information, used by CCPD, allows the registration method to achieve good accuracy when the surface is not very detailed where the drift of points is not constrained by the irregularities of the shape.

\begin{figure}
 \centering
	 \includegraphics[width = 0.18\textwidth]{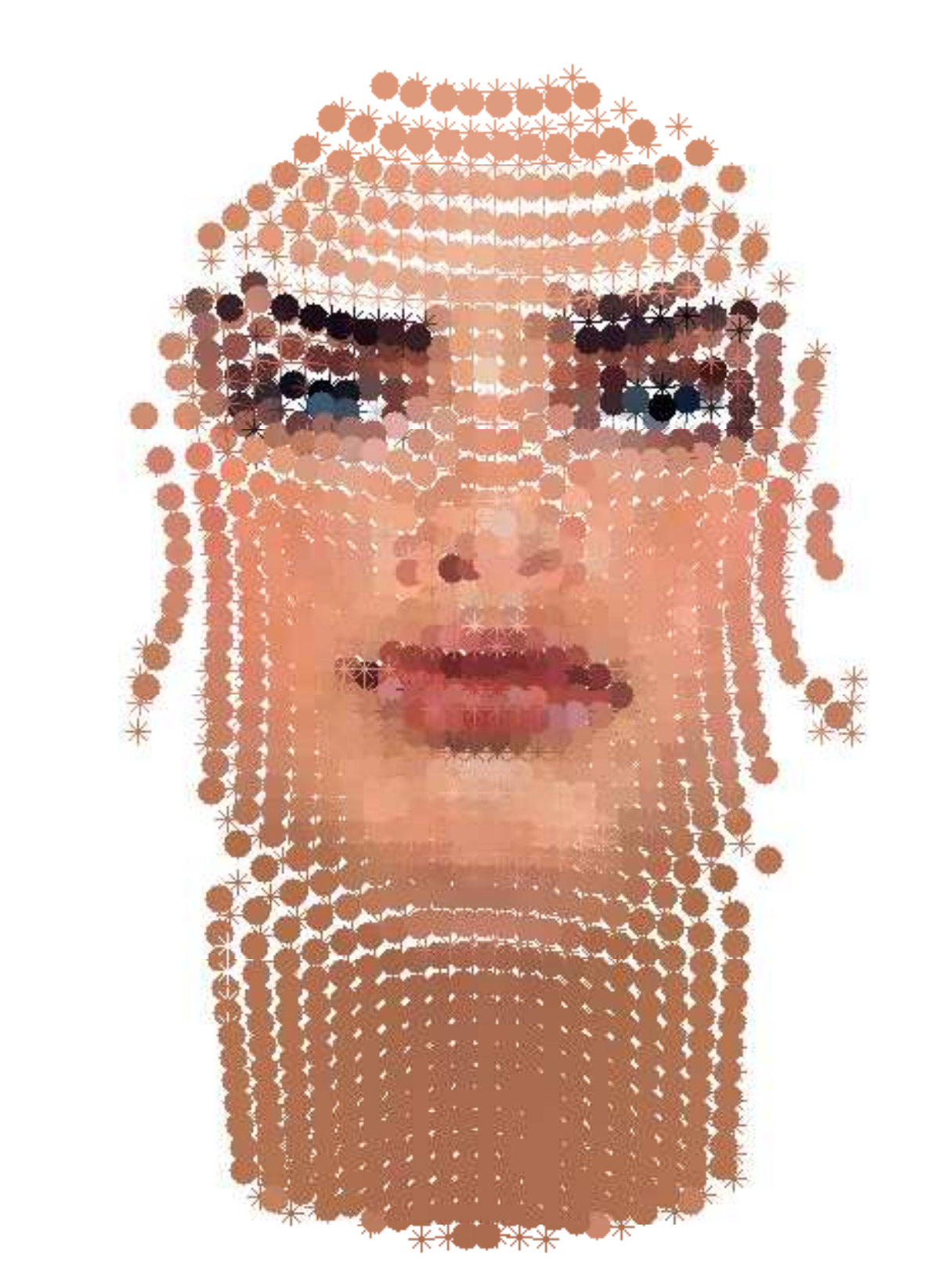}
	 \includegraphics[width = 0.18\textwidth]{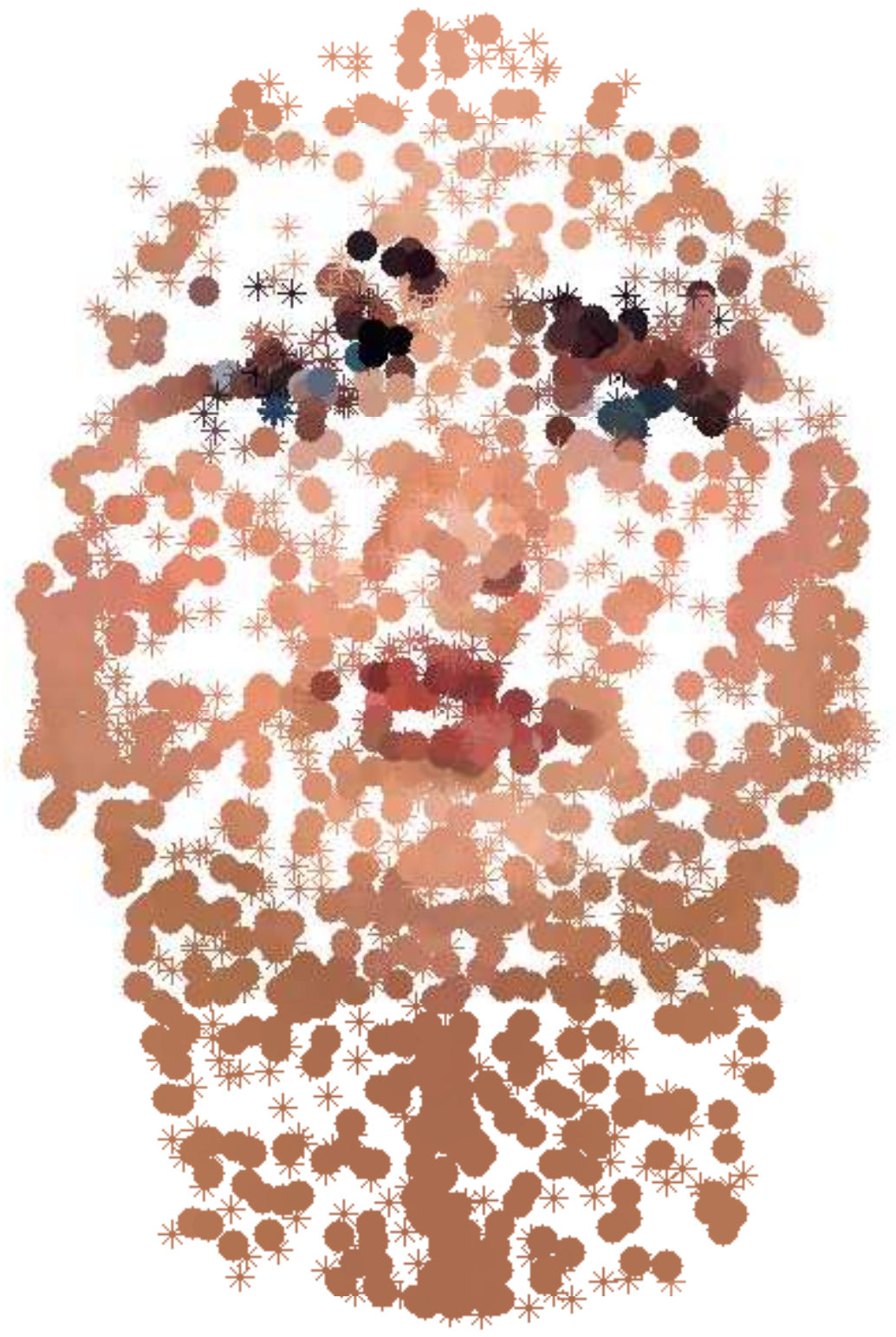}
	 \includegraphics[width = 0.18\textwidth]{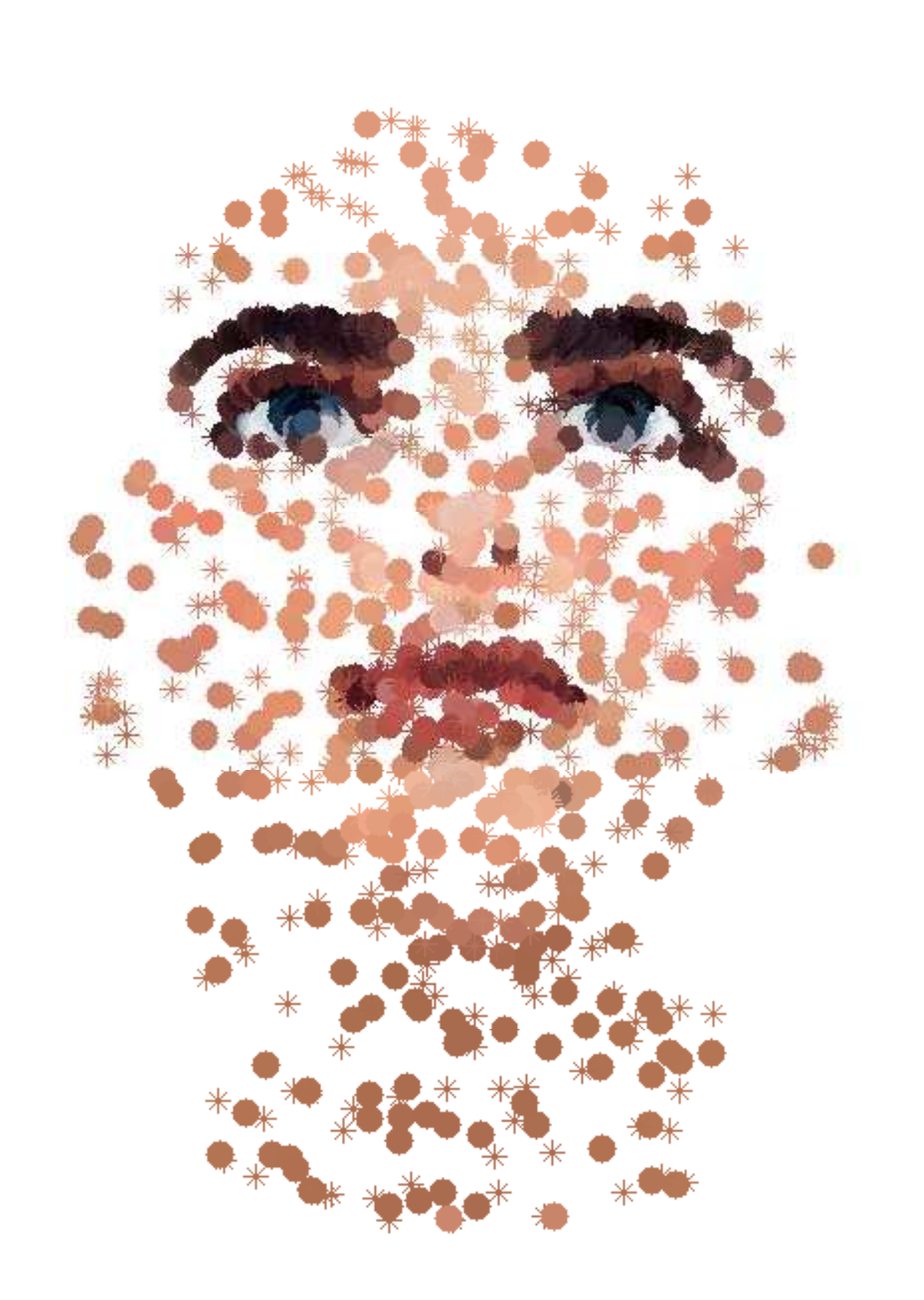}
	 \includegraphics[width = 0.18\textwidth]{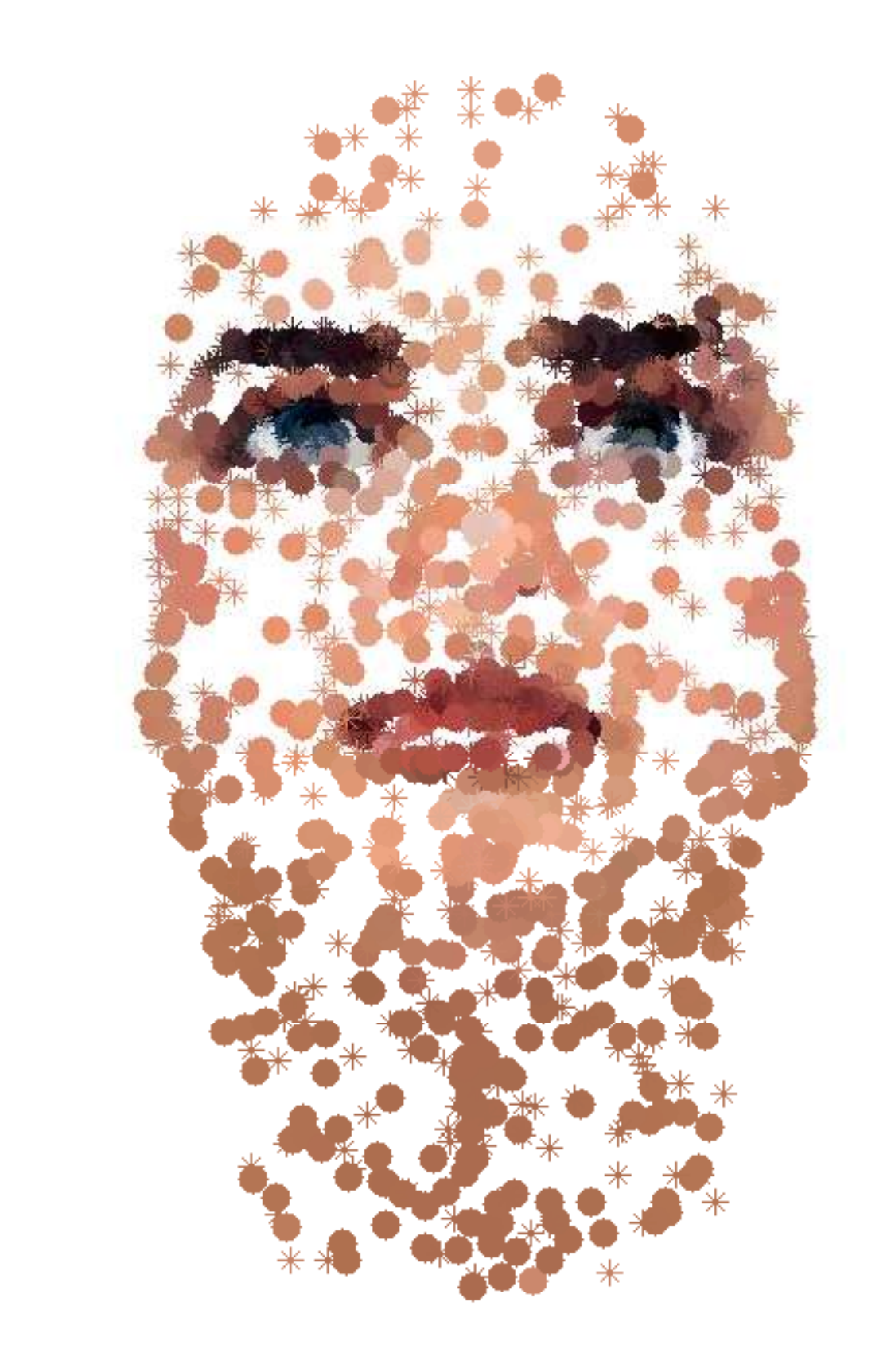}
	 \includegraphics[width = 0.18\textwidth]{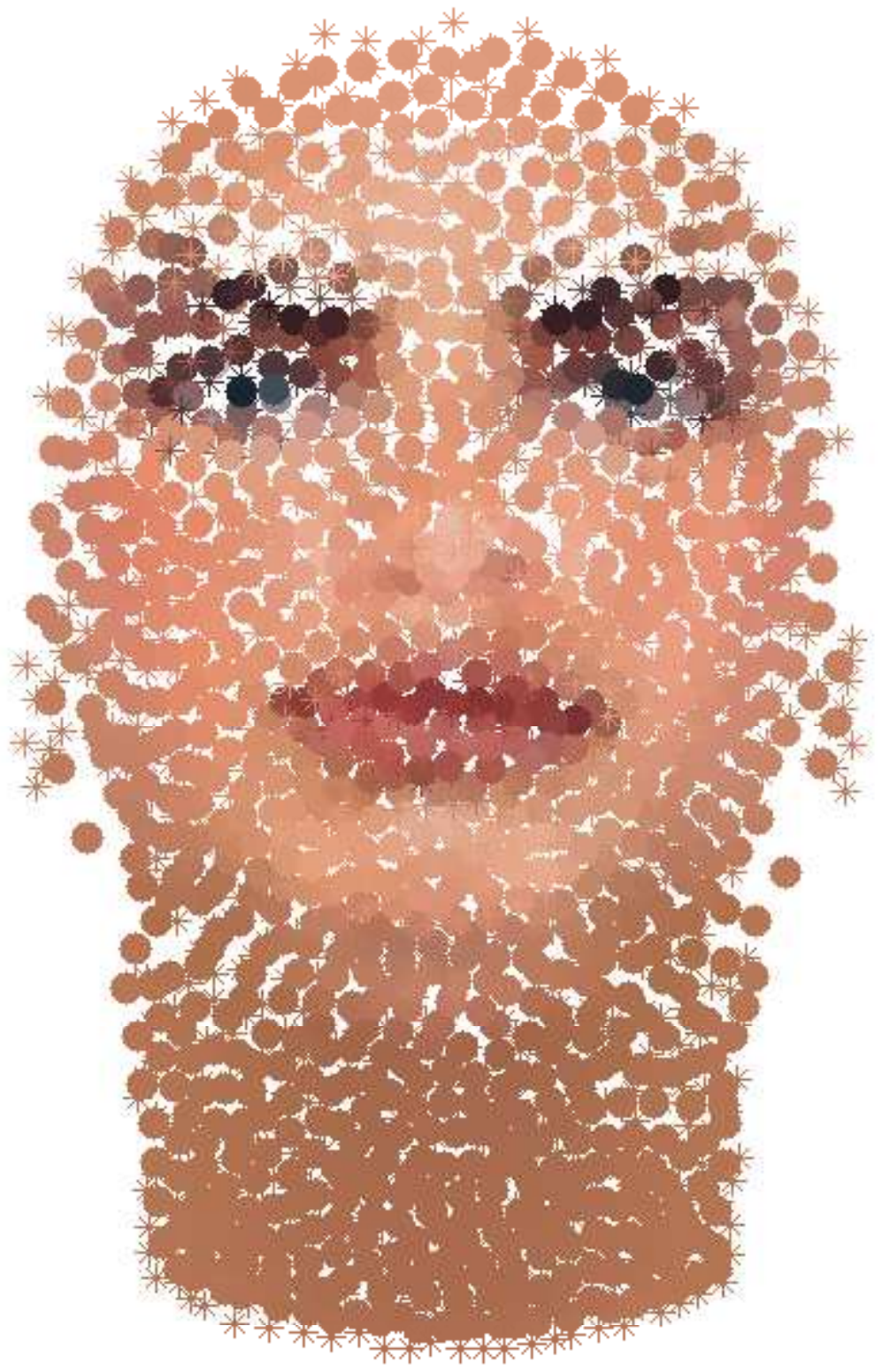}
\\
	 \includegraphics[width = 0.18\textwidth]{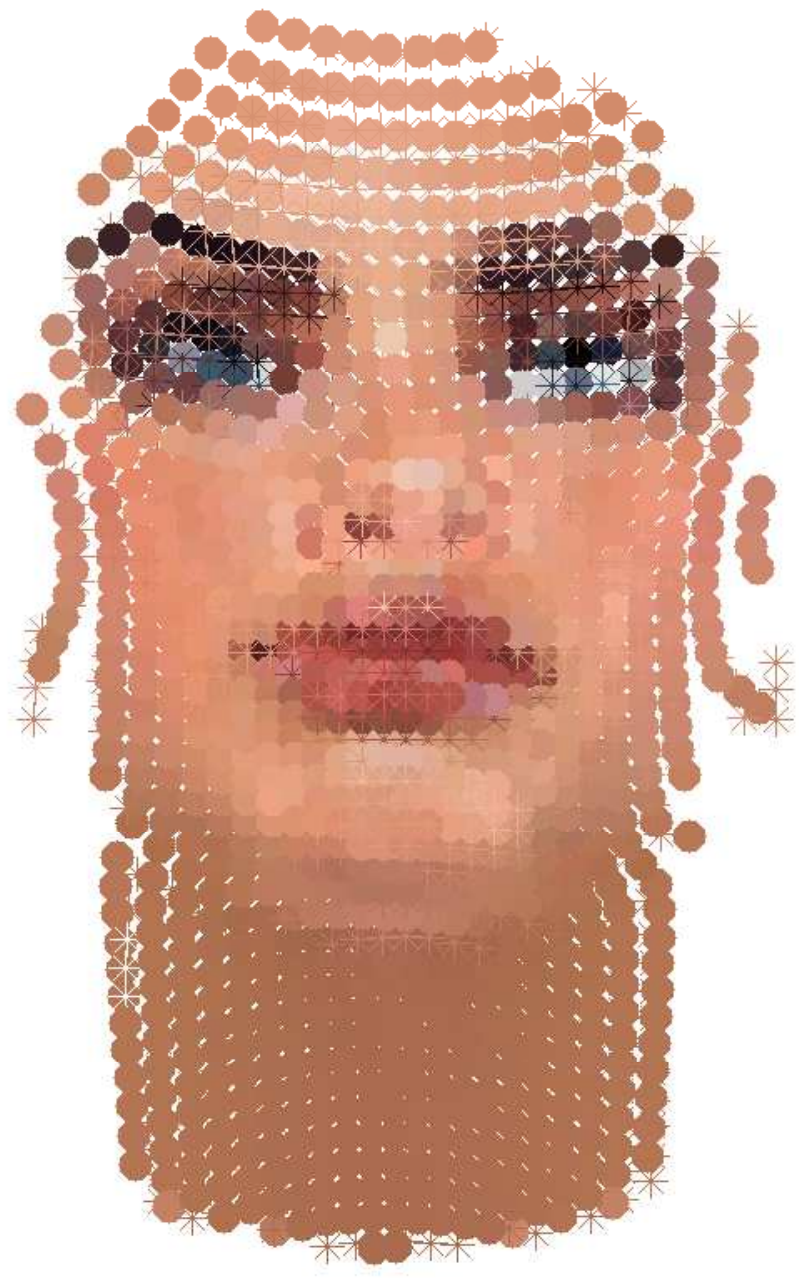}
	 \includegraphics[width = 0.18\textwidth]{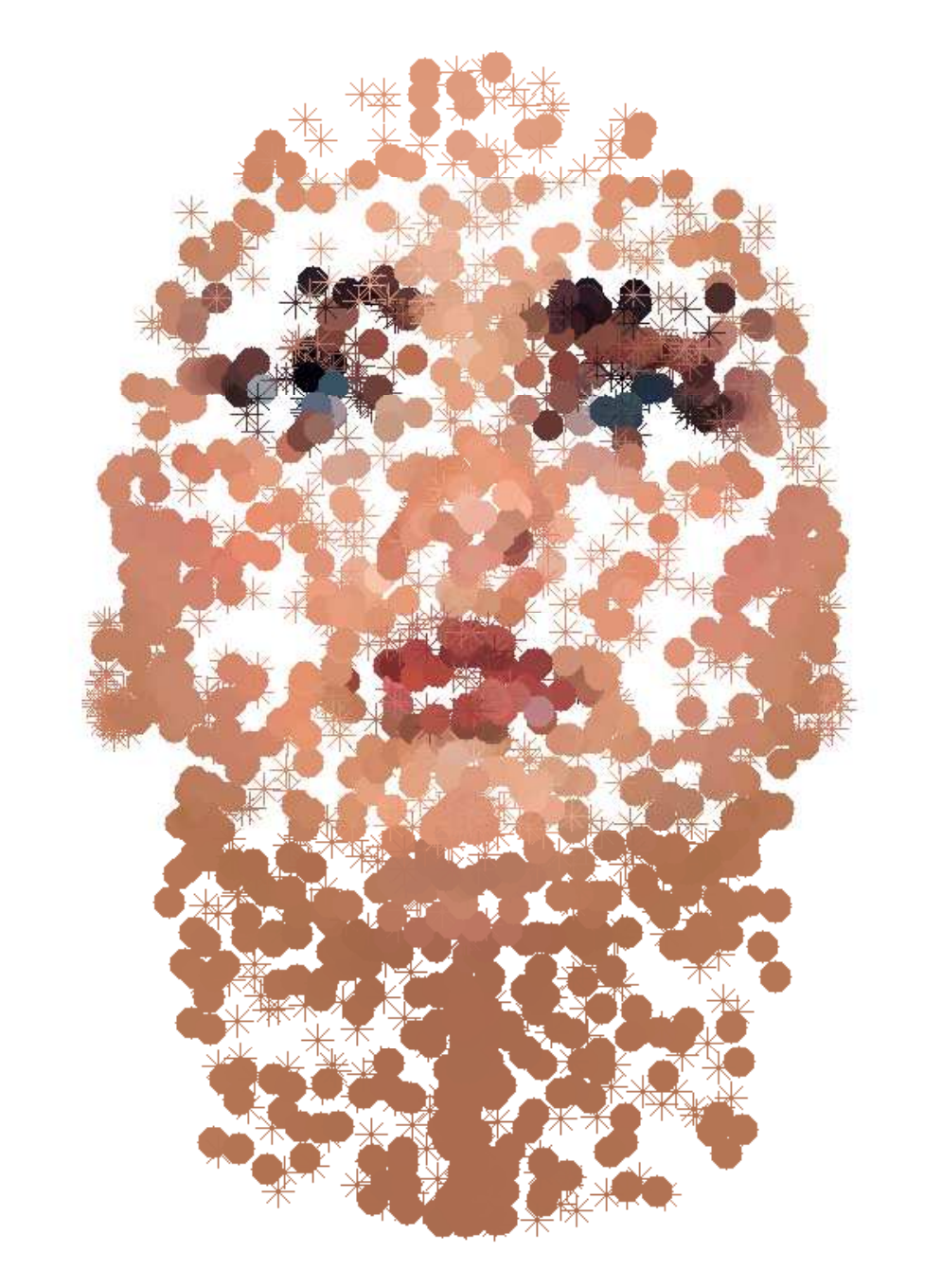}
	 \includegraphics[width = 0.18\textwidth]{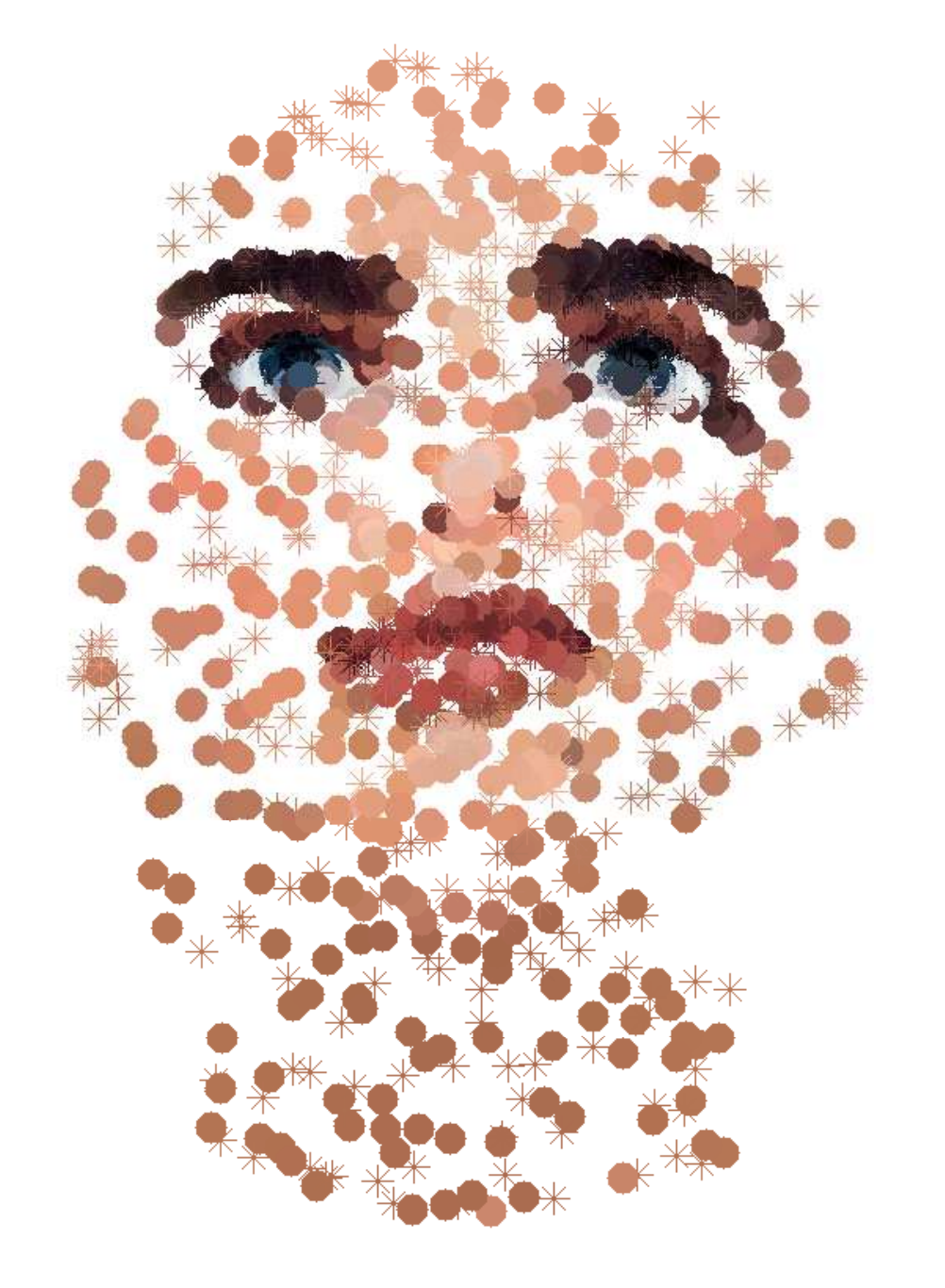}
	 \includegraphics[width = 0.18\textwidth]{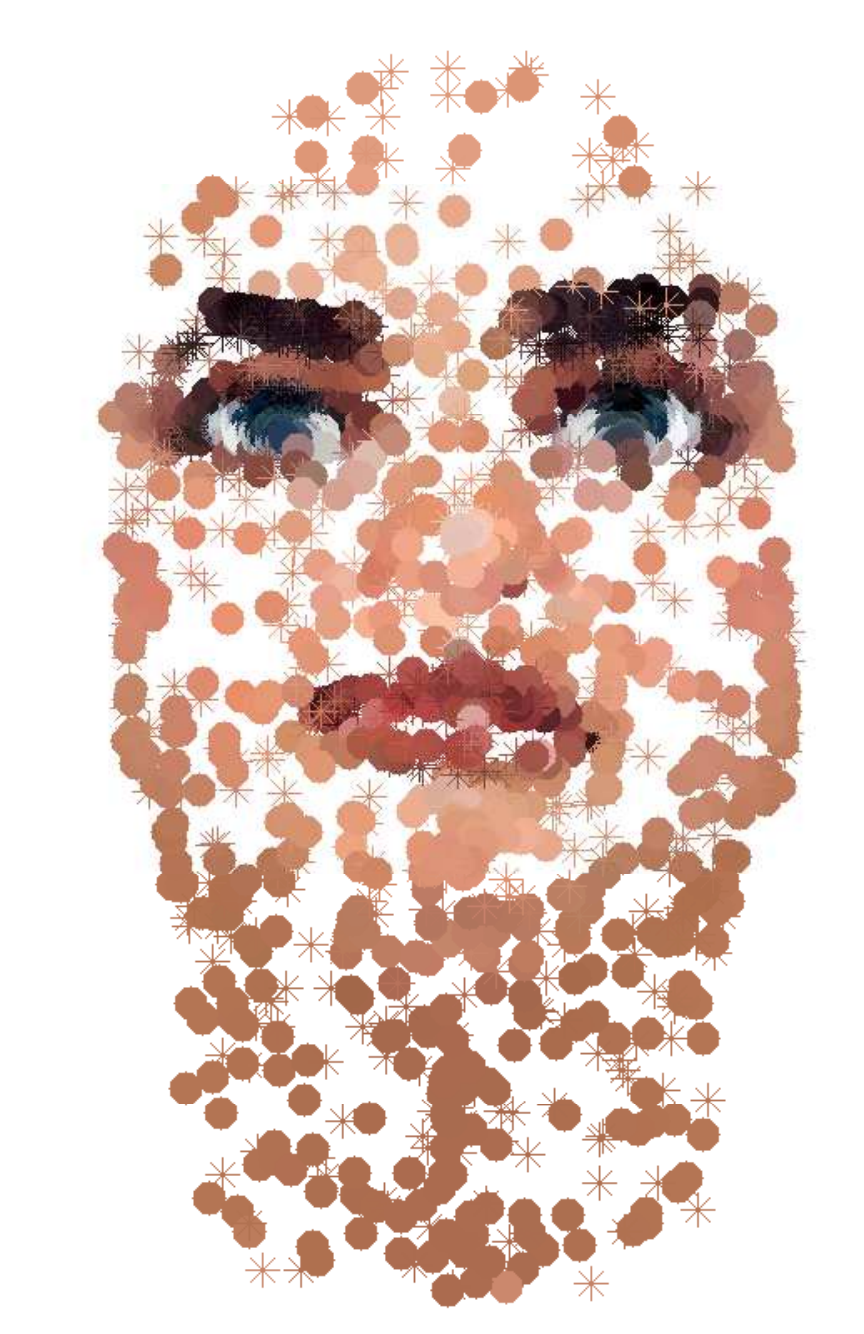}
	 \includegraphics[width = 0.18\textwidth]{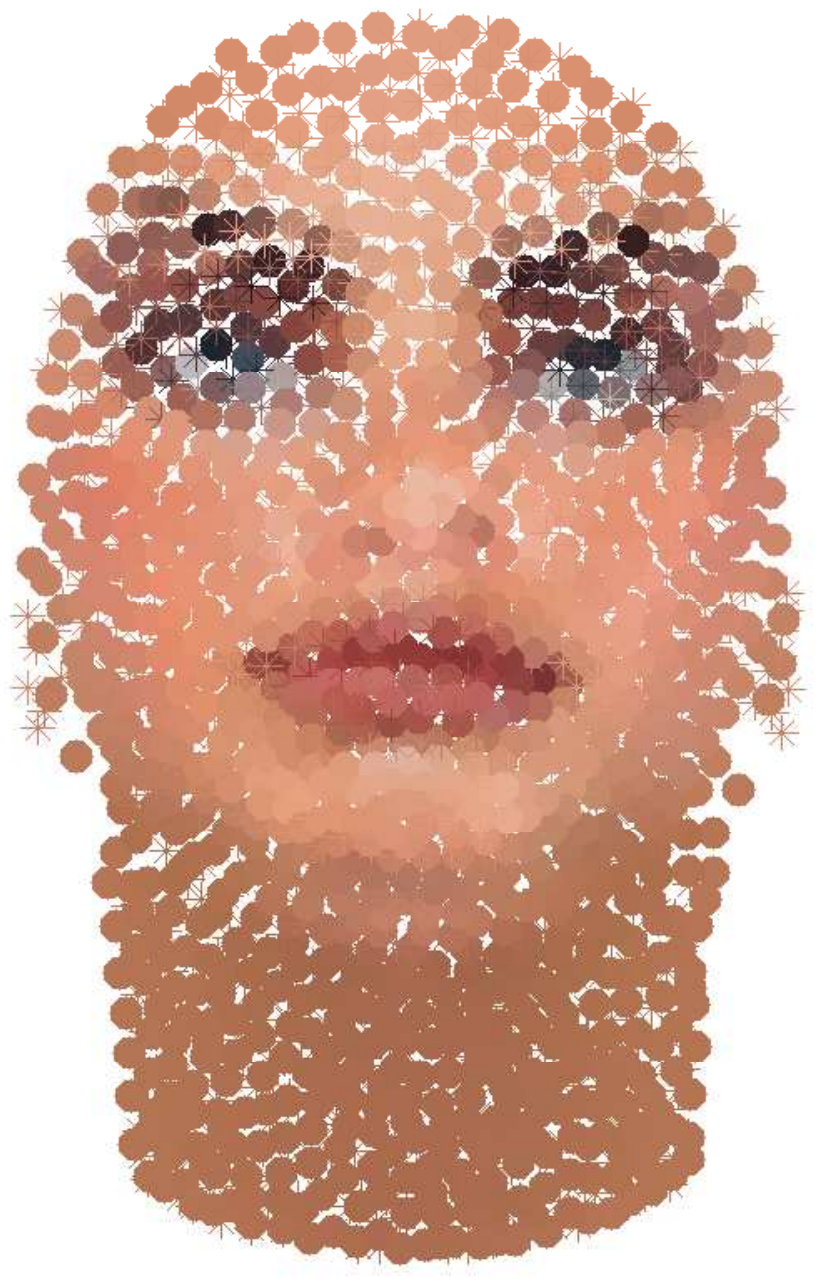}
\\
	\caption{Non-rigid registration result of face shape for a 1000 point sampling. The first row shows CCPD, and the original CPD in the second. Columns show different sampling algorithms that are from left to right, bilinear, normal-based, color-based, NC-based, GNG.}
  \label{fig:reg:face:1000}
\end{figure}

Using the data sampled, the non-rigid registration methods are qualitatively evaluated by visual inspection. Figure \ref{fig:reg:face:1000} shows the face shape for CCPD and the original CPD with 1000 points sampled with the different methods. More experiments have been performed with 250 and 500 data (which correspond to the experimentation in \cite{Saval-Calvo2015c} \cite{Saval-Calvo2016a}), but are not included as the results are similar to the presented experiment. 

\begin{figure}
 \centering
\scalebox{0.95}{
	 \includegraphics[width = 0.18\textwidth]{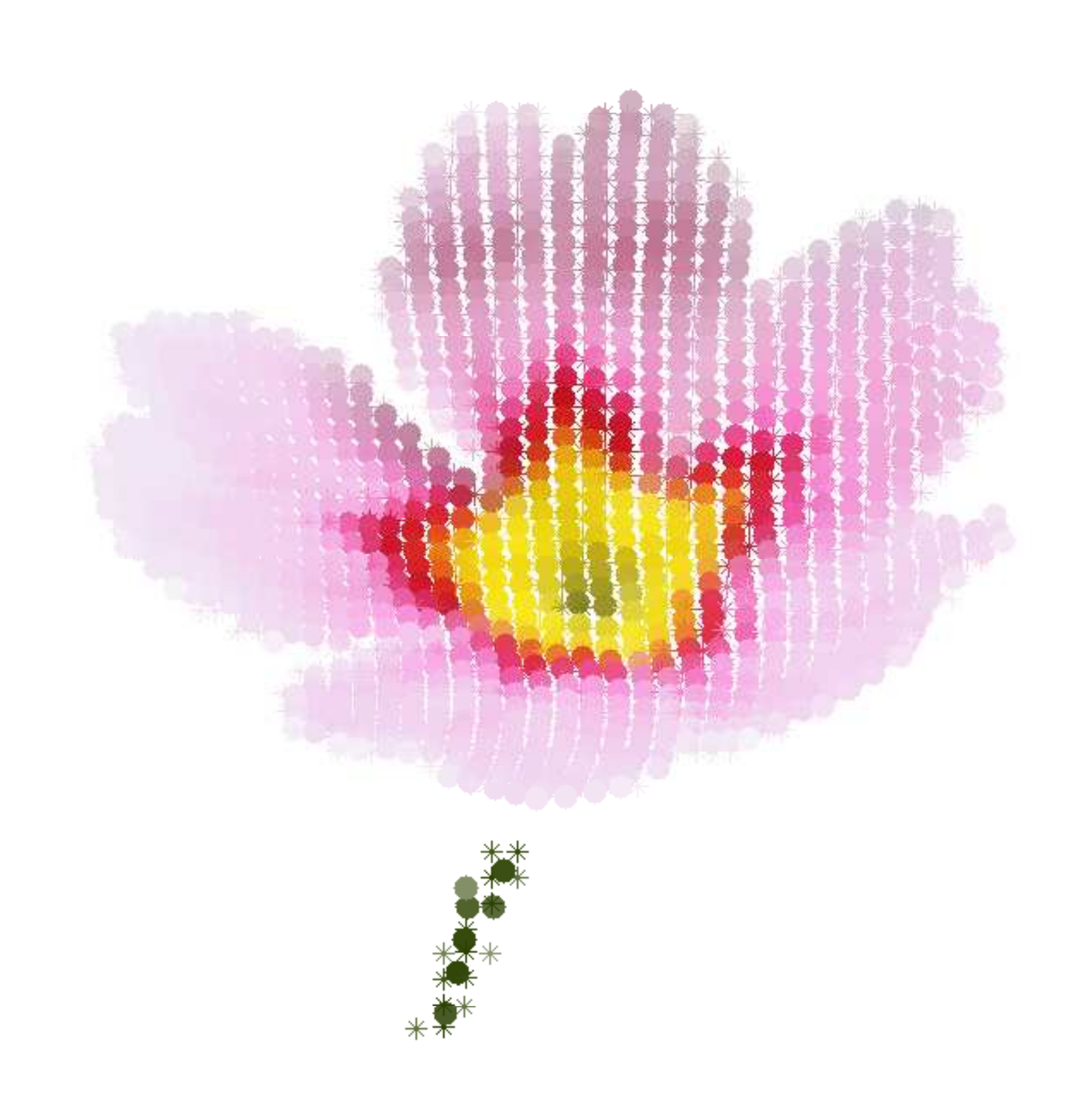}
	 \includegraphics[width = 0.18\textwidth]{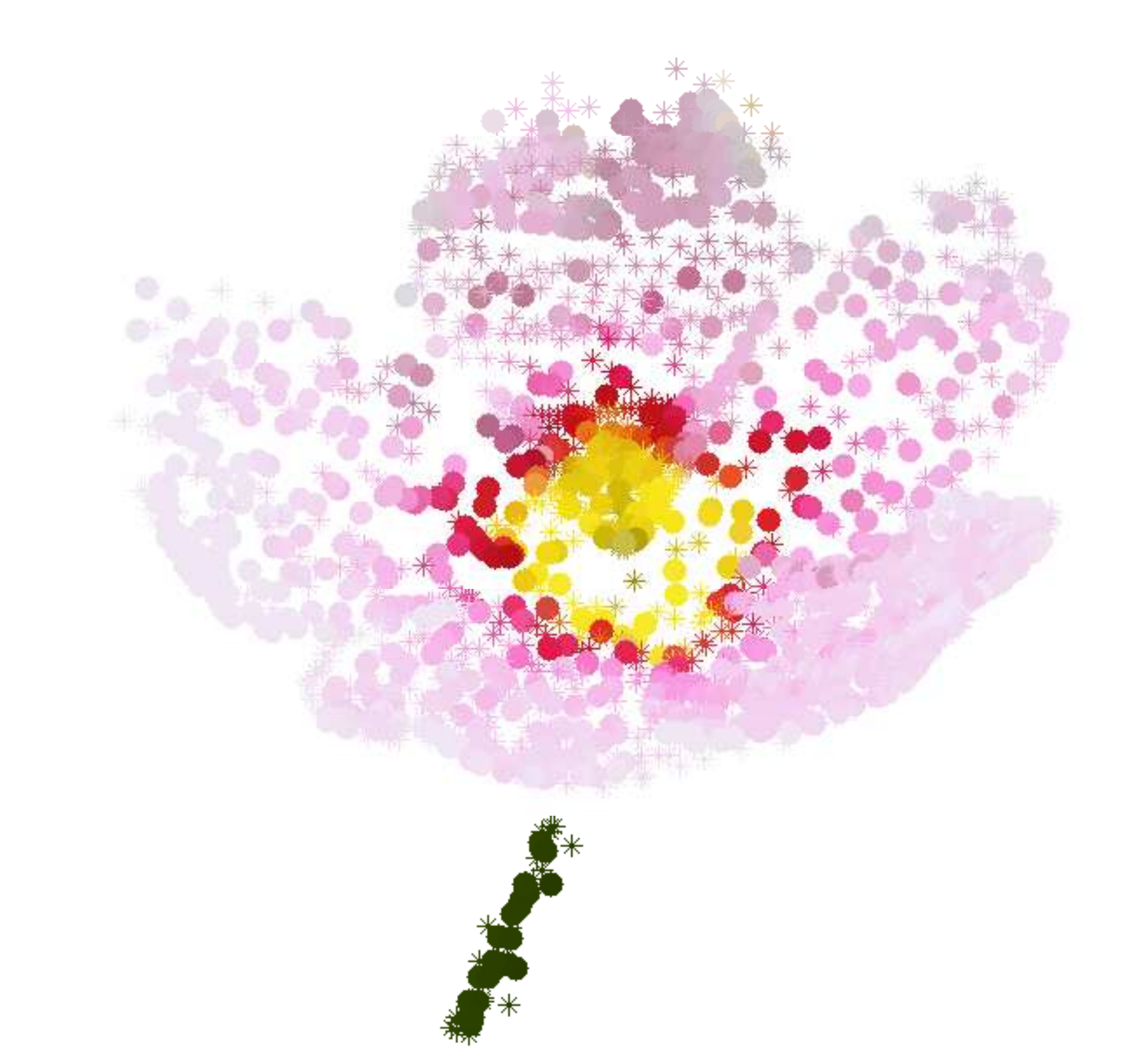}
	 \includegraphics[width = 0.18\textwidth]{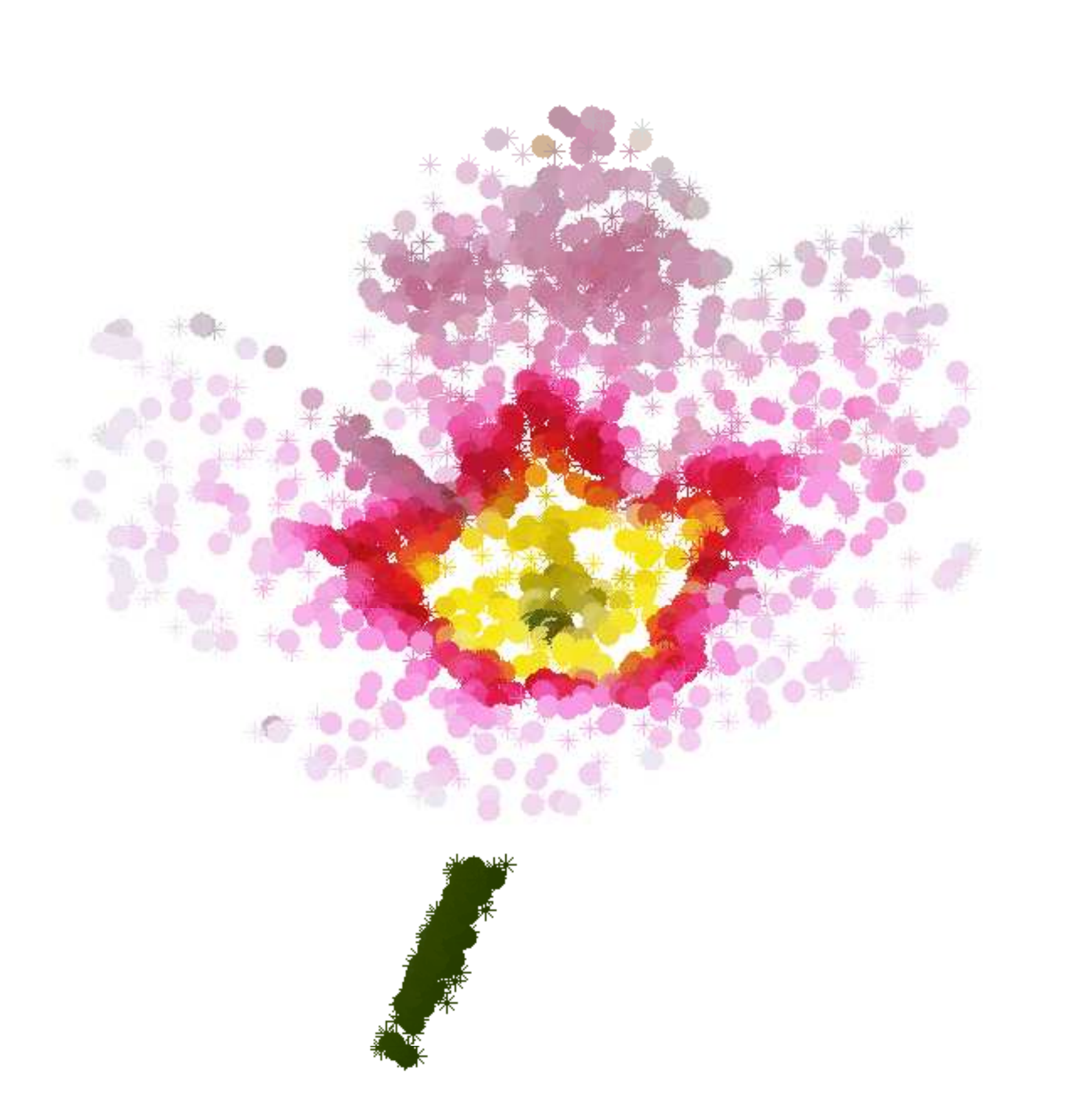}
	 \includegraphics[width = 0.18\textwidth]{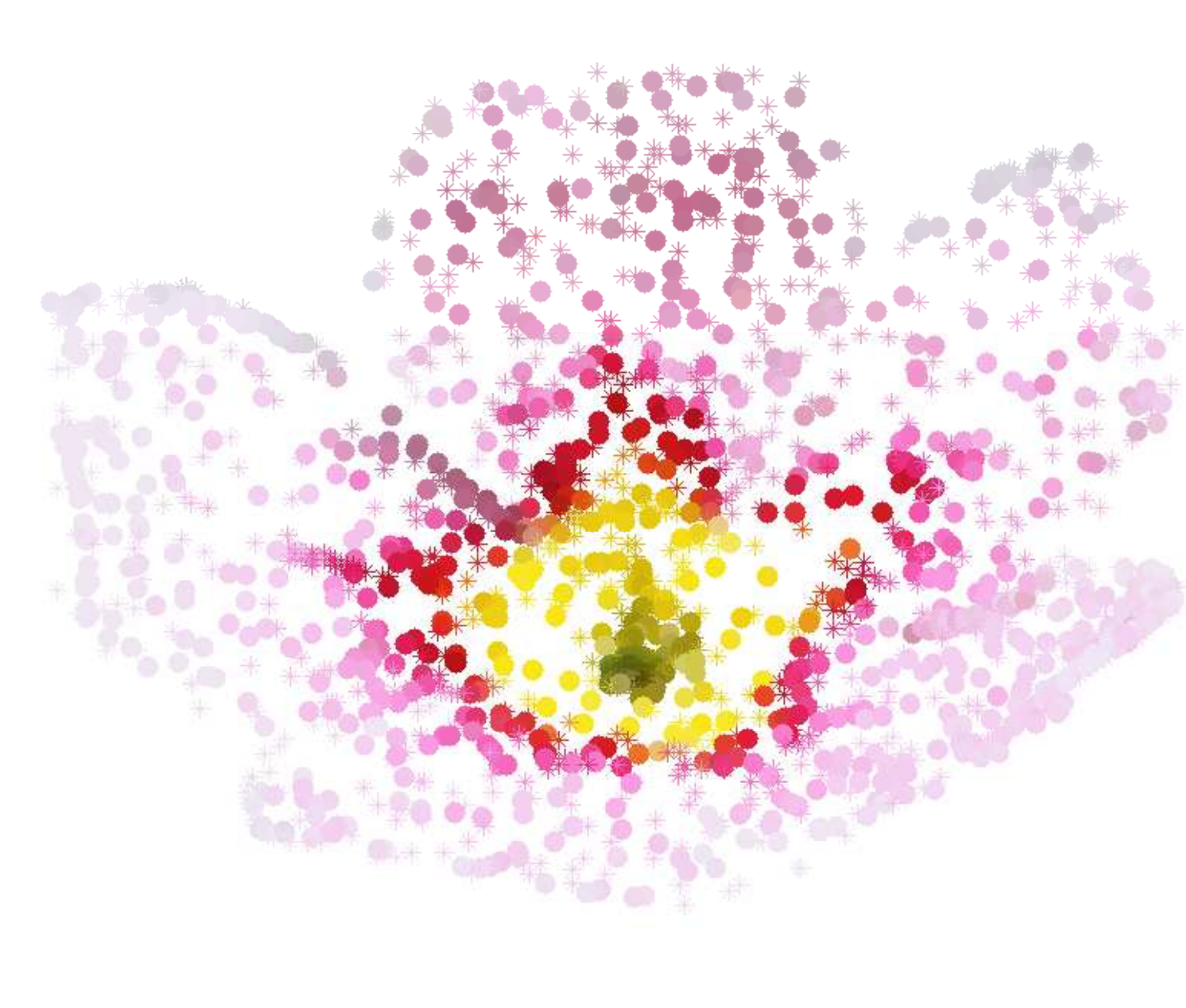}
	 \includegraphics[width = 0.18\textwidth]{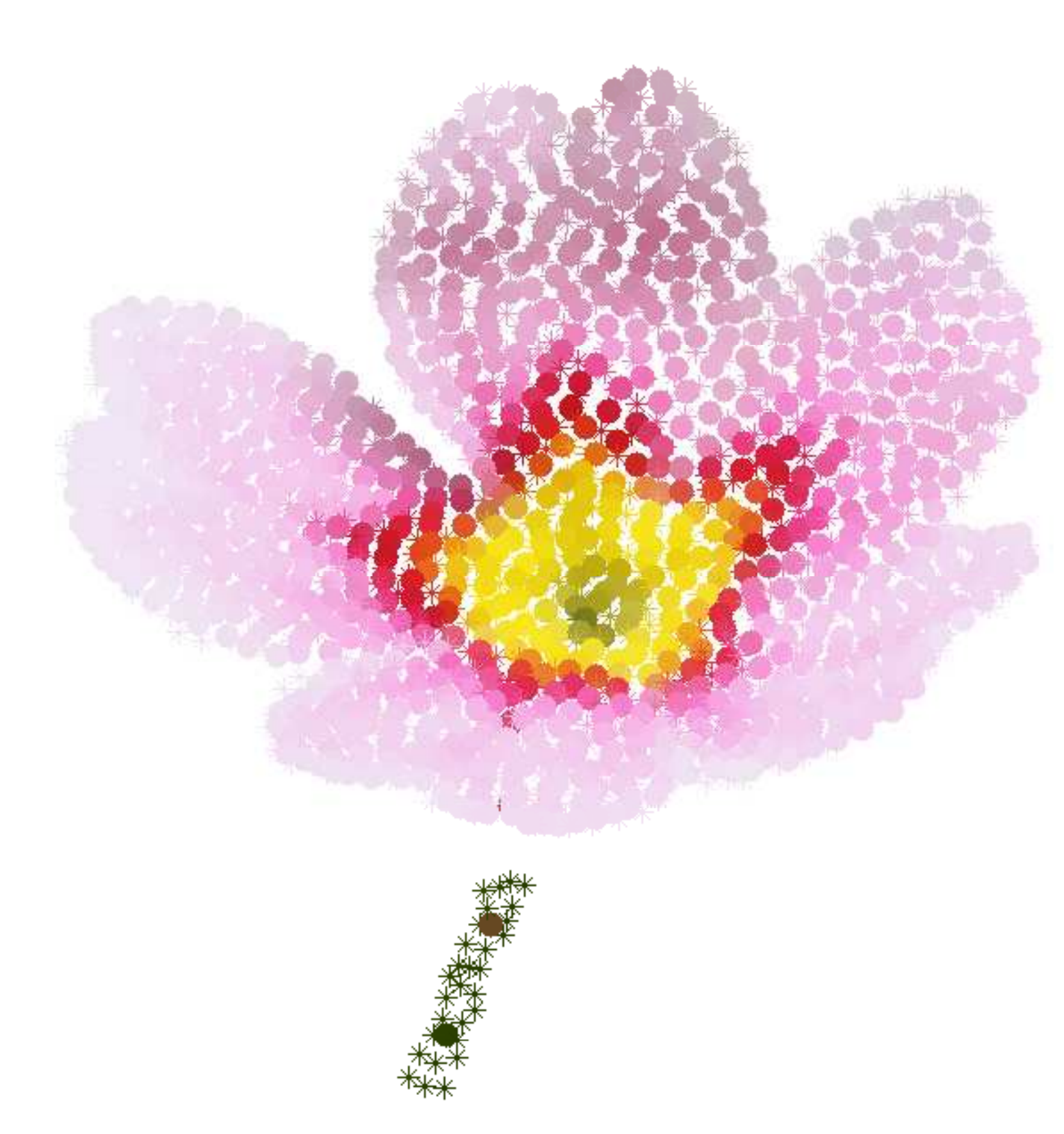}}
\\
\scalebox{0.95}{		
	 \includegraphics[width = 0.18\textwidth]{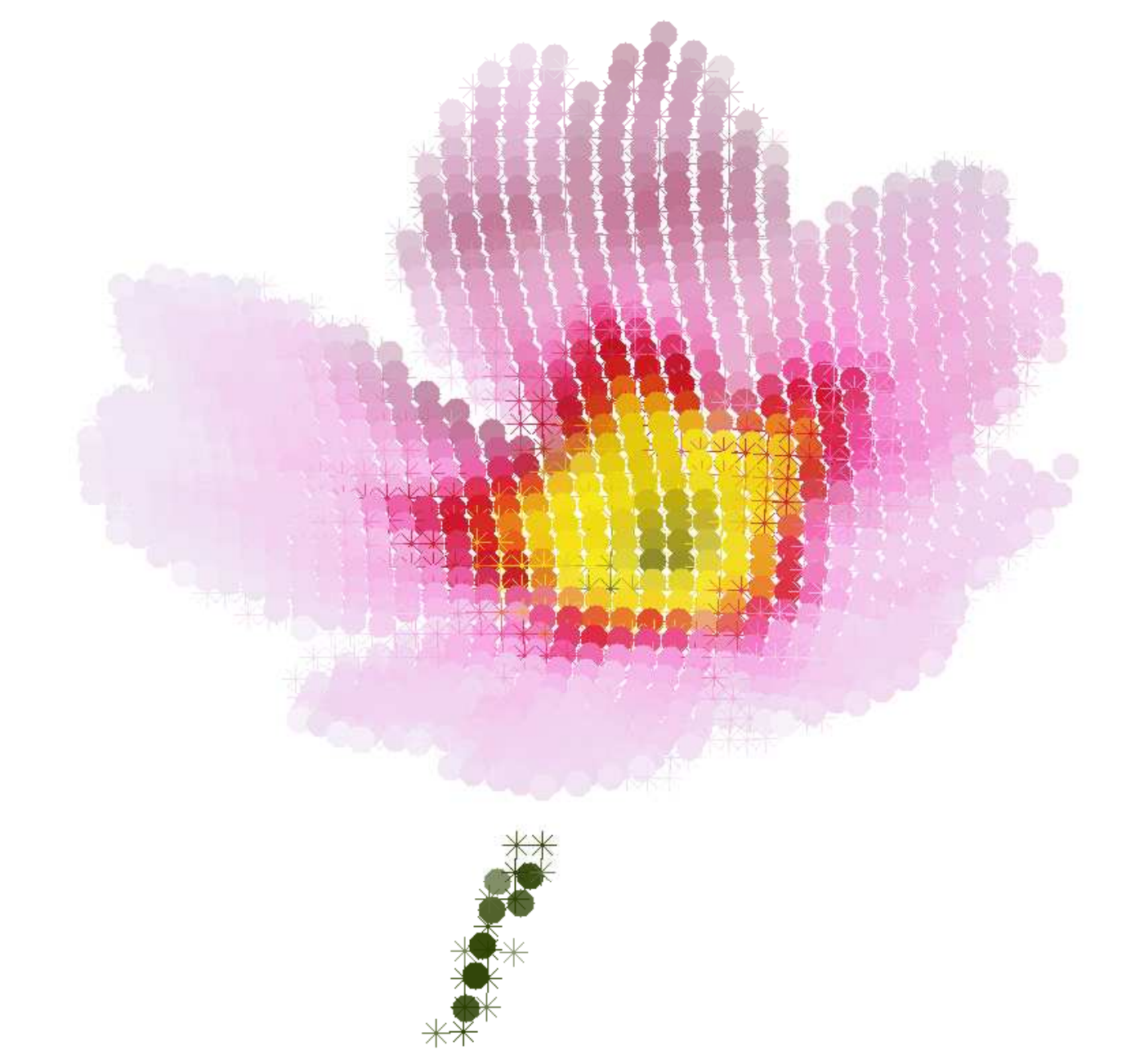}
	 \includegraphics[width = 0.18\textwidth]{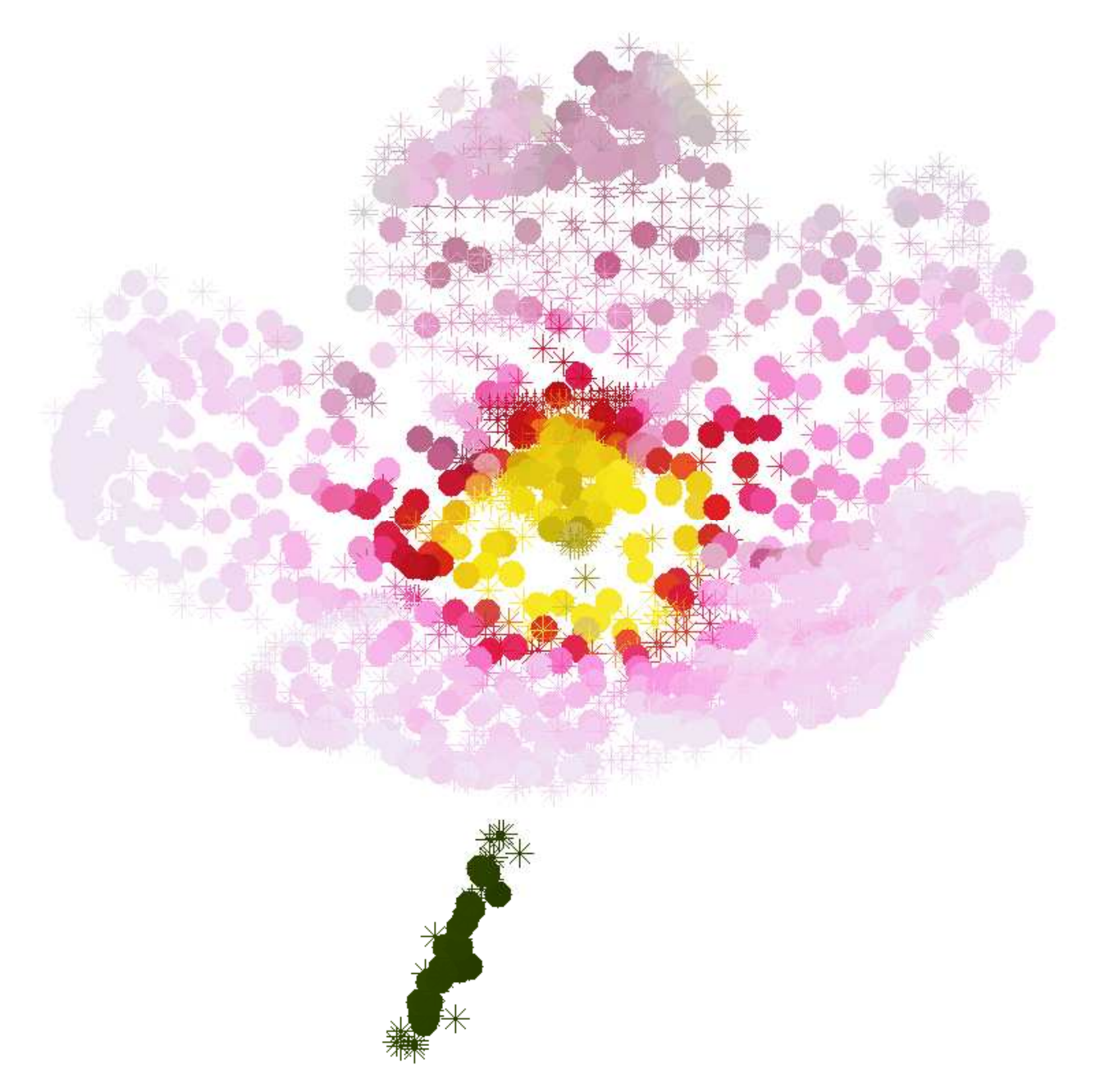}
	 \includegraphics[width = 0.18\textwidth]{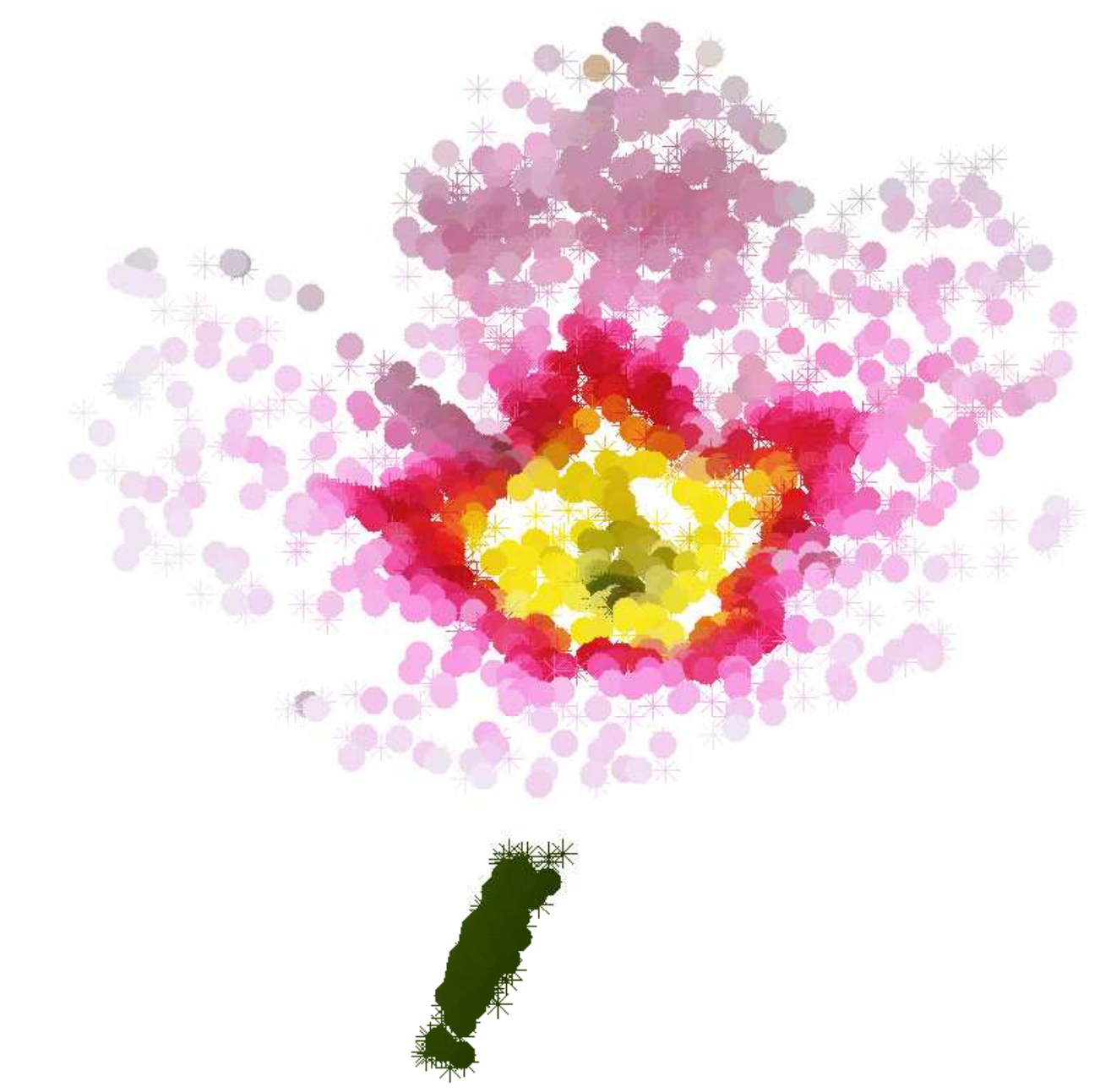}
	 \includegraphics[width = 0.18\textwidth]{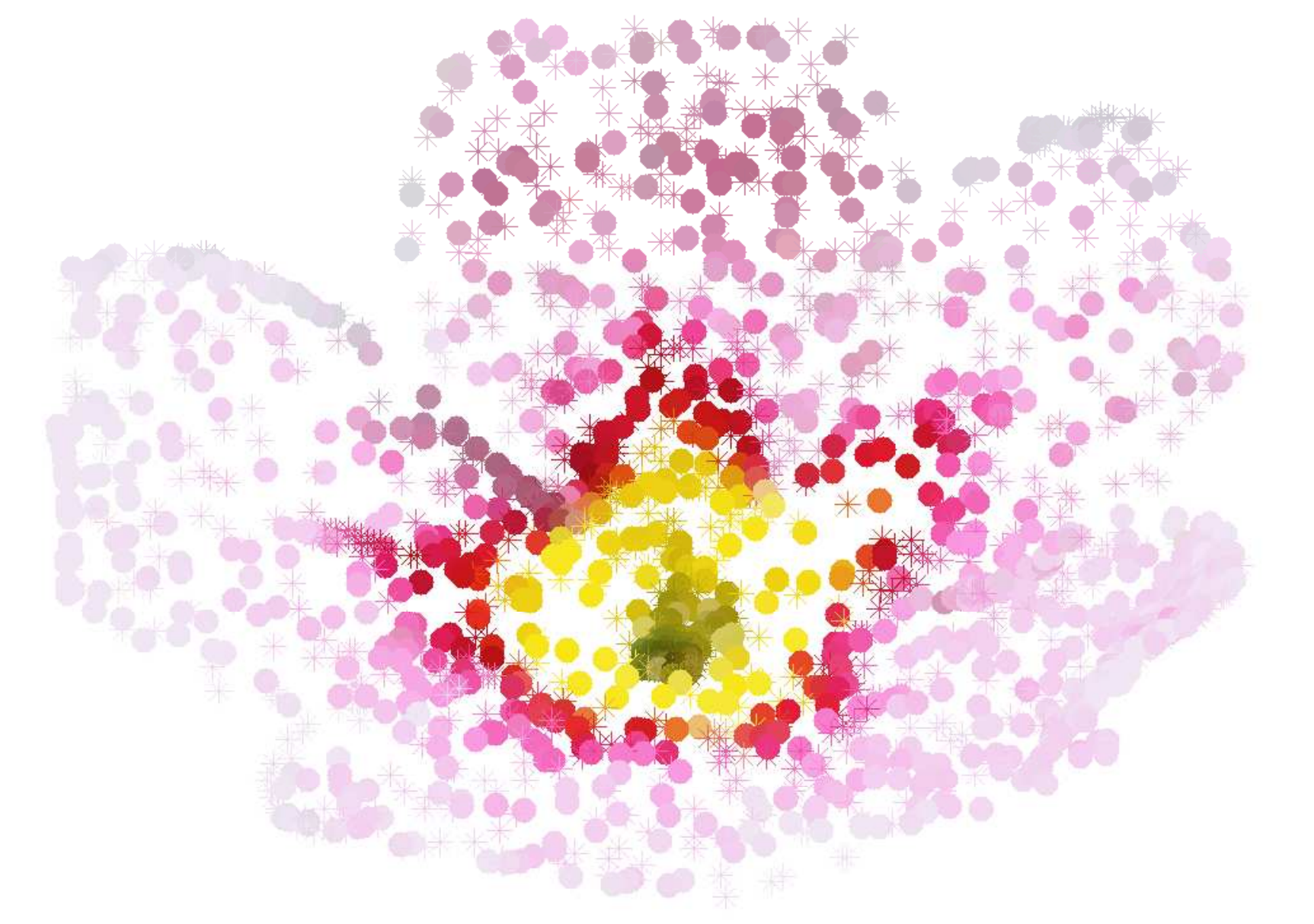}
	 \includegraphics[width = 0.18\textwidth]{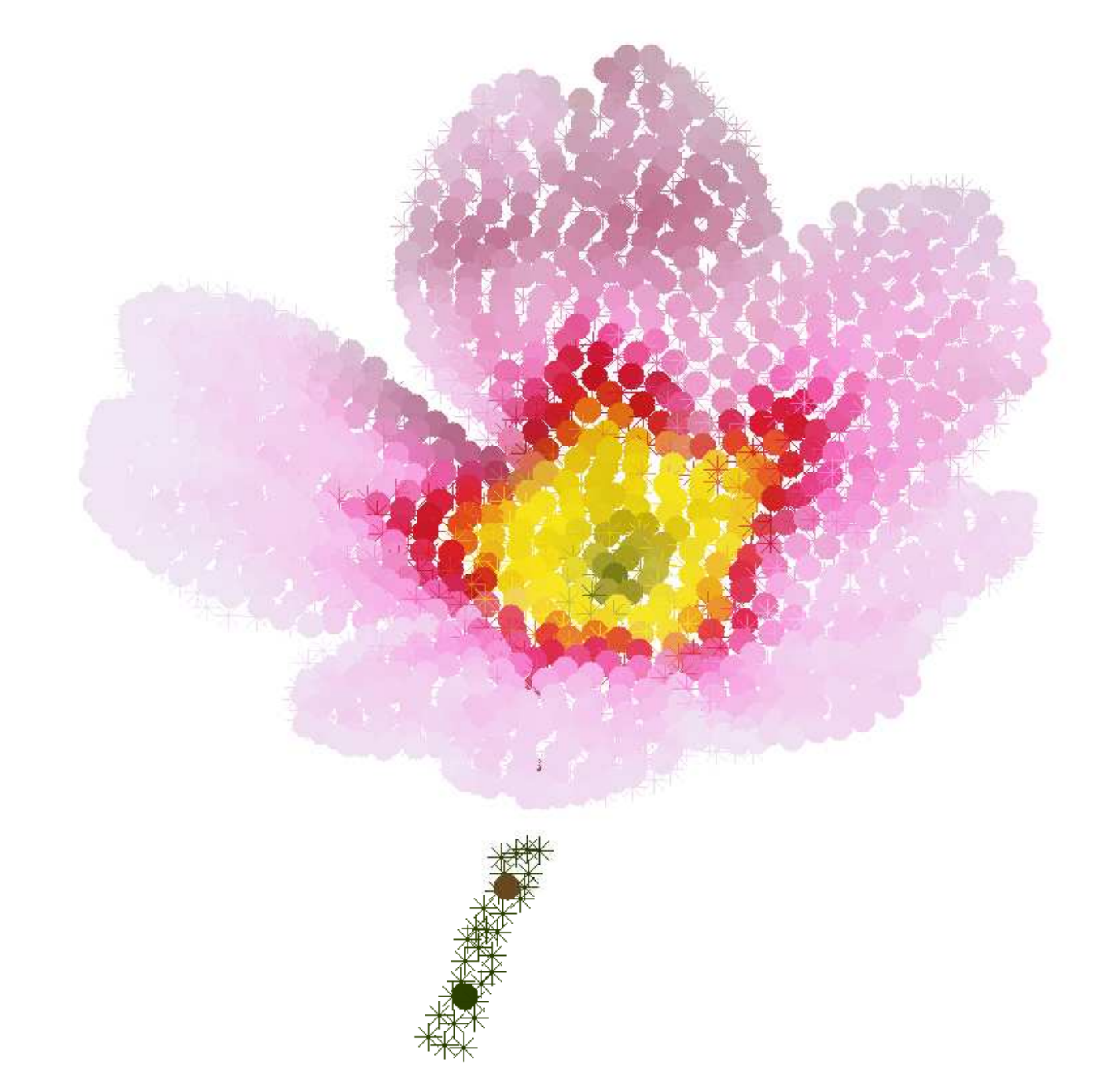}}
\\
	\caption{Non-rigid registration result of flower shape for a 1000 point sampling. The first row shows CCPD, and the original CPD in the second. Columns shows different sampling algorithms that are from left to right, bilinear, normal-based, color-based, NC-based, GNG.}
  \label{fig:reg:plant:1000}
\end{figure}

Figure \ref{fig:reg:plant:1000} shows the flower shape registration results for CCPD and the original CPD with the same point sampling (similarly, more experiments have been done with similar results). The figures show the registration for the second deformation (right of Fig. \ref{fig:face:models} and \ref{fig:plant:models}) of each shape as it is the larger one, and hence, the most difficult in terms of registration procedure. For each figure, the first row presents the CCPD method and the second the original CPD. From left to right, the sampling techniques are: bilinear, normal-based, color-based, NC-based, and GNG. 

\begin{figure}
 \centering
	 \includegraphics[width = 0.45\textwidth]{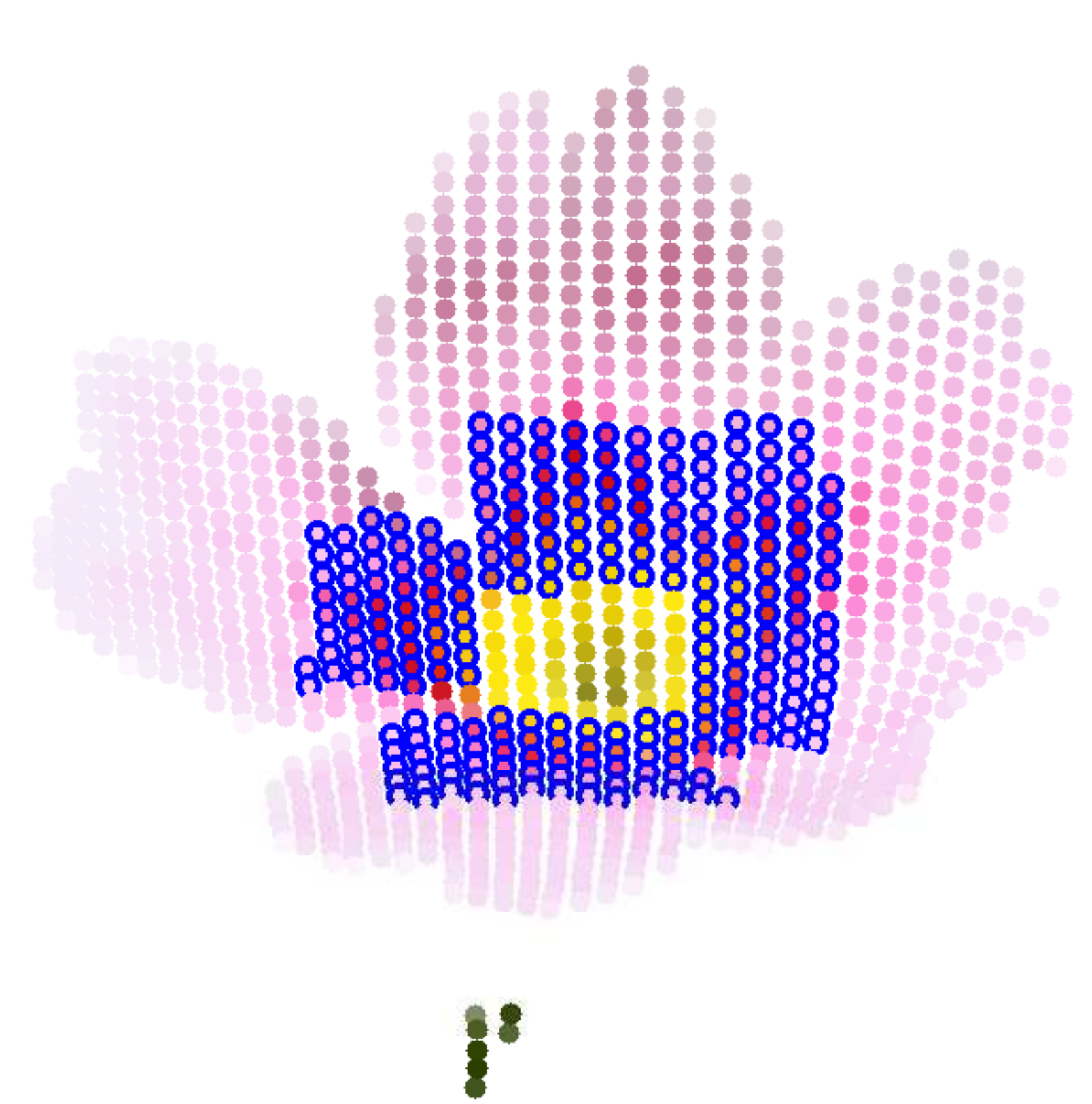}		
	 \qquad \includegraphics[width = 0.45\textwidth]{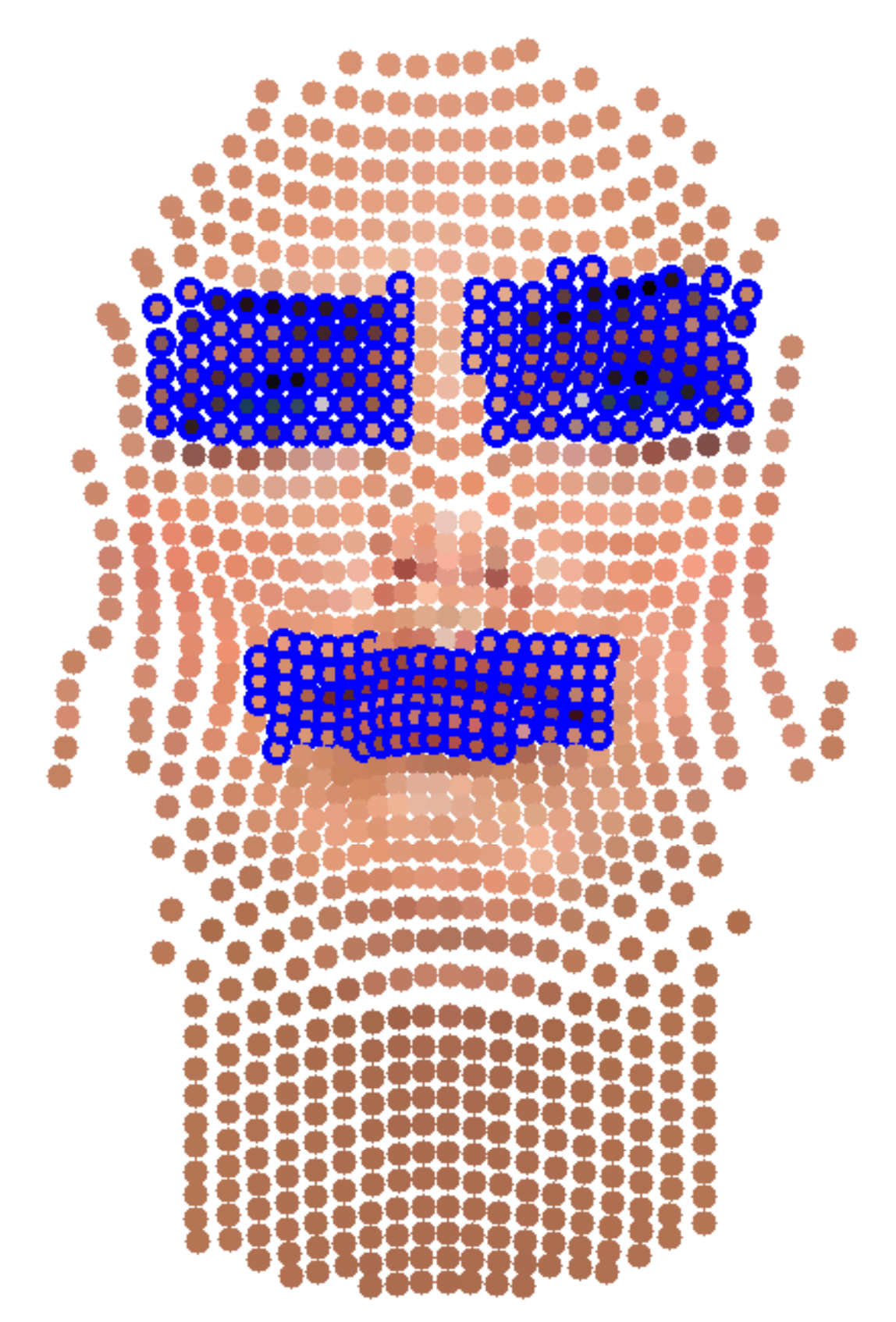}
	\caption{Example of Region of Interest for both shapes. The ROI are highlighted with blue circles.}
  \label{fig:ROI}
\end{figure}

\begin{figure}
  \centering
	 \includegraphics[width = 0.3\textwidth]{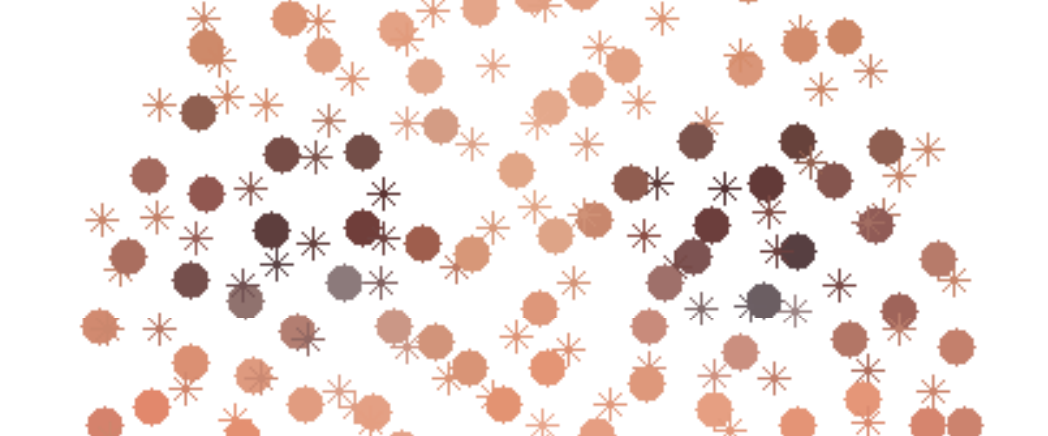}
	 \includegraphics[width = 0.3\textwidth]{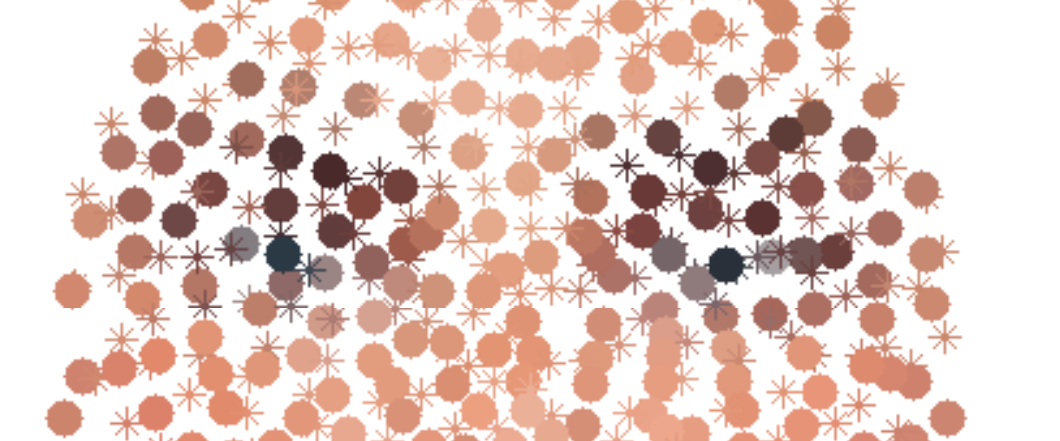}
	 \includegraphics[width = 0.3\textwidth]{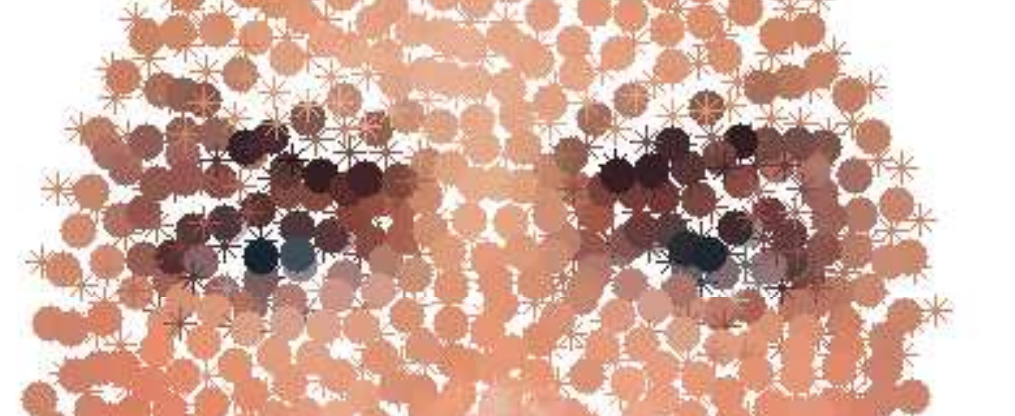}
\\
	 \includegraphics[width = 0.3\textwidth]{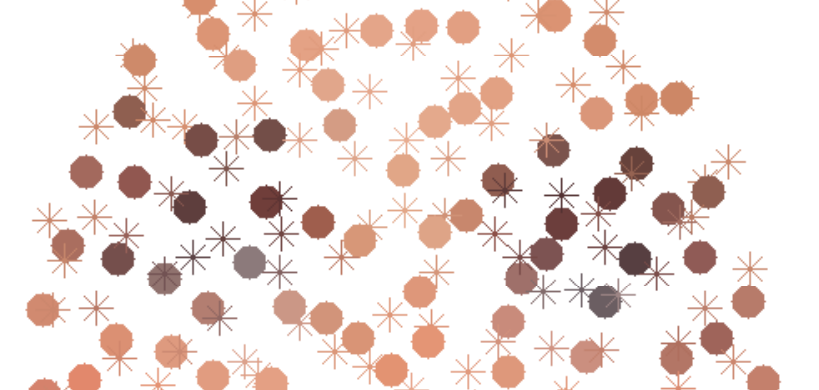}
	 \includegraphics[width = 0.3\textwidth]{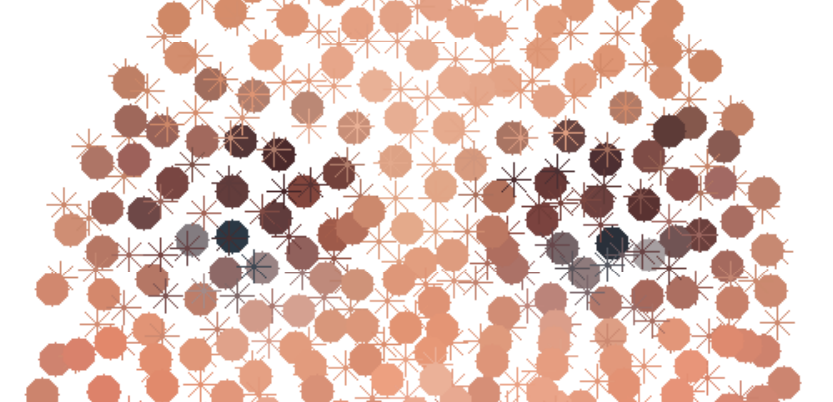}
	 \includegraphics[width = 0.3\textwidth]{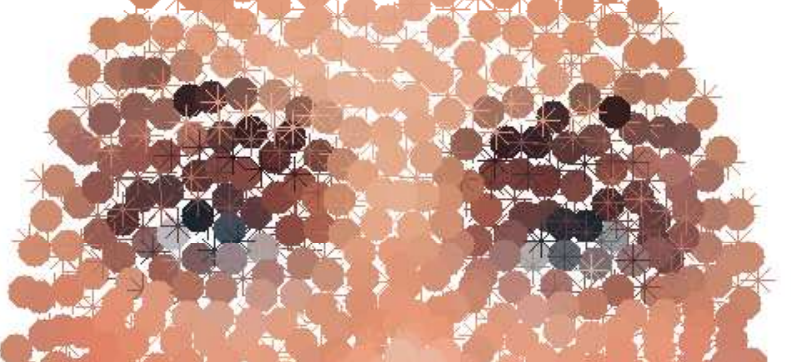}

	\caption{Enlarged example of the ROI for the face sampled with GNG. The first row shows the CCPD and the second the original CPD. \mscc{The data size is, from left to right, 250, 500, and 1000 points for the GNG.}}

%
%
  \label{fig:face:detail:gng}
\end{figure}

\begin{figure}
  \centering
\scalebox{1}{
	 \includegraphics[width = 0.3\textwidth]{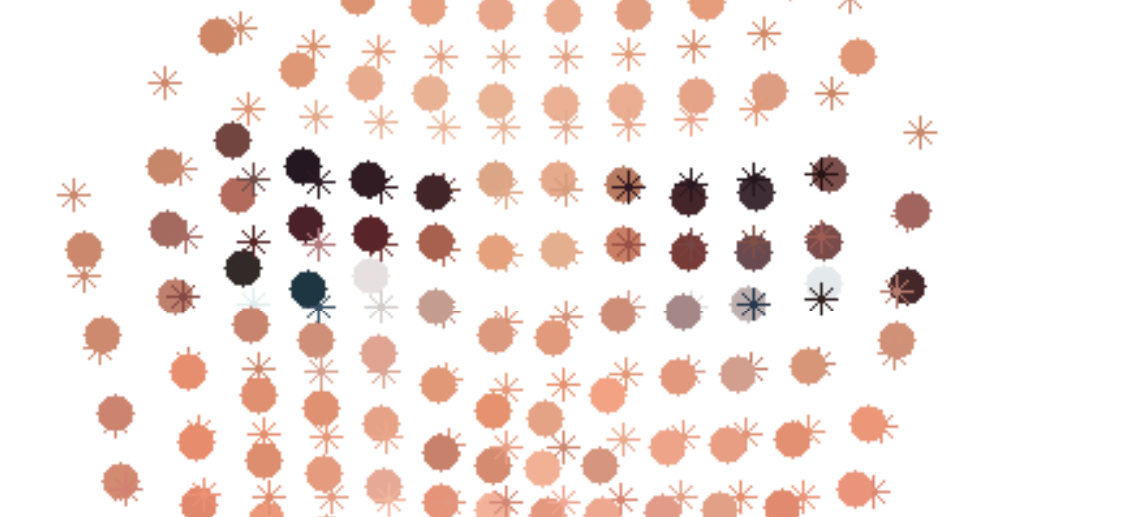}
	 \includegraphics[width = 0.3\textwidth]{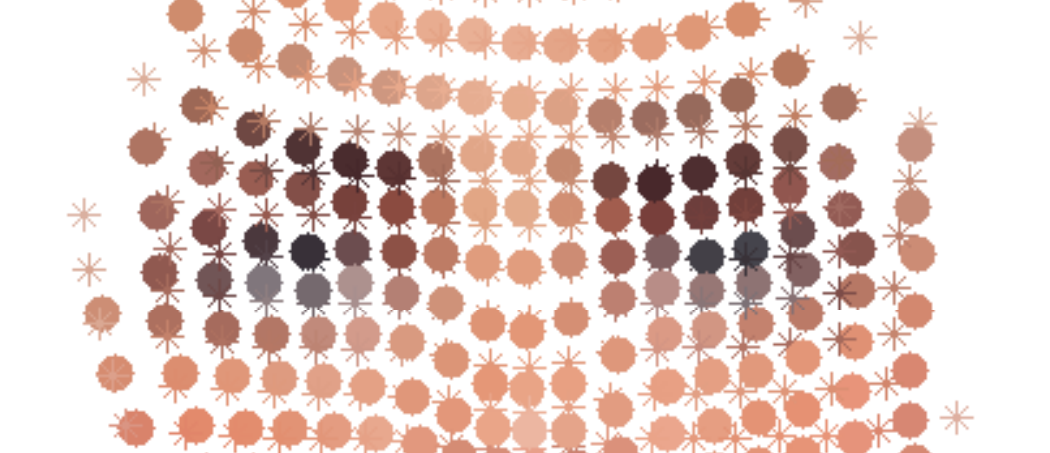}
	 \includegraphics[width = 0.3\textwidth]{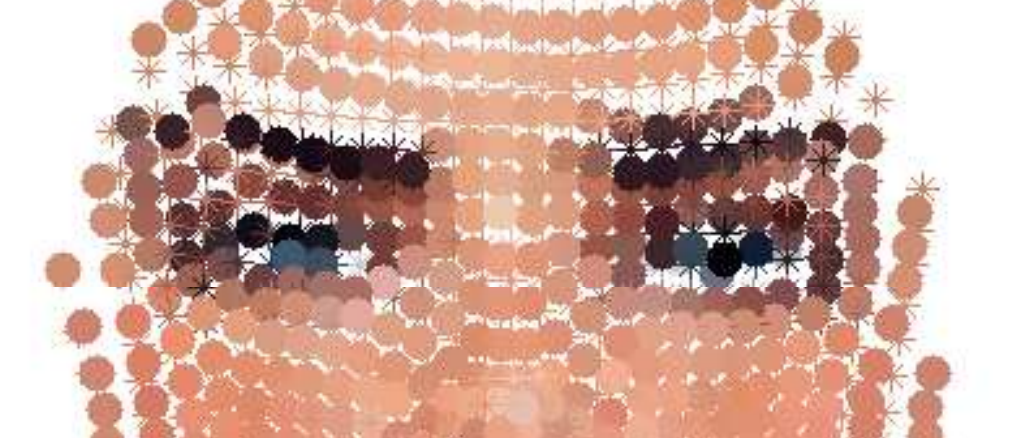}}
\\
\scalebox{1}{
	 \includegraphics[width = 0.3\textwidth]{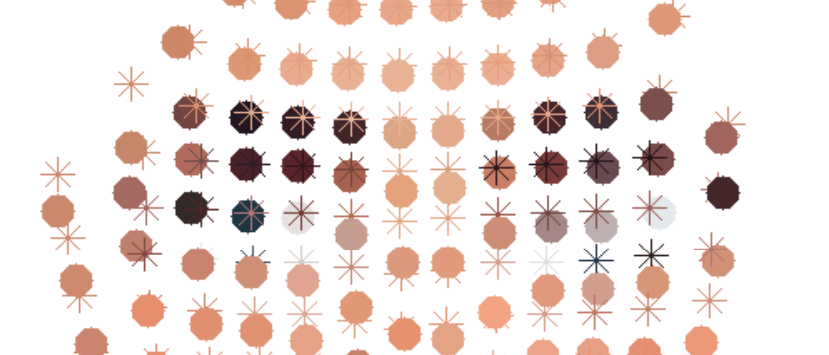}
	 \includegraphics[width = 0.3\textwidth]{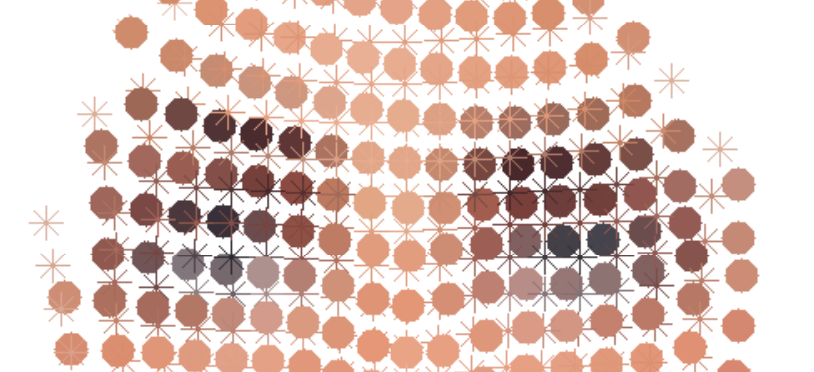}
	 \includegraphics[width = 0.3\textwidth]{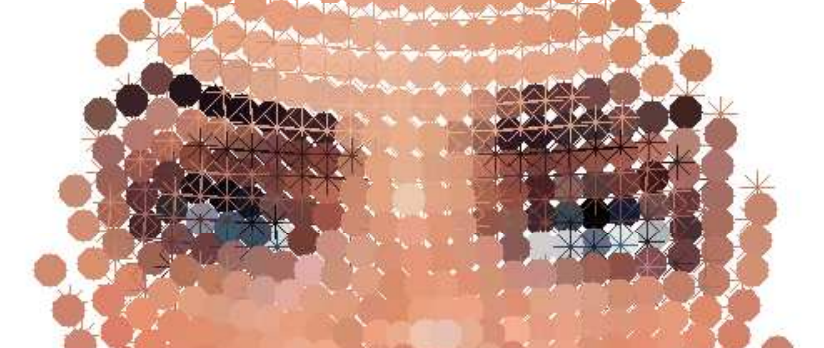}}
\caption{Enlarged example of the ROI for the face sampled with bilinear sampling. The first row shows the CCPD and the second the original CPD. \mscc{ The data size is, from left to right, 250, 500, and 1000 points for the bilinear.}}%

%
%
  \label{fig:face:detail:bilinear}
\end{figure}

\begin{figure}
  \centering
\scalebox{1}{
	 \includegraphics[width = 0.3\textwidth]{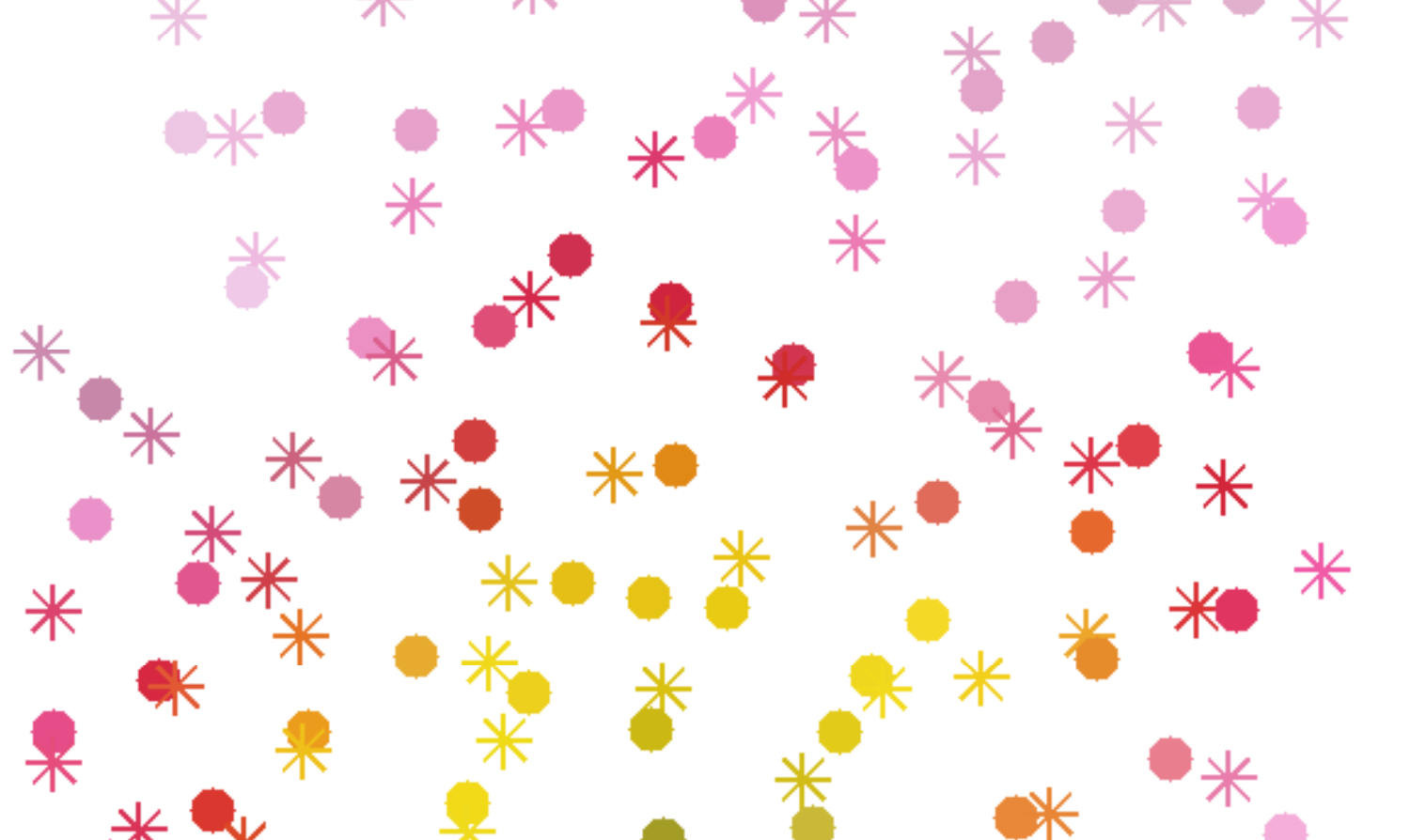}
	 \includegraphics[width = 0.3\textwidth]{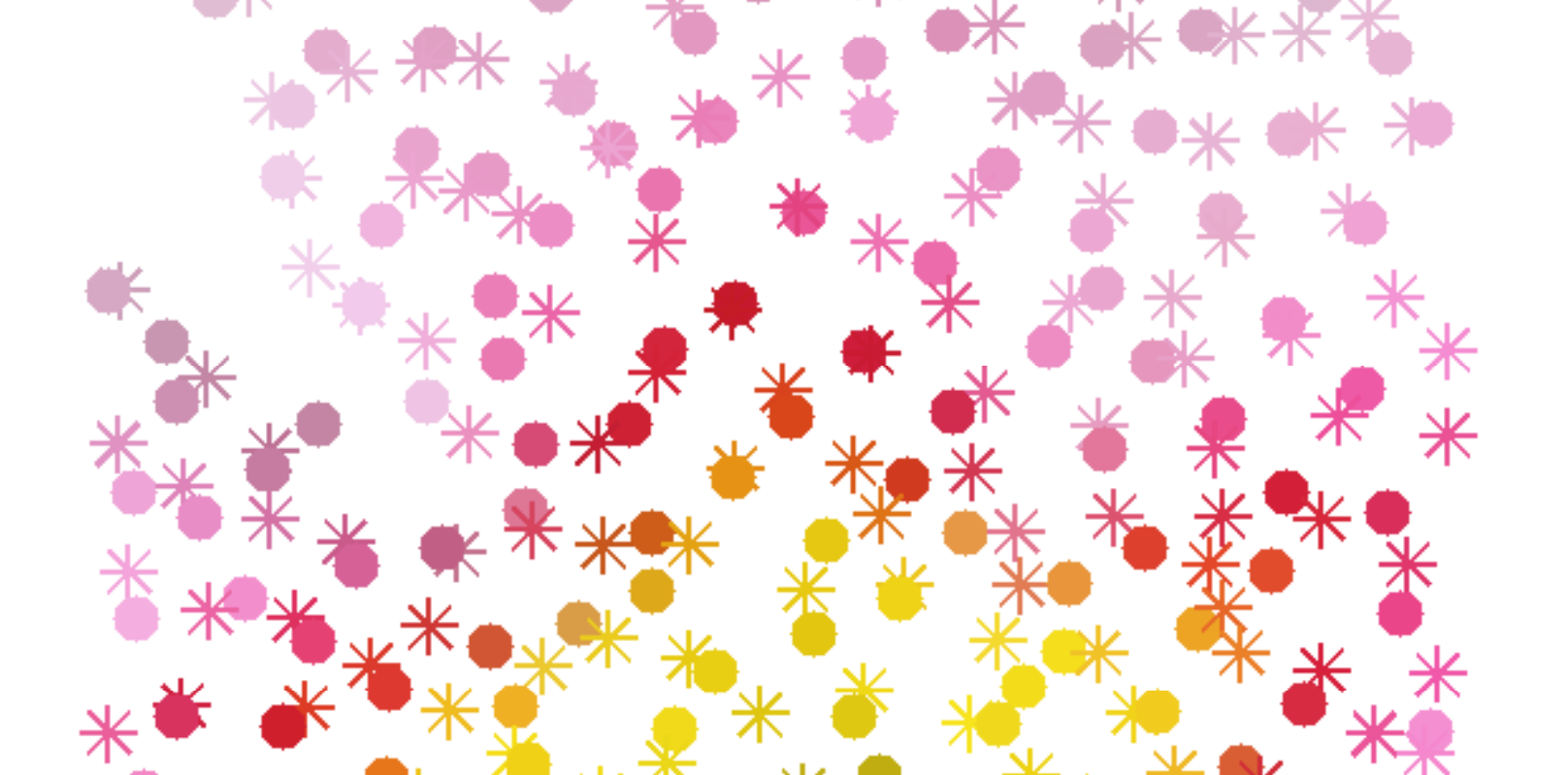}
	 \includegraphics[width = 0.28\textwidth]{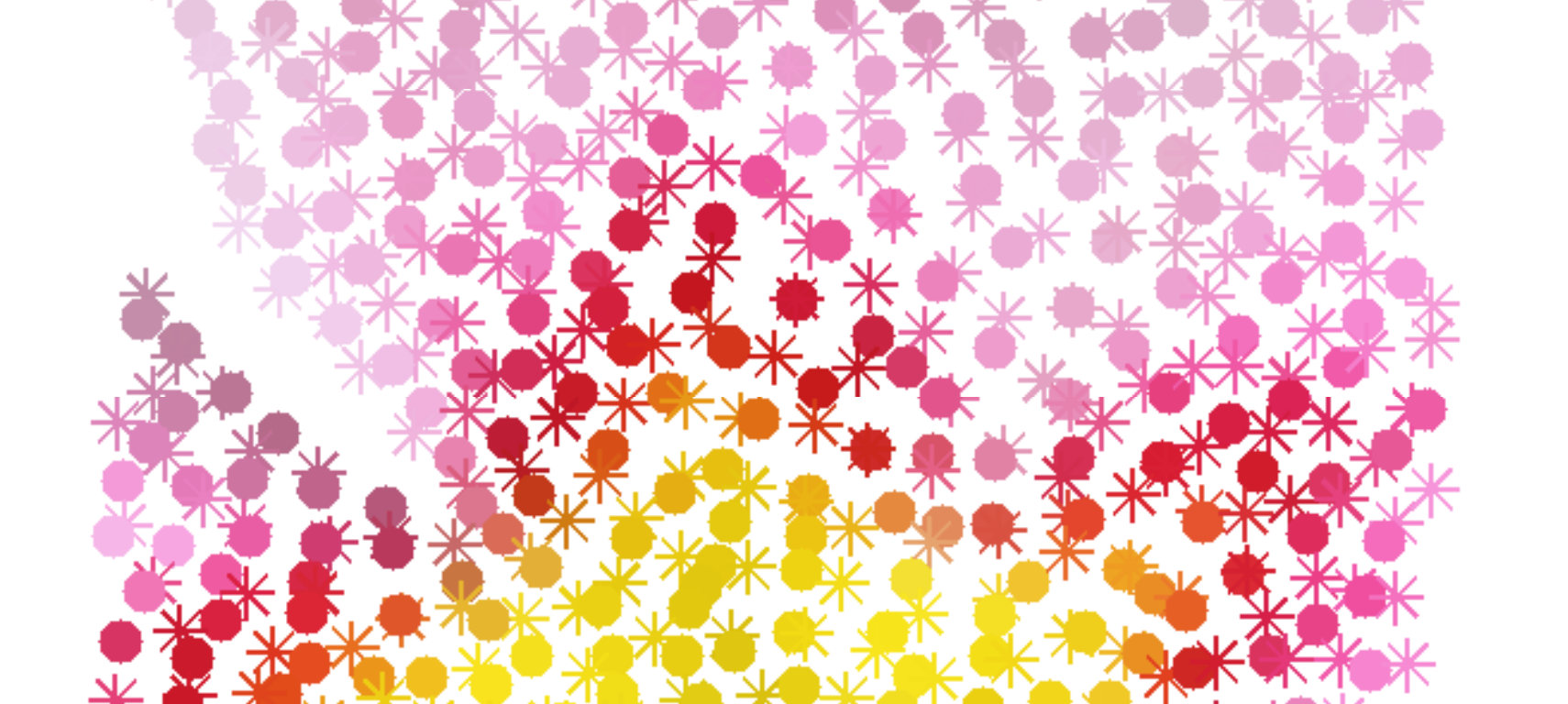}}
\\
\scalebox{1}{
	 \includegraphics[width = 0.3\textwidth]{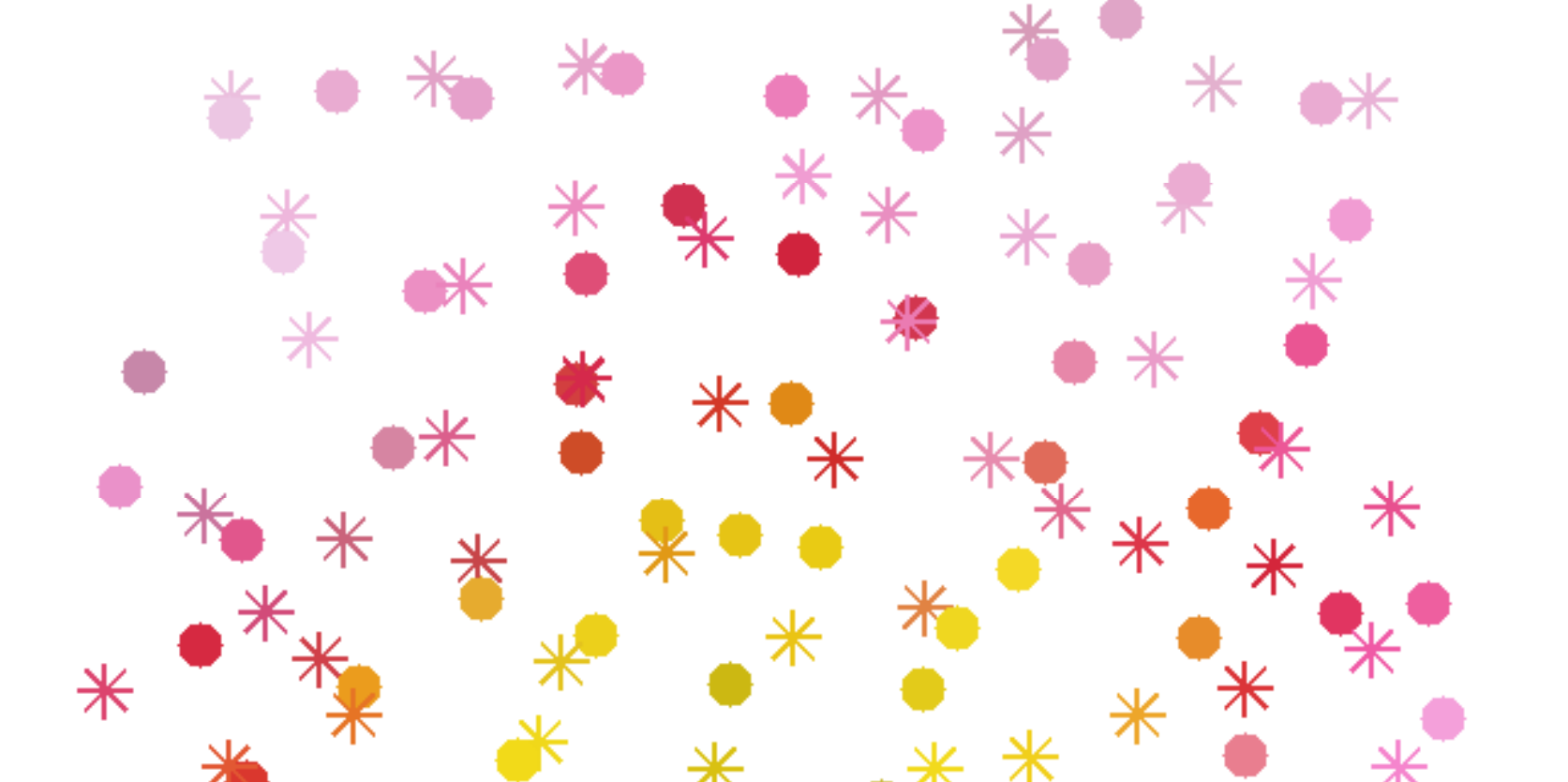}
	 \includegraphics[width = 0.3\textwidth]{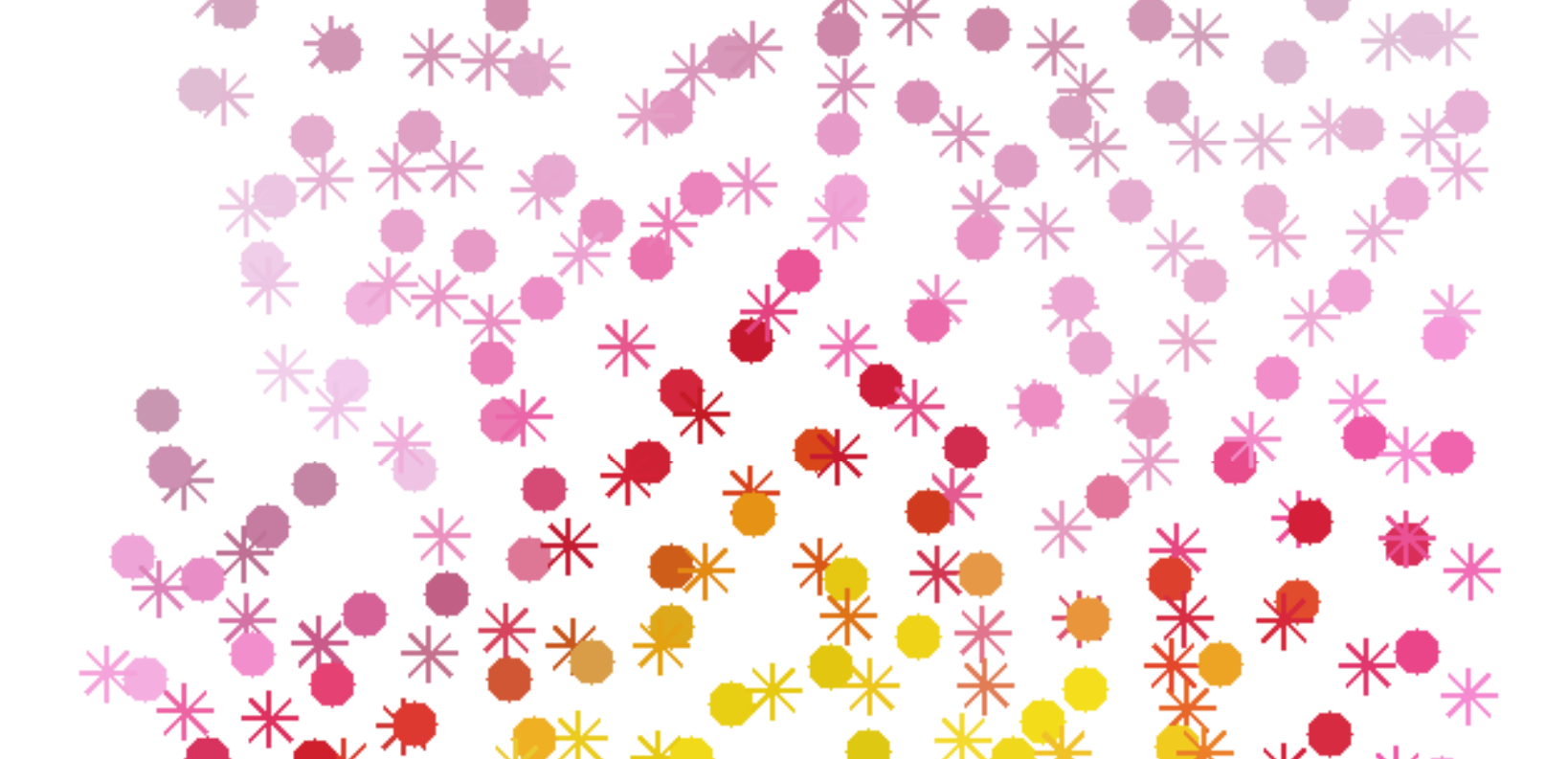}
	 \includegraphics[width = 0.3\textwidth]{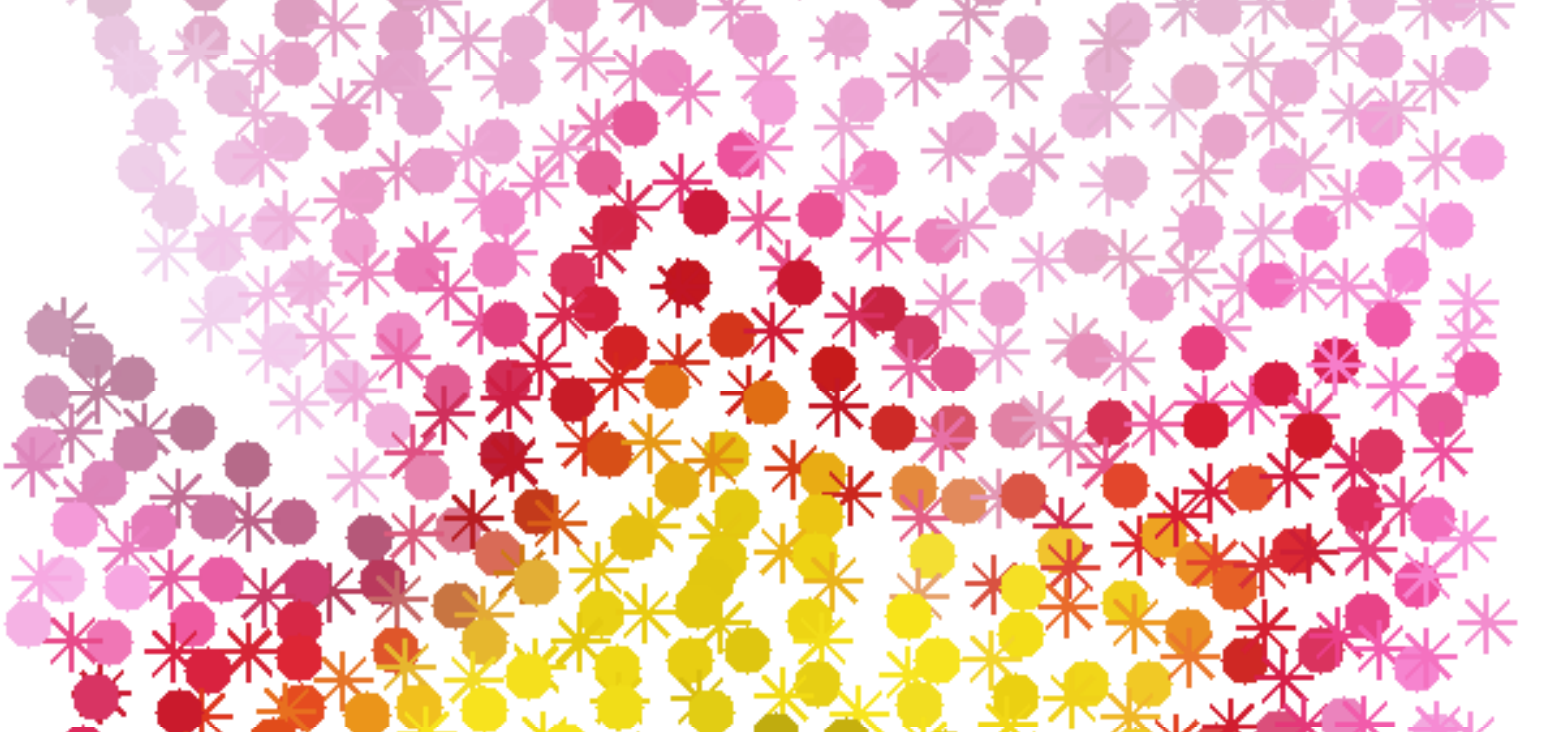}}

	\caption{Enlarged example of the ROI for the flower sampled with GNG. The first row shows the CCPD and the second the original algorithm. \mscc{The data size is, from left to right, 250, 500, and 1000 points for the GNG.}}

%
%
  \label{fig:plant:detail:gng}
\end{figure}

\begin{figure}
  \centering

\scalebox{1}{
	 \includegraphics[width = 0.3\textwidth]{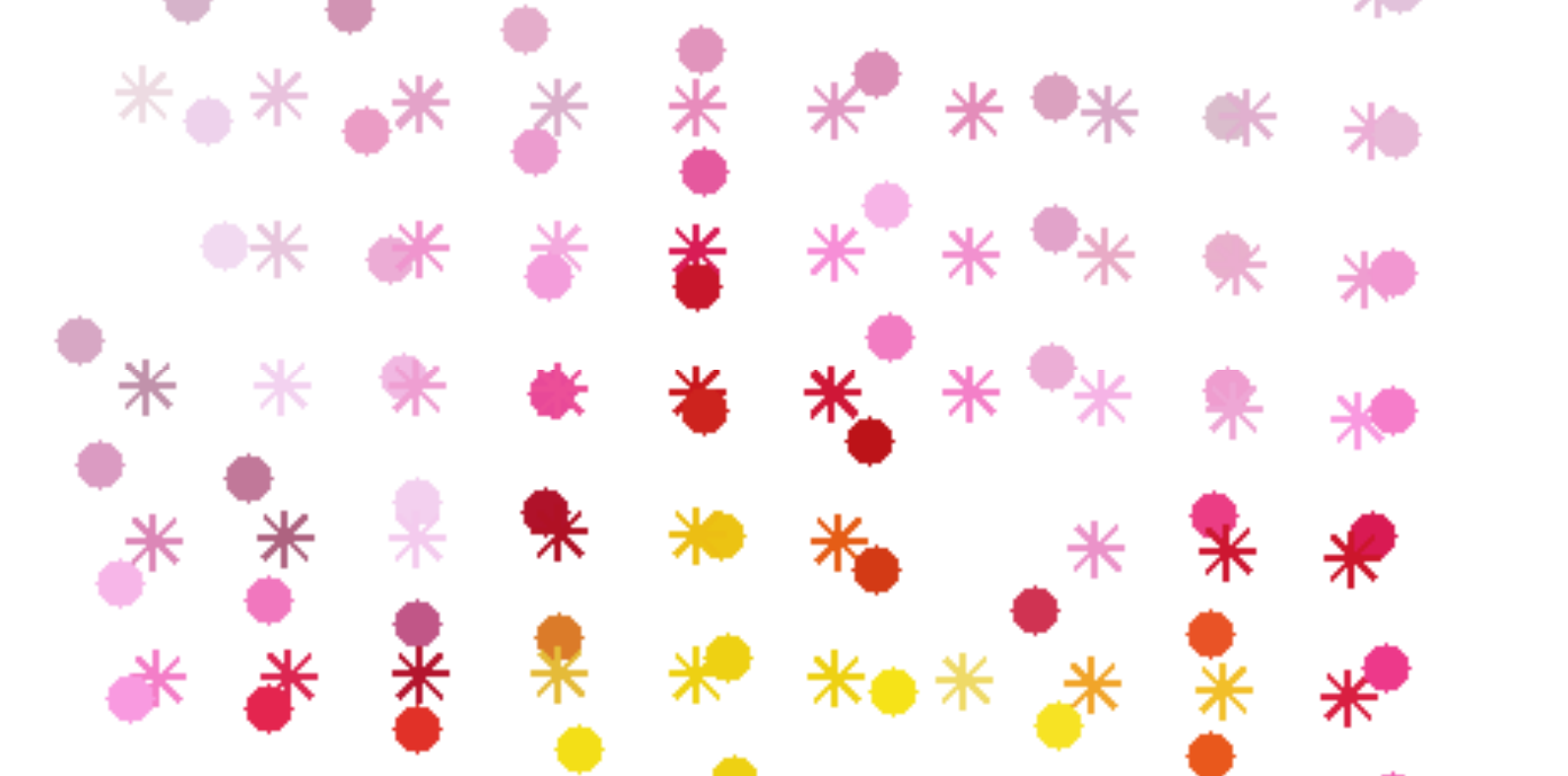}
	 \includegraphics[width = 0.3\textwidth]{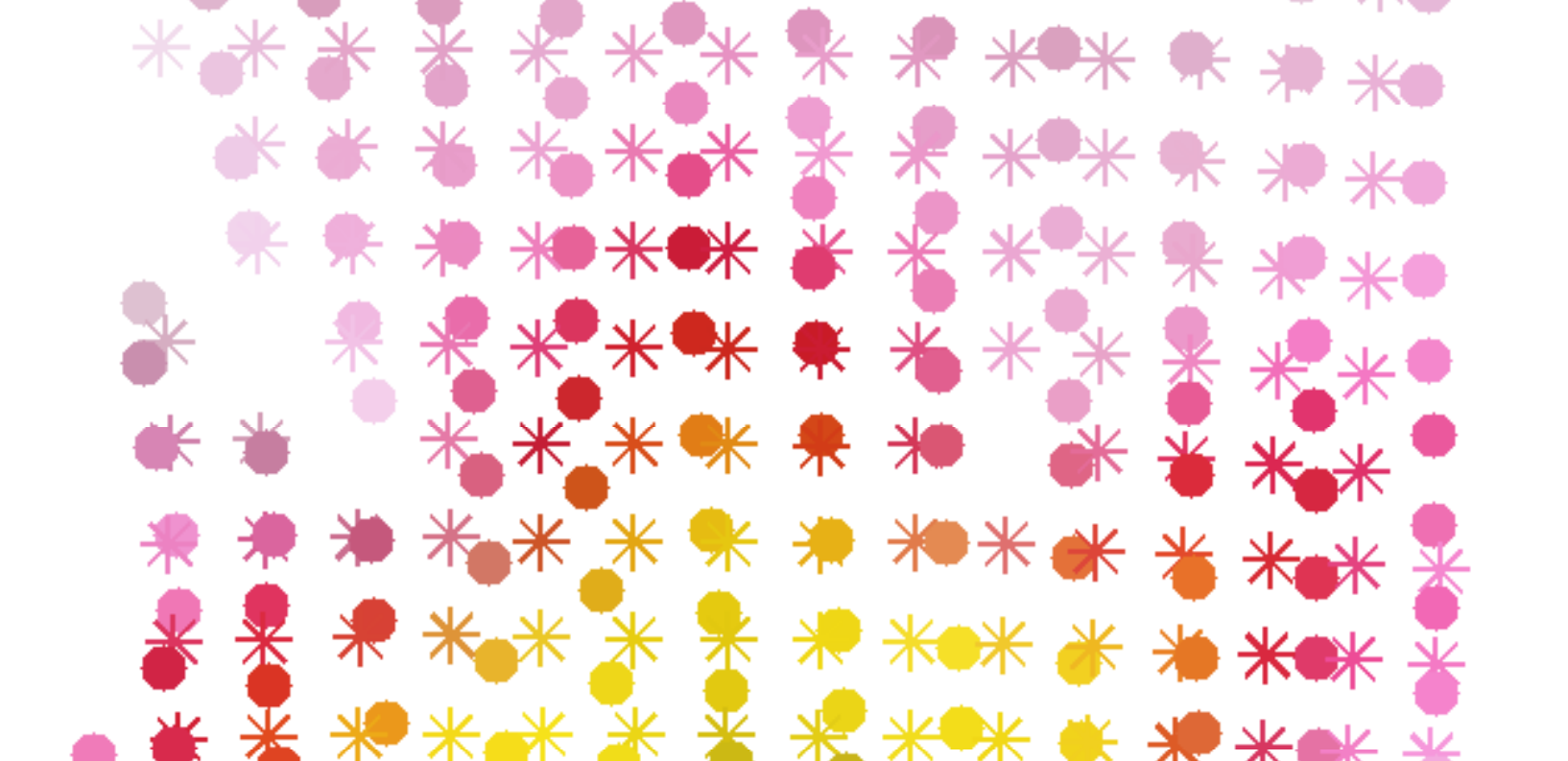}
	 \includegraphics[width = 0.3\textwidth]{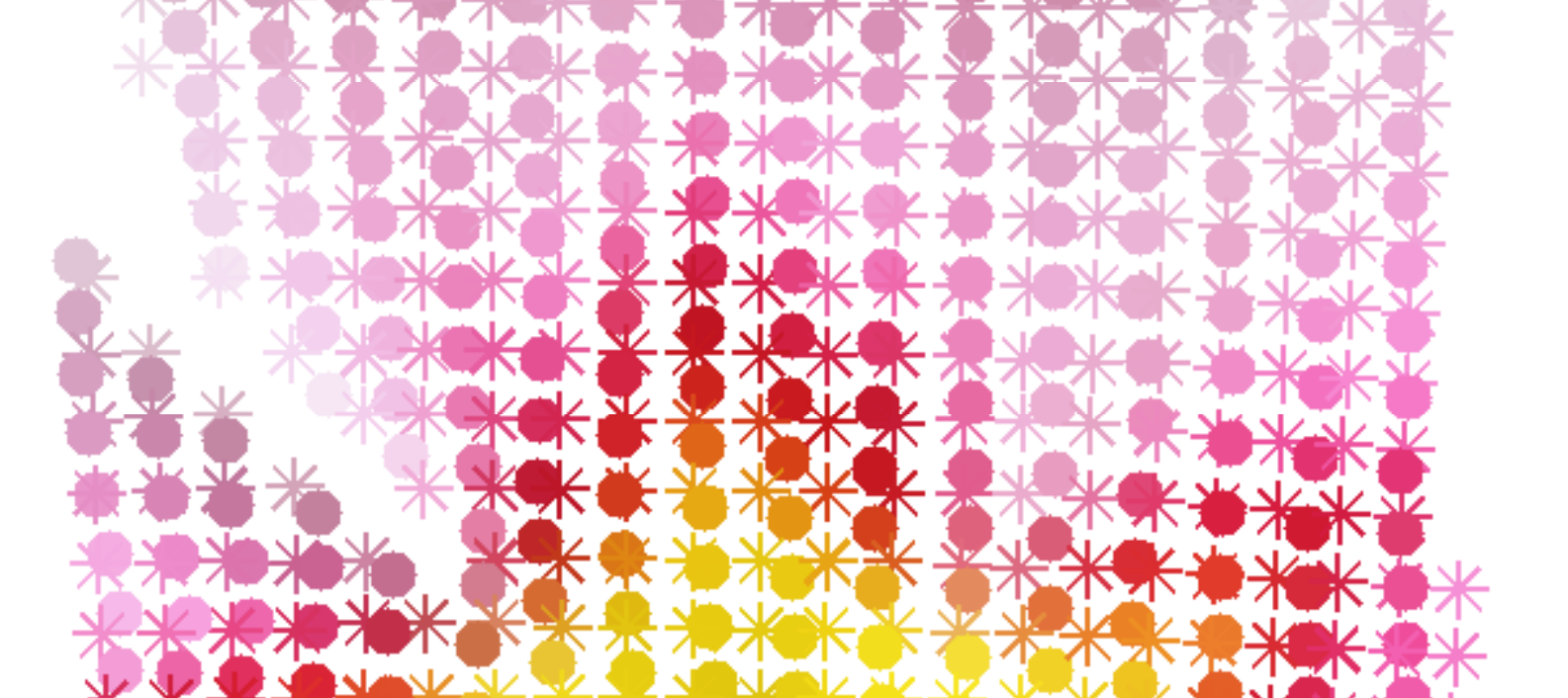}}
\\
\scalebox{1}{
	 \includegraphics[width = 0.3\textwidth]{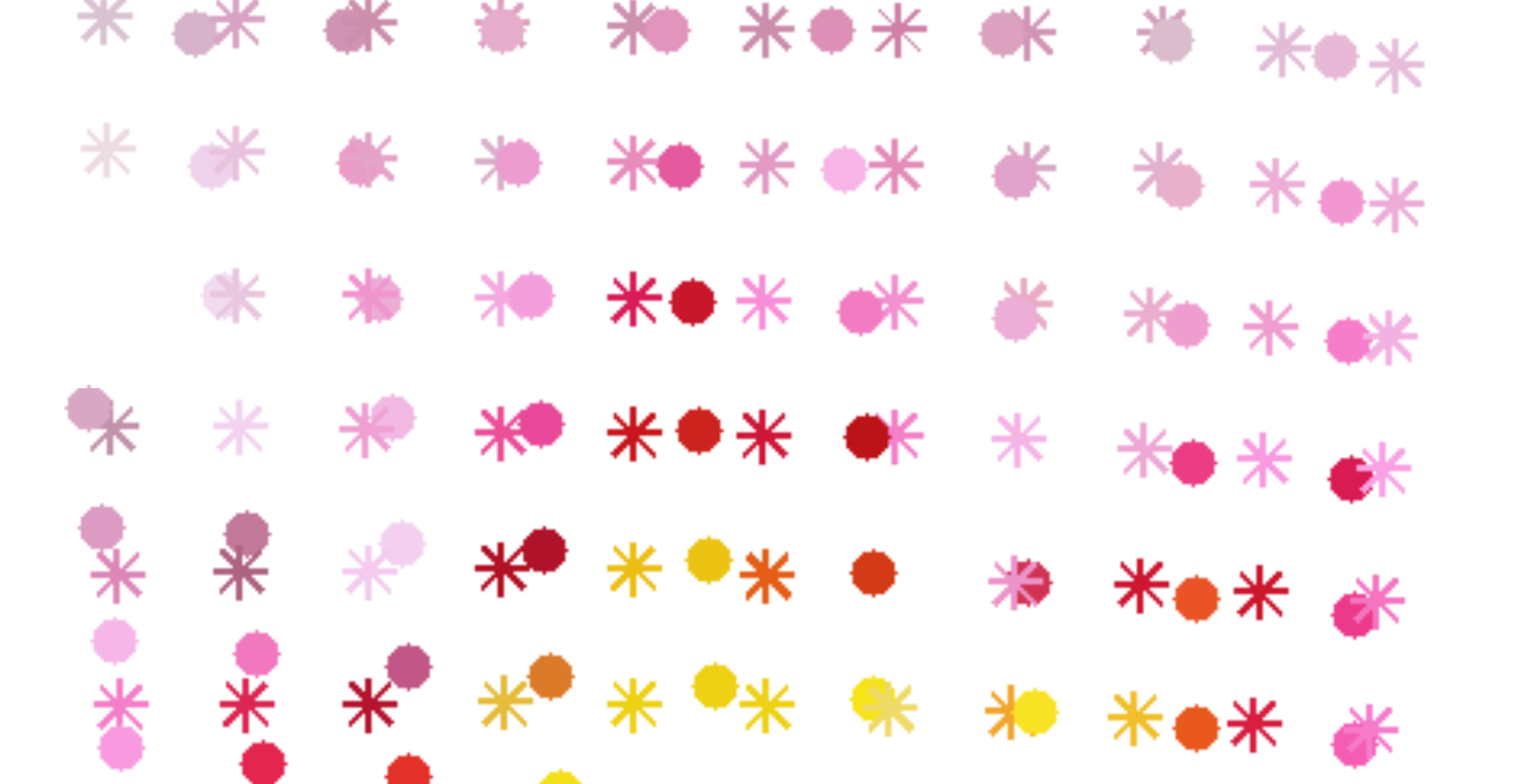}
	 \includegraphics[width = 0.3\textwidth]{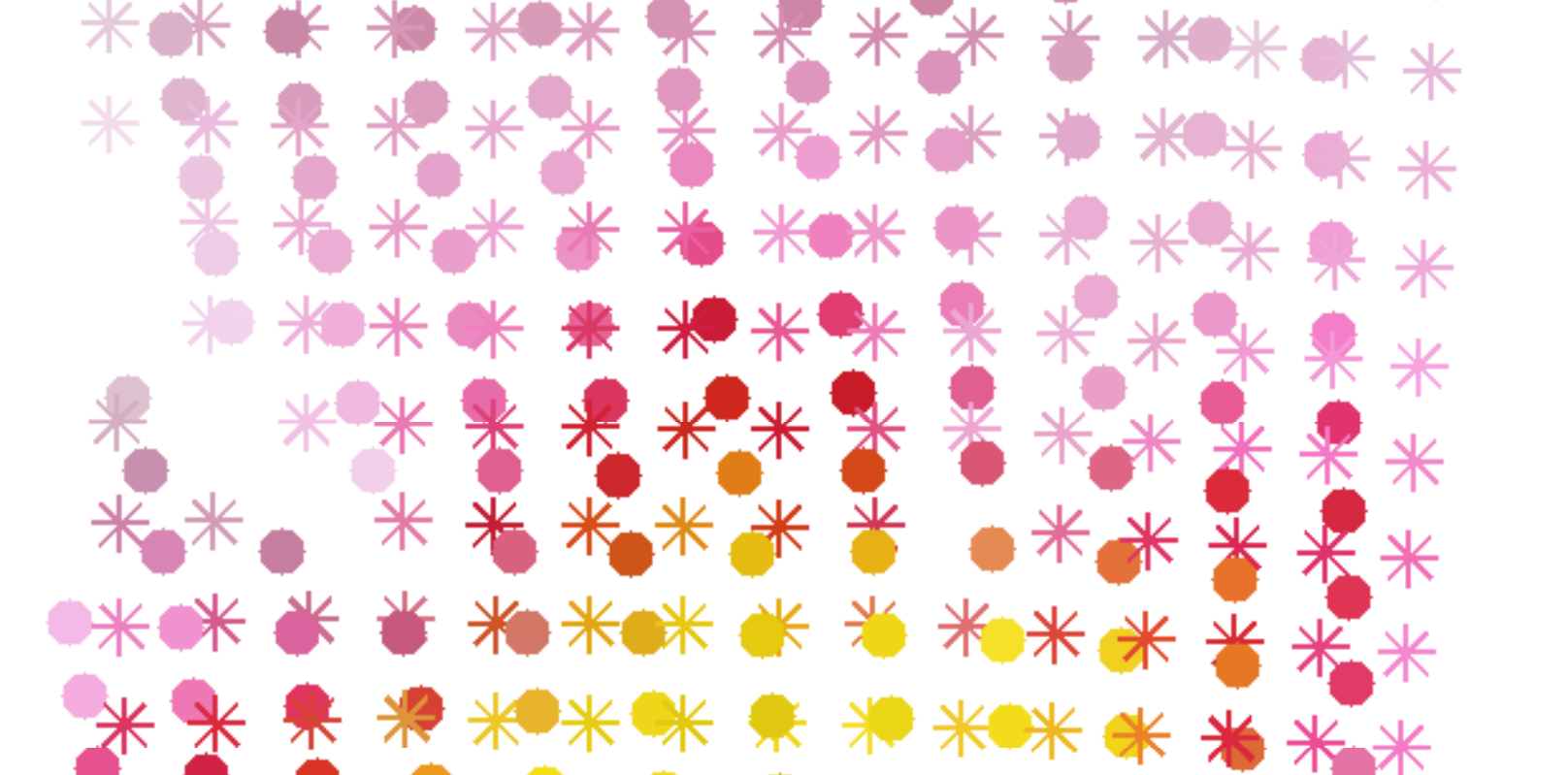}
	 \includegraphics[width = 0.3\textwidth]{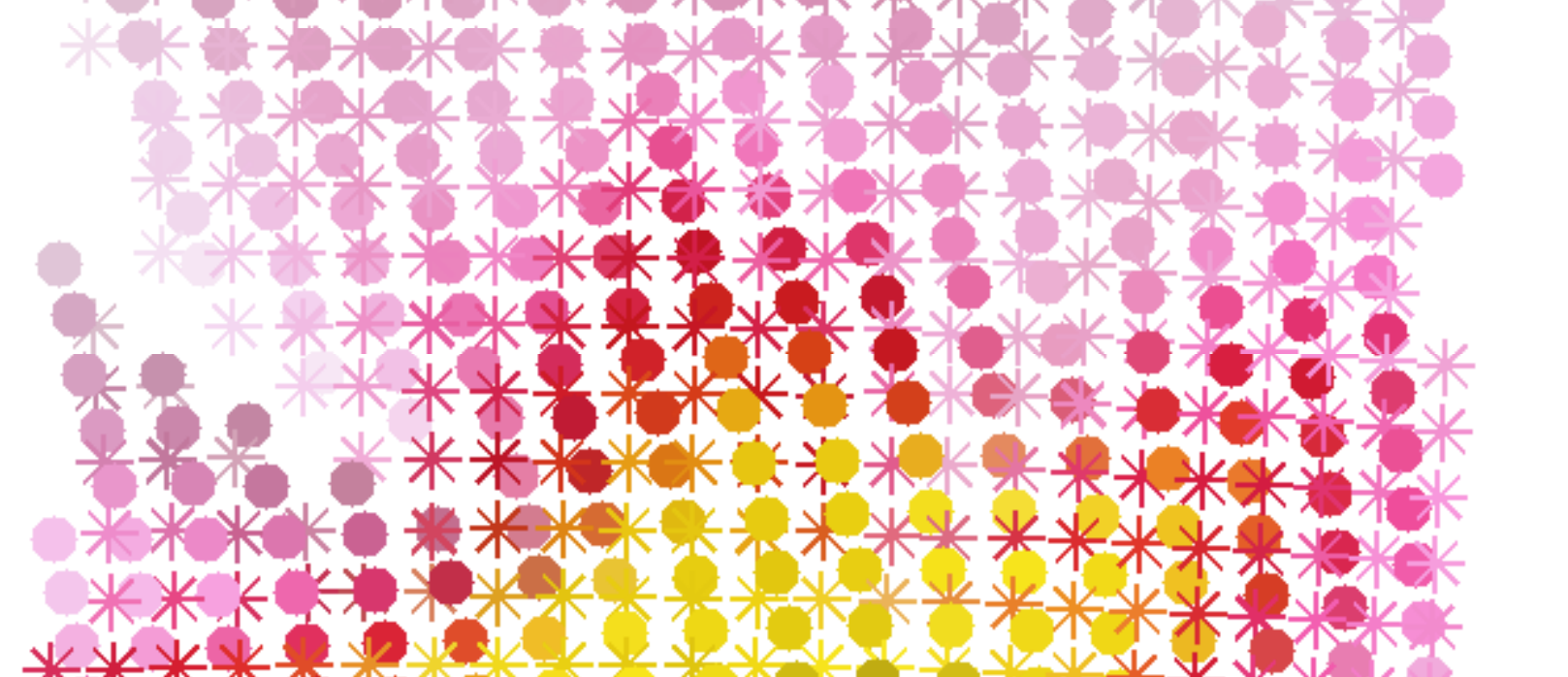}}

	\caption{Enlarged example of the ROI for the flower sampled with bilinear sampling. The first row shows the CCPD and the second the original algorithm. \mscc{The data size is, from left to right, 250, 500, and 1000 points for the bilinear.}}
	
  \label{fig:plant:detail:bilinear}
\end{figure}

To analyse the registration, we will pay special attention to a specific Region-of-Interest (ROI) for each model (i.e. those parts that are the aim of the study) depicted in Figure \ref{fig:ROI}. In the face, the ROI will correspond to the mouth and eyebrows as they are the parts which are mainly displaced. The ROI in the flower will correspond to the central part, pink and yellow, as they do not deform in color unlike the rest of the leaves (i.e. the deformation produces an enlargement of the tip of leaves, but the center remains the same). This simulates the growth of a flower, where not all parts grow in the same way. Figure \ref{fig:reg:face:1000} and \ref{fig:reg:plant:1000} show the registration result for all different sampling techniques using CCPD (first row) and CPD (second row). Figures \ref{fig:face:detail:gng} and \ref{fig:face:detail:bilinear} present a detailed view of this analysis for the face shape, and Figure \ref{fig:plant:detail:gng} and \ref{fig:plant:detail:bilinear} for the flower shape. For both shapes the first figure shows the registration using a GNG sampled dataset and the second the bilinear sampled dataset.

We can make several conclusions from the results of the experiments for the face shape. From Figures \ref{fig:reg:face:1000}, \ref{fig:face:detail:gng} and \ref{fig:face:detail:bilinear} we can see that the proposed CCPD achieves better alignment. If we pay attention to the eyebrows area, it can be seen that the alignment of CCPD results is better as it takes into account the color. In the detailed figures, it is easier to perceive this situation.

The flower shape has similar behavior to the face in the registration results. When the data comes from either color-based or NC-based, both CCPD and CPD achieves similar results. Moreover, when the data has been sampled using GNG or bilinear, the proposed CCPD achieves higher registration accuracy than the original CPD. As the deformation in this shape is not isometric, the tips of some leaves are the parts that get larger, CCPD moves the points differently at the tip of the leaves than the ROI, achieving accurate results. However, as CPD moves coherently, the points shrink all together (the registration is from the larger to the original position) such that ROI ends with a wrong color alignment. This situation is presented in Figure~\ref{fig:plant:detail:gng} and Figure~\ref{fig:plant:detail:bilinear}. 
 
Figures~\ref{fig:plant:detail:gng} and \ref{fig:plant:detail:bilinear} show a detail of the registration to visually evaluate the accuracy of both methods. It is easy to appreciate that CCPD achieves better results than the original version in the alignment. 

Finally, processing time of the registration process has been evaluated and shown in Table \ref{tab:time:face} for the face data, and Table \ref{tab:time:plant} for the flower. The original CPD always achieves lower times due to the number of operations. To calculate the posterior probability in CCPD, it is necessary to estimate for each point both color and location probability. Moreover, the convergence is not the same in both methods, as CCPD commonly needs more iterations to achieve a more accurate result. The time of both tables is presented in seconds, and is shown for each sampling method. The columns are: first, the sampling method; from second to fourth, the sampling rates for the first deformation; and from fifth to seventh, the three sampling rates for the second deformation. As a rule-of-thumb, the CCPD computing times are about 3 times longer thatn the CPD times. 
 
\begin{table}
  \centering
  \caption{Average processing time in seconds CCPD and CPD in face shape.}
  \scalebox{0.95}{
    \begin{tabular}{r|rrr|rrr}
    \hline
          & \multicolumn{3}{c}{Face deformation 1} & \multicolumn{3}{c}{Face deformation 2} \\
    \hline
    \multicolumn{7}{c}{CCPD} \\
    \hline
      &  250  &  500  &  1000  &  250  &  500  &  1000 \\
    \hline
    bilinear & 11.7988 & 58.4475 & 560.6209 & 30.2463 & 98.0815 & 654.7234 \\
    normals & 17.0563 & 114.3599 & 555.5070 & 24.3791 & 62.2356 & 873.7409 \\
    color & 11.7048 & 58.2177 & 703.9807 & 19.7996 & 69.2475 & 809.8889 \\
    NC    & 23.2824 & 141.5593 & 465.4034 & 25.8104 & 70.3541 & 700.6084 \\
    GNG   & 35.1574 & 121.0136 & 541.7851 & 31.5702 & 181.0353 & 698.8645 \\
    \hline
    \multicolumn{7}{c}{CPD} \\
    \hline
    bilinear & 3.4144 & 13.3339 & 58.8231 & 9.0456 & 44.0501 & 175.9423 \\
    normals & 7.5732 & 31.0907 & 173.2305 & 12.1042 & 45.0169 & 181.4115 \\
    color & 5.501 & 27.6411 & 174.1651 & 7.9162 & 29.5334 & 135.8508 \\
    NC    & 12.4572 & 36.2307 & 149.8286 & 7.1914 & 42.8539 & 193.599 \\
    GNG   & 11.3055 & 43.2056 & 171.6053 & 11.6725 & 44.3024 & 183.5948 \\
    \hline
    \end{tabular}}
  \label{tab:time:face}%
\end{table}%

\begin{table}
  \centering
  \caption{Average time processing CCPD and CPD in flower shape.}
  \scalebox{0.95}{
    \begin{tabular}{r|rrr|rrr}
    \hline
          & \multicolumn{3}{c}{Flower deformation1} & \multicolumn{3}{c}{Flower deformation2} \\
    \hline
    \multicolumn{7}{c}{CCPD} \\
    \hline
      &  250  &  500  &  1000  &  250  &  500  &  1000 \\
    \hline
	bilinear & 33.8404 & 100.1836 & 560.1712 & 13.4003 & 153.0062 & 562.0116 \\
    normals & 18.9529 & 74.6561 & 369.0443 & 20.0593 & 121.5026 & 537.5028 \\
    color & 13.0186 & 129.9349 & 458.0764 & 21.498 & 74.1095 & 361.1847 \\
    NC    & 52.0435 & 91.3152 & 388.7584 & 66.038     & 274.563    & 267.593 \\
    GNG   & 20.7716 & 112.4218 & 443.8667 & 18.5236 & 66.7437 & 615.4628 \\
    \hline
    \multicolumn{7}{c}{CPD} \\
    \hline
    bilinear & 9.8136 & 43.4829 & 175.2797 & 11.2072 & 43.2063 & 172.0116 \\
    normals & 11.4153 & 45.7999 & 181.3168 & 11.8425 & 42.7337 & 172.5745 \\
    color & 11.5049 & 42.8385 & 175.7153 & 9.9246 & 42.8615 & 175.9582 \\
    NC    & 11.79 & 43.2504 & 174.0308 & 11.2697 & 43.7694 & 174.9237 \\
    GNG   & 11.8023 & 45.2732 & 176.094 & 11.1835 & 43.7019 & 171.5101 \\
    \hline
    \end{tabular}}
  \label{tab:time:plant}%
\end{table}%

\subsection{Real data experimentation}\label{exp:real}
\begin{figure}
  \centering
	\includegraphics[width = 0.28\textwidth]{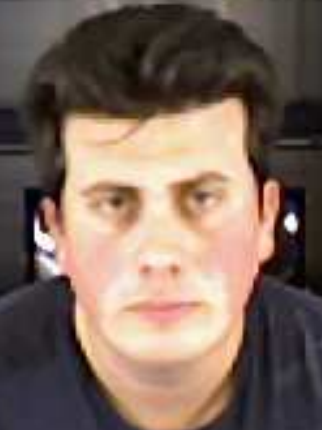} \quad
	 \includegraphics[width = 0.28\textwidth]{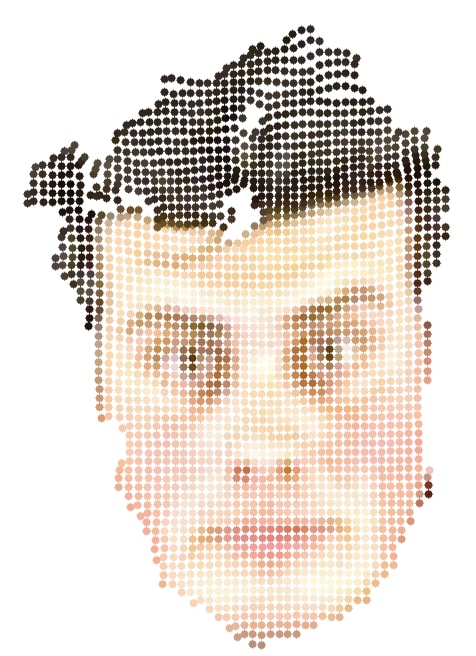} \quad
	 \includegraphics[width = 0.25\textwidth]{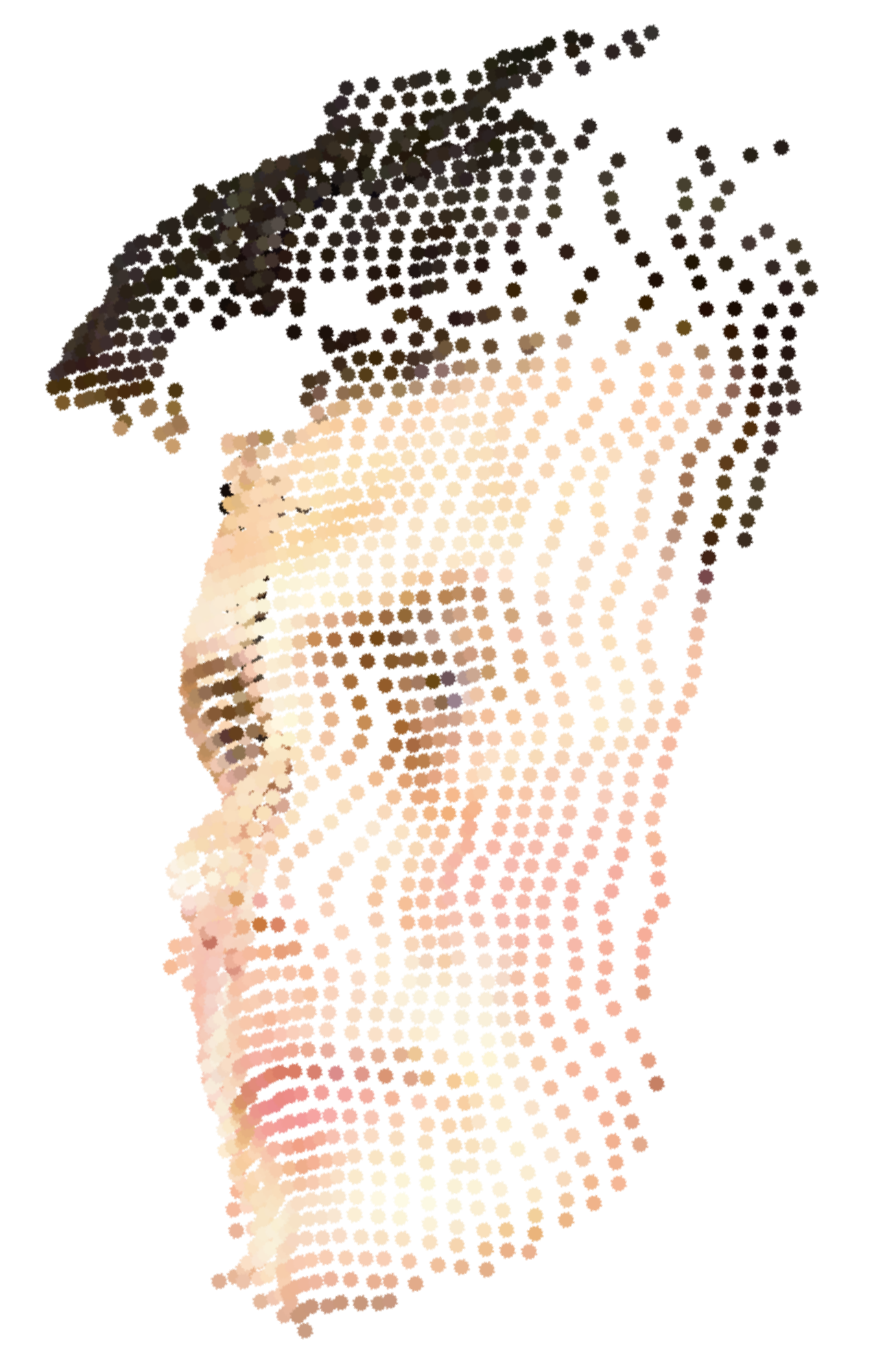}
	
	\includegraphics[width = 0.28\textwidth]{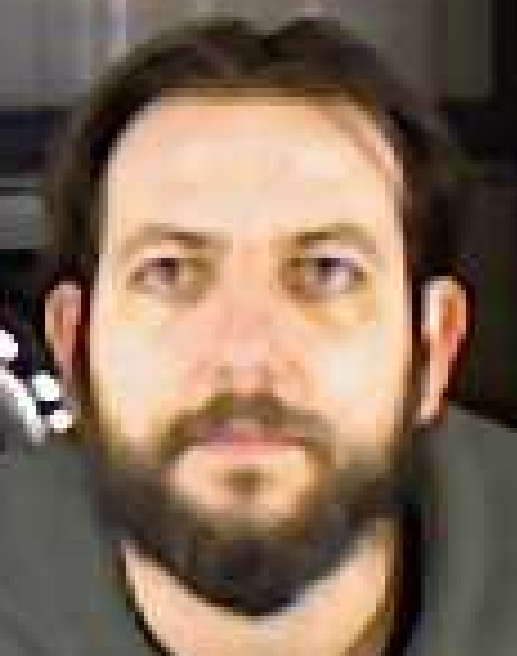} \quad
	 \includegraphics[width = 0.28\textwidth]{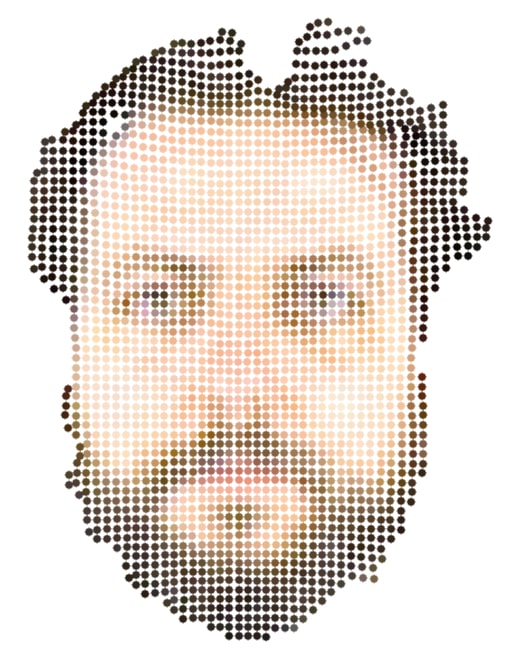} \quad
	 \includegraphics[width = 0.25\textwidth]{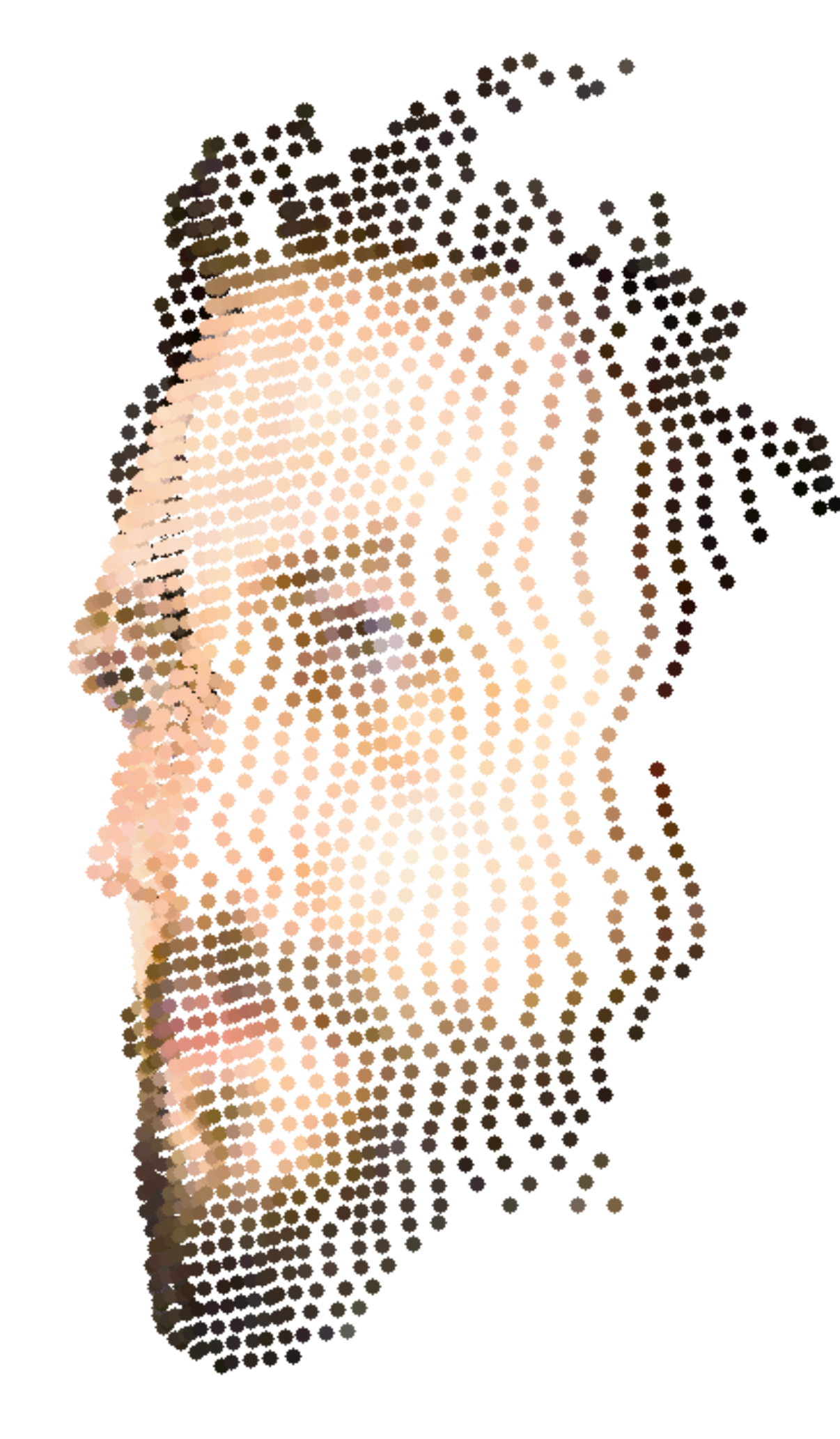}

	\includegraphics[width = 0.28\textwidth]{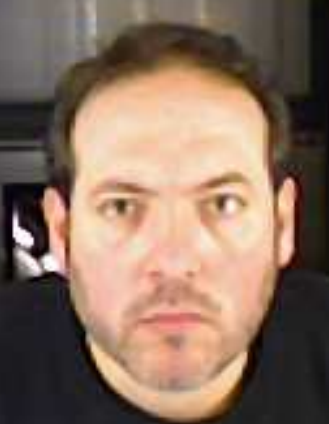} \quad
	 \includegraphics[width = 0.28\textwidth]{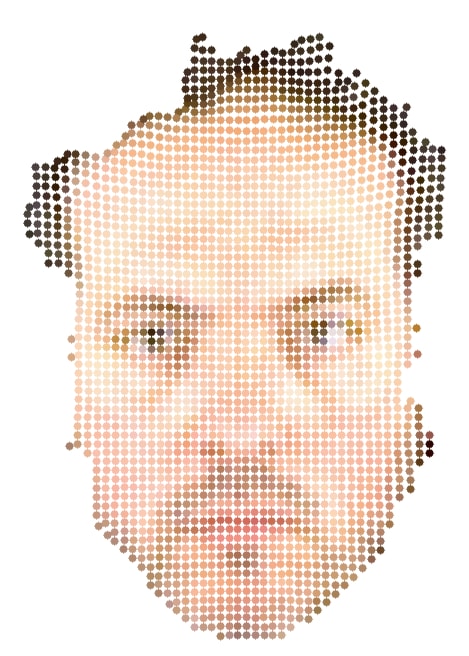} \quad
	 \includegraphics[width = 0.25\textwidth]{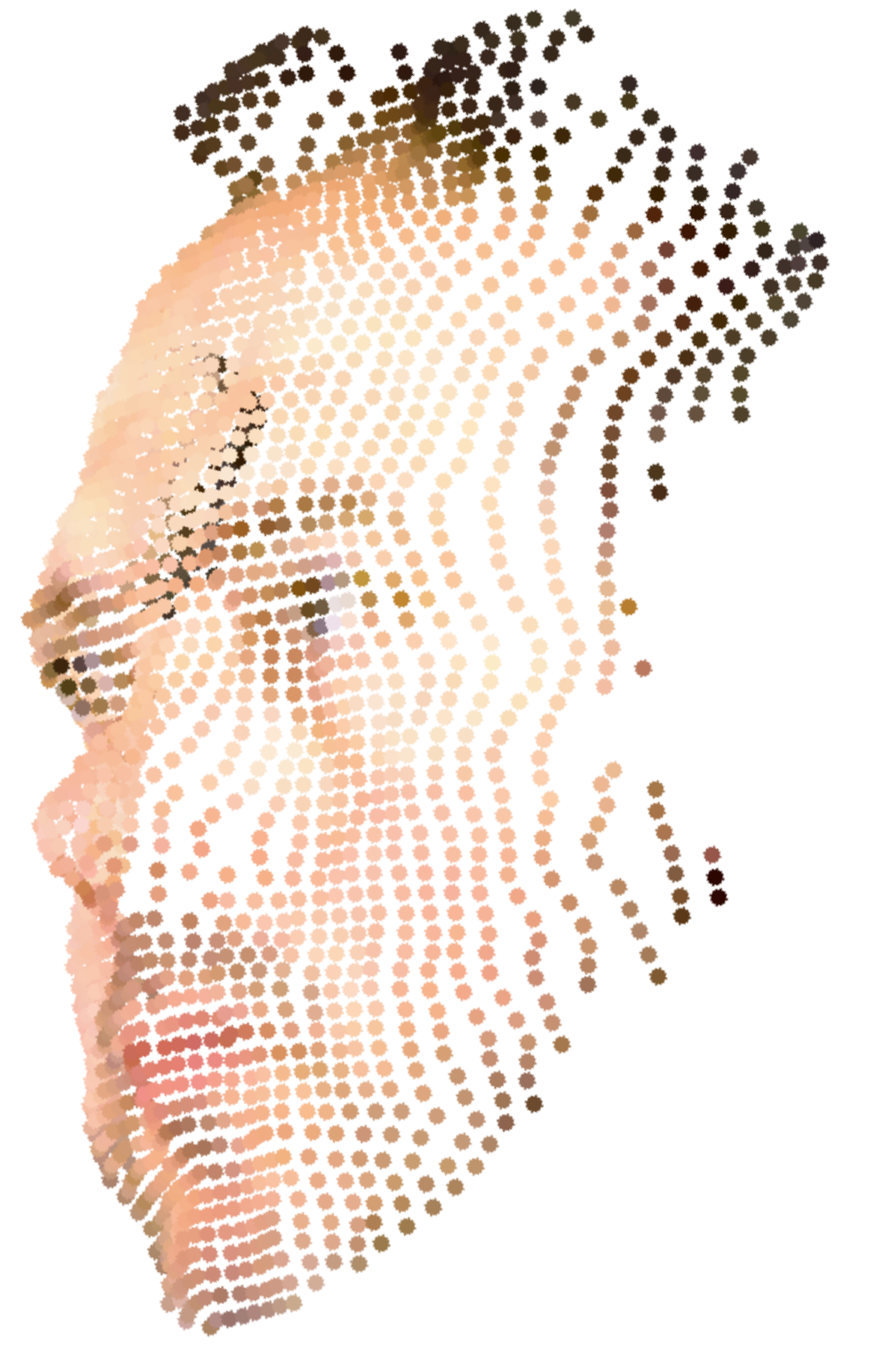}
	 	
		\caption{ \msc{Real data used for the non-rigid experiments. The first column shows the original color images, the second and third show the 3D point cloud data from front and side of the faces.} }
  \label{fig:ccpd:real:data}
\end{figure}

To evaluate the method in real conditions, experiments with data from a general purpose RGB-D sensor has been carried out. In this case, a face with different expressions is used to evaluate the non-rigid registration using CCPD, against CPD. Due to the absence of ground truth, the data will be visually evaluated to analyse the performance of both methods. Figure \ref{fig:ccpd:real:data} shows the data used in this experimentation.

\begin{figure}
  \centering
	 \includegraphics[width = 0.2\textwidth]{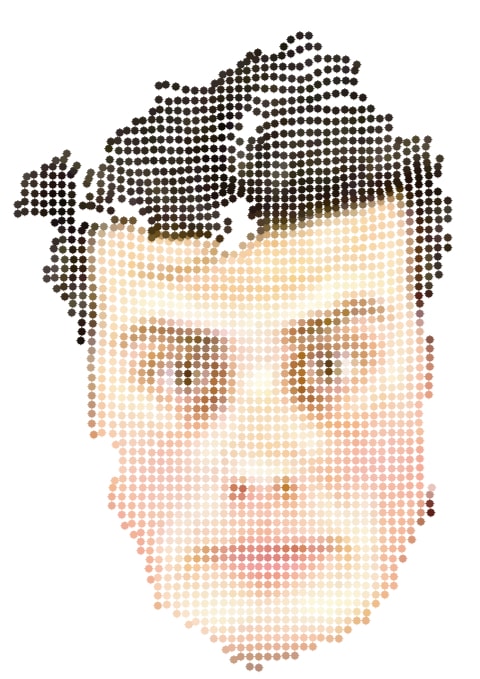}
	 \includegraphics[width = 0.2\textwidth]{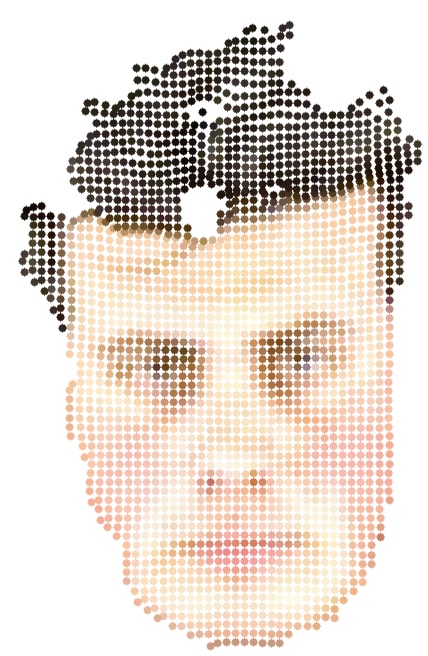}
	 \includegraphics[width = 0.2\textwidth]{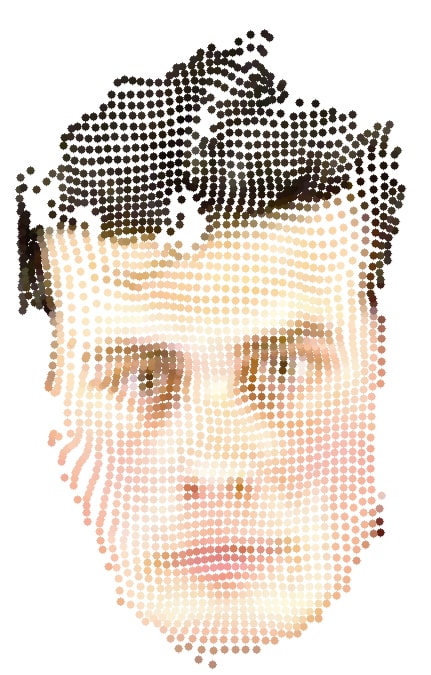}
	 \includegraphics[width = 0.2\textwidth]{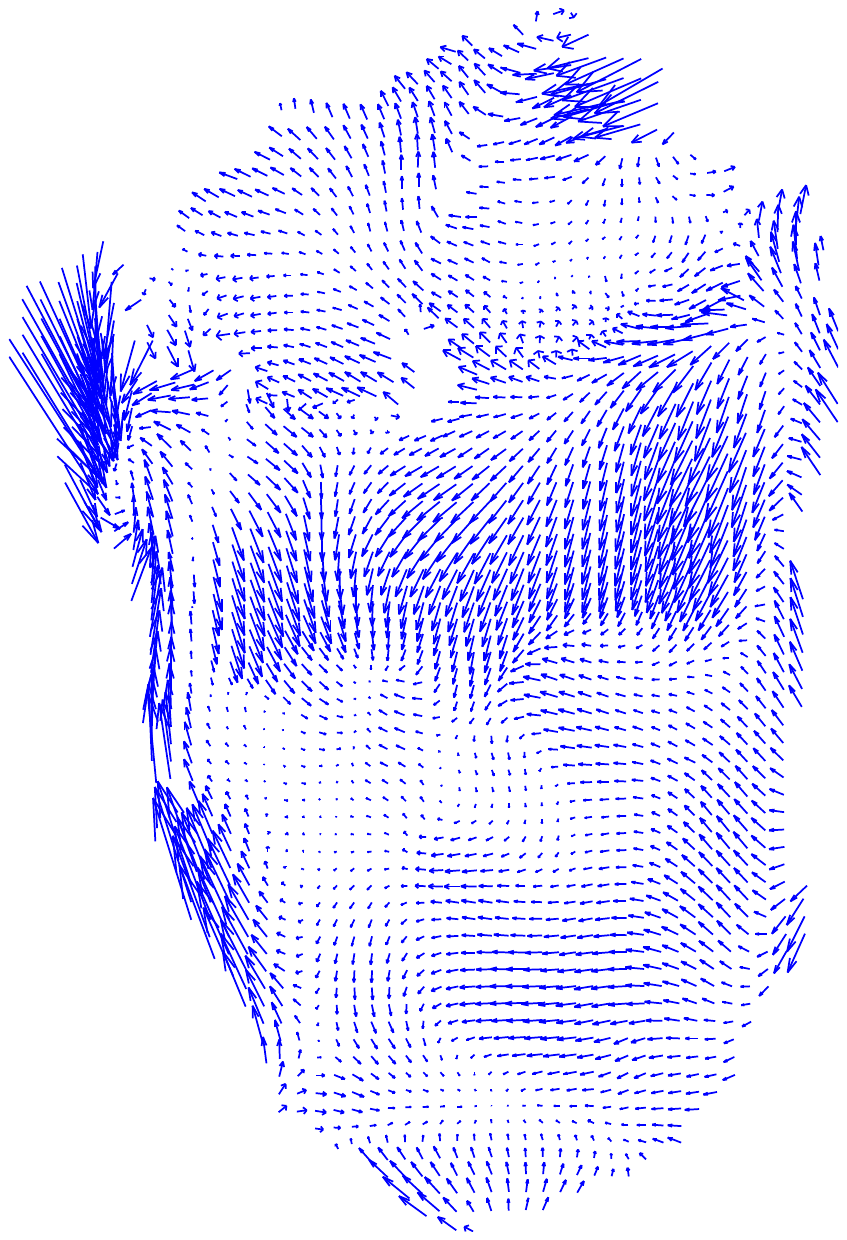}
	\\
	
	 \includegraphics[width = 0.2\textwidth]{./face-cpd-face-PS-Victor-origin.jpg}
	 \includegraphics[width = 0.2\textwidth]{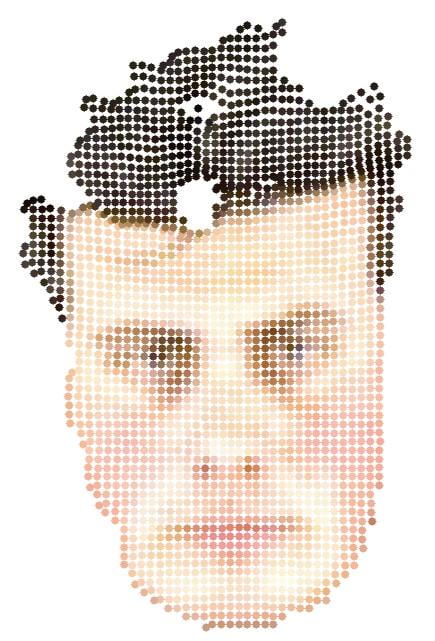}
	 \includegraphics[width = 0.2\textwidth]{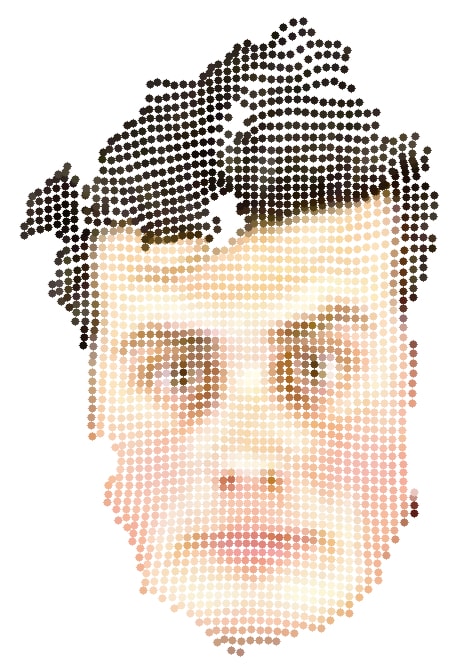}
	 \includegraphics[width = 0.2\textwidth]{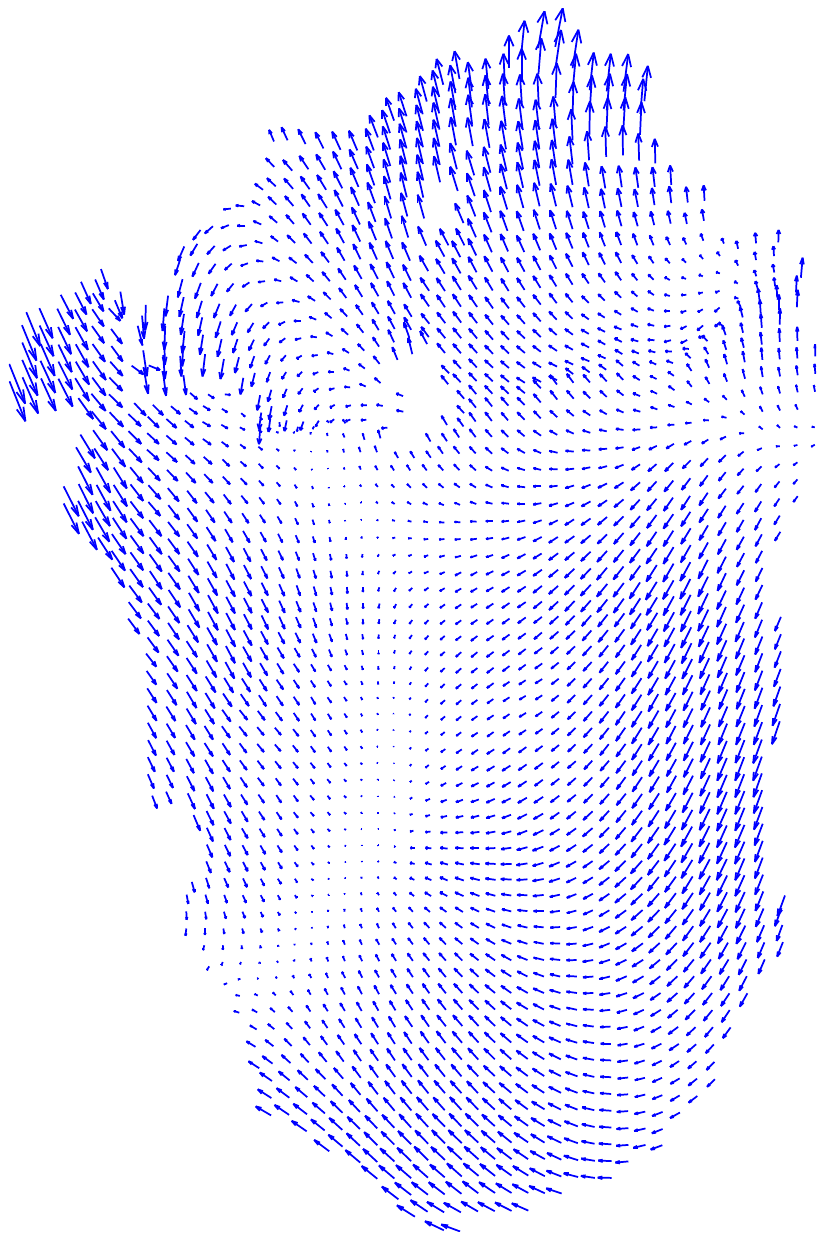}
	
		\caption{Real data registration, eyebrow rising. The top row is CCPD, and the second CPD. From left to right, original deformation, target shape, registered shape, and data flow}
  \label{fig:real:victor}
\end{figure}

Figure~\ref{fig:real:victor} shows an eyebrow rising deformation. The target is a neutral expression and the deformation is a surprise expression. The registration results of CCPD accurately aligns the shapes. The right column shows the data flow. It clearly shows the movement of the eye region downward, from the surprise expression to the neutral one. In this case, CPD only takes into account the location, so it cannot align properly the eyebrows, resulting in a wrong homogeneous displacement.

\begin{figure}
  \centering
	 \includegraphics[width = 0.2\textwidth]{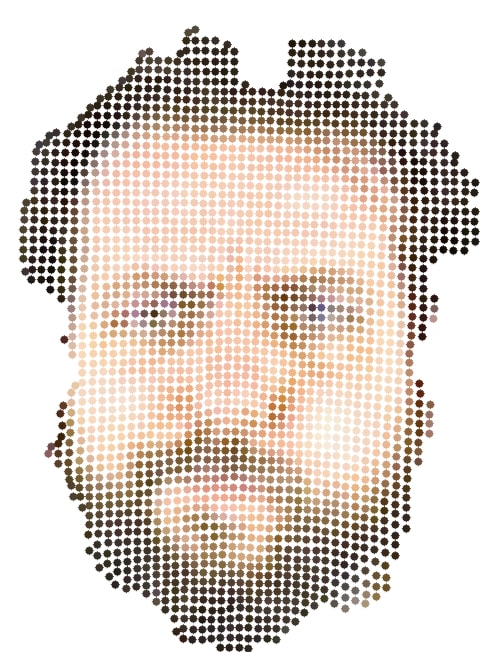}
	 \includegraphics[width = 0.2\textwidth]{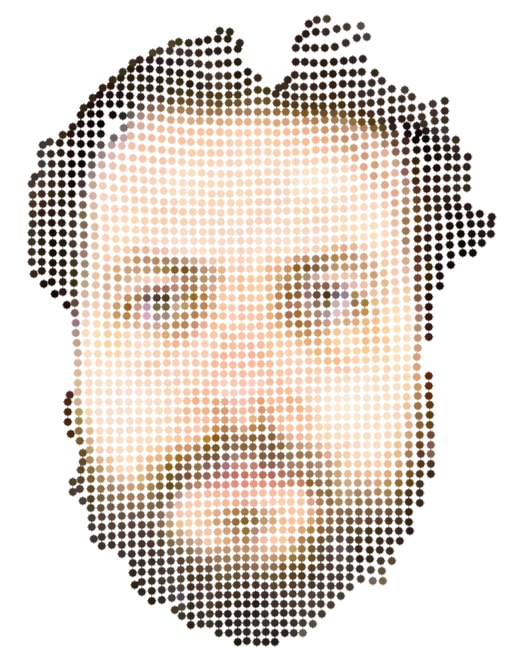}
	 \includegraphics[width = 0.2\textwidth]{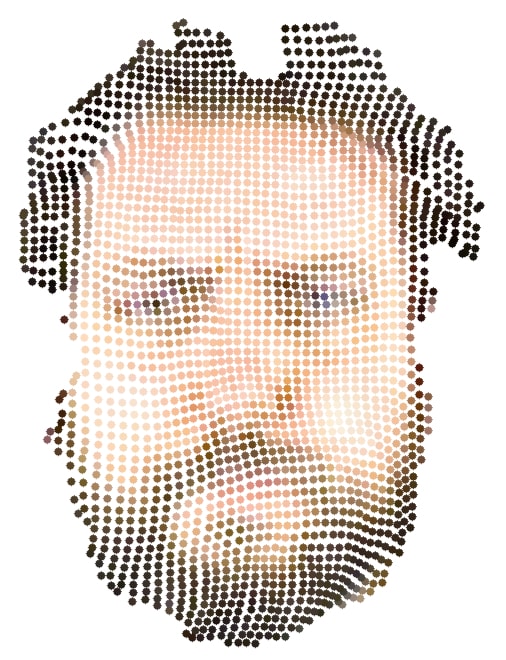}
	 \includegraphics[width = 0.2\textwidth]{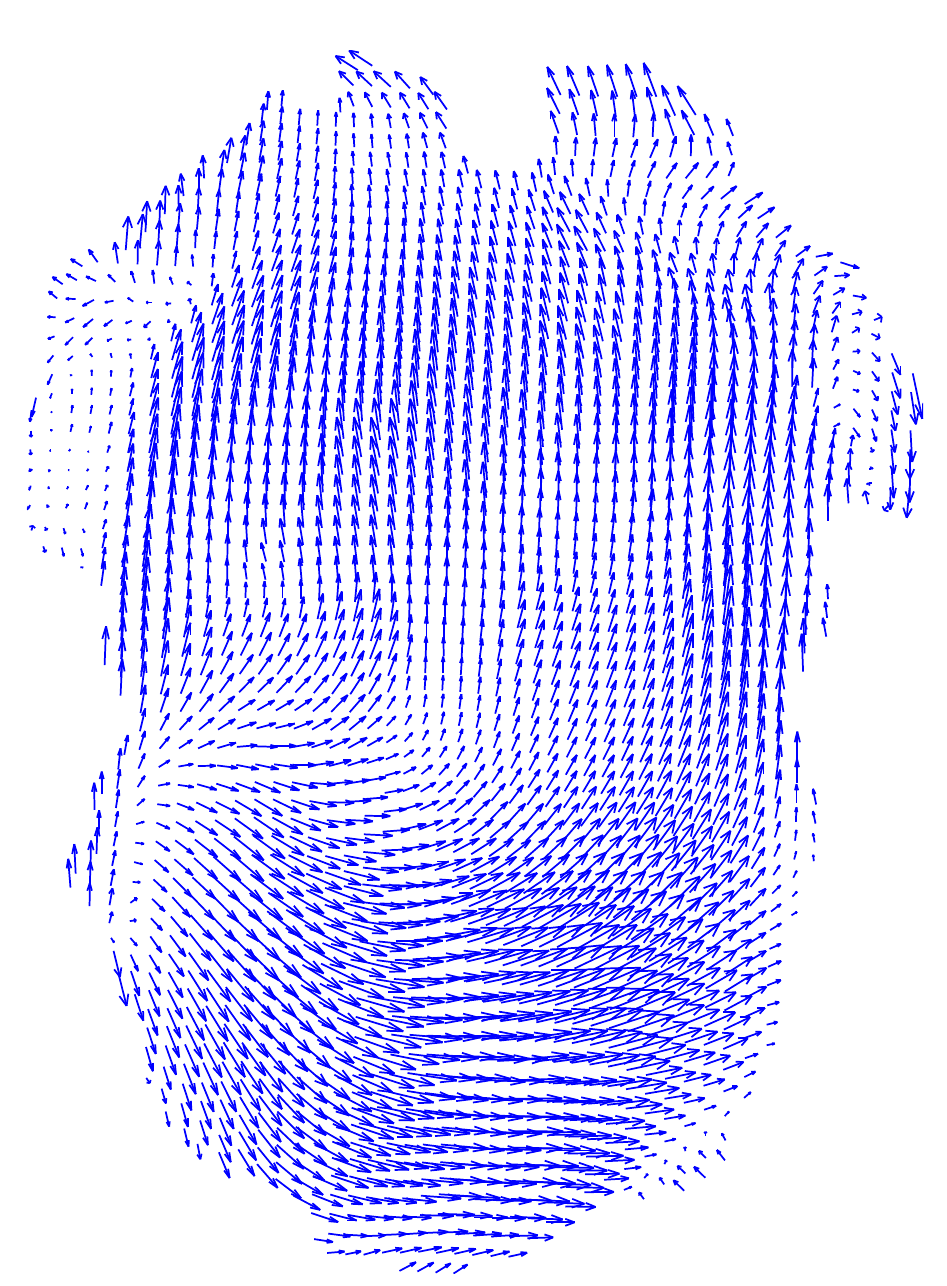}
	\\
	
	 \includegraphics[width = 0.2\textwidth]{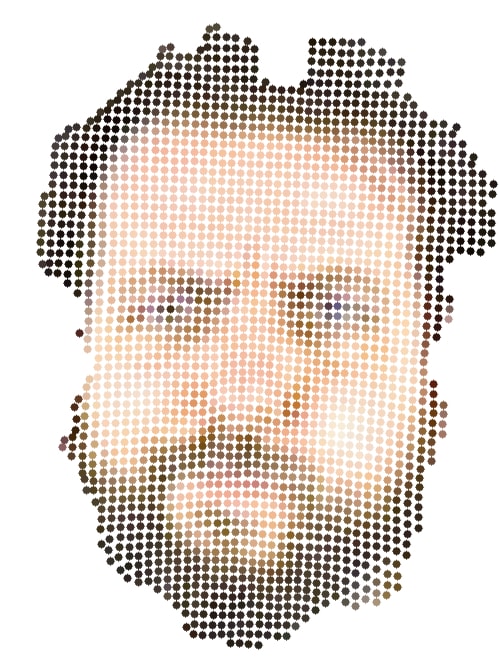}
	 \includegraphics[width = 0.2\textwidth]{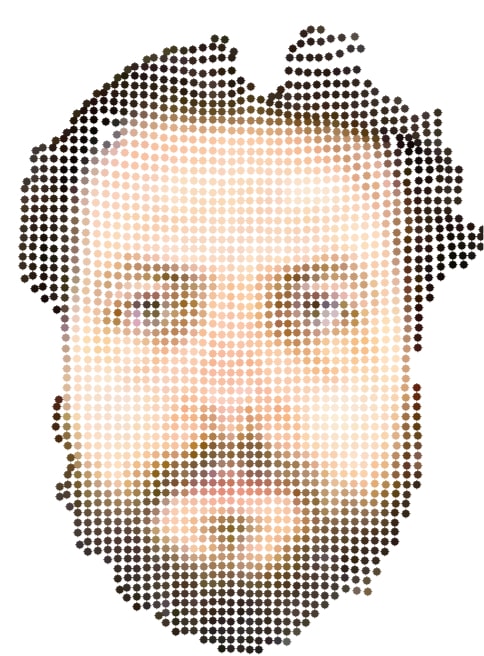}
	 \includegraphics[width = 0.2\textwidth]{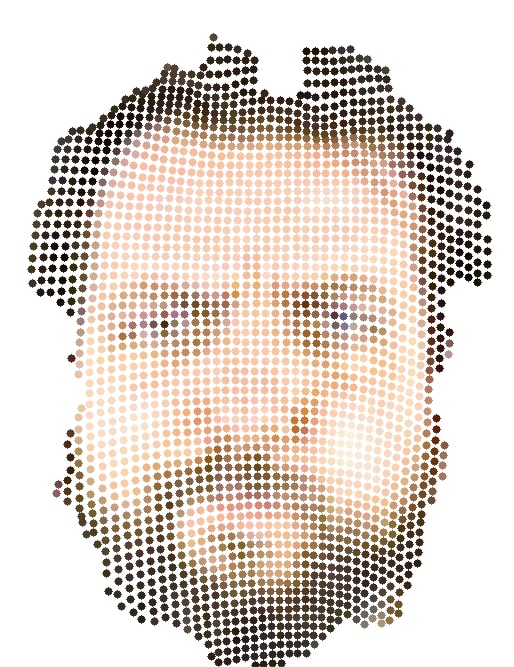}
	 \includegraphics[width = 0.2\textwidth]{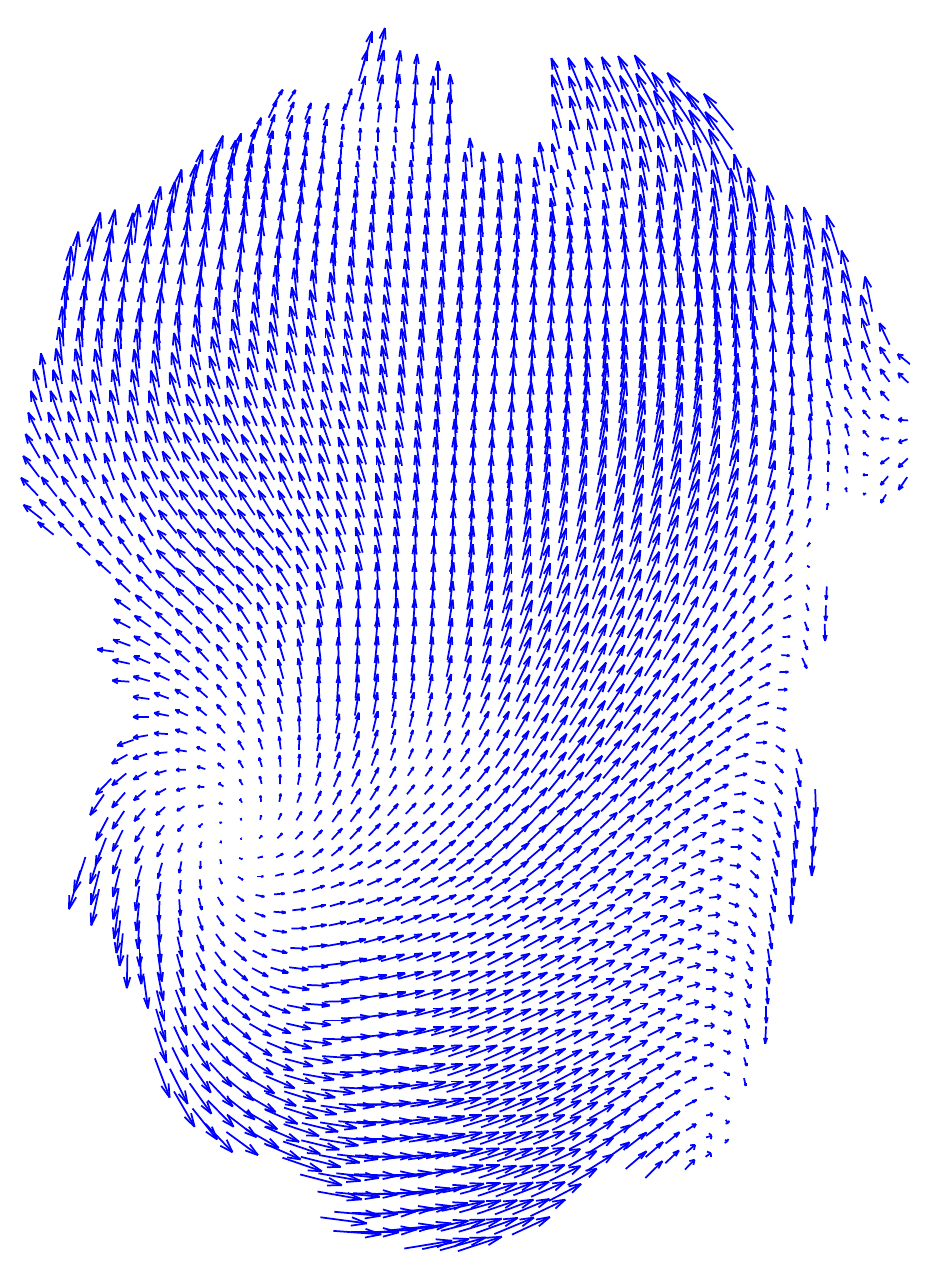}
	
		\caption{Real data registration, left cheek inflation. The top row is CCPD, and the second CPD. From left to right, original deformation, target shape, registered shape, and data flow}
  \label{fig:real:marcelo}
\end{figure}

Figure~\ref{fig:real:marcelo} shows a cheek inflating deformation. The person inflates one cheek so the mouth also moves to the side, the target is a neutral expression. The CCPD outperforms the registration of CPD as it uses the beard color to properly align and move the points into a correct location, where correct means the color of both $X$ and $Y$ registered are the most similar over the data. CPD, despite the good result, produces an inaccurate registration because it can only use the location information.  

\begin{figure}
  \centering
	 \includegraphics[width = 0.2\textwidth]{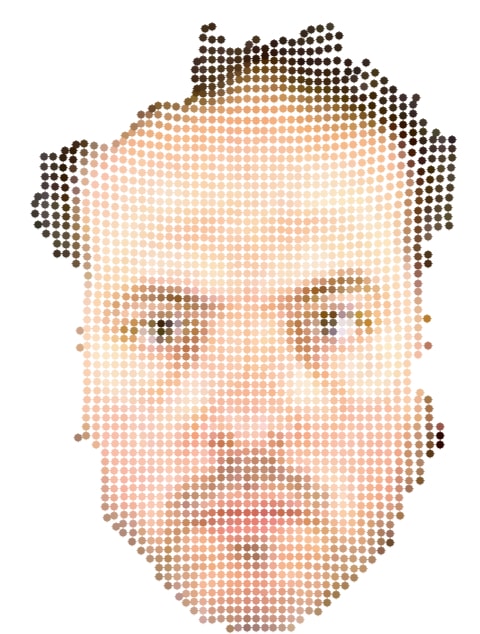}
	 \includegraphics[width = 0.2\textwidth]{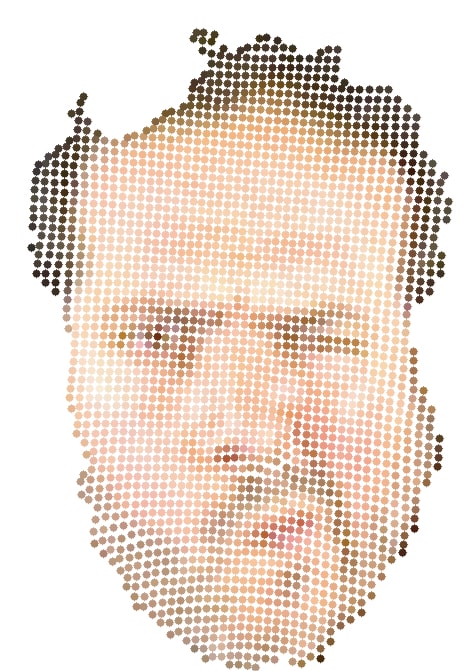}
	 \includegraphics[width = 0.2\textwidth]{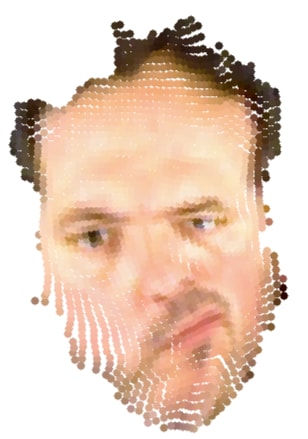}
	 \includegraphics[width = 0.2\textwidth]{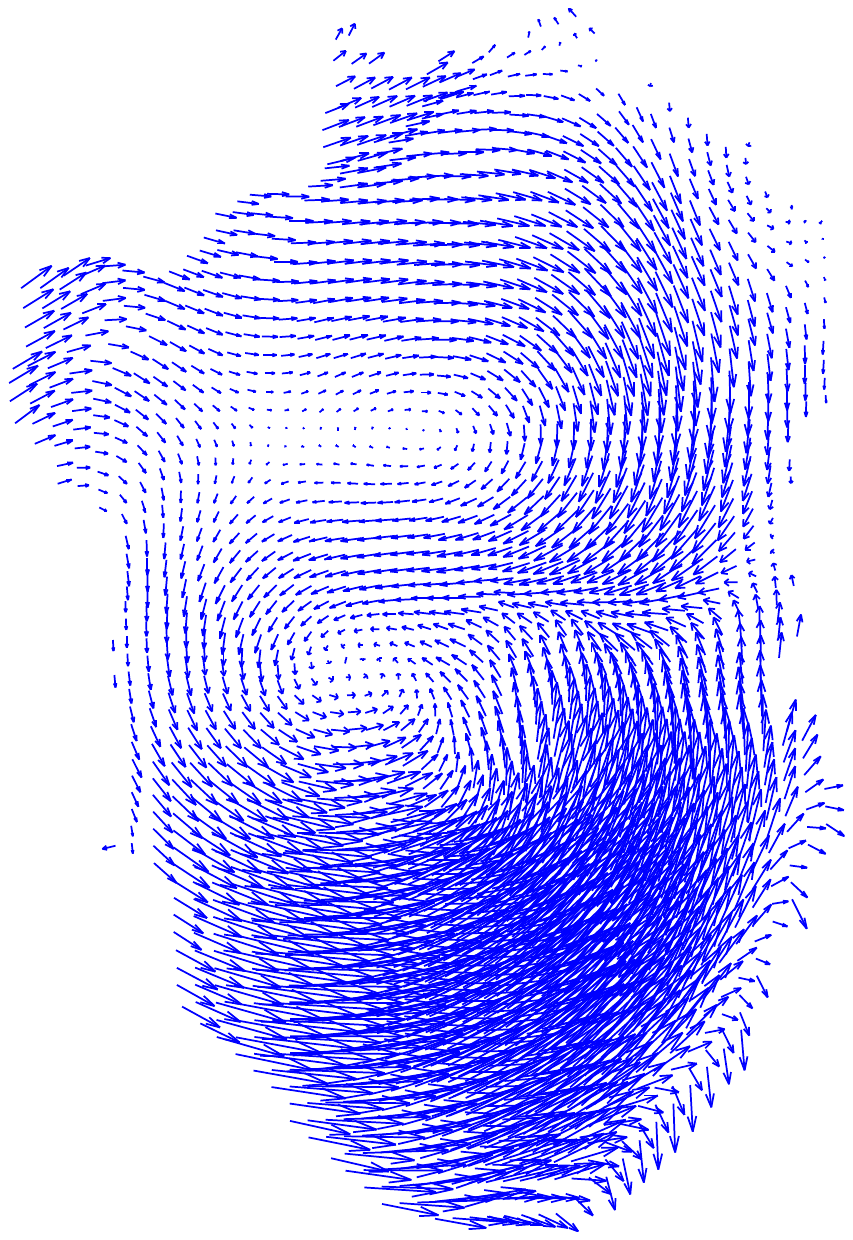}
	\\
	
	 \includegraphics[width = 0.2\textwidth]{./face-cpd-face-PS-Jorge-target.jpg}
	 \includegraphics[width = 0.2\textwidth]{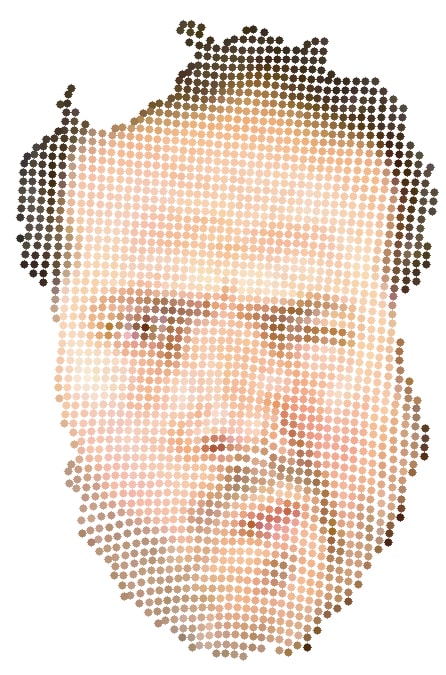}
	 \includegraphics[width = 0.2\textwidth]{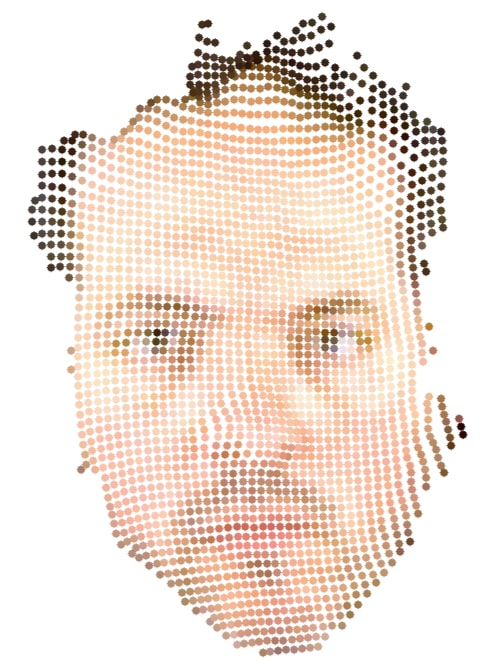}
	 \includegraphics[width = 0.2\textwidth]{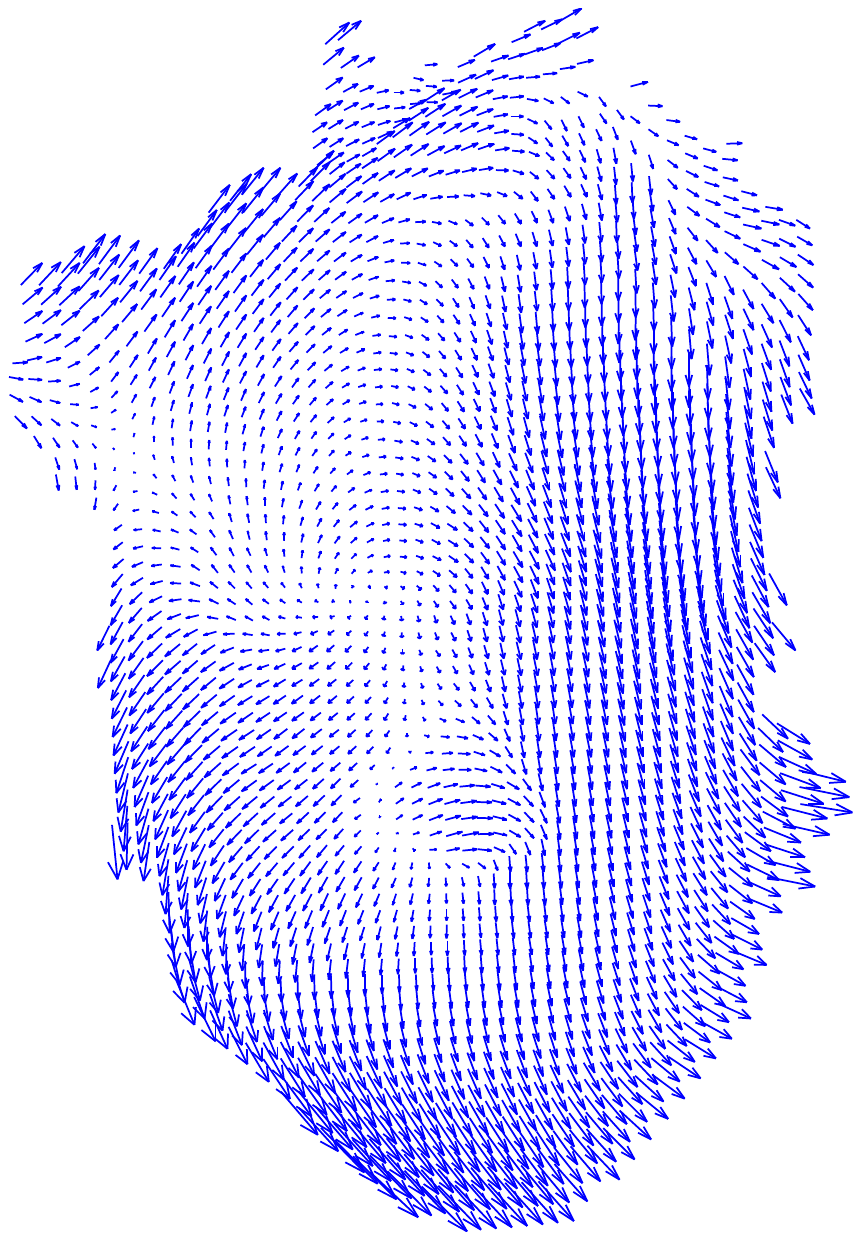}
	
		\caption{Real data registration, large deformation. The top row is CCPD, and the second CPD. From left to right, original deformation, target shape, registered shape, and data flow}
  \label{fig:real:jorge}
\end{figure}

Figure~\ref{fig:real:jorge} presents a large deformation. Here the face is highly deformed to one side and closing an eye. CCPD aligns the points better because the registered point set results in a correct location. CPD, however, cannot correctly move the points resulting in an inaccurate result (points registered have different color). 

\section{Discussion and conclusions} \label{sec:conclusions}
In this paper, a novel non-rigid registration approach called Color Coherent Point Drift (CCPD) is presented. This proposal, based on the well-know CPD, introduces color information in the correspondence estimation of non-rigid registration. The combination of color and location (3D position) information in the estimated correspondence improves the result in the presence of noise, missing data and outliers. 

In general terms, the proposed CCPD algorithm outperforms the original CPD in most cases. The new input, color, provides information that disambiguates situations where the 3D space provides the wrong correspondences. For example, a flower that grows is used because some parts remain the same but the tips of the leaves expand. Here, CPD returns a coherent movement which  moves the center points to a wrong position, while CCPD keeps the color in a good registration. 

The experiments included three parts: synthetic simple subjects, synthetic realistic subjects and real data. The simple subjects are those used in the original CPD but with added color information (a fish and a face). The realistic subjects have been obtained using Blensor, and the real data has been acquired using a Primesense Carmine RGB-D sensor. The first experiments with a fish and a face shape show how the proposed method is able to overcome noise, outliers, missing data and large deformations. To evaluate the outliers and missing data, first the registered dataset $Y$ is aligned to the target dataset $X$, this second set with outliers (points in $X$ without correspondences in the registered set $Y$), providing similar result for both CCPD and CPD. Secondly, missing data evaluation has been carried out by removing points in $X$, so that there are points in $Y$ without correspondences in $X$. In this evaluation, for the fish shape, CCPD had 4.82 times lower RMS error than CPD in registration accuracy for 21.9\% missing data and 23.1 times lower for 58.2\% missing data. For the face, CCPD had 24 times lower RMS error on average for all missing data tests than the original method. For a large deformation evaluation, a square shape was registered to the fish shape, obtaining better alignment by CCPD than by CPD for the RMS error (23.1 times lower RMS). In the case of the face, the large deformation moves the eyebrow up while the rest of face remains the same, which forces a non-coherent movement in a specific region. It has been visually evaluated with CCPD outperforming CPD. 

\msc{Experimental results show that a balanced adjustment of both color and location parameters, using the proposed CCPD meets the requirements of these registration problems, dealing with difficult data conditions (very high levels of noise and outliers in color or location space), approaching to the optimal solution. Nevertheless, including color information (CCPD) improves the registration process even taking into account very difficult color input data conditions. In the worst case, in presence of corrupt color data, the CCPD can become the original CPD with either large $\sigma_C$ or assigning 0 to $w_C$.}

For realistic data experiments, two subjects have been evaluated, a flower and a face. Both subjects have two deformations, one larger than the other. The face changes shape with expression. Ten eyebrow, and mouth are the regions that mainly deform, which can be treated as elastic deformations. The flower, with the growth of some leaves, can be seen as a free deformation as the subject changes both size and topology as new points appear in the deformation. CCPD has been evaluated and compared to CPD using the data provided by five downsampling methods which were used in previous works. The results have been visually evaluated, showing more accurate registration for the proposed method in most cases. The subjects, for all data (each downsampling method), are aligned not only by the point distribution, but also with a coherence in the color space (similar colors are aligned together). 

The real data includes three face deformations, from smaller to larger, returning more accurate registration results for the proposed method. The deformations of the shapes are better aligned by CCPD because the flow of the points is more similar and coherent to the expected (expected by visual inspection), by aligning the points using the shape and color information.

Generalization for multiple (e.g.: include topology along with color and location) spaces combination is the next step to be done. Moreover, evaluating biological growth using CCPD is a short term future work that will provide a very useful tool for many applications. As long term future work, we are interested in modifying the method to accelerate the process by comparing neighbor points instead of the whole data set. Moreover, an implementation of the method in a massive parallel processing GPU is proposed as future work to speed up the process.

\section*{References}

\bibliography{referencesFinal}

\end{document}